\documentclass[lettersize,journal]{IEEEtran}
\usepackage{amsmath,amsfonts}
\usepackage{array}
\usepackage{textcomp}
\usepackage{stfloats}
\usepackage{multirow}
\usepackage{multicol}
\usepackage{xcolor}
\usepackage{url}
\usepackage{verbatim}
\usepackage{graphicx}
\usepackage{tabularx}
\usepackage{tablefootnote}
\usepackage{amsthm}
\usepackage{hyperref}
 \hypersetup{
     colorlinks=true,
     linkcolor=blue,
     filecolor=blue,
     citecolor = blue,      
     }
\usepackage{caption}
\newtheorem{innercustomgeneric}{\customgenericname}
\providecommand{\customgenericname}{}
\newcommand{\newcustomtheorem}[2]{%
  \newenvironment{#1}[1]
  {%
   \renewcommand\customgenericname{#2}%
   \renewcommand\theinnercustomgeneric{##1}%
   \innercustomgeneric
  }
  {\endinnercustomgeneric}
}

\newcustomtheorem{customthm}{Theorem}
\newcustomtheorem{customlemma}{Lemma}

\usepackage{lipsum,afterpage,refcount}

\usepackage{algpseudocode}% http://ctan.org/pkg/algorithmicx
\usepackage{algorithm}% http://ctan.org/pkg/algorithm
\usepackage{algcompatible}
\def\BibTeX{{\rm B\kern-.05em{\sc i\kern-.025em b}\kern-.08em
    T\kern-.1667em\lower.7ex\hbox{E}\kern-.125emX}}
\usepackage{balance}

\usepackage{tikz}
\usepackage{textcomp}
\usepackage{hyperref}
\usepackage{lipsum}
\newcommand\copyrighttext{%
  \footnotesize \textcopyright 2024 IEEE. Permission from IEEE must be obtained for all other uses in any current or future media. This article has been accepted for publication in IEEE Journal of Selected Topics in Signal Processing. 
  DOI: \href{https://ieeexplore.ieee.org/abstract/document/10462147}{10.1109/JSTSP.2024.3374593}}
\newcommand\copyrightnotice{%
\begin{tikzpicture}[remember picture,overlay]
\node[anchor=south,yshift=10pt] at (current page.south) {\fbox{\parbox{\dimexpr\textwidth-\fboxsep-\fboxrule\relax}{\copyrighttext}}};
\end{tikzpicture}%
}

\usepackage{amsmath,amsfonts,bm}
\usepackage{amsthm}
\usepackage{amssymb}
\usepackage{algorithm}
\usepackage{algpseudocode}

\usepackage{graphicx}
\usepackage{subfigure}

\newtheorem{lemma}{Lemma}
\newtheorem{thm}{Theorem}
\newtheorem{definition}{Definition}
\newtheorem{corollary}{Corollary}
\newtheorem{rmk}{Remark}

\newtheorem{assumption}{Assumption}
\newtheorem{property}{Property}
\def\bfzero{{\boldsymbol{0}}}

\def\bfa{{\boldsymbol a}}
\def\bfb{{\boldsymbol b}}
\def\bfc{{\boldsymbol c}}

\def\bfe{{\boldsymbol e}}

\def\bfg{{\boldsymbol g}}

\def\bfp{{\boldsymbol p}}

\def\bfr{{\boldsymbol r}}
\def\bfs{{\boldsymbol s}}
\def\bft{{\boldsymbol t}}
\def\bfu{{\boldsymbol u}}
\def\bfv{{\boldsymbol v}}
\def\bfw{{\boldsymbol w}}
\def\bfx{{\boldsymbol x}}
\def\bfy{{\boldsymbol y}}
\def\bfz{{\boldsymbol z}}
\def\bfmu{{\boldsymbol \mu}}

\def\bfSg{{\boldsymbol\Sigma}}

\def\bfLM{{\boldsymbol\Lambda}}

\def\bfA{{\boldsymbol A}}
\def\bfB{{\boldsymbol B}}
\def\bfC{{\boldsymbol C}}
\def\bfD{{\boldsymbol D}}

\def\bfH{{\boldsymbol H}}
\def\bfI{{\boldsymbol I}}

\def\bfM{{\boldsymbol M}}

\def\bfP{{\boldsymbol P}}
\def\bfQ{{\boldsymbol Q}}
\def\bfR{{\boldsymbol R}}
\def\bfS{{\boldsymbol S}}
\def\bfT{{\boldsymbol T}}
\def\bfU{{\boldsymbol U}}
\def\bfV{{\boldsymbol V}}
\def\bfW{{\boldsymbol W}}

\def\bfZ{{\boldsymbol Z}}

\newcommand{\R}{\mathbb{R}}
\newcommand{\tcr}{\textcolor{red}}

\DeclareMathOperator\erf{erf}

\newcommand{\be}{\begin{equation}}
\newcommand{\ee}{\end{equation}}

\def\eqref#1{equation~\ref{#1}}
\def\1{\bm{1}}

\DeclareMathAlphabet{\mathsfit}{\encodingdefault}{\sfdefault}{m}{sl}
\SetMathAlphabet{\mathsfit}{bold}{\encodingdefault}{\sfdefault}{bx}{n}

\begin{document}
\title{How does promoting the minority fraction affect generalization? A theoretical study of the one-hidden-layer neural network on group imbalance}

\author{Hongkang~Li,~\IEEEmembership{Student Member,~IEEE;}
        ~Shuai~Zhang,~\IEEEmembership{Member,~IEEE;}
        ~Yihua Zhang,~\IEEEmembership{Student Member,~IEEE;}
        ~Meng~Wang,~\IEEEmembership{Senior Member,~IEEE;}
        ~Sijia~Liu,~\IEEEmembership{Senior Member,~IEEE;}
        ~and~Pin-Yu~Chen,~\IEEEmembership{Senior Member,~IEEE}% <-this % stops a space
	\thanks{The authors Hongkang Li and Dr. Meng Wang are with the Dept. of Electrical, Computer, and Systems Engineering,   Rensselaer Polytechnic Institute.  Email:   \{lih35, wangm7\}@rpi.edu. The author Dr. Shuai Zhang is with New Jersey Institute of Technology. Email: sz457@njit.edu. The authors Yihua Zhang and Dr. Sijia Liu are with the Dept. of Computer Science and Engineering, Michigan State University.  Email:   \{zhan1908, liusiji5\}@msu.edu. The author Dr. Pin-Yu Chen is with IBM Research. Email: pin-yu.chen@ibm.com.
 }}
%\author{IEEE Publication Technology Department
%\thanks{Manuscript created October, 2020; This work was developed by the IEEE Publication Technology Department. This work is distributed under the \LaTeX \ Project Public License (LPPL) ( http://www.latex-project.org/ ) version 1.3. A copy of the LPPL, version 1.3, is included in the base \LaTeX \ documentation of all distributions of \LaTeX \ released 2003/12/01 or later. The opinions expressed here are entirely that of the author. No warranty is expressed or implied. User assumes all risk.}}

%\markboth{Journal of \LaTeX\ Class Files,~Vol.~18, No.~9, September~2020}%
%{How to Use the IEEEtran \LaTeX \ Templates}

\maketitle
\copyrightnotice
\begin{abstract}
Group imbalance has been a known problem in empirical risk minimization (ERM), where the achieved     high \textit{average} accuracy is accompanied by    low accuracy in a \textit{minority} group. 
Despite   algorithmic efforts to improve the minority group accuracy, a theoretical    generalization analysis of ERM on individual groups remains elusive. By formulating the group imbalance problem with the Gaussian Mixture Model, this paper quantifies the impact of individual groups on the sample complexity, the convergence rate, and the average and group-level testing performance. Although our theoretical framework is centered on binary classification using a one-hidden-layer neural network, to the best of our knowledge, we provide the first theoretical analysis of the group-level  generalization of ERM in addition to the commonly studied average generalization performance. Sample insights of our theoretical results include that when  all group-level co-variance is in  the medium regime and all mean are close to zero,  the learning performance is most desirable in the sense of a small  sample complexity, a fast  training rate, and a high  average and group-level testing accuracy. Moreover, we show that increasing the fraction of the minority group in the training data does not necessarily improve the generalization performance of the minority group.  Our theoretical results are validated on  both synthetic and empirical datasets, such as CelebA and CIFAR-10 in image classification.
\end{abstract}


\section{Introduction}
 %Deep neural networks  \cite{LBH15} have demonstrated superior empirical performance    in various applications such as speech recognition \cite{KSH12} and computer vision \cite{GMH13, HZRS16}. 

Training neural networks with empirical risk minimization (ERM)  is a common practice to reduce the average loss  of a machine learning task evaluated on a dataset. However, recent findings \cite{BG18, MPL19, SRKL20, SKHL20, YWLZ22} have shown empirical evidence about a critical challenge of ERM, known as \textit{group imbalance}, where  a well-trained model that has high average  accuracy   may have significant errors on  the minority group that infrequently appears in the data. Moreover, the group attributes that determine the majority and minority groups are usually hidden and unknown during the training. %One way to increase the number of training samples is to  generate synthetic data 
The training set can be augmented by %synthetic data generated from 
data augmentation methods \cite{SK19} with varying performance, such as %but not limited to 
%geometric transformation (like 
cropping and rotation \cite{KSH12}, noise injection \cite{MSJU18}, and  generative adversarial network (GAN)-based   methods \cite{GIMXW14}. %The resulting %change of the test 
%performance varies for  augmentation methods and  tasks. 

 %%Distributionally robust optimization (DRO) assumes that the data are drawn from a few different distributions, and the training objective is to reduce worse-group training loss %worse-case   error over all potential distributions instead of the average error by ERM. DRO methods are either more computionally expensivie than solving ERM or require information that are not easily avaiable such as to which distribution each training sample belongs.  %Another line of work focuses on spurious correlations, which xxxx . Spurious corerelations is one of the reasons that could lead to the phonomia of low average error and high worse-group error, and the mitigation methods can downsample the majority group or remove the spurisos fueatures in solving ERM.  

%Te augmentations listed in this
%survey are geometric transformations, color space transformations, kernel flters, mixing
%images, random erasing, feature space augmentation, adversarial training, GAN-based

%Among the above efforts in handling multiple groups of data,  %mulitple distributions in the data, 
As ERM is a prominent  method  and enjoys great empirical success,  it is important to  characterize  the impact of ERM on group imbalance theoretically.
%a theoretical study %of the impact of multiple data distributions on 
%of 
%the learning performance of ERM  theoretically when the data contains multiple groups with diverse patterns, so as to %  is, while such analysis is necessary to
%evaluate ERM and compare with other methods like DRO formally.
However,
the  technical difficulty of analyzing the nonconvex ERM problem of neural networks results from the concatenation of nonlinear functions across layers, and the existing generalization analyses of ERM  often require strong assumptions and focus on the average performance of all data. %make overly simplistic assumptions and only focus on the average generalization performance.  %rather than the generalization on individual groups. %do not characterize the impact of different groups on the learning performance.
%In fact, compared with great empirical successes of ERM in learning deep neural networks, the sample complexity for  its generalization performance, i.e.,   the required number of training data to learn a model with high accuracy on the testing data, % of learning neural networks
%is much less investigated, 
% The existing works often make simplifying assumptions and do not characterize the impact of different groups on the learning performance.  %assumption of 
%only consider independent and identically distributed (i.i.d.)  training samples from one same distribution, which cannot  diverse features of different groups of data. 
For example, the neural tangent kernel type of analysis \cite{ADHL19,ALS19,ALL19,CG19, LL18, JGH18,LWLC22, SLW24} linearizes the neural network around the random initialization. % to remove the nonconvex interactions across layers. %The resulting ERM becomes  convex, and the convexity simplifies the analysis.
The generalization results are %distribution free and % results apply to an arbitrary unknown data distribution and 
  independent of the feature distribution and cannot be exploited to characterize the impact of individual groups.
  Ref.~\cite{LL18} provides the sample complexity analysis %studies the  generalization error of over-parameterized one-hidden-layer networks  
when the data comes from the mixtures of well-separated distributions but still %. However, \cite{LL18}  requires strong separation among different groups, %and as the number of group increases, each distribution has to be more concentrated. 
%and does not 
cannot characterize the learning performance of individual groups. 
  %Another line of works %The standard Gaussian distrbution is another common assumption in some recent works 
%\cite{DLT18,GMMM20, MMN18, SWL21, KWLS21, AL22, LWLC23, AL23, LWML23, LWLW23, DYMG23} considers one-hidden-layer neural networks because the ERM problem is already highly nonconvex, and the analytical complexity increases tremendously when the number of hidden layers increases. In some of these works \cite{DLT18,GMMM20, MMN18}, the input features are usually assumed to be i.i.d. samples drawn from the standard Gaussian distribution, and this data model cannot differentiate the majority and minority groups. 
In %other works that are usually categorized as ``feature learning'' type of analysis
another line of works \cite{SWL21, KWLS21, AL22, LWLC23, AL23, LWML23, CHZB23, LWLW23, DYMG23, LWLC24, ZLYC24}, people make data assumptions that the labels are determined merely by some input features and are irrelevant to other features or model parameters. The generalization analysis characterizes how the neurons learn important features. Our work follows the line of works \cite{ZSJB17, ZWLC20, FCL20, ZWLC21_self, LZW22, ZLWL23}, where the label of each data is generated by both the input distribution and the ground-truth model so that group imbalance can be characterized.  %and unimportant ones that areinput features are formulated as separable to directly determine the labels.}   %when the number of groups increases. %In contrast, this paper does not require any data separation and can handle an arbitrary number of Gaussian distributions with arbitrary mean and variance.
% Nevertheless, the learning performance clearly depends on the input data distribution in practice. \cite{LBOM98} states that the learning method converges faster if the inputs are whitened to be the standard Gaussian. Batch normalization \cite{IS15} modifies the mean and variance in each layer   and is a popular practical method to achieve fast and stable convergence. Various explanations such as  \cite{BGSW18,CPM20,STIM18} have been proposed to explain the enormous success of Batch normalization,  but little consensus exists on the exact   mechanism.

 \textbf{Contribution}: To the best of our knowledge, \textit{this paper provides the first theoretical  
 characterization of both the average and group-level generalization of a one-hidden-layer neural network trained by ERM on data generated from a mixture of distributions. }%that can overlap with each other. } 
 This paper considers the binary classification problem with the cross entropy loss function, % using one-hidden-layer neural networks,  % This paper %studies the generalization performance of neural networks in the `
 %following the ``teacher-student'' setup  \cite{GASK19,ZSJB17, ZSD17, ZYWG19, ZWXL20, FCL20, ZWLC20}, 
 with  training data   generated by a ground-truth neural network with  known architecture and unknown weights. The optimization problem  is challenging due to a high non-convexity from the multi-neuron architecture and the non-linear   sigmoid activation.%, and the learning is performed on a student network  with the same architecture. 

 Assuming the  features follow a Gaussian Mixture Model (GMM), where samples of each group are generated from a Gaussian distribution with an arbitrary mean vector and co-variance matrix,  this paper quantifies   the impact of   individual   groups  on the sample complexity,  the training convergence rate, and the average and group-level test error.  The training algorithm is the gradient descent following a tensor initialization and   converges linearly. %to the ground-truth model. 
 Our key results include %does not require any data separation and can handle an arbitrary 

\begin{figure}[t]
    \centering
    \subfigure[]{
        \begin{minipage}{0.46\textwidth}
        \centering
        \includegraphics[width=1\textwidth,height=0.46\textwidth]{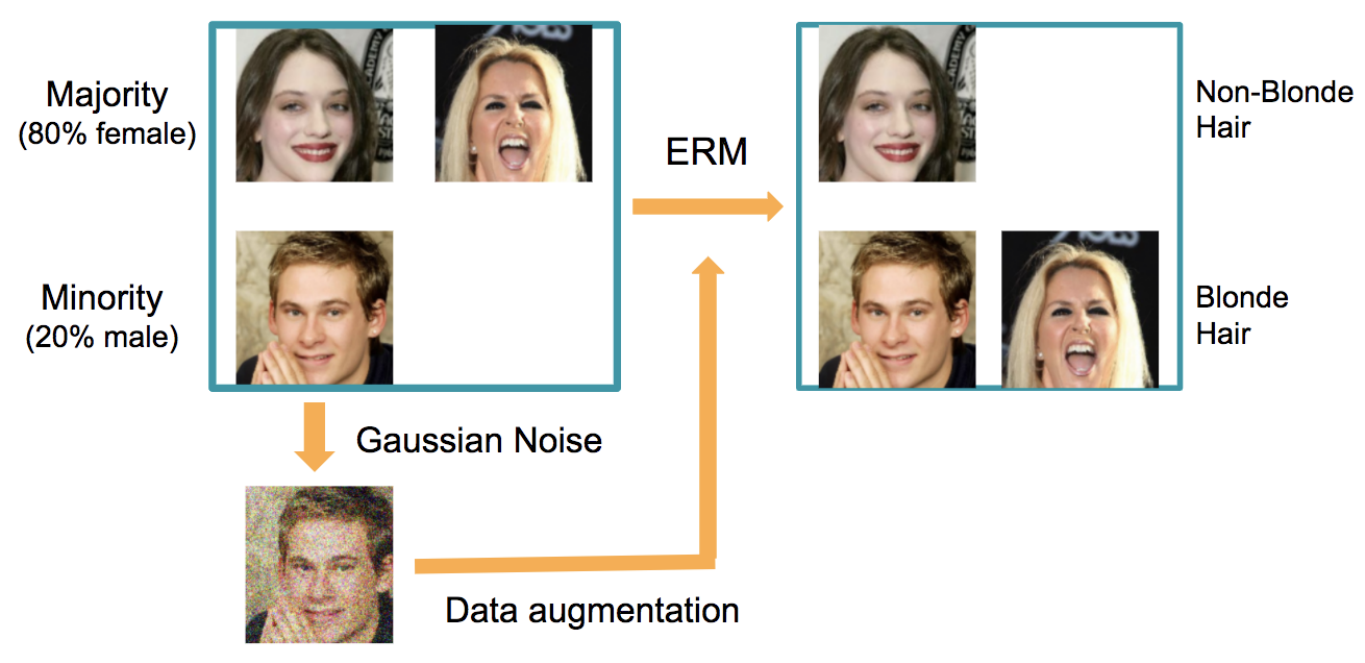}
        \end{minipage}
    }
    ~
    \subfigure[]{
        \begin{minipage}{0.4\textwidth}
        \centering
        \includegraphics[width=1\textwidth,height=0.53\textwidth]{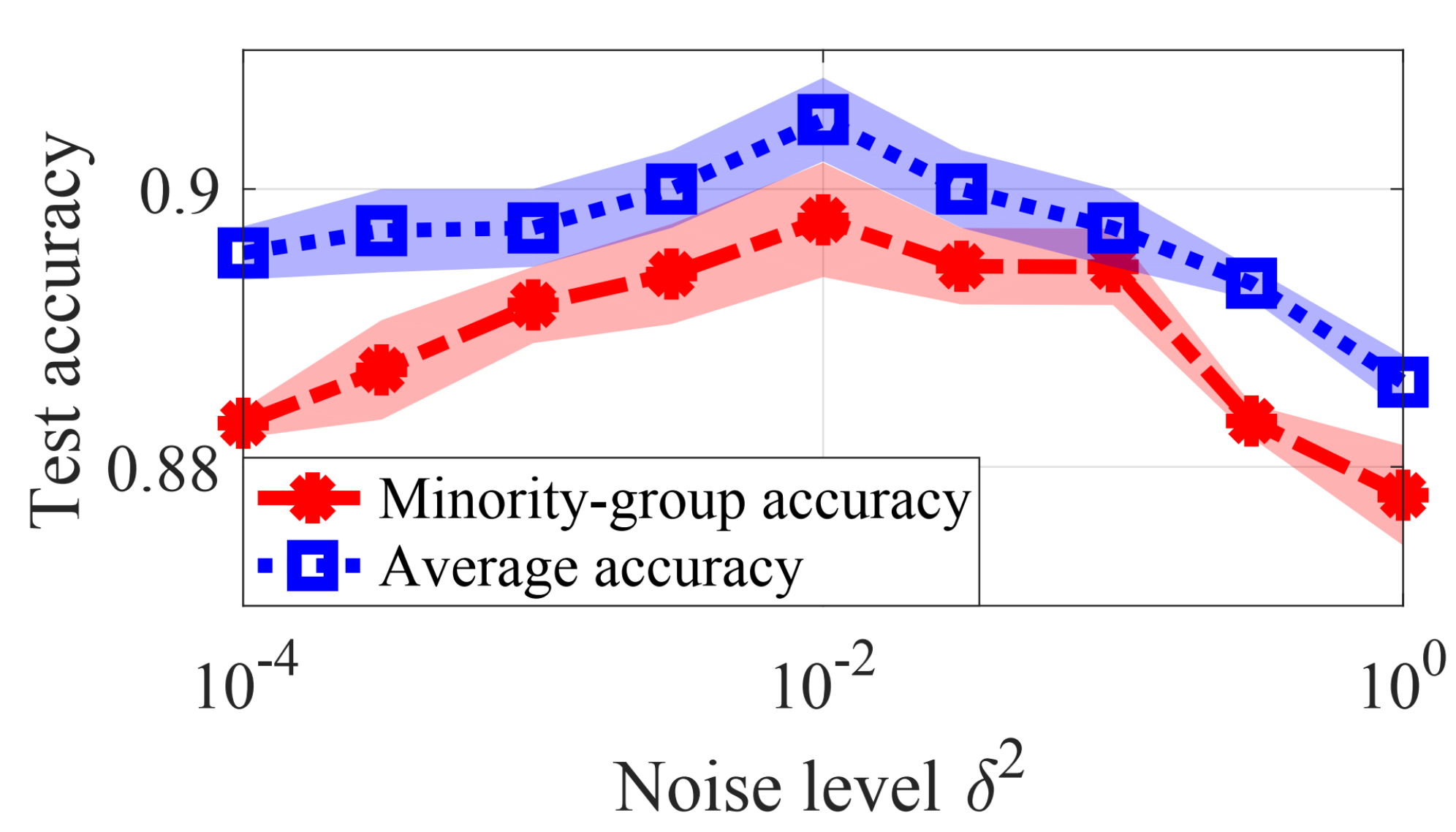}
        \end{minipage}
    }
    \vspace{-2mm}
    \caption{Group imbalance  experiment. (a) Binary classification on CelebA dataset using Gaussian   augmentation to control the minority group co-variance. (b) Test accuracy against the augmented noise level.}
    \label{figure: celebA_noise}
\end{figure}

(1) \textit{Medium-range group-level co-variance enhances the learning performance}. %When the norm of a co-variance matrix of one group approaches zero or increases from some constant value, 
When a group-level co-variance deviates from the medium regime, %either increasing or decreasing, 
the learning performance degrades in terms of higher sample complexity, slower convergence in training, and worse average and group-level generalization performance.  As shown in Figure \ref{figure: celebA_noise}(a), we introduce Gaussian augmentation to control the co-variance level of the minority group in the CelebA dataset \cite{LLWT15}. The learned model achieves the highest test accuracy when the co-variance is at the medium level, see Figure \ref{figure: celebA_noise}(b). %Thus, the learning performance is enhanced when each Gaussian has some constant level of co-variance. 
Another  implication is that the diverse performance of different data augmentation methods %on different tasks  
might   partially result from the different  group-level co-variance introduced by these methods. Furthermore, although our setup does not directly model the  batch normalization approach \cite{IS15} that modifies the mean and variance in each layer    to achieve fast and stable convergence, our result provides a theoretical insight that co-variance indeed affects the learning performance.

%that although data argumentation may improve the learning performance when the data have medium-level covariance after augmentation, while the learning performance can become worse if an argumentation method introduces a significant level of variance in the data.   This partially explains why different argumentation methods many have different impacts. 

(2) \textit{Group-level mean shifts from zero hurt the learning performance}. When a group-level mean deviates from zero, the sample complexity increases,  the algorithm converges slower,  and both the average and   group-level test error increases. Thus, the learning performance is improved if each distribution is zero-mean. This paper provides a similar theoretical insight to  practical tricks such as  whitening \cite{LBOM98}, subgroup shift \cite{KSGB22, MDM21}, population shift \cite{AS21, GMBT22}  and the pre-processing of making data zero-mean \cite{LBBH98}, that data mean affects the learning performance. %Although our setup does not directly explain some practical tricks such as subgroup shift \cite{KSGB22, MDM21}, population shift \cite{AS21, GMBT22}  and the pre-processing of making data zero-mean \cite{LBBH98}, this paper provides a similar theoretical insight that data mean affects the learning performance. }

(3) \textit{Increasing the fraction of the minority group  in the training data does not always improve its  generalization performance}. The generalization performance is also affected by the mean and co-variance of individual groups. % If a minority group has larger variance than all other groups, increasing the fraction of the minority group   may hurt the generalization performance in some cases. 
In fact,  increasing the   fraction of the minority group in the training data can have a completely opposite impact %on the generalization performance 
in different datasets.  %    our experiments on the CelebA dataset  in Figure \ref{figure: celebA_group} indicate that the testing accuracy increases as the minority-group percentage increases when the female group is the minority. In contrast, when the male group is the minority, the test accuracy deceases as its fraction in the training data increases. 

\section{Background and Related Work}
\textbf{Improving the minority-group performance with known group attributes}. With known  group attributes, %various training methods have been proposed to improve the minority group, where  
%One approach   generates synthetic data for the minority group. Examples   include %but not limited to 
%geometric transformation (like cropping and rotation), noise injection and  generative adversarial network (GAN)-based   methods. The resulting %change of the test %performance vary for  augmentation methods and  datasets. Another 
%one common idea is to  
%one line of works attempts to improve 
distributionally robust optimization (DRO) \cite{SKHL20} minimizes the worst-group training loss %in the training stage 
instead of solving ERM.  %can increase the number of training samples, especially for the minority group, with the hope of decreasing the minority group error. 
 DRO   is  more computationally expensive than  ERM and does not always outperform ERM in the minority-group test  error.  
%\textbf{Worst-group error through spurious correlations}. 
%Some works focus on the special data model of 
Spurious
correlations \cite{SRKL20} can be viewed as one reason of group imbalance, where strong associations between labels and irrelevant features exist in   training samples. %The model of spurious correlations can be viewed as one of the reasons that can lead to group imbalance. % because neural networks can learn the spurious correlations to obtain high average accuracy but perform poorly on the minority group.   
Different from %Various approaches have been proposed to 
the approaches that address spurious correlations, such as down-sampling the majority \cite{GLJG17, BMM18}, up-weight the minority group \cite{BL19}, and removing spurious features \cite{GPLT19, ZWSP13},  this paper %differs from these works because it 
does not require the special model of spurious correlations and any group attribute information. %, which is only one of the reasons  for group imbalance. Moreover, this paper considers ERM without any group attribute information. %  and study the general setup of multiple Gaussian distributions from different groups. 

\textbf{Imbalance learning and long-tailed learning} focus on learning from imbalanced data with a long-tailed distribution, which means that a few classes of the data make up the majority of the dataset, while the majority of classes have little data samples \cite{CBHK02, SXK18, MDD19, KJS20, CBLL20, LGLW21, FHLS21, YJSG22, PHHY22}. Some works \cite{SXK18, PHHY22} claimed that naively increasing the number of the minority does not always improve the generalization. Therefore, some recent works develop novel oversampling and data augmentation methods \cite{LGLW21, CBLL20, YJSG22} that can promote the minority fraction by generating diverse and context-rich minority data. However, there are very limited theoretical explanations of how these techniques affect the generalization.

\textbf{Generalization performance with the standard Gaussian input for one-hidden-layer neural networks.} % This paper follows  the teacher-student setup where the network width can be an arbitrary number.  %In the teacher-student setup of one-hidden-layer neural networks, 
%This line of works %follow the teacher-student setup and 
%considers one-hidden-layer neural networks only. 
\cite{DLTPS17,GLM17, LY17,SS17}   consider  infinite   training samples. % without any sample complexity analysis. % for one-hidden-layer neural networks   so that the training and test accuracy coincide and can be analyzed simultaneously.
%When the number of training samples is finite, 
\cite{ZSJB17} characterize the sample complexity   %, i.e., the required number of samples, of learning 
%one-hidden-layer  
of fully connected neural networks with smooth activation functions. % and show that gradient iterations following a tensor initialization can converge to the ground-truth model linearly. %with both sample complexity and computational complexity linear in the input dimension and logarithmic in the precision.
\cite{ZYWG19, ZWXL20, ZLWL23} extend  to the non-smooth ReLU activation for fully-connected and convolutional neural networks, respectively. % of one-hidden-layer ReLU networks with multiple neurons using gradient descent and accelerated gradient descent, respectively.
\cite{FCL20} analyzes the cross entropy loss function for binary classification problems.  %The challenge of analyzing   %Compared with other common loss functions such as the squared loss, 
%the cross entropy loss function  results from the complicated forms and the saturation phenomenon of its Gradient and Hessian. 
\cite{ZWLC20} analyzes the generalizability  of graph neural networks for both regression and binary classification problems. One-hidden-layer case of neural network pruning and self-training are also studied in \cite{ZWLC21_sparse} and \cite{ZWLC21_self}, respectively.

\textbf{Theoretical characterization of learning performance from other input distributions for one-hidden-layer neural networks.} % A few recent works analyze 
%The learning performance with input distribution beyond the standard Gaussian distribution are analyzed in one-hidden-layer neural networks only.
\cite{YO19}
 analyzes the %TPlateau Phenomenon of 
 training loss % that the   decrease of the risk slows down significantly partway and speeds up again in one-hidden-layer neural networks 
 with %inputs drawn from 
 a single Gaussian with an arbitrary co-variance.  \cite{MKUZ20} quantifies the % provides analytical  equations for %the size-one minibatch 
SGD evolution %in a perceptron %(i.e. a single-node network) 
trained on the Gaussian mixture model. When the hidden layer only contains one neuron,  \cite{DLT18} analyzes rotationally invariant distributions.  With an infinite number of neurons and an infinite input dimension, 
\cite{MMN18} analyzes the generalization error based on the mean-field analysis for distributions like Gaussian Mixture with the same mean. %for a large class of distributions, including a mixture of Gaussian distributions with the same mean. 
\cite{GMMM20} considers inputs with low-dimensional structures. % and compares neural networks with kernel methods.
No sample complexity is provided in all these works.  %To the best of knowledge, only 
%\cite{LL18} provides the sample complexity analysis %studies the  generalization error of over-parameterized one-hidden-layer networks  
%when the data come  from the mixtures of well-separated distributions. However, \cite{LL18}  requires strong separation among different groups, and each group needs to be more concentrated when the number of groups increases. In contrast, this paper does not require any data separation and can handle an arbitrary number of Gaussian distributions with arbitrary mean and variance. % but the separation requirement excludes Gaussian distributions and Gaussian mixture models. 

%Our sample complexity is in the order of $d\log^2d$.  Both results are in the same order as the state-of-the-art algorithmic convergence   and sample complexity analyses when the input data follow the standard Gaussian \cite{FCL20}.

%\tcr{a list of major  notations at the end of Introduction}\\
\textbf{Notations}:% Vectors are in bold lowercase, matrices and tensors in
%are bold uppercase. Scalars are in normal fonts. For instance, 
$\bfZ$ is a matrix with $Z_{i,j}$ as the $(i,j)$-th entry. $\bfz$ is a vector with $z_i$ as the $i$-th entry.
$[K]$ %($K>0$) 
denotes the set including integers from $1$ to $K$. $\bfI_d$ % \in\mathbb{R}^{d\times d}$ 
and $\bfe_i$ represent the identity matrix in $\mathbb{R}^{d \times d}$ and the $i$-th standard basis vector, respectively. 
 $\delta_i(\bfZ)$  denotes the $i$-th largest singular value of $\bfZ$. The matrix norm  $\|\bfZ\|=\delta_1(\bfZ)$. %$\bfA\succeq0$ means $\bfA$ is a positive semi-definite (PSD) matrix. $\bfA^\frac{1}{2}$ means that $\bfA=(\bfA^\frac{1}{2})^2$. 
%We follow the convention that 
$f(x)=O(g(x))$ (or $\Omega(g(x))$, $\Theta(g(x)))$ means that $f(x)$ increases at most, at least, or in the order of $g(x)$, respectively. 

\section{Problem Formulation and Algorithm}\label{sec:formulation}

We consider the classification problem with an unbalanced dataset using fully connected neural networks over $n$ independent training examples $\{ (\bfx_i, y_i) \}_{i=1}^N$ from a data distribution. The learning algorithm is to minimize the empirical risk function via gradient descent (GD). In what follows, we will present the data model and neural network model considered in this paper.

\textbf{Data Model}. 
Let $\bfx \in \mathbb{R}^d$ and $y\in \mathbb{R}$ denote the input feature and label, respectively.
We consider an unbalanced dataset that consists of $L$ ($L\geq 2$) 
groups of data,  where the feature $\bfx$ in the group $l$ ($l\in [L]$) is drawn from a multi-variate Gaussian distribution with mean  $\bfmu_l\in \R^{d}$, and covariance $\bfSg_l\in \R^{d\times d}$. Specifically, $\bfx$ follows the Gaussian mixture model  (GMM)  \cite{P94, HK13, MV10, RV17},   denoted as
%\begin{equation}
$\bfx \sim \sum_{l=1}^L\lambda_l\mathcal{N}(\bfmu_l,\bfSg_l)$\footnote{We consider this data model inspired by existing works on group imbalance and practical datasets. Details can be found in Appendix \ref{subsec: gmm}.}. $\lambda_l\in (0,1)$ is the probability of sampling from distribution-$l$ and represents the expected fraction of group-$l$ data.   $\sum_{l=1}^L\lambda_l=1$.
%%Let $\lambda_l\in (0,1)$ denote the fraction of group-$l$ data, and  
 %$\sum_{l=1}^L\lambda_l=1$. 
 Group $l$ is defined as a minority group if $\lambda_l$ is less than  $1/L$. %$\lambda_{l'}$ for other groups $l'$. 
%we assume the input features $\bfx_i$ are generated from one of $L$   multi-variate Gaussian distribution with mean  $\bfmu_l\in \R^{d}$, and covariance $\bfSg_l$ for all $l\in [L]$, and
 % the percentage of data generated from $\mathcal{N}(\bfmu_l,\bfSg_l)$ is $\lambda_l\in (0,1)$ with 
 % Equivalently, 
%\label{notation_GMM} \end{equation}
%and 
We use $\Psi=\{\lambda_l, \bfmu_l, \bfSg_l, \forall l\}$ to denote all parameters of the mixture model\footnote{  In practice,  $\Psi$ can be estimated by the EM algorithm \cite{RW84} and the moment-based method \cite{HK13}. The EM algorithm returns model parameters within Euclidean distance $O((\frac{d}{n})^\frac{1}{2})$ when the number of mixture components $L$ is known. When $L$ is unknown,  one usually %chooses some estimate $\bar{L}$
% as the number of components and $\bar{L}$
% can be much larger than $L$. In this over-specified
  over-specifies an estimate $\bar{L}>L$, then the estimation error by the EM algorithm scales as $O((\frac{d}{n})^\frac{1}{4})$. Please refer to 
\cite{HN16,DHKJ20,DHKw20} for details.}.  %$\zeta_l=(\det(\bfSg_l))^\frac{K}{2}\|\bfSg_l^{-1}\|^K,\ l\in[L]$, $\zeta=\max_{l\in[L]}\{\zeta_l\}$.
We consider binary classification with label $y$ generated by a  ground-truth neural network with unknown weights $\bfW^*=[\bfw^*_1,...,\bfw^*_K]\in \R^{d\times K}$ and sigmoid activation\footnote{The results can be generalized to     any activation  function $\phi$  with bounded  $\phi$, $\phi'$ and $\phi''$, where $\phi'$ is even. %(3) $\phi$ can be linearly mapped to the range $[0,1]$. 
Examples %of such activation 
include $\tanh$ and $\erf$.}. function $\phi(x)=\frac{1}{1+\exp(-x)}$, where\footnote{Our data model is reduced to  logistic regression in the special case that  $K=1$. We mainly study the more challenging case when $K>1$, %which is much more challenging than $K=1$ since 
because the learning problem becomes highly non-convex when  there are multiple neurons in the network.} % $\bfw^*_j \in \R^{d}$ ($j\in [K]$), where  %the teacher committee machine through %output $y$ is mapped to $\{0,1\}$ by
 \begin{equation} 
 \mathbb{P}(y=1|\bfx)=H(\bfW^*,\bfx):=\frac{1}{K}\sum_{j=1}^K\phi({\bfw_j^*}^\top\bfx).\label{cla_model}
 \end{equation}

\textbf{Learning model}. 
 Learning is performed over a   neural network that has the same architecture
as in (\ref{cla_model}), which is a one-hidden-layer fully connected neural network\footnote{All the weights in the second layer are assumed to be
  fixed   to  facilitate the analysis. This is a standard assumption % which has also been made in other papers on 
  in theoretical generalization analysis \cite{ZYWG19, FCL20, ZWLC20}.} with its weights   denoted by $\bfW\in \R^{d\times K}$.
% \tcr{
% \begin{equation} 
% \mathbb{P}(y=1|\bfx\sim\mathcal{N}(\bfmu_1,\bfSg_1))=H(\bfW^{(1)},\bfx):=\frac{1}{K}\sum_{j=1}^K\phi({\bfw^{(1)}_j}^\top\bfx).\label{cla_model}
% \end{equation}
% \begin{equation} 
% \mathbb{P}(y=1|\bfx\sim\mathcal{N}(\bfmu_2,\bfSg_2))=H(\bfW^{(2)},\bfx):=\frac{1}{K}\sum_{j=1}^K\phi({\bfw^{(2)}_j}^\top\bfx).\label{cla_model}
% \end{equation}
% }
% the empirical risk function is 
%\tcr{
%\begin{equation}
%f_n(\bfW)=\frac{1}{N}\sum_{i=1}^N %\ell(\bfW;\bfx_i,y_i)\label{empirical}
%\end{equation}
% \begin{equation}
%     f_n^{(1)}(\bfW)=\frac{1}{N_1}\sum_{i\in\mathcal{S}_1} \ell(\bfW;\bfx_i,y_i)
% \end{equation}
% \begin{equation}
%     f_n^{(2)}(\bfW)=\frac{1}{N_2}\sum_{i\in\mathcal{S}_2} \ell(\bfW;\bfx_i,y_i)
% \end{equation}
%}
%To estimate   $\bfW^*$ from  $N$   training samples $\{\bfx_i,y_i\}_{i=1}^N$,  
Given  $n$   training samples $\{\bfx_i,y_i\}_{i=1}^n$ where $\bfx_i$ follows the GMM model, and $y_i$ is from (\ref{cla_model}), we aim to find the model weights via solving the empirical risk minimization (ERM), where %minimizing the nonconvex empirical risk 
$f_n(\bfW)$ is the empirical risk,  % from  $n$   training samples $\{\bfx_i,y_i\}_{i=1}^n$,  
\be
\min_{\bfW \in \R^{d\times K}} f_n(\bfW):=\frac{1}{n}\sum_{i=1}^n \ell(\bfW;\bfx_i,y_i), \label{eqn:problem}
\ee
where $\ell(\bfW;\bfx_i,y_i)$ is the cross-entropy loss function, i.e., 
\begin{equation}
\begin{aligned}
 \ell(\bfW;\bfx_i,y_i) 
=&-y_i\cdot\log(H(\bfW,\bfx_i))\\
&-(1-y_i)\cdot\log(1-H(\bfW,\bfx_i)).\label{cross-entropy}
\end{aligned}
\end{equation}
%(\ref{eqn:problem}) is nonconvex due to the activation function. 
%In a model recovery setting, we are given $n$ training samples $\{\bfx_i,y_i\}_{i=1}^n$ that are 

Note that for any permutation matrix $\bfP$, $\bfW \bfP$ corresponds %to the weights of a neural network obtained by 
permuting neurons of a network with weights $\bfW$. Therefore,  $H(\bfW,\bfx)=H(\bfW \bfP,\bfx)$, and $f_n(\bfW \bfP)=f_n(\bfW)$.   The estimation is considered successful if one finds any column permutation of $\bfW^*$.

The average generalization performance of a learned model $\bfW$ is evaluated by the average risk
\be \label{eqn:barf}
\bar{f}(\bfW)= \mathbb{E}_{\bfx \sim \sum_{l=1}^L\lambda_l\mathcal{N}(\bfmu_l,\bfSg_l) } \ell(\bfW;\bfx_i,y_i),
\ee 
and the generalization performance on group $l$ is evaluated by the group-$l$ risk 
\be \label{eqn:barfl}
\bar{f}_l(\bfW)= \mathbb{E}_{\bfx \sim   \mathcal{N}(\bfmu_l,\bfSg_l) } \ell(\bfW;\bfx_i,y_i).
\ee

 \textbf{Training Algorithm}. 
  Our algorithm starts from an initialization $\bfW_0 \in \R^{d\times K}$ computed based on the tensor initialization method (Subroutine 1 in  in Appendix) and then updates the iterates $\bfW_t$ using   gradient descent   with the step size\footnote{Algorithm 1 employs a constant step size. One can potentially speed up the convergence, i.e., reduce $v$, by using a variable step size. We leave the corresponding theoretical analysis for future work.} $\eta_0$. The computational complexity of tensor initialization is $O(Knd)$. The per-iteration complexity of the gradient step is $O(Knd)$. We defer the details of Algorithm 1 in Supplementary Material.

\setcounter{algorithm}{0}
\begin{algorithm}

\begin{algorithmic}[1]
\caption{Our ERM learning algorithm}\label{gd}
\STATE{\textbf{Input: }} 
 Training data $\{(\bfx_i,y_i)\}_{i=1}^n$, the step size $\eta_0=O\Big( \big( \sum_{l=1}^L\lambda_l (\|\tilde{\bfmu}_l\|_\infty+\|\bfSg_l^\frac{1}{2}\|)^2 \big)^{-1}\Big)$, the total number of iterations $T$
\STATE{\textbf{Initialization: }}$\bfW_{0}\leftarrow$ Tensor initialization method via Subroutine \ref{TensorInitialization}
\STATE{\textbf{Gradient Descent:}} for $t=0,1,\cdots,T-1$
\vspace{-0.1in}
%\textcolor{red}{$$\bfW_{t+1}=\bfW_t-\eta_0\widetilde{\nabla f_n(\bfW_t)}=\bfW_t-\eta_0\Big(\nabla f_n(\bfW^*)+\frac{1}{n}\sum_{i=1}^n \nu_i\Big)$$}
\begin{equation}\label{eqn:gradient}
\begin{aligned}
\bfW_{t+1}&=\bfW_t-\eta_0 \cdot \frac{1}{n} \sum_{i=1}^n ( \nabla  l(\bfW, \bfx_i, y_i) +\nu_i )\\
&= \bfW_t-\eta_0\Big(\nabla f_n(\bfW)+\frac{1}{n}\sum_{i=1}^n \nu_i\Big)
\end{aligned}
\end{equation}
\vspace{-0.1in}
\STATE{\textbf{Output: }} $\bfW_T$
\end{algorithmic}
\end{algorithm}

\section{Main Theoretical Results}\label{sec: theory}  
% \textbf{Main Theory}.
We will formally present our main theory below, and  the insights  are summarized  in Section \ref{sec:insights}. For the convenience of presentation, some quantities are defined here, and all of them can be viewed as constant. Define $\sigma_{\max}=\max_{l\in[L]}\{\|\bfSg_l\|^\frac{1}{2}\}$, $\sigma_{\min}=\min_{l\in[L]}\{\|\bfSg_l^{-1}\|^{-\frac{1}{2}}\}$. Let $\tau =\sigma_{\max}/\sigma_{\min}$. We assume $\tau=\Theta(1)$%\footnote{This is a mild assumption because many practical datasets can be approximated with a minor loss of information by a low-rank dataset which has $\tau$ being $\Theta(1)$. It is verified in Section A.1 in the Appendix using the CelebA dataset.}
, indicating that $\sigma_{\max}$ and $\sigma_{\min}$ are in the same order\footnote{Note that it is a very mild assumption that $\sigma_{\min}$ is not very close to zero, or equivalently, $\tau=\Theta(1)$. We verify this in Appendix \ref{subsec: sigma_min}.}. 
Let $\delta_i(\bfW^*)$ denote the $i$-th largest singular value of $\bfW^*$. Let   $\kappa=\frac{\delta_1(\bfW^*)}{\delta_K(\bfW^*)}$, and define $\eta=\prod_{i=1}^K(\delta_i(\bfW^*)/\delta_K(\bfW^*))$.

 %\tcr{If possible, add some discussion of the order of rho and D}
\begin{thm}\label{thm1}
%Consider the binary classification problem with one-hidden-layer fully connected neural network as in (\ref{cla_model}). Suppose  Assumption \ref{assumption1} holds, then 
There exist $\epsilon_0\in(0,\frac{1}{4})$ and positive value functions $\mathcal{B}(\Psi)$ (sample complexity parameter), $q(\Psi)$ (convergence rate parameter), and  $\mathcal{E}_w(\Psi)$, $\mathcal{E}(\Psi)$, $\mathcal{E}_l(\Psi)$ (generalization parameters) such that   as long as the sample size $n$ satisfies
\begin{equation}
%\begin{aligned}
n \geq n_{\textrm{sc}}:=poly(\epsilon_0^{-1}, \kappa, \eta, \tau, K, \delta_1(\bfW^*)) \mathcal{B}(\Psi) d\log^2{d}
  , \label{final_sp}
%\end{aligned}
\end{equation}
we have that with probability at least $1-d^{-10}$, the iterates $\{\bfW_t\}_{t=1}^T$ returned by Algorithm 1 with step size $\eta_0=O\Big( \big(\sum_{l=1}^L\lambda_l(\|\bfmu_l\|+\|\bfSg_l\|^\frac{1}{2})^2\big)^{-1}\Big)$  converge  linearly with a statistical error to a critical point $\widehat{\bfW}_n$ with the rate of convergence $v$,  %$v=1-K^{-2}q(\boldsymbol{\lambda},\bfM,\bfS)$, 
i.e.,   
\begin{equation}
\begin{aligned}
 ||\bfW_t-\widehat{\bfW}_n||_F 
\leq  &v(\Psi)^t||\bfW_0-\widehat{\bfW}_n||_F\\
&+\frac{\eta_0\xi}{1-v(\Psi)}\sqrt{dK\log n/n},
\label{linear convergence}
\end{aligned}
\end{equation}
\begin{equation}\label{eqn:v} 
 v(\Psi)=1-K^{-2}q(\Psi),
\end{equation}
where $\xi \geq 0$ is the upper bound of the entry-wise additive noise in the gradient computation.
%where $\mathcal{B}(\boldsymbol{\lambda},\bfM,\boldsymbol{\sigma},\bfW^*)$ and $v(\boldsymbol{\lambda},\bfM,\boldsymbol{\sigma},\bfW^*)$ satisfies Corollary \ref{prop: Bv}.
%\begin{equation}
%\begin{aligned}
%&||\bfW_t-\widehat{\bfW}_n| _F\\
%\leq& \Big(1-\frac{\sum_{l=1}^L\lambda_l\frac{\sigma_l^2}{\sigma_{\max}^2\eta\kappa^2}\rho(\frac{{\bfW^*}^\top\bfmu_l}{\sigma_l\delta_K(\bfW^*)}, \sigma_l\delta_K(\bfW^*))}{C\cdot D_2(\boldsymbol{\lambda},\bfM,\boldsymbol{\sigma})K^{2}}\Big)^t||\bfW_0-\widehat{\bfW}_n||_F\label{linear convergence}
%\end{aligned}
%\end{equation}
%\tcr{need to show the constant under the power t is between 0 and 1 to guarantee convergence.}

Moreover, there exists a permutation matrix $\bfP^*$ such that % the distance between $\bfW^*\bfP^*$ and $\widehat{\bfW}_n$ is bounded by
\begin{equation}
\begin{aligned}
 ||\widehat{\bfW}_n-\bfW^*\bfP^*||_F \leq &\mathcal{E}_w(\Psi)\cdot\text{poly}(\kappa,\eta, \tau, \delta_1(\bfW^*))\\
 &\cdot\Theta\Big( K^{\frac{5}{2}}(1+\xi)\cdot\sqrt{d\log{n}/n}\Big).\label{eqn: w_bound}
 \end{aligned}
  \end{equation}
The average population risk $\bar{f}$ and the group-l risk $\bar{f}_l$ satisfy
\begin{equation}
\begin{aligned}
    \bar{f}\leq &\mathcal{E}(\Psi)\cdot\text{poly}(\kappa,\eta, \tau, \delta_1(\bfW^*))\\
    &\cdot\Theta\Big( K^{\frac{5}{2}}(1+\xi)\cdot\sqrt{d\log{n}/n}\Big)\label{eqn: f_bound}
\end{aligned}
\end{equation}
\begin{equation}
\begin{aligned}
    \bar{f}_l\leq &\mathcal{E}_l(\Psi)\cdot\text{poly}(\kappa,\eta, \tau, \delta_1(\bfW^*))\\
    &\cdot\Theta\Big( K^{\frac{5}{2}}(1+\xi)\cdot\sqrt{d\log{n}/n}\Big)\label{eqn: fl_bound}
\end{aligned}
\end{equation}
  \end{thm}
  
 The closed-form expressions of $\mathcal{B}$, $q$, $\mathcal{E}_w$, $\mathcal{E}$, and $\mathcal{E}_l$ %for all   $(\boldsymbol{\lambda},\bfM,\bfS,\bfW^*)$ 
 are  %in (\ref{B_closedform}) and (\ref{q_closedform}) 
 in Section D of the supplementary material and skipped here. % due to the complicated form. % but do not include them here for a compact representation.} 
  %To characterize how the parameters of the Gaussian mixture model affects the learning performance, we
The quantitative impact of the GMM model parameters $\Psi$ on the learning performance varies in different regimes and can be derived from Theorem \ref{thm1}. The following corollary summarizes the impact of   $\Psi$ on the learning performance in some sample regimes. 
  %\AtBeginEnvironment{tabular}{\footnotesize}

\begin{table*}[t]
    \centering
    %\vspace{2mm}
    \caption{Impact of GMM parameters on the learning performance in sample regimes}
    \renewcommand\arraystretch{2}
    \begin{tabular}{|c|p{2.2cm}|p{2cm}|p{2cm}|p{2cm}|p{2cm}|}
    
    \hline
        & \multicolumn{2}{|c|}{ $\bfSg_l$ changes}   & \multirow{2}{*}{$\bfmu_l$ changes} &\multicolumn{2}{|c|}{ $\lambda_l$ changes,  constant $\|\bfSg_j\|$'s, equal   $\|\bfmu_j\|$'s}
    \\
    \cline{2-3}
    \cline{5-6}
    
    & { $\|\bfSg_l\|=o(1)$} & {$\|\bfSg_l\|=\Omega(1)$} &  &   {if $\|\bfSg_l\|=\sigma^2_{\min}$}  &  { if $\|\bfSg_l\|=\sigma^2_{\max}$}  
    \\
    \hline
     
    { $\mathcal{B}(\Psi)$, sample complexity $n_{sc}$} & {$O(|\bfSg_l\|^{-3})$}& {  $O\|\bfSg_l\|^{3})$} & {$O(\textrm{poly}(\|\bfmu_l\|))$\footnote{\afterpage{\footnotetext[7]{poly($\|\bfmu_l\|$) is $\|\bfmu_l\|^4$ for $\|\bfmu_l\|\leq 1$; $\|\bfmu_l\|^{12}$ for $\|\bfmu_l\|> 1$.}}}} 
    & { $O(\frac{1}{(1+\lambda_l)^2})$}&  { $O(1)-\frac{\Theta(1)}{(1+\lambda_l)^2}$}
    \\
    \hline
 { convergence rate $v(\Psi)\propto -q(\Psi)$} &{ $1- \Theta (\|\bfSg_l\|^3)$} & {  $1- \Theta (\frac{1}{1+\|\bfSg_l\|})$} & {  $1- \Theta (\frac{1}{\|\bfmu_l\|^2+1})$} & { $\Theta(\frac{1}{1+\lambda_l})$}&  {$1-\Theta(\frac{1}{1+\lambda_l})$}
    \\
    \hline
     { $\mathcal{E}_w(\Psi)$, $\|\widehat{\bfW}_n-\bfW^*\bfP\|_F$}   & { $O(1)-\Theta(\|\bfSg_l\|^3)$}& { $O(\sqrt{\|\bfSg_l\|})$} &{$O(1+\|\bfmu_l\|)$} & {$O(\frac{1}{1+\sqrt{\lambda_l}})$}&  {$O(1+\sqrt{\lambda_l})$}
     \\
    \hline
     {  $\mathcal{E}(\Psi)$, average risk $\bar{f}$}   & {  $O(1)-\Theta(\|\bfSg_l\|^3)$} & {  $O(\|\bfSg_l\|)$} & {  $O(1+\|\bfmu_l\|^2)$}& {  $O(\frac{1}{1+\lambda_l})$}&  {  $O(1)-\frac{\Theta(1)}{1+\lambda_l}$}
    \\
    \hline
    {  $\mathcal{E}_l(\Psi)$, group-$l$ risk $\bar{f}_l$}  & {  $O(1)-\Theta(\|\bfSg_l\|^3)$} & {  $O(\|\bfSg_l\|)$} &  {  $O(1+\|\bfmu_l\|^2)$}&{ 
    $O(\frac{1}{1+\sqrt{\lambda_l}})$} &  {  $O(1+\sqrt{\lambda_l})$}
    \\
    \hline
 %   \multicolumn{3}{p{13.5cm}}{Probability of containing class-relevant nodes in the sampled neighbors of one node.} %Normalized sampling probability of class-relevant nodes; $\alpha \geq 1$.
    %A larger $\alpha$ indicates a higher probability}  
    % \\
  %  \hline
%    $\beta$ & \multicolumn{3}{p{12cm}}{Pruning rate of model weights; $\beta \in [0,1-1/L)$; $\beta=0$ means no pruning;}
  %  \\
%    \hline
 %   \hline
    \end{tabular}
    \label{tbl:results}
    \vspace{-1mm}
\end{table*}

 \begin{corollary}\label{cor}
  When we vary one parameter of group $l$ for any $l\in[L]$ of  the GMM model $\Psi$ and fix all the others, the learning performance degrades in the sense that the sample complexity $n_{sc}$, the convergence rate $v$,  $\|\widehat{\bfW}_n-\bfW^*\bfP\|_F$,  average risk $\bar{f}$ and group-$l$ risk $\bar{f}_l$ all increase (details summarized in Table \ref{tbl:results}), as long as any of the following conditions happens,
 % \begin{itemize}
 %\item 
 
 (i) $\|\bfSg_l\|$ approaches $0$; \quad  
   % \item 
 (ii) $\|\bfSg_l\|$ increases from some constant; \quad
   %\item 
 (iii) $\|\bfmu_l\|$ increases from $0$,
  
 %\item  
 (iv) $\lambda_l$ decreases, provided that   $\|\bfSg_l\|=\sigma^2_{\min}$, i.e., group $l$ has the smallest group-level co-variance, where $\|\bfSg_j\|$ are all constants, and $\|\bfmu_i\|=\|\bfmu_j\|$ for all $i, j\in[L]$.
  
 % \item
 (v) $\lambda_l$ increases, provided that   $\|\bfSg_l\|=\sigma^2_{\max}$, i.e., group $l$ has the largest group-level co-variance, where $\|\bfSg_j\|$ are all constants, and $\|\bfmu_i\|=\|\bfmu_j\|$ for all $i,j\in[L]$.
 
%  \end{itemize}
  \end{corollary}

%\begin{equation}
%\begin{aligned}
%||\widehat{\bfW}_n-\bfW^*||_F&\leq\Omega\Big(\frac{K^{\frac{5}{2}}\sqrt{ D_2(\boldsymbol{\lambda},\bfM,\boldsymbol{\sigma})}\sigma_{\max}}{\sum_{l=1}^L\lambda_l\frac{\sigma_l^2}{\eta\kappa^2}\rho(\frac{{\bfW^*}^\top\bfmu_l}{\sigma_l\delta_K(\bfW^*)}, \sigma_l\delta_K(\bfW^*))}\\
%&\cdot\sqrt{\frac{d\log{n}}{n}}\Big)
%\end{aligned}
%\end{equation}

%Table \ref{tbl:results} summarizes the results in some sample regimes by %the quantitative impact of the GMM model parameter on the learning performance, 
%quantify  the impact the distribution parameters of one group in the GMM on the learning performance in  by 
%varying one parameter while keeping others fixed.

To the best of our knowledge, 
Theorem \ref{thm1} provides the first   characterization of the sample complexity, learning rate,  and generalization performance under the Gaussian mixture model. It   also firstly  characterizes the per-group generalization performance % of individual groups 
in addition to the average generalization. % performance.  % theoretical guarantee of learning one-hidden-layer neural networks with the input following the Gaussian mixture model. 
%Although we consider the sigmoid activation in this paper, our results   apply to any   activation  function $\phi$ provided that $\phi'$ is an even function, and   $\phi$, $\phi'$ and $\phi''$ are bounded. %(3) $\phi$ can be linearly mapped to the range $[0,1]$. 
%Examples %of such activation 
%include $\tanh$ and $\erf$.  % The implications of Theorem \ref{thm1} include the following aspects.

% to simplify the form of $\mathcal{B}(\Psi)$, $q(\Psi)$, $\mathcal{E}_w(\Psi)$, $\mathcal{E}(\Psi)$ and $\mathcal{E}_l(\Psi)$ in Theorem \ref{thm1}.  % Gaussian mixture model on the sample complexity $n_{\textrm{sc}}$ and the convergence rate $v$.   We start with  diagonal covariance matrices in Corollary \ref{prop: diag_Cor}, as it is more straightforward to see the impact of the mean and variance of individual features in this case,  and generalize to arbitrary covariance matrices in Corollary \ref{prop: Bv}.  %$1-K^{-2}v(\boldsymbol{\lambda},\bfM,\boldsymbol{\sigma},\bfW^*)$ 
%discussed in Theorem \ref{thm1} as follows.

 \subsection{Theoretical  Insights}\label{sec:insights}
%Before formally stating our result in Theorem \ref{thm1}, 
We summarize the crucial implications of Theorem \ref{thm1} and Corollary \ref{cor} as follows. %Table \ref{tbl:results} highlights the impact of individual parameters parameter of one group on the learning performance quantitatively.  

\textbf{(P1). Training convergence and generalization guarantee}. % Convergence rate and estimation accuracy}: 
%When gradients are accurate, % (i.e., $\xi=0$),
The iterates $\bfW_t$ converge  to a critical point $\widehat{\bfW}_n$ linearly, and the distance between $\widehat{\bfW}_n$  and $\bfW^*\bfP^*$ is $O(\sqrt{d\log n/n})$ for a certain permutation matrix $\bfP^*$. % With the noise in the gradient, 
When the computed gradients contain noise, there is an additional error term of $O(\xi\sqrt{d\log n/n})$, where $\xi$ is the noise level ($\xi=0$ for noiseless case). Moreover, the average risk of all groups and the risk of each individual group are both $O((1+\xi)\sqrt{d\log n/n})$.

%For example, when   $n$ is $\Theta(d\log^2d)$, %\footnote{$f(x)=O(g(x))$ means there exist $c'>0$ such that $f(x)\leq c' g(x)$ holds when $x$ is sufficiently large. $f(x)=\Omega(g(x))$ means there exist $c''>0$ such that $f(x)\geq c'' g(x)$ holds when $x$ is sufficiently large. $f(x)=\Theta(g(x))$ means there exists $c>0$ and $C>0$ such that $cg(x)\leq f(x)\leq Cg(x)$ holds when $x$ is sufficiently large.}, 
%the  error of estimating $\bfW^*$ decays as $O(\frac{1+\xi}{\sqrt{\log d}})$. Moreover, the average risk of all groups and the risk of each invidial group are all

\textbf{(P2). Sample complexity.} For a given GMM, the sample complexity %for accurate estimation 
is  $\Theta(d\log^2{d})$, where $d$ is the feature dimension. This result is in the same order as the sample complexity for  the standard Gaussian input in \cite{FCL20} and \cite{ZSJB17}. %, indicating that our method can handle input from the Gaussian mixture model without increasing the order of the sample complexity.   % Because $\bfW^*$ has $dK$ parameters, where   $K$ is the number of the nodes in the hidden layer, our sample complexity
Our bound is  almost order-wise optimal with respect to $d$ because the degree of freedom is $dK$. The   additional multiplier of $\log^2{d}$ results from the concentration bound in the proof technique. We focus on the dependence   on the feature dimension $d$ and treat the network width $K$ as constant. The sample complexity in \cite{FCL20} and \cite{ZSJB17} is also $d\cdot \text{poly}(K, \log d)$.

\textbf{(P3). Learning performance is improved at a medium regime  of group-level co-variance}. 
On the one hand, when %everything else is fixed and 
$\|\bfSg_l\|$ is  $\Omega(1)$,  the learning performance degrades as $\|\bfSg_l\|$ increases in the sense that    the   sample complexity $n_{sc}$,  the convergence rate $v$, the estimation error of $\bfW^*$, the  average risk  $\bar{f}$, and  the group-$l$ risk $\bar{f}_l$  all increase.  
  This is due to the saturation of the loss and gradient when the samples have a large magnitude.   
On the other hand, when  %everything else is fixed and 
$\|\bfSg_l\|$ is   $o(1)$,    the learning performance also degrades when $\|\bfSg_l\|$ approaches zero. The intuition %for the poor learning performance $\|\Sigma_l\|$ approaching zero 
is that in this regime,  the input  data are concentrated on a few vectors,  and the optimization problem does not have a benign landscape.

\textbf{(P4). Increasing the fraction of the minority group  data does not always improve the generalization}, while the performance also depends on the mean and co-variance of individual groups. %We consider the case that
Take $\|\bfSg_j\|=\Theta(1)$ for all group $j$, and $\|\bfmu_j\|$ is the same for all $j$ as an example (columns 5 and 6 of Table \ref{tbl:results}).  When   $\|\bfSg_l\|$ is the smallest among all groups, increasing $\lambda_l$ improves the learning performance. When  $\|\bfSg_l\|$ is the largest among all groups, increasing $\lambda_l$ actually degrades the performance. The intuition is that from (P3), the  learning performance is enhanced at a medium regime of group-level co-variance. Thus, increasing the fraction of a group with a medium level of co-variance improves the performance, while  increasing the fraction of a group with large co-variance  degrades the learning performance. 
%\tcr{We discuss the impact of the group percentage under the condition that $\|\bfSg_j\|=\Omega(1), j\in[L]$ and $\|\bfmu_i\|=\|\bfmu_j\|, i,j\in[L]$.} When group $l$ has the  smallest $\|\bfSg_l\|$ among all groups, increasing $\lambda_l$ improves the learning performance. 
 %When group $l$ has the largest $\|\bfSg_l\|$ among all groups, increasing $\lambda_l$ improves the learning performance.  The intuition is that a medium level of variance can benefit the generalization.
 Similarly, when augmenting the training data, an argumentation method that introduces medium variance could improve the learning performance, while an argumentation method that introduces a significant level of variance could hurt the learning performance. 

\textbf{(P5). Group-level mean shifts from zero degrade the learning performance}. %When everything else is fixed, 
The learning performance degrades as $\|\bfmu_l\|$ increases.   An intuitive explanation of the degradation is that %$$  when some mean or variance is large, 
   some training samples %$\bfx$
have a significant   large magnitude such that %$H(\bfW^*,\bfx)$ is very close to 1   because 
the sigmoid function saturates.

 \subsection{Proof Idea and Technical Novelty}

 \vspace{5mm}
 \subsubsection{Proof Idea}
  Different from the analysis of logistic regression for generalized linear models, our paper deals with more technical challenges of nonconvex optimization due to the multi-neuron architecture, the GMM model, and a more complicated activation and loss. The establishment of Theorem \ref{thm1} consists of three key lemmas. 
  
  \begin{lemma}\label{lemma: convexity_informal}(informal version)
  As long as the number of training samples is larger than $\Omega(dK^5 \log^2 d)$, the empirical risk function is strongly convex in
the neighborhood of $\bfW^*$ (or a permutation of $\bfW^*$). The size of the convex region is characterized by the Gaussian mixture distribution.
  \end{lemma}
  
  The main proof idea of Lemma \ref{lemma: convexity_informal} is to show that %although the ERM in (\ref{eqn:problem})  is  nonconvex, %for one-hidden-layer neural networks, 
  the nonconvex empirical risk $f_n(\bfW)$ in a small neighborhood around $\bfW^*$ (or any permutation $\bfW^*\bfP$) is almost convex with a sufficiently large $n$. The difficulty is to find a positive lower bound of the smallest singular value of $\nabla^2\bar{f}(\bfW)$, which should also be a function of the GMM. Then, we can obtain $\nabla^2 f_n(\bfW)$ from $\nabla^2 \bar{f}(\bfW)$ by concentration inequalities. 
  
\begin{lemma}\label{lemma: convergence_informal}(informal version)
  If initialized in the convex region, the gradient descent algorithm converges
linearly to a critical point $\widehat{\bfW}_n$, which is close to $\bfW^*$ (or any permutation of $\bfW^*$), and the distance  is diminishing as the number of training samples increases.
  \end{lemma}
  
  Given the locally strong convexity, Lemma \ref{lemma: convergence_informal} provides the linear convergence to a critical point. The convergence rate is determined by the GMM.  %The distance between the critical point and the ground truth $\bfW^*$ (or $\bfW^*\bfP$) is diminishing as the number of training samples increases.}%Then if $\bfW_0$ can be initialized in any of these local regions, gradient-based iterates can be proved to converge to  $\bfW^*$ (or   $\bfW^*\bfP$). 
  \begin{lemma}\label{lemma: tensor bound_informal}(informal version)
   Tensor Initialization Method initializes $\bfW_0\in\mathbb{R}^{d\times K}$ around $\bfW^*$ (or a permutation of $\bfW^*$).
  \end{lemma}
  The idea of tensor initialization is to first find quantities (see $\bfQ_j$ in Definition 1) in the  supplementary material)   which are proven to be functions of tensors of  $\bfw^*_i$. Then the method approximates these quantities numerically using training samples and then applies the tensor decomposition method  on the estimated quantities to  obtain   $\bfW_0$, which is an estimation of $\bfW^*$.  %Because $\bfQ_j$   can only be estimated from training samples, tensor decomposition does not return $\bfw^*_i$ exactly but provides a close approximation,
  %With a large number of training samples $n$, the estimation $\bfW_0$ %gets closer to $\bfW^*$ and 
  %can be proved to be in the local convex region. 

  Combining the above three lemmas together, one can derive the required sample complexity and the upper bound of $\bar{f}$ and $\bar{f}_l$ in (\ref{final_sp}), (\ref{eqn: f_bound}), and (\ref{eqn: fl_bound}), respectively. The idea is first to compute the sample complexity bound such that the tensor initialization method initializes $\bfW_O$ in the local convex region by Lemma \ref{lemma: tensor bound_informal}. Then the final sample complexity is obtained by comparing two sample complexities from Lemma \ref{lemma: convexity_informal} and \ref{lemma: tensor bound_informal}.

  By further looking into the order of the terms $\mathcal{B}(\Psi)$, $v(\Psi)$, $\mathcal{E}(\Psi)$, $\mathcal{E}_w(\Psi)$, and $\mathcal{E}_l(\Psi)$ in several cases of $\Psi$, % using Taylor expansion,
  Theorem \ref{thm1} leads to Corollary \ref{cor}. To be more specific, we only vary parameters $\bfSg_l$, or $\bfmu_l$, or $\lambda_l$ following the cases in Table \ref{tbl:results}, while fixing all other parameters of $\Psi$. We apply the Taylor expansion to approximate the terms and derive error bounds with the Lipschitz smoothness of the loss function.

\vspace{5mm}
\subsubsection{Technical Novelty}
 Our algorithmic and analytical framework is built upon  some recent works on the generalization analysis of one-hidden-layer neural networks,  %have analyzed a  training algorithm based on  tensor initialization and gradient descent, 
  see, e.g., \cite{ZSJB17, ZYWG19, FCL20, ZWLC20, ZWLC21_sparse}, which  %The existing works in this line of research all 
assume that $\bfx_i$ follows the standard Gaussian distribution % When the data follow GMM,  the existing analyses for both the local convexity and tensor initialization do not hold and 
and cannot be directly  extended to GMM. % in (\ref{notation_GMM}).
This paper makes new technical contributions from the following  aspects. 
%First, we develop new concentration bounds for the sample complexity analysis under the GMM model, while  the matrix concentration inequalities in the existing works only hold for the standard Gaussian distribution. 
%Second, %the existing analysis to bound the Hessian of the population risk function does not extend to the Gaussian mixture model. 
%new tools are developed to explicitly quantify the impact of   $\Psi$ on the landscapes of   risk functions.  

\textbf{First, we characterize the local convex region near $\bfW^*$ for the GMM model.} To be more specific, we explicitly characterize the positive lower bound of the smallest singular value of $\nabla^2 \bar{f}(\bfW)$ with respect to $\Psi$, while existing results either only hold for standard Gaussian data \cite{ZSJB17, FCL20, ZWLC21_sparse, ZWLC21_self}, or can only show $\nabla^2 \bar{f}(\bfW)$ is positive definite regardless the impact of $\Psi$ \cite{ADHL19}.

\textbf{Second, new tools, including matrix concentration bounds are developed to explicitly quantify the impact of $\Psi$ on the sample complexity.}

\textbf{Third, we investigate and provide the order of the bound for sample complexity, convergence rate, generalization error, average risk, and group-$l$ risk in terms of $\Psi$ for the first time} in the line of research of model estimation \cite{ZSJB17, FCL20, ZWLC20, ZWLC21_sparse, ZWLC21_self}, which is also a novel result for the case of Gaussian inputs.

\textbf{Fourth, we design and analyze new tensors %($\bfQ_j$ in (\ref{eqn:mj}))
for the mixture model to initialize properly}, while the previous tensor methods in \cite{ZSJB17, ZYWG19, FCL20, ZWLC20} utilize the rotation invariant property that only holds for zero mean Gaussian.  %  Due to the page limit, we skip the details of the algorithm and the proof  to focus on the presentation of our results. 

\section{Numerical Experiments}\label{sec: experiment}
\subsection{Experiments on Synthetic datasets}
\begin{figure}[ht]
    \centering
        \begin{minipage}{0.4\textwidth}
        \centering
        \includegraphics[width=0.8\textwidth, height=1.5in]{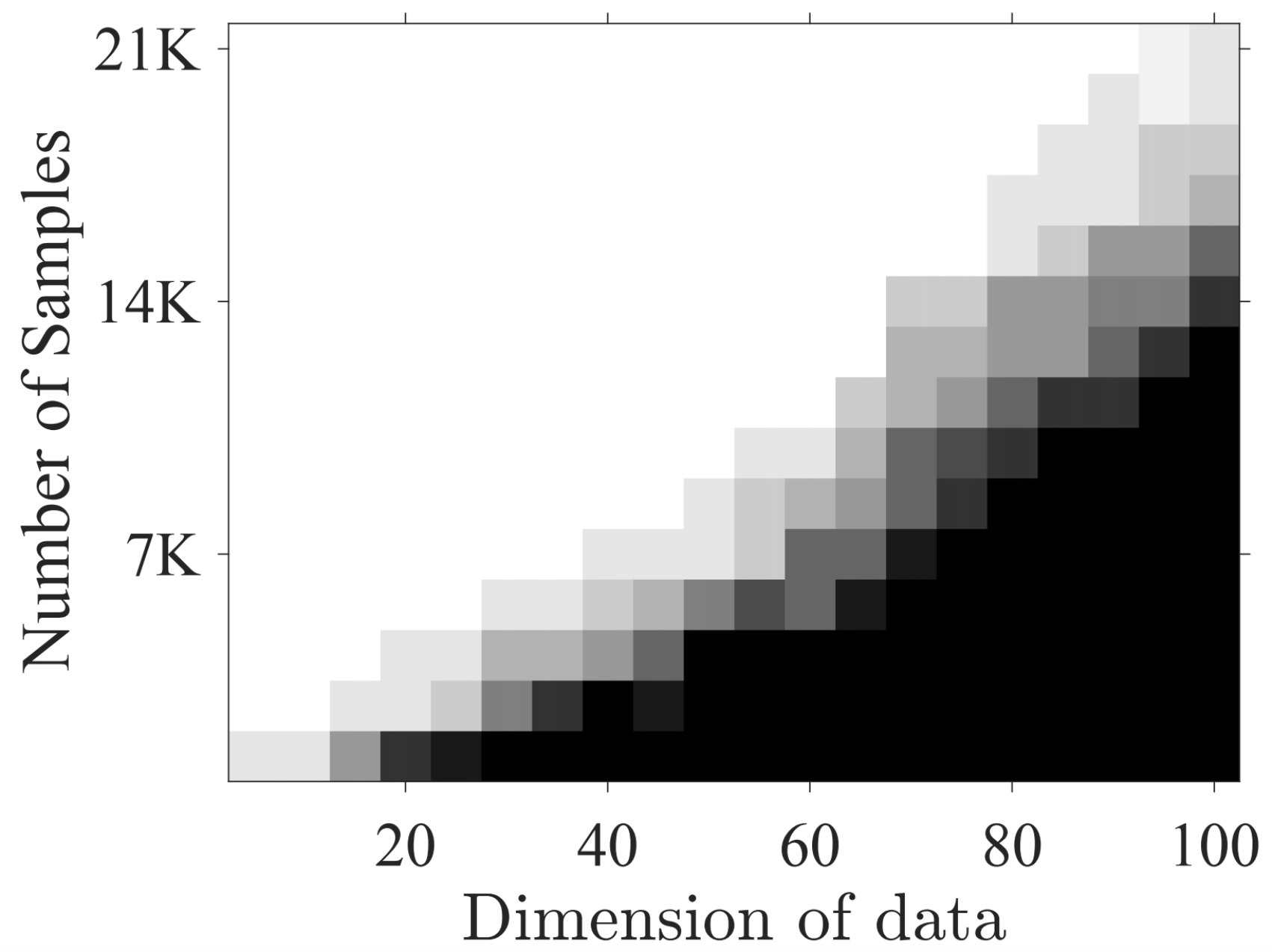}
        \end{minipage}
        \caption{The sample complexity when the feature dimension changes}\label{figure: sample_complexity}
\end{figure}

\begin{figure}[ht]
    \centering

    ~
    \subfigure[]{
        \begin{minipage}{0.21\textwidth}
        \centering
        \includegraphics[width=1\textwidth,height=1.18in]{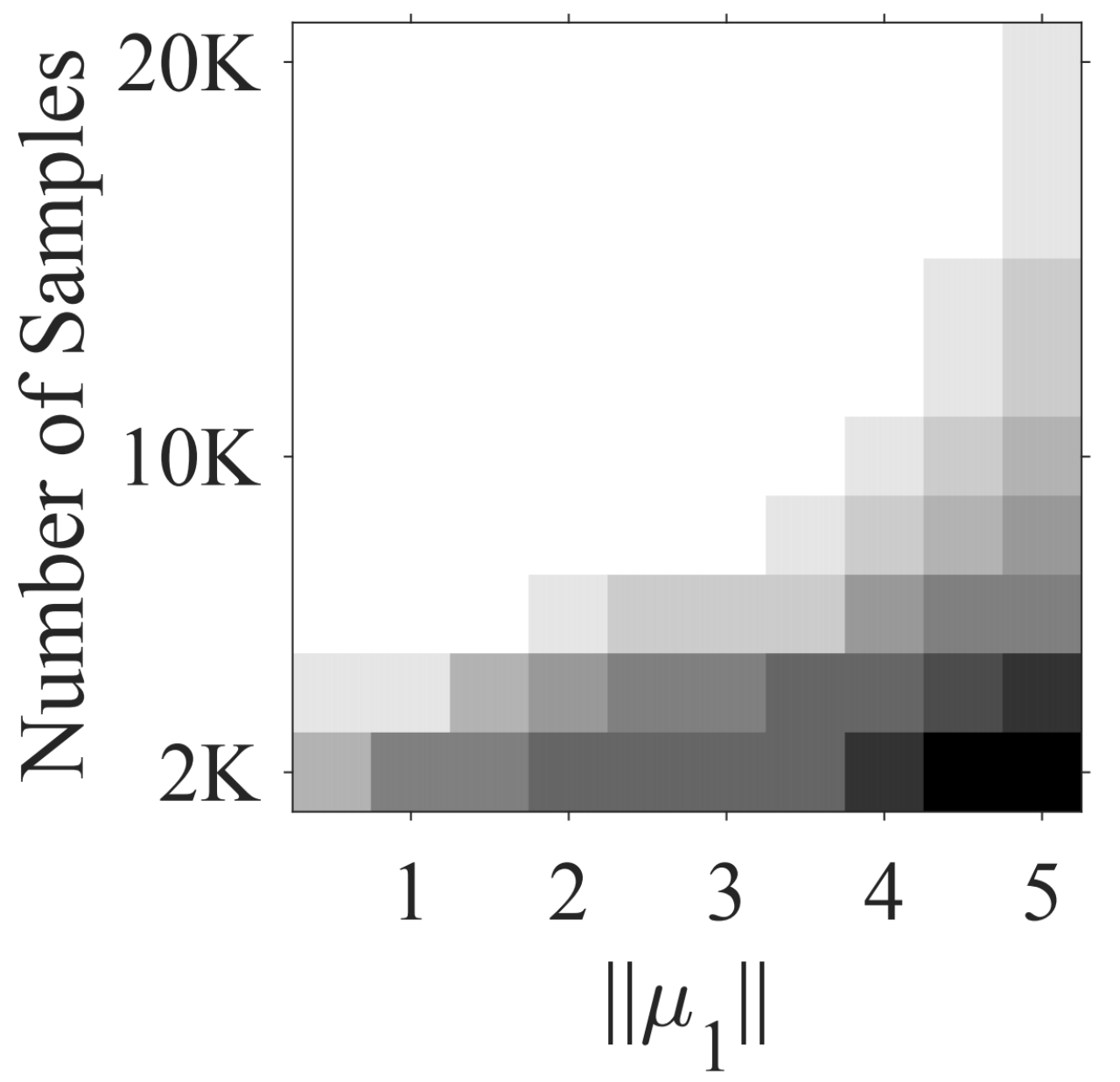}
        \end{minipage}
    }
    ~
    \subfigure[]{
        \begin{minipage}{0.22\textwidth}
        \centering
        \includegraphics[width=1\textwidth,height=1.2in]{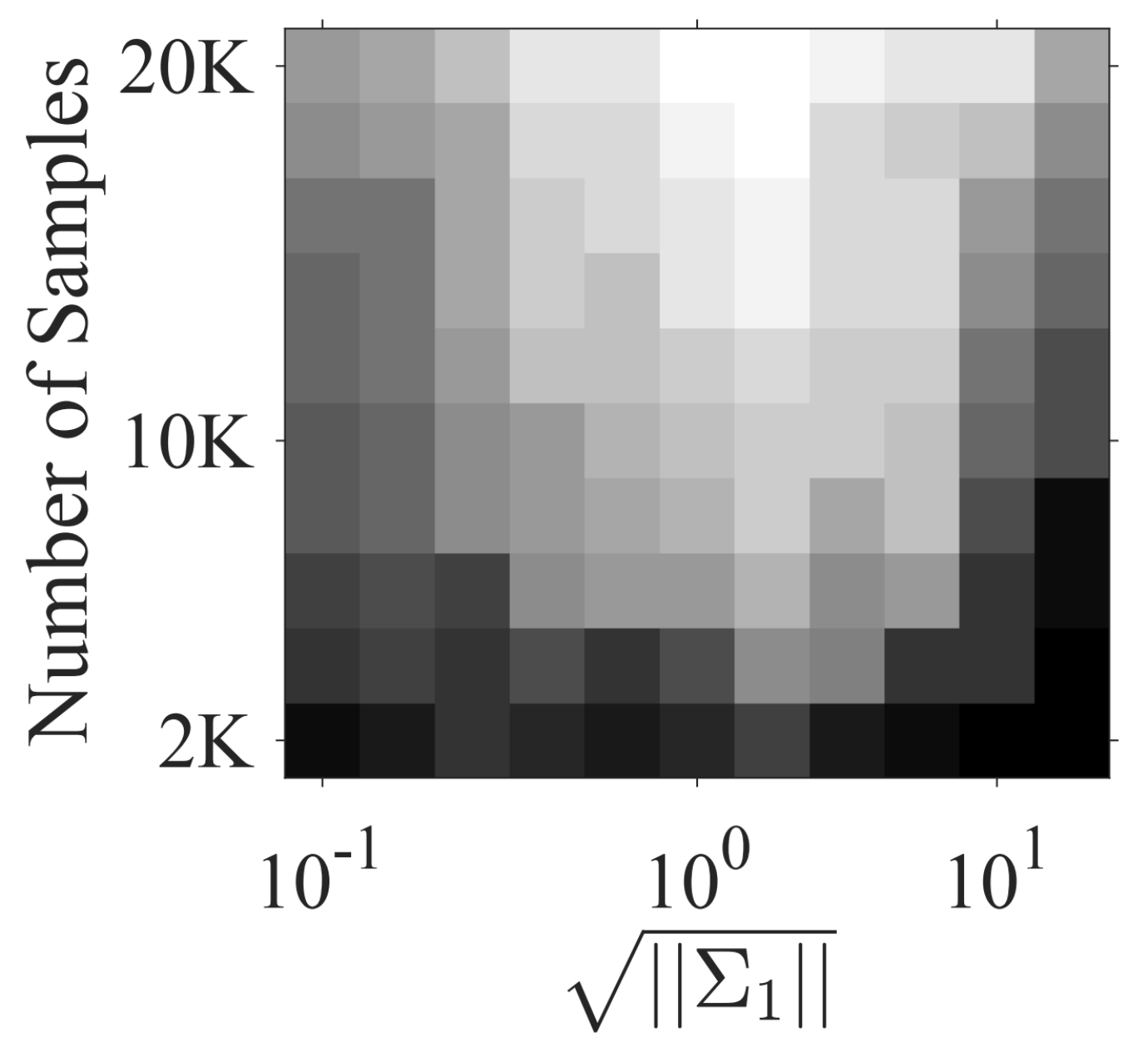}
        \end{minipage}
    }
    \caption{ The sample complexity (a)  when one mean changes, (b)   when one co-variance changes.}
    \label{figure: spmusigma}
\end{figure}

We first verify the theoretical bounds in  Theorem \ref{thm1} on synthetic data.   Each entry of $\bfW^*\in \R^{d\times K}$  is generated from $\mathcal{N}(0,1)$. The training data $\{\bfx_i, y_i\}_{i=1}^n$ is generated using %(\ref{notation_GMM})
the GMM model and  (\ref{cla_model}). If not otherwise specified, $L=2$, $d=5$, and $K=3$\footnote{Like \cite{ZSJB17, ZYWG19, FCL20}, we consider a small-sized network in synthetic experiments to reduce the computational time, especially for computing the sample complexity in Figure \ref{figure: spmusigma}. Our results hold for large networks too. }. To reduce the computational time, we randomly initialize near $\bfW^*$ instead of computing the tensor initialization\footnote{The existing methods based on tensor initialization all use random initialization in synthetic experiments to reduce the computational time. See \cite{FCL20, ZYWG19, ZWLC20, ZWLC21_self} as examples. We compare tensor initialization and local random initialization numerically in Section B of the supplementary material and show that they have the same performance.}.   %\footnote{The codes are available at \url{https://github.com/lohek330/Gaussian-mixture-distribution}}. 
\textbf{Sample complexity}. We first study the impact of $d$ on the sample complexity. 
  Let $\bfmu_1 =\bf1$  in $\R^d$ and let $\bfmu_2=\bf0$.  Let $\bfSg_1=\bfSg_2=\bfI$. $\lambda_1=\lambda_2=0.5$. % We  vary $d$ and evaluate the sample complexity bound in (\ref{final_sp}) with respect to $d$. 
  We randomly initialize $M$ times and let   $\widehat{\bfW}_n^{(m)}$ denote the output of Algorithm 1 in the $m$th trail.  Let $\bar{\bfW}_n$ denote the mean values of all $\widehat{\bfW}_n^{(m)}$, and let $V_W=\sqrt{\sum_{m=1}^M||\widehat{\bfw}_n^m-\bar{\bfW}_n||^2/M}$ denote the variance.  An experiment is successful if $V_W\leq10^{-3}$ and fails otherwise. $M$  
 is set to $20$.
   %We vary $d$ and the number of samples $n$.  
 For each pair of $d$ and $n$,    $20$ independent sets of  $\bfW^*$ and the corresponding training samples are generated. Figure \ref{figure: sample_complexity} shows the success rate  of these independent experiments. A black block means  that all the experiments fail. A white block means  that they   all succeed. The sample complexity is indeed almost linear in $d$, as predicted by (\ref{final_sp}). % Moreover, the coefficient $n/d$   can be large depending on the problem setup.
  
\iffalse

\begin{wrapfigure}{r}{0.5\linewidth}
\centering
%\begin{minipage}{0.6\linewidth}
\centering
\includegraphics[width=0.8\linewidth]{cla4.jpg}
%\end{minipage}\hfill
\caption{\tcr{The sample complexity against the feature dimension $d$}}\label{fig:ntod}
\end{wrapfigure}
\fi

We   next   study the impact on the sample complexity of the GMM model. %Set $d=5$.  % when the mean and variance in the Gaussian mixture model change. %\tcr{$\|\bfW_0-\bfW^*\|_F/\|\bfW^*\|_F\leq 0.1$. change?}
In Figure \ref{figure: spmusigma} (a),    $\bfSg_1=\bfSg_2=\bfI$, and let $\bfmu_1=\mu \cdot\boldsymbol{1}$, $\bfmu_2=-\boldsymbol{1}$.  $\|\bfmu_1\|$ varies from $0$ to $5$. Figure \ref{figure: spmusigma}(a) shows that when the mean increases, the sample complexity increases. % This coincides with our theoretical analyses in Section \ref{sec: theory}. 
In Figure \ref{figure: spmusigma} (b), we fix   $\bfmu_1=\boldsymbol{1}$, $\bfmu_2=-\boldsymbol{1}$, and let $\bfSg_1=\sigma^2\bfI$ and $\bfSg_2=\bfI$.   $\sigma$ varies  from $10^{-1}$ to $10^1$. The  sample complexity increases both when $\|\bfSg_1\|$ increases and when $\|\bfSg_1\|$ approaches zero. All results match   predictions in Corollary \ref{cor}. % Section \ref{sec: theory}.

\begin{figure}[ht]
    \centering
    \subfigure[]{
        \begin{minipage}{0.35\textwidth}
        \centering
        \includegraphics[width=1\textwidth,height=1.1in]{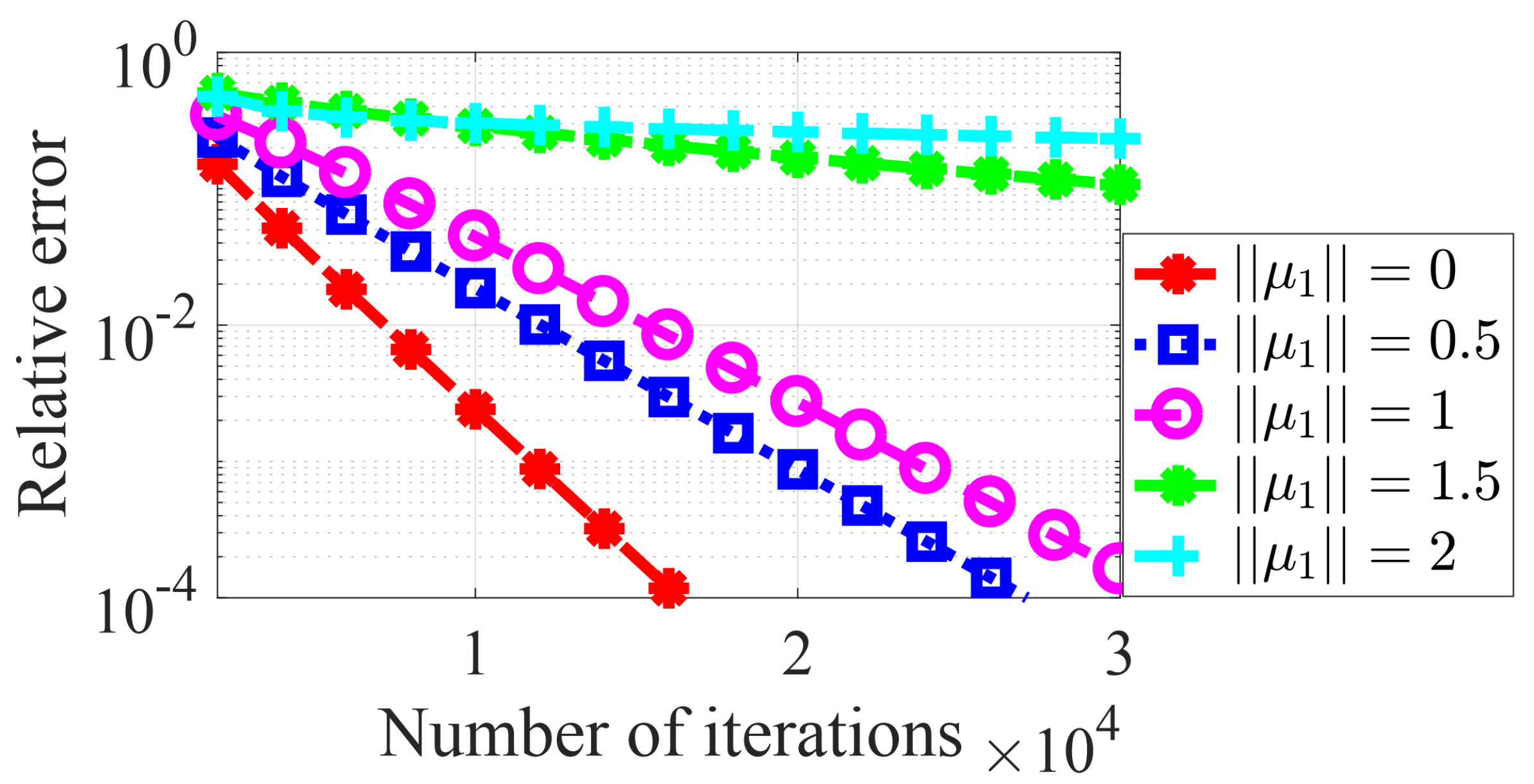}
        \end{minipage}
    }
    ~
    \subfigure[]{
        \begin{minipage}{0.35\textwidth}
        \centering
        \includegraphics[width=1\textwidth,height=1.1in]{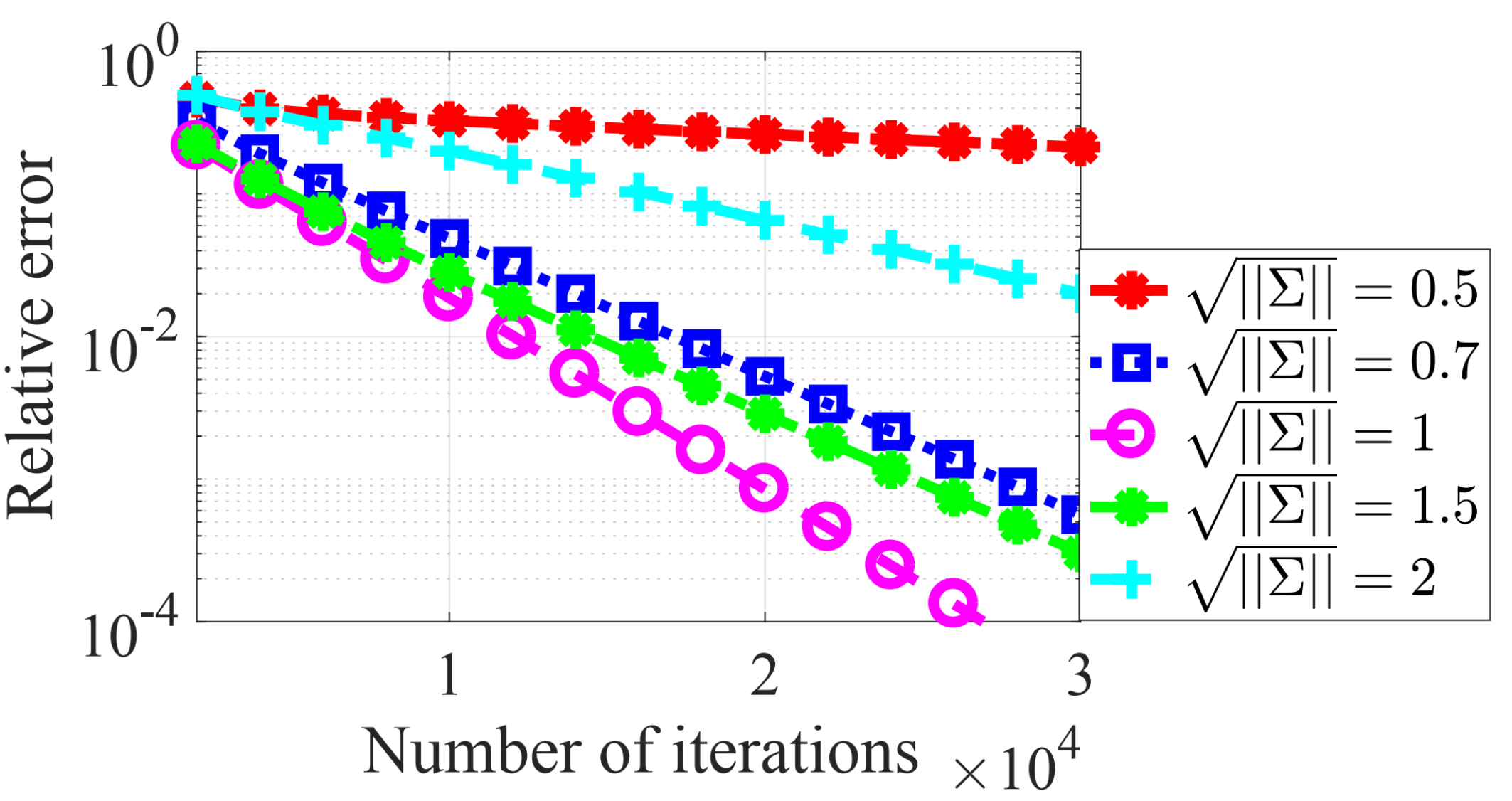}
        \end{minipage}
    }
    
    \caption{(a) The convergence rate with different $\bfmu_1$. (b) The convergence rate with different $\bfSg$. (c) Convergence rate when the number of neurons $K$ changes.}
    \label{figure: convergence_musigma}
\end{figure}

\begin{figure}[ht]
    \centering
    \subfigure[]{
        \begin{minipage}{0.22\textwidth}
        \centering
        \includegraphics[width=1\textwidth,height=1.05in]{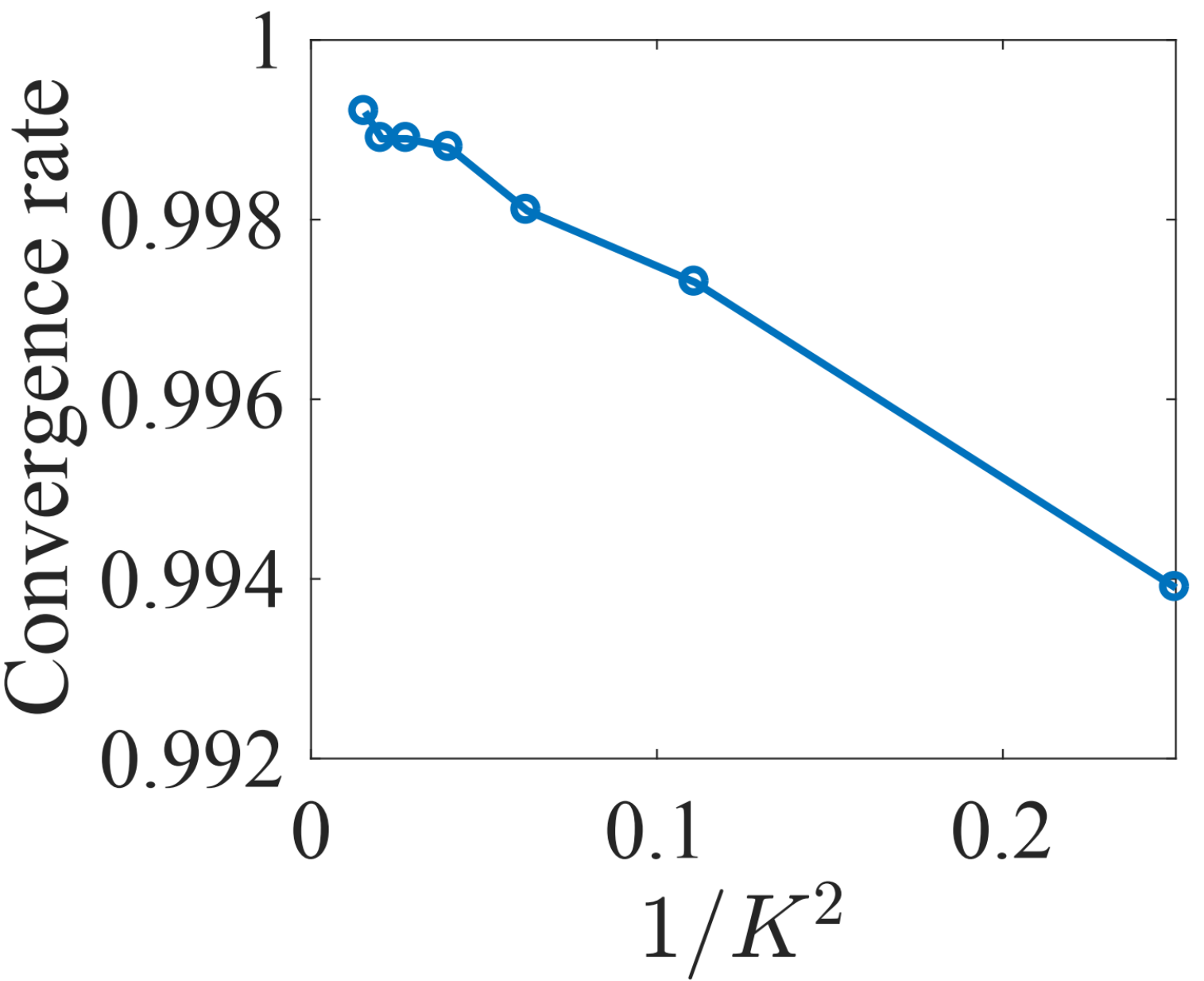}
        \end{minipage}
    }
    ~
    \subfigure[]{
        \begin{minipage}{0.22\textwidth}
        \centering
        \includegraphics[width=1\textwidth,height=1.1in]{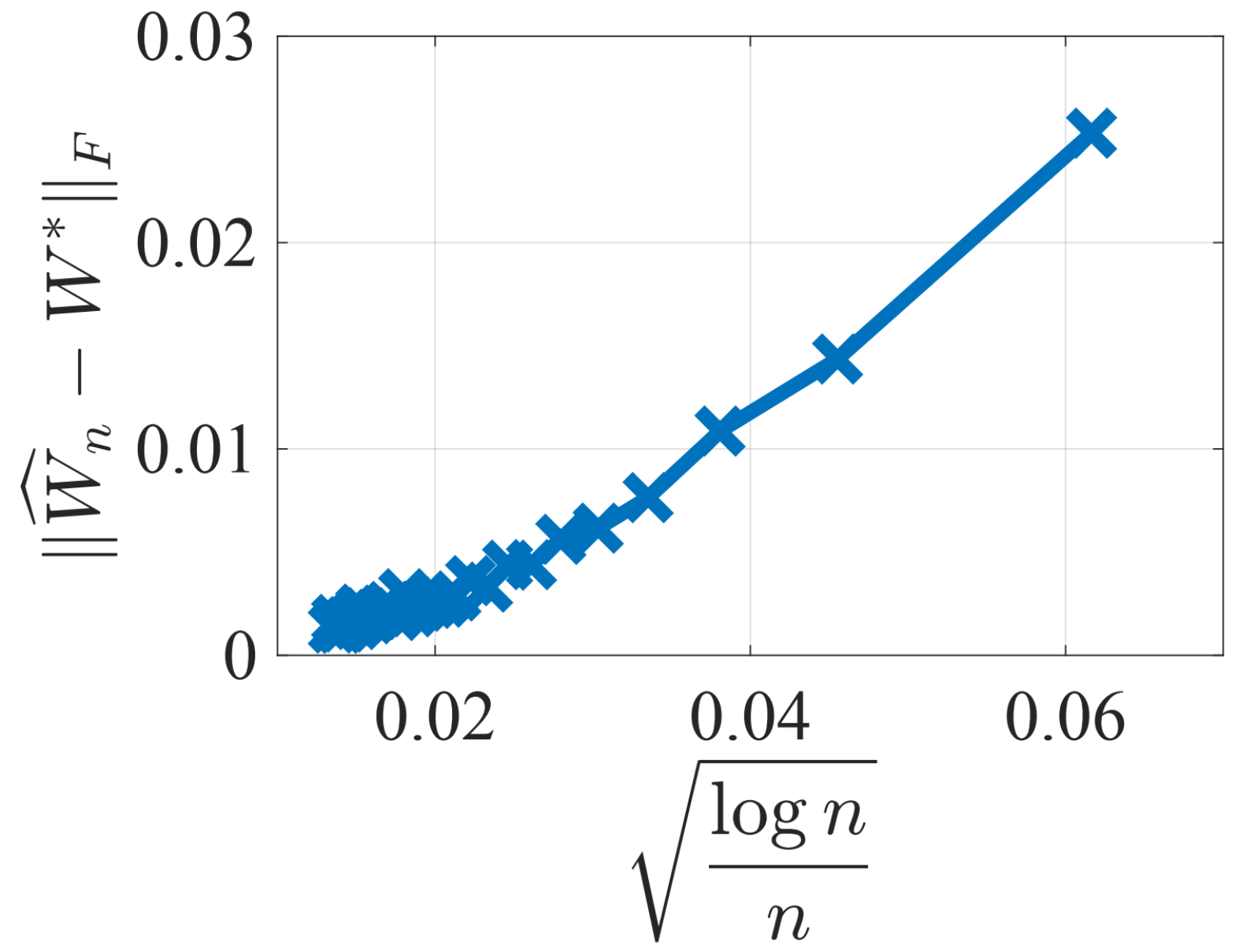}
        \end{minipage}
    }
    \caption{(a) Convergence rate when the number of neurons $K$ changes. (b) The relative error of the learned model when $n$ changes.}\label{figure: conv_K_n}
\end{figure}

\iffalse
\begin{figure}[htbp]
\centering
\begin{minipage}{0.22\textwidth}
\centering
\includegraphics[width=3 in]{spmu.jpg}
\end{minipage}\hfill
\begin{minipage}{0.22\textwidth}
\centering
\includegraphics[width=3 in]{spsigma.jpg}
\end{minipage}\hfill
\caption{The sample complexity against $\sigma$}\label{fig:sigma}
\end{figure}\\
\fi
\textbf{Convergence analysis}. 
We next study the convergence rate of Algorithm 1. %$d$ is fixed as 5.
%\tcr{add a figure to show linear convergence} We fix $d=5$ and $\|\bfW_0-\bfW^*\|_F/\|\bfW^*\|\leq 0.1$. 
 Figure \ref{figure: convergence_musigma}(a) shows the impact of $\|\bfmu_l\|$. %denoted by the Gaussian mixture model on the convergence rate. 
  $\lambda_1=\lambda_2=0.5$, $\bfmu_1=-\bfmu_2=C\cdot\boldsymbol{1}$ for a positive $C$, and $\bfSg_1=\bfSg_2=\bfLM^\top\bfD\bfLM$. Here $\bfLM$ is generated by computing the left-singular vectors of a $d\times d$ random matrix from the Gaussian distribution.  $\bfD=\text{diag}(1,1.1,1.2,1.3,1.4)$.  $n=1\times 10^4$. %The value of $C$ is selected such that
 %We vary $C$ and let $\tilde{\mu}=\max_{l}\|\tilde{\bfmu_l}\|_\infty$.  The sample complexity 
  Algorithm 1 always converges linearly when $\|\bfmu_1\|$ changes. Moreover, as $\|\bfmu_1\|$ increases, Algorithm 1 converges slower. %as predicted by our theoretical analyses in Section \ref{sec: theory}. 
  Figure \ref{figure: convergence_musigma} (b) shows the impact of the variance of the Gaussian mixture model. $\lambda_1=\lambda_2=0.5$, $\bfmu_1=\boldsymbol{1}$, $\bfmu_2=-\boldsymbol{1}$, $\bfSg_1=\bfSg_2=\bfSg=\sigma^2\cdot\bfLM^\top\bfD\bfLM$. $n=5\times 10^4$. We change $\|\bfSg\|$ by changing $\sigma$. Among the values  we test, Algorithm 1 converges fastest when $\|\bfSg\|=1$. The convergence rate slows down when $\|\bfSg\|$ increases or decreases from 1. All results are consistent with the predictions in Corollary \ref{cor}. %The result is consistent with our theoretical results in Section \ref{sec: theory}. %We can find that the convergence rate will increase if $\mu$ or $\sigma$ increases. In
  We then study the impact of $K$ on the convergence rate. %verify the convergence rate in (\ref{linear convergence}), which shows that $v= 1-\Theta(K^{-2})$. We set  
 $\lambda_1=\lambda_2=0.5$, $\bfmu_1=\boldsymbol{1}$, $\bfmu_2=-\boldsymbol{1}$, $\bfSg_1=\bfSg_2=\bfI$. %$K$ ranges from $2$ to $8$. 
  Figure \ref{figure: conv_K_n} (a) shows that, as predicted by (\ref{eqn:v}), the convergence rate is linear in $-1/K^2$. %which verifies our theoretical result of the convergence rate of $1-\Theta(K^{-2})$ in \eqref{linear convergence}.

\textbf{Average and group-level generalization performance}. %We   evaluate 
The distance between   $\widehat{\bfW}_n$ returned by Algorithm 1  and $\bfW^*$ is measured by $||\widehat{\bfW}_n-\bfW^*||_F$. %$d$ is   5. %\tcr{$\|\bfW^*\|_F=1$ Do you normalize $\bfW^*$ in all the experiments}. 
$n$ ranges from $2\times 10^3$ to $6\times 10^4$. $\bfSg_1=\bfSg_2=9\bfI$, $\bfmu_1=\boldsymbol{1}$, $\bfmu_2=-\boldsymbol{1}$. %We randomly pick an initial point such that $\|\bfW_0-\bfW^*\|_F/\|\bfW^*\|=0.1$.
%In Fig.~\ref{fig:distance}, we show the distance between $\widehat{\bfW}_n$ returned by Algorithm \ref{gd}  and $\bfW^*$, measured by the error $||\widehat{\bfW}_n-\bfW^*||_F$. 
Each point in Figure \ref{figure: conv_K_n} (b) is averaged over $20$   experiments of different $\bfW^*$ and %the corresponding 
training set. %$\|\bfW^*\|_F$ is normalized to 1. 
The error is indeed linear in $\sqrt{\log(n)/n}$, as predicted by (\ref{linear convergence}).

\begin{figure}[ht]
    ~
    \subfigure[]{
        \begin{minipage}{0.2\textwidth}
        \centering
        \includegraphics[width=1.1\textwidth, height=1.1in]{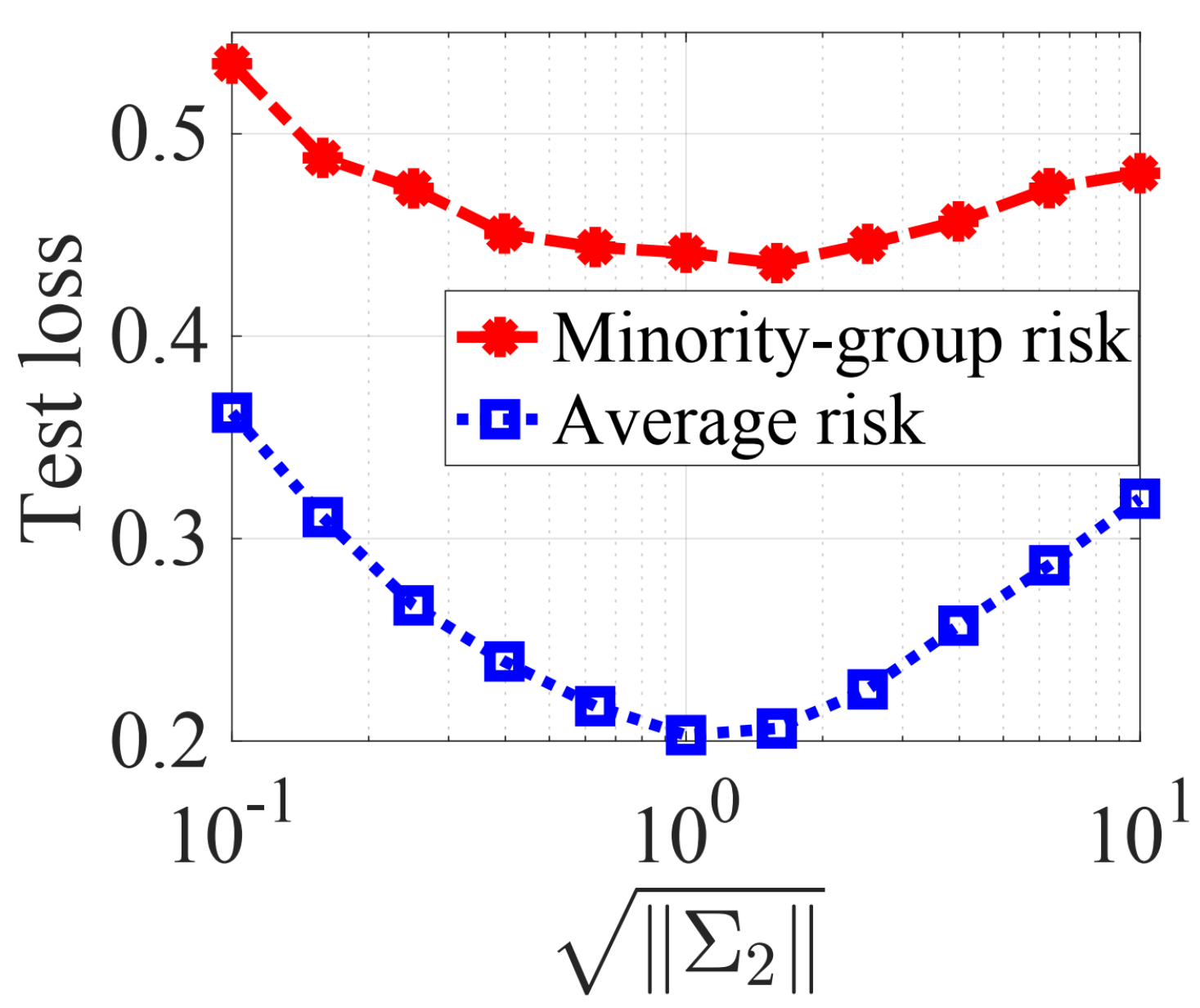}
        \end{minipage}
    }
    ~
    \subfigure[]{
        \begin{minipage}{0.2\textwidth}
        \centering
        \includegraphics[width=1.1\textwidth, height=1.1in]{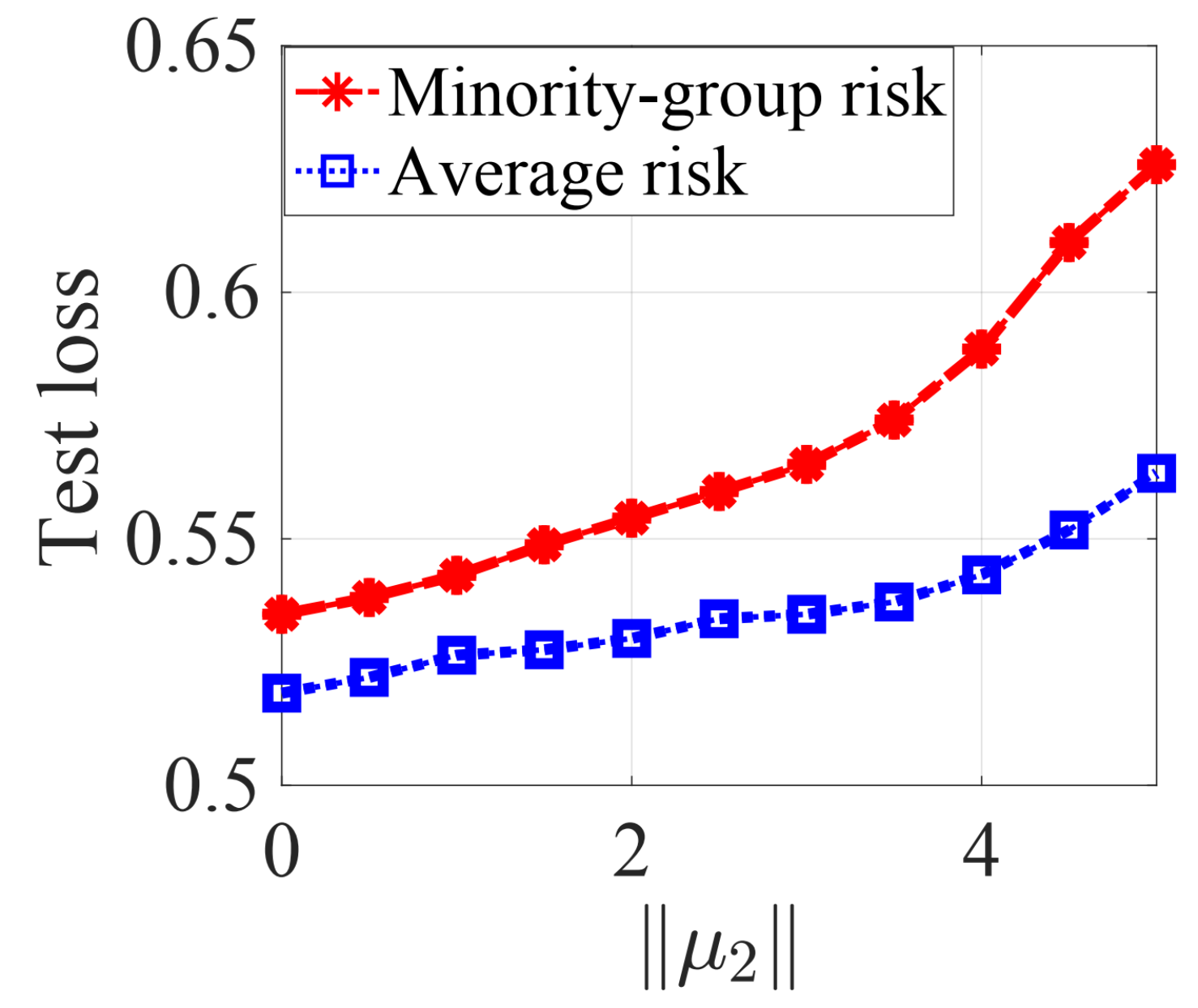}
        \end{minipage}
    }
    \caption{(a) The cross-entropy test loss when the co-variance of the minority group changes. (b) The cross-entropy  test loss  when the mean of the   minority group changes.}
    \label{figure: loss_mu_sigma}
\end{figure}

 We evaluate the impact of one mean/co-variance of the minority group on the generalization. $n=2\times 10^4$. Let $\lambda_1=0.8$, $\lambda_2=0.2$, $\bfmu_1=2\cdot\boldsymbol{1}$, $\bfSg_1=\bfI$. First,  we let $\bfmu_2=(\mu_2-2)\cdot\boldsymbol{1}$ and $\bfSg_2=\bfI$.     
%We then verify how the distribution parameters of data affect the generalization error.
%Let the data dimension $d=5$ and the number of samples $n=2\times 10^4$. We first fix the group ratio and only change the mean or variance of the minority group. 
%Set $\lambda_1=0.8$, $\lambda_2=0.2$, $\bfmu_1=-\boldsymbol{1}$, $\bfSg_1=\bfI$. If we let $\bfmu_2=\mu_2\cdot\boldsymbol{1}$ and $\bfSg_2=\bfI$, 
Figure \ref{figure: loss_mu_sigma} (b) shows that both the average risk and the group-$2$ risk increase as $\mu_2$ increases,  consistent with  (P5). Then we set $\bfmu_2=-2\cdot\boldsymbol{1}$, $\bfSg_2=\sigma_2^2\cdot\bfI$. Figure \ref{figure: loss_mu_sigma} (a) indicates that both the average   and the group-$2$ risk will first decrease and then increase as $\|\Sigma\|_2$ increases,  consistent with  (P3). % which verifies the insight (P3).

\begin{figure}[ht]
    \centering
    \subfigure[]{
        \begin{minipage}{0.3\textwidth}
        \centering
        \includegraphics[width=1\textwidth, height=0.5\textwidth]{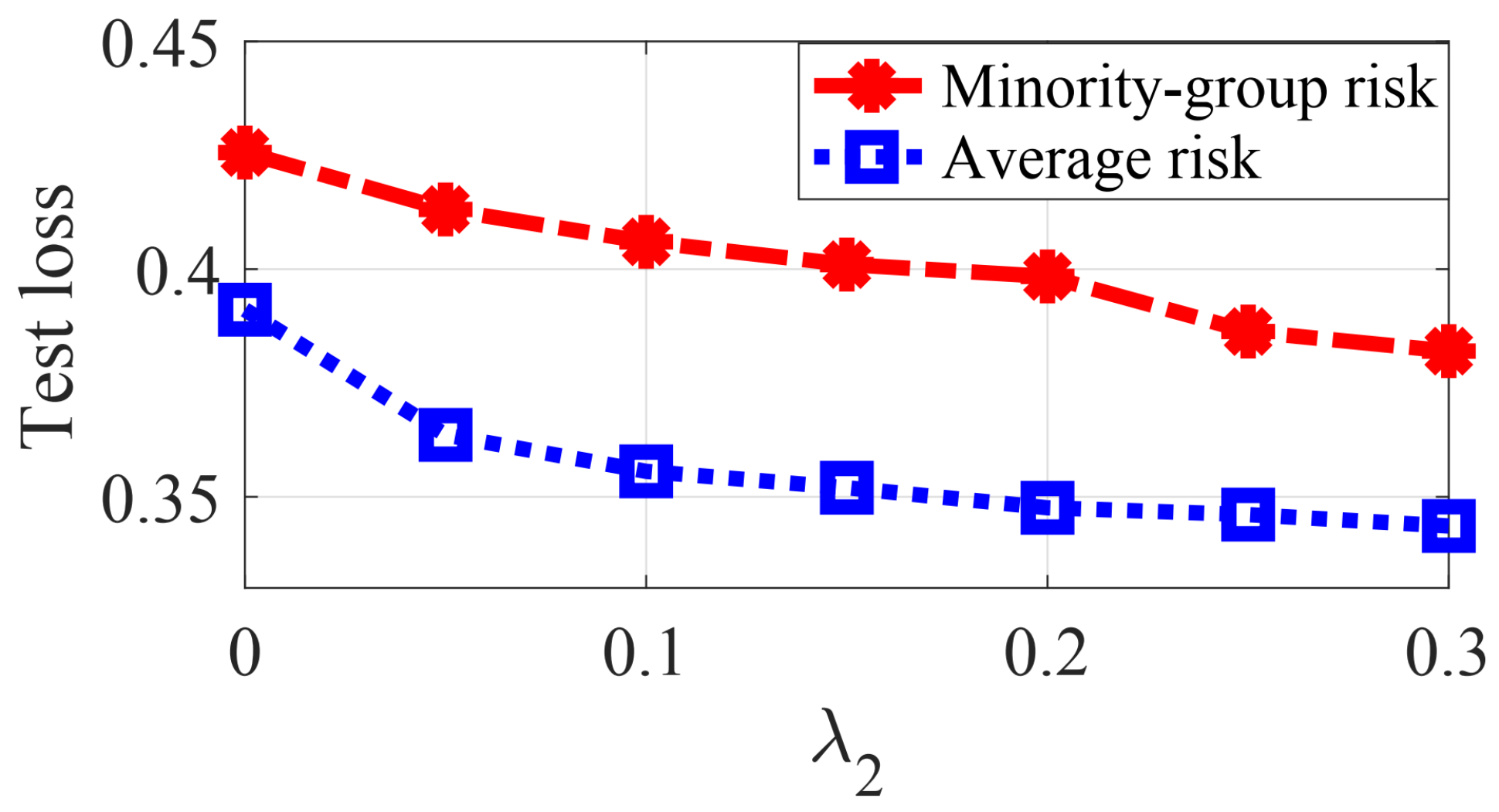}
        \end{minipage}
    }
    ~
    \subfigure[]{
        \begin{minipage}{0.3\textwidth}
        \centering
        \includegraphics[width=1\textwidth,height=0.48\textwidth]{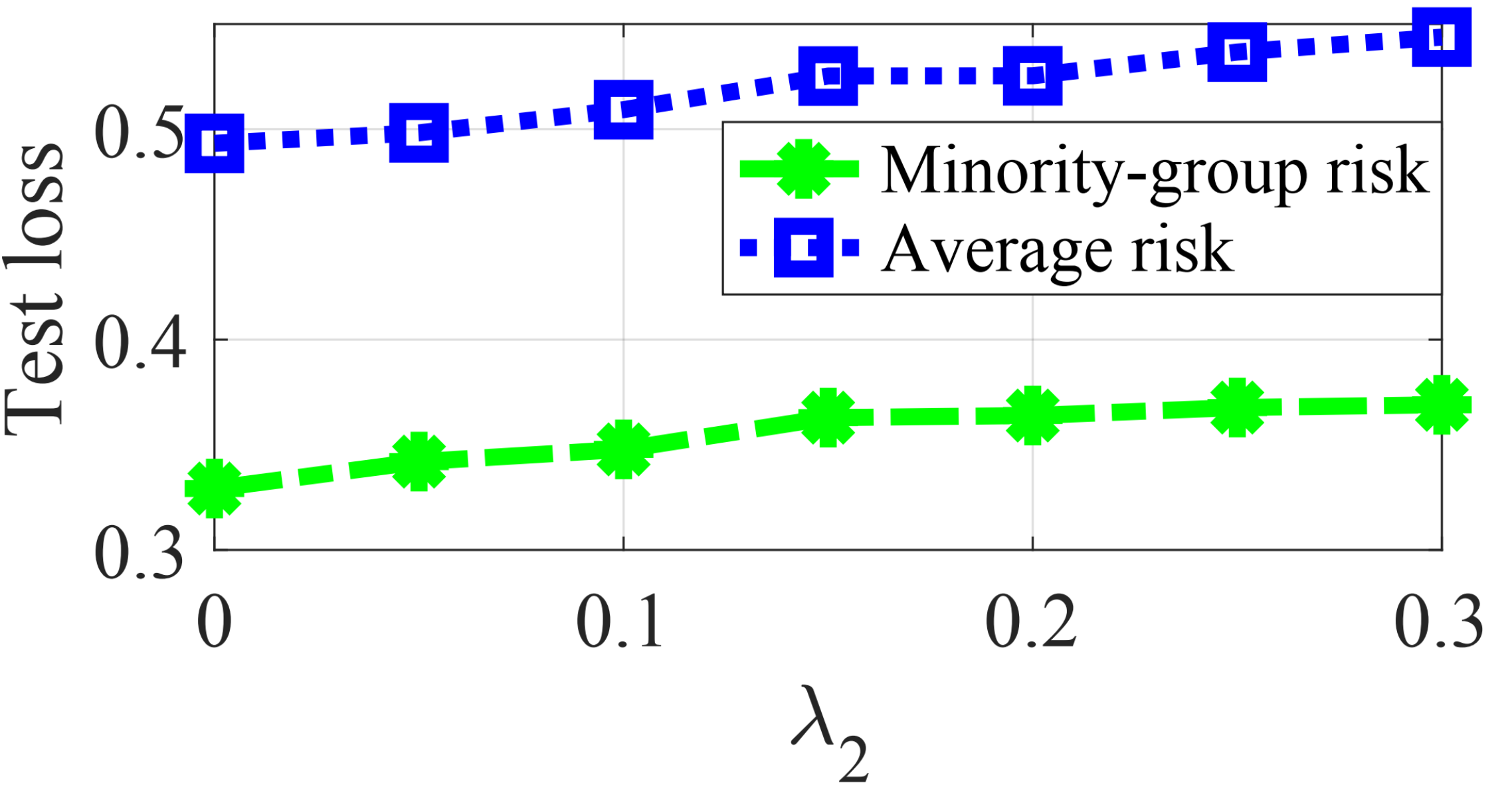}
        \end{minipage}
    }
    \caption{The test loss (cross entropy loss) of synthetic data with different $\lambda_2$ values. (a) Group 2 has a smaller level of co-variance. (b) Group 2 has a larger level of co-variance.}
    \label{figure: lambda2_v}
\end{figure}

Next, we study the impact of increasing the fraction of the minority group. 
%We then fix the mean and variance and only change the group ratio. Let group 2 be the minority group. 
$\bfmu_1=\bfmu_2=0$. Let group 2 be the minority group. In Figure \ref{figure: lambda2_v} (a), $\bfSg_1=10\cdot\bfI$ and $\bfSg_2=\bfI$, the minority group has a smaller level of co-variance. Then when $\lambda_2$ increases from 0 to 0.5, both the average   and group-$2$ risk decease. 
In Figure \ref{figure: lambda2_v} (b),
$\bfSg_1=\bfI$ and $\bfSg_2=10\cdot\bfI$, and the minority group has a higher-level of co-variance. Then when $\lambda_2$ increases from 0 to 0.3, both the average   and group-$2$ risk increase. As predicted by insight (P4), increasing $\lambda_2$ does not necessarily improve the generalization of group $2$. 

%reflect different trends of test loss under different conditions of data variance. The test loss decreases as $\lambda_2$ increases if $\bfSg_1=10\cdot\bfI$ and $\bfSg_2=\bfI$, while the loss increases as $\lambda_2$ increases if $\bfSg_1=\bfI$ and $\bfSg_2=10\cdot\bfI$. This is in accordance with the insight P4. 

\subsection{Image classification on dataset CelebA}\label{sec: celebA_main}
   %The total number of raw training data is set to $5K$. 
%We select $5000$ images  from CeleA with two groups, male and female.
We choose the attribute ``blonde hair'' as the binary classification label. ResNet 9 \cite{HZRS16} is selected to be the learning model here because it was applied in many simple computer vision tasks \cite{WWYJ18, DBAH20}. %,  due to the simple task and the small training size. % small amount of data used for training,  to avoid overfitting. 
To study the impact of co-variance, %how the performance is related to the data variance, 
we pick $4000$ female (majority) and $1000$ male (minority) images and implement Gaussian data augmentation to create additional $300$   images for the male group. Specifically, we select $300$ out of $1000$ male images and add i.i.d. noise drawn from $\mathcal{N}(0, \delta^2)$  to  every entry. The test set includes $500$ male and $500$ female images.  %We generate $300$ extra data of the minority group by adding different levels of Gaussian noise to the raw images to control the group variance. 
Figure \ref{figure: celebA_noise} shows that when $\delta^2$ increases, i.e., when the co-variance of the minority group increases, % with the noise level in the  increasing, 
both the minority-group and average test accuracy increase first and then decrease, coinciding with our insight (P3). 

\begin{figure}[ht]
    \centering
    \subfigure[]{
        \begin{minipage}{0.3\textwidth}
        \centering
        \includegraphics[width=1\textwidth,height=0.56\textwidth]{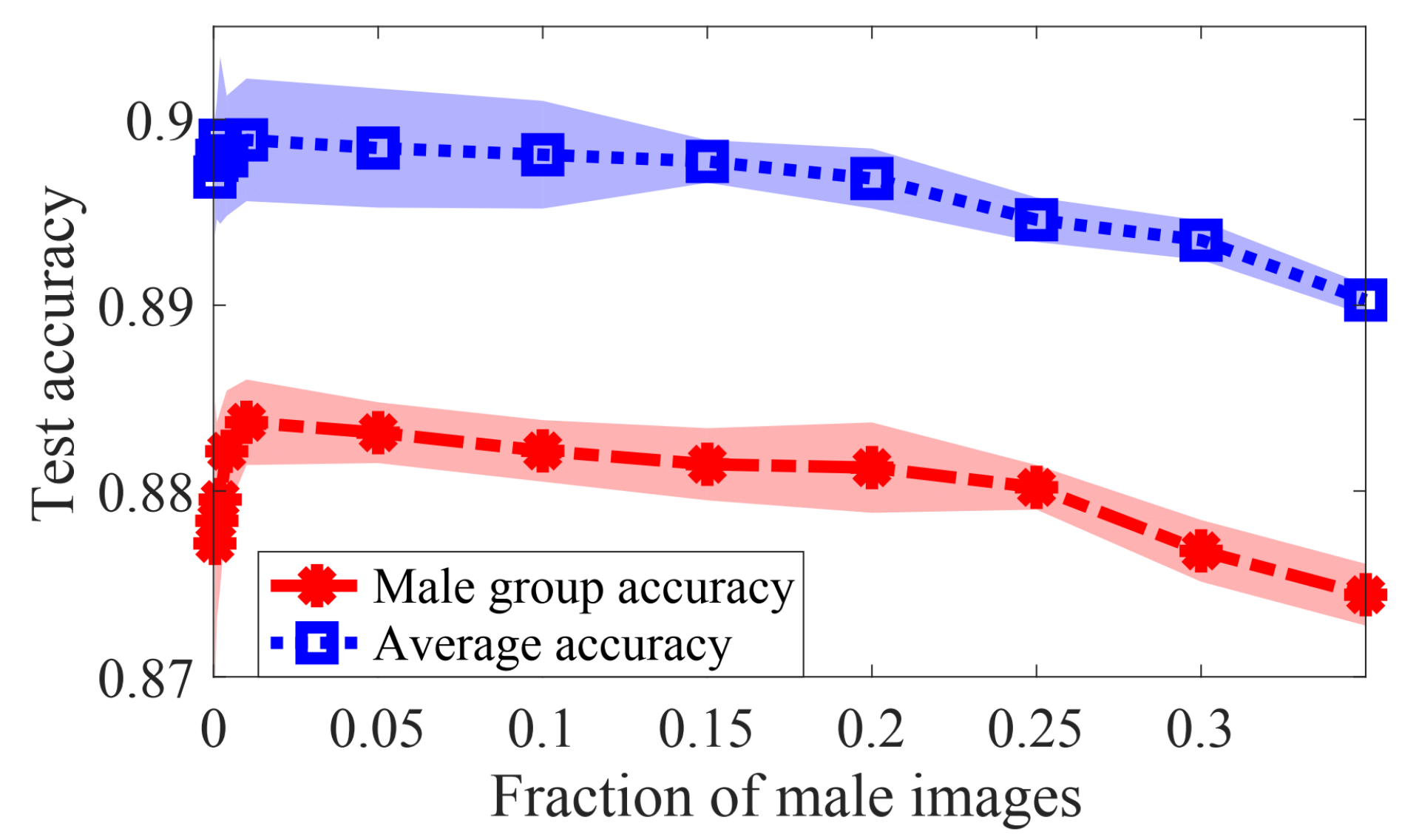}
        \end{minipage}
    }
    ~
    \subfigure[]{
        \begin{minipage}{0.3\textwidth}
        \centering
        \includegraphics[width=1\textwidth,height=0.53\textwidth]{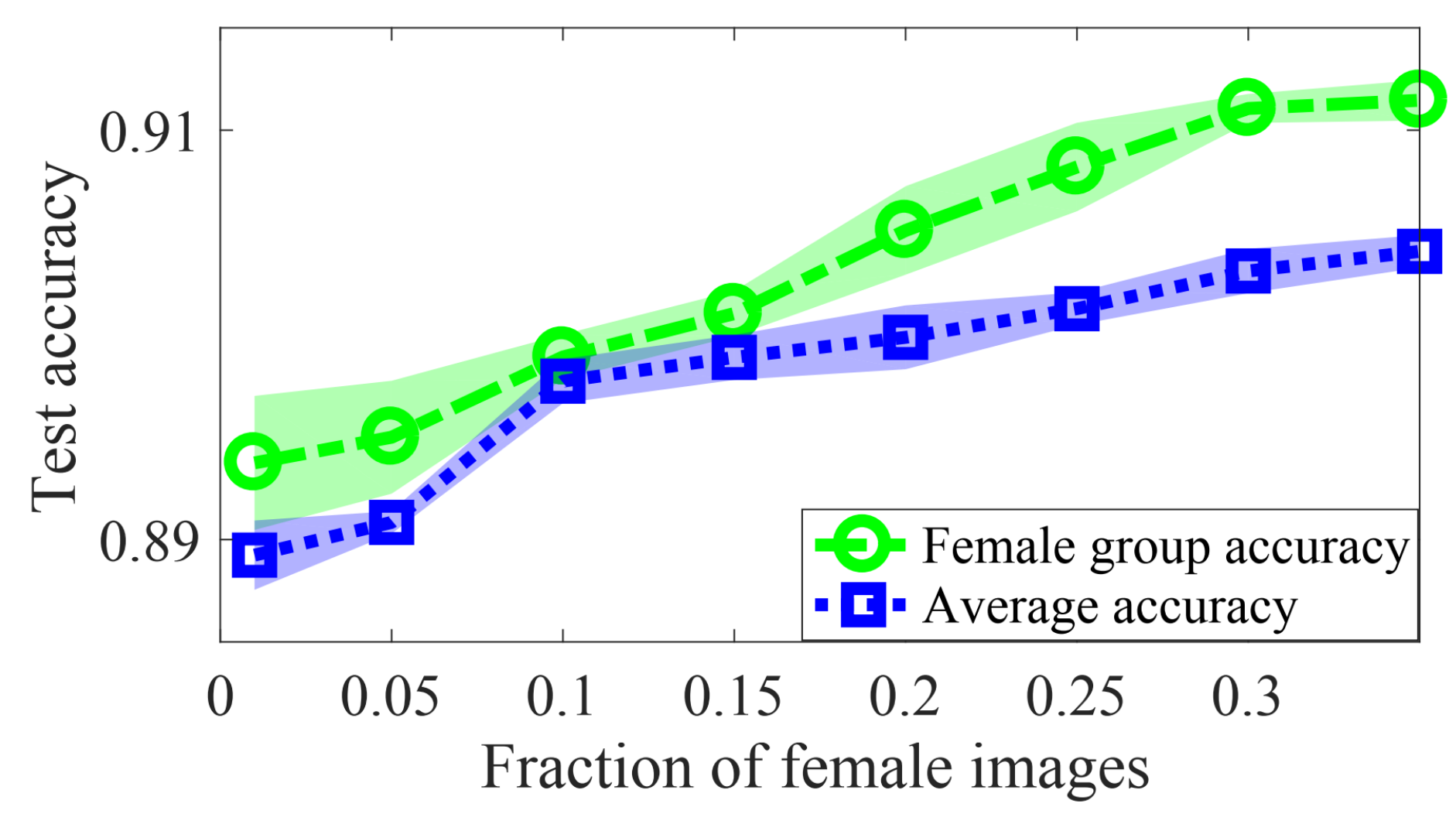}
        \end{minipage}
    }
       \vspace{-2mm}
    \caption{The test accuracy on CelebA dataset has opposite trends when the minority group fraction increases.  (a) Male group is the minority (b) Female group is the minority}\label{figure: celebA_group}
\end{figure}

Then we fix the total number of training data to be $5000$ and vary the fractions of the two groups.  From Figure \ref{figure: celebA_group}(a)\footnote{%The test accuracy in  Figure \ref{figure: celebA_group}(a)   drops a little when the minority fraction is very close to zero, less than $0.01$. Because the
 In Figure \ref{figure: celebA_group}(a), when the  minority fraction is  less than $0.01$,   the minority group distribution is almost removed from the Gaussian mixture model. Then the $O(1)$ constants in the last column of Table \ref{tbl:results} have some minor changes, and the order-wise analyses do not reflect the minor fluctuations in this regime. 
} and (b), we observe opposite trends if we increase the fraction of the minority group in the training data with the male being the minority and the female being the minority. The norm of covariance of the male and female group in the feature space is $5.1833$ and $4.9716$, respectively. This is consistent with Insight (P4).  %ratio of the male or the female group as the minority, respectively. 
Due to space limit, our results on the CIFAR10 dataset are deferred to Section A in the supplementary material. 

\iffalse
\begin{figure}[htbp]
\centering
%\begin{minipage}{0.90\linewidth}
 \centering
\includegraphics[width=0.85\linewidth]{figures/celebA_noise.png}
\caption{The test loss of CelebA data with different levels of added noise.}
\label{figure: celebA_noise}
\end{figure}
\begin{figure}[htbp]
\centering
%\begin{minipage}{0.90\linewidth}
 \centering
\includegraphics[width=0.85\linewidth]{figures/celebA_male.jpg}
\caption{The test accuracy of CelebA data with different ratio of the male.}
\label{figure: celebA_male}
\end{figure}
\begin{figure}[htbp]
\centering
%\begin{minipage}{0.90\linewidth}
 \centering
\includegraphics[width=0.85\linewidth]{figures/celebA_female.jpg}
\caption{The test accuracy of CelebA data with different ratio of the female.}
\label{figure: celebA_female}
\end{figure}
\fi

\section{Conclusions and Future Directions}\label{sec: conclusion_lim_future}
%This paper analyzes the theoretical performance guarantee of learning one-hidden-layer neural networks for binary classification when 
This paper provides a novel theoretical framework for characterizing neural network generalization with group imbalance. The group imbalance is formulated using the Gaussian mixture model.  
This paper explicitly quantifies the impact of each group on the sample complexity, convergence rate, and the average and the group-level   generalization. The learning performance is enhanced when the group-level covariance is at a medium regime, and the group-level mean is close to zero. Moreover, increasing the fraction of minority group does not guarantee improved group-level generalization. 

%We develop an algorithm that converges linearly to a model that has a diminishing difference from the teacher model that has guaranteed generalizability. We also provide   explicit characterizations of the impact of the input distribution on the sample complexity and convergence rate. %This is the first theoretical characterization of the impact of input distributions on the learning performance.

One future direction is to extend the analysis to 
 multiple-hidden-layer neural networks and multi-class classification. %, as well as ReLU activation function. 
 Because of the concatenation of nonlinear activation functions, the analysis of the landscape of the empirical risk and the design of a proper initialization is more challenging and   requires the development of new tools. Another future direction is to analyze other robust training methods, such as DRO. We see no ethical or immediate negative societal consequence of our work.% that for any value below, the 

\section{Acknowledgments}
\noindent This research is supported in part by NSF 1932196, AFOSR FA9550-20-1-0122, and Rensselaer-IBM AI Research Collaboration (http://airc.rpi.edu), part of the IBM AI Horizons Network (http://ibm.biz/AIHorizons).

\appendix

\subsection{Definitions}

\begin{definition}\label{def: rho}
($\rho$-function). Let $\bfz\sim\mathcal{N}(\bfu,\bfI_d)\in\mathbb{R}^d$. Define $\alpha_q(i,\bfu,\sigma)=\mathbb{E}_{z_i\sim\mathcal{N}(u_i,1)}[\phi'(\sigma\cdot z_i)z_i^q]$ and $\beta_q(i,\bfu,\sigma)=\mathbb{E}_{z_i\sim\mathcal{N}(u_i,1)}[\phi'^2(\sigma\cdot z_i)z_i^q]$, $\forall\ q\in\{0,1,2\}$, where $z_i$ and $u_i$ is the $i$-th entry of $\bfz$ and $\bfu$, respectively. Define $\rho(\bfu,\sigma)$ as
\begin{equation}
\begin{aligned}
\rho(\bfu,\sigma)=\min_{i,j\in[d],j\neq i}\{&(u_{j}^2+1)(\beta_0(i,\bfu,\sigma)-\alpha_0(i,\bfu,\sigma)^2), \\
&\beta_2(i,\bfu,\sigma)-\frac{\alpha_2(i,\bfu,\sigma)^2}{u_i^2+1}\}
\end{aligned}\label{rho_function}
\end{equation}
\end{definition}

\begin{definition}\label{def: D}
(D-function). Given the Gaussian Mixture Model and any positive integer $m$,   define $D_m(\Psi)$ as
\begin{equation}D_m(\Psi)=\sum_{l=1}^L\lambda_l(\frac{\|{\bfmu}_l\|}{\|\bfSg_l^{-1}\|^{-\frac{1}{2}}}+1)^m,
\end{equation}
\end{definition}

$\rho$-function is defined to compute the lower bound of the Hessian of the population risk with Gaussian input. $D$-function is a normalized parameter for the means and variances. It is lower bounded by 1. $D$-function is an increasing function of $\|\bfmu_l\|$ and a decreasing function of $\sigma_l$. %$\bfQ_j, j=1,2,3$ are tensors of $\bfw_i^*$ which can be used to estimate $\bfw_i^*$ via tensor decomposition.

\subsection{Proof of Lemma \ref{lemma: convexity_informal}}
We first restate the formal version of Lemma \ref{lemma: convexity_informal} in the following.

\begin{customlemma}{1}\label{lemma: convexity}(Strongly local convexity)
\normalsize Consider the classification model with FCN (\ref{cla_model}) and the sigmoid activation function. There exists a constant $C$ such that as long as the sample size 
\begin{equation}
\begin{aligned}
n\geq &C_1\epsilon_0^{-2}\cdot \big(\sum_{l=1}^L\lambda_l(\|\bfmu_l\|+\|\bfSg_l^\frac{1}{2}\|)^2\big)^2\\
&\cdot\Big(\sum_{l=1}^L\lambda_l\frac{\|\bfSg_l^{-1}\|^{-1}}{\eta\tau^K\kappa^2}\rho(\frac{{\bfW^*}^\top\bfmu_l}{\delta_K(\bfW^*)\|\bfSg_l^{-1}\|^{-\frac{1}{2}}}, \\
&\delta_K(\bfW^*)\|\bfSg_l^{-1}\|^{-\frac{1}{2}})\Big)^{-2}dK^5\log^2{d}
\label{sp_1}
\end{aligned}
\end{equation}
for some constant $C_1>0$,  $\epsilon_0\in(0,\frac{1}{4})$, and any fixed permutation matrix $\bfP \in\mathbb{R}^{K\times K}$
we have for all $\bfW\in\mathbb{B}(\bfW^*\bfP,r)$,
\begin{equation}
\begin{aligned}
&\Omega\Big(\frac{1-2\epsilon_0}{K^2}\sum_{l=1}^L\lambda_l\frac{\|\bfSg_l^{-1}\|^{-1}}{\eta\tau^K\kappa^2}\rho(\frac{{\bfW^*}^\top\bfmu_l}{\delta_K(\bfW^*)\|\bfSg_l^{-1}\|^{-\frac{1}{2}}}, \\
&\delta_K(\bfW^*)\|\bfSg_l^{-1}\|^{-\frac{1}{2}})\Big)\cdot\bfI_{dK}\\
&\preceq\nabla^2f_n(\bfW)
\preceq C_2 \sum_{l=1}^L\lambda_l(||\tilde{\bfmu}_l||_\infty+\|\bfSg_l^\frac{1}{2}\|)^2\cdot\bfI_{dK}
\end{aligned}
\end{equation}
with probability at least $1-d^{-10}$ for some constant $C_2>0$.
\end{customlemma}

\subsubsection{Useful lemmas}
Lemmas \ref{lm: rho}, \ref{lm: fraction bound}, \ref{lm: smoothness}, \ref{lm: convex}, and \ref{lm: bernstein} are required for the proof.
\begin{lemma}\label{lm: rho}
\normalsize
\begin{equation}
\begin{aligned}&\mathbb{E}_{\bfx\sim\frac{1}{2}\mathcal{N}(\bfmu,\bfI_d)+\frac{1}{2}\mathcal{N}(-\bfmu,\bfI_d)}\Big[(\sum_{i=1}^k \bfr_i^\top \bfx\cdot\phi'(\sigma\cdot x_i))^2\Big] \\&\geq \rho(\bfmu,\sigma)||\bfR||_F^2,
\end{aligned}\end{equation}
where $\rho(\bfmu,\sigma)$ is defined in Definition \ref{def: rho} and $\bfR=(\bfr_1,\cdots,\bfr_k)\in\mathbb{R}^{d\times k}$ is an arbitrary matrix.\\
\end{lemma}

\begin{lemma}\label{lm: fraction bound} \normalsize With the FCN model (\ref{cla_model}) and the Gaussian Mixture Model,  for any permutation matrix $\bfP$, for some constant $C_{12}>0$, we have
we have
\begin{equation}
\begin{aligned}
&\mathbb{E}_{\bfx\sim\sum_{l=1}^L\lambda_l\mathcal{N}(\bfmu_l,\bfSg_l)}\Big[ \sup_{\bfW\neq \bfW'\in\mathbb{B}(\bfW^*\bfP,r)}||\nabla^2\ell(\bfW,\bfx)\\
&-\nabla^2\ell(\bfW',\bfx)||/||\bfW-\bfW'||_F\Big]\\
\lesssim&  d^{\frac{3}
{2}}K^{\frac{5}{2}}\sqrt{\sum_{l=1}^L\lambda_l(\|\bfmu_l\|_\infty+\|\bfSg_l\|)^2\sum_{l=1}^L\lambda_l(\|\bfmu_l\|_\infty+\|\bfSg_l\|)^4}
\end{aligned}
\end{equation}
\end{lemma}

\begin{lemma}\label{lm: smoothness}
(Hessian smoothness of population loss) \normalsize In the FCN model (\ref{cla_model}),  for any permutation matrix $\bfP$, we have
\begin{equation}
\begin{aligned}
&||\nabla^2\bar{f}(\bfW)-\nabla^2\bar{f}(\bfW^*\bfP)||
\lesssim  K^{\frac{3}{2}}\cdot ||\bfW-\bfW^*\bfP||_F\\
&\cdot\Big(\sum_{l=1}^L\lambda_l(\|\bfmu_l\|+\|\bfSg_l^\frac{1}{2}\|)^4\sum_{l=1}^L\lambda_l(\|\bfmu_l\|+\|\bfSg_l^\frac{1}{2}\|)^8\Big)^{\frac{1}{4}}\label{smooth}
\end{aligned}
\end{equation}
\end{lemma}

\begin{lemma}\label{lm: convex}
(Local strong convexity of population loss) \normalsize In the FCN model $(\ref{cla_model})$, for any permutation matrix $\bfP$, if $||\bfW-\bfW^*\bfP||_F\leq r$ for an $\epsilon_0\in(0,\frac{1}{4})$, then, 
\begin{equation}\label{upper&lower}
\begin{aligned}
&\frac{4(1-\epsilon_0)}{K^2}\sum_{l=1}^L\lambda_l\frac{\|\bfSg_l^{-1}\|^{-1}}{\eta\tau^K\kappa^2}\rho(\frac{{\bfW^*}^\top\bfmu_l}{\delta_K(\bfW^*)\|\bfSg_l^{-1}\|^{-\frac{1}{2}}},\delta_K(\bfW^*)\\
&\cdot\|\bfSg_l^{-1}\|^{-\frac{1}{2}})\cdot\bfI_{dK}\preceq\nabla^2\bar{f}(\bfW)\preceq \sum_{l=1}^L\lambda_l(\|\bfmu_l\|+\bfSg_l^\frac{1}{2})^2\cdot\bfI_{dK}
\end{aligned}
\end{equation}
\end{lemma}

\begin{lemma}\label{lm: bernstein}
\normalsize In the FCN model $(\ref{cla_model})$, for any permutation matrix $\bfP$, as long as $n\geq C'\cdot dK\log{dK}$ for some constant $C'>0$, we have
\begin{equation}\label{distance}
\begin{aligned}
&\sup_{\bfW\in\mathbb{B}(\bfW^*\bfP,r)}||\nabla^2 f_n(\bfW)-\nabla^2\bar{f}(\bfW)||\\
\leq &\sum_{l=1}^L\lambda_l(\|\bfmu_l\|+\|\bfSg_l^\frac{1}{2}\|)^2\sqrt{\frac{dK\log{n}}{n}})
\end{aligned}
\end{equation}
with probability at least $1-d^{-10}$.
\end{lemma}
\vspace{5mm}
We next show the proof of Lemma \ref{lemma: convexity}.

\subsubsection{Proof}\label{subsecc: pf lm1}
\normalsize 
\noindent From Lemma \ref{lm: convex} and \ref{lm: bernstein}, with probability at least $1-d^{-10}$, 
\begin{equation}
\begin{aligned}
    \nabla^2 f_n(\bfW)&\succeq \nabla^2 \bar{f}(\bfW)-||\nabla^2 \bar{f}(\bfW)-\nabla^2 f_n(\bfW)||\cdot\bfI\\
    &\succeq \Omega\Big(\frac{(1-\epsilon_0)}{K^2}\sum_{l=1}^L\lambda_l\frac{\|\bfSg_l^{-1}\|^{-1}}{\eta\tau^K\kappa^2}\rho(\frac{{\bfW^*}^\top\bfmu_l}{\delta_K(\bfW^*)\|\bfSg_l^{-1}\|^{-\frac{1}{2}}},\\
    &\delta_K(\bfW^*)\|\bfSg_l^{-1}\|^{-\frac{1}{2}})\Big)\cdot\bfI\\
    &-O\Big(C_6\cdot \sum_{l=1}^L\lambda_l(\|\bfmu_l\|+\|\bfSg_l^\frac{1}{2}\|)^2\sqrt{\frac{dK\log{n}}{n}}\Big)\cdot\bfI
\end{aligned}
\end{equation}
As long as the sample complexity is set to satisfy
\begin{equation}
\begin{aligned}
&\sum_{l=1}^L\lambda_l(\|\bfmu_l\|+\|\bfSg_l^\frac{1}{2}\|)^2\cdot\sqrt{\frac{dK\log{n}}{n}}\leq \frac{\epsilon_0}{K^2}\sum_{l=1}^L\lambda_l\frac{\|\bfSg_l^{-1}\|^{-1}}{\eta\tau^K\kappa^2}\\
&\cdot\rho(\frac{{\bfW^*}^\top\bfmu_l}{\delta_K(\bfW^*)\|\bfSg_l^{-1}\|^{-\frac{1}{2}}},\delta_K(\bfW^*)\|\bfSg_l^{-1}\|^{-\frac{1}{2}})\cdot\bfI
\end{aligned}
\end{equation}
i.e.,
\begin{equation}
\begin{aligned}
n\gtrsim &\epsilon_0^{-2}\cdot \big(\sum_{l=1}^L\lambda_l(\|\bfmu_l\|+\|\bfSg_l^\frac{1}{2}\|)^2\big)^2\\
&\cdot\Big(\sum_{l=1}^L\lambda_l\frac{\|\bfSg_l^{-1}\|^{-1}}{\eta\tau^K\kappa^2}\rho(\frac{{\bfW^*}^\top\bfmu_l}{\delta_K(\bfW^*)\|\bfSg_l^{-1}\|^{-\frac{1}{2}}},\\
&\delta_K(\bfW^*)\|\bfSg_l^{-1}\|^{-\frac{1}{2}})\cdot\bfI\Big)^{-2}dK^5\log^2{d}
\end{aligned}
\end{equation}
for some constant $C_1>0$, then we have the lower bound of the Hessian with probability at least $1-d^{-10}$. 
\begin{equation}\label{lowerbound_lm1}
\begin{aligned}
    &\nabla^2 f_n(\bfW) \succeq \Omega\Big(\frac{1-2\epsilon_0}{K^2}\sum_{l=1}^L\lambda_l\frac{\|\bfSg_l^{-1}\|^{-1}}{\eta\tau^K\kappa^2}\\
    &\cdot\rho(\frac{{\bfW^*}^\top\bfmu_l}{\delta_K(\bfW^*)\|\bfSg_l^{-1}\|^{-\frac{1}{2}}},\delta_K(\bfW^*)\|\bfSg_l^{-1}\|^{-\frac{1}{2}})\Big)\cdot\bfI
\end{aligned}
\end{equation}By (\ref{upper&lower}) and (\ref{distance}), we can also derive  the upper bound as follows, 
\begin{equation}
\begin{aligned}\label{upperbound_lm1}
    ||\nabla^2 f_n(\bfW)||&\leq ||\nabla^2 \bar{f}(\bfW)||+||\nabla^2 f_n(\bfW)-\nabla^2 \bar{f}(\bfW)||\\
    &\lesssim \sum_{l=1}^L\lambda_l(\|\bfmu_l\|+\|\bfSg_l^\frac{1}{2}\|)^2\\
    &+\sum_{1=1}\lambda_l(\|\bfmu_l\|+\|\bfSg_l^\frac{1}{2}\|)^2\sqrt{\frac{dK\log{n}}{n}}\\
    &\lesssim \sum_{l=1}^L\lambda_l(\|\bfmu_l\|+\|\bfSg_l^\frac{1}{2}\|)^2
\end{aligned}
\end{equation}
Combining (\ref{lowerbound_lm1}) and (\ref{upperbound_lm1}), we have
\begin{equation}
\begin{aligned}
&\Omega\Big(\frac{1-2\epsilon_0}{K^2}\sum_{l=1}^L\lambda_l\frac{\|\bfSg_l^{-1}\|^{-1}}{\eta\tau^K\kappa^2}\rho(\frac{{\bfW^*}^\top\bfmu_l}{\delta_K(\bfW^*)\|\bfSg_l^{-1}\|^{-\frac{1}{2}}},\\
&\delta_K(\bfW^*)\|\bfSg_l^{-1}\|^{-\frac{1}{2}})\Big)\cdot\bfI\preceq\nabla^2f_n(\bfW)\\
\preceq &\sum_{l=1}^L\lambda_l(||\tilde{\bfmu}_l||_\infty+\|\bfSg_l^\frac{1}{2}\|)^2\cdot\bfI
\end{aligned}
\end{equation}
with probability at least $1-d^{-10}$.

\subsection{Proof of Lemma \ref{lemma: convergence}}
We restate the formal version of Lemma \ref{lemma: convergence_informal} in the following.
\begin{customlemma}{2}\label{lemma: convergence}(Linear convergence of gradient descent)
\normalsize Assume the conditions in Lemma \ref{lemma: convexity} hold. Given  any fixed permutation matrix $\bfP\in\mathbb{R}^{K\times K}$, if the local convexity of $\mathbb{B}(\bfW^*\bfP,r)$  holds, 
there exists a critical point in $\mathbb{B}(\bfW^*\bfP,r)$ for some constant $C_3>0$,   and $\epsilon_0\in(0,\frac{1}{2})$, such that 
\begin{equation}
\begin{aligned}&||\widehat{\bfW}_n-\bfW^*\bfP||_F\\
\lesssim &\frac{K^{\frac{5}{2}}\sqrt{ \sum_{l=1}^L\lambda_l(\|\bfmu_l\|+\|\bfSg_l^\frac{1}{2}\|)^2}(1+\xi)\cdot \sqrt{d\log{n}/n}}{\sum_{l=1}^L\lambda_l\frac{\|\bfSg_l^{-1}\|^{-1}}{\eta\tau^K\kappa^2}\rho(\frac{{\bfW^*}^\top\bfmu_l}{\delta_K(\bfW^*)\|\bfSg_l^{-1}\|^{-\frac{1}{2}}}, \delta_K(\bfW^*)\|\bfSg_l^{-1}\|^{-\frac{1}{2}})}\label{critical_groundtruth}
\end{aligned}
\end{equation}

\noindent If the initial point $\bfW_0\in\mathbb{B}(\bfW^*\bfP, r)$, the gradient descent linearly converges to $\widehat{\bfW}_n$, i.e.,
\begin{equation}
\begin{aligned}
&||\bfW_t-\widehat{\bfW}_n||_F\leq ||\bfW_0-\widehat{\bfW}_n||_F\cdot\Big(1-\\
&\Omega\big(\frac{\sum_{l=1}^L\frac{\lambda_l\|\bfSg_l^{-1}\|^{-1}}{\eta \tau^K\kappa^2}\rho(\frac{{\bfW^*}^\top\bfmu_l}{\delta_K(\bfW^*)\|\bfSg_l^{-1}\|^{-\frac{1}{2}}},\delta_K(\bfW^*)\|\bfSg_l^{-1}\|^{-\frac{1}{2}})}{K^2\sum_{l=1}^L\lambda_l(\|\bfmu_l\|+\|\bfSg_l^\frac{1}{2}\|)^2}\big)\Big)^t
\end{aligned}\label{convergence_rate_initial}
\end{equation}
with probability at least $1-d^{-10}$.
  
\end{customlemma}

\subsubsection{A useful lemma}
\begin{lemma}\label{lm: gradient_Bernstein}
If $r$ is defined in (\ref{radius}) for some constant $C_3>0$ and $\epsilon_0\in(0,\frac{1}{4})$, then with probability at least $1-d^{-10}$, we have\footnote{$\nabla \tilde{f}_n(\bfW)$ is defined as $\frac{1}{n} \sum_{i=1}^n ( \nabla  l(\bfW, \bfx_i, y_i) +\nu_i )$ in algorithm \ref{gd}}
\begin{equation}
\begin{aligned}&\sup_{\bfW\in\mathbb{B}(\bfW^*\bfP,r)}||\nabla \tilde{f}_n(\bfW)-\nabla \tilde{f}(\bfW)||\\
\lesssim&\sqrt{K \sum_{l=1}^L\lambda_l(\|\bfmu_l\|+\|\bfSg_l\|)^2}\sqrt{\frac{d\log{n}}{n}}(1+\xi)
\end{aligned}\end{equation}
, where $\bfP$ is a permutation matrix. 
\end{lemma}

\vspace{5mm}
We next show the proof of Lemma \ref{lemma: convergence}.

\subsubsection{Proof}
\noindent Following the proof of Theorem 2 in \cite{FCL20}, first, we have Taylor's expansion of $f_n(\widehat{\bfW}_n)$

\begin{equation}
    \begin{aligned}
        f_n(\widehat{\bfW}_n)=&f_n(\bfW^*\bfP)+\left\langle \nabla \tilde{f}_n(\bfW^*\bfP),\text{vec}(\widehat{\bfW}_n-\bfW^*\bfP)\right\rangle\\
        +&\frac{1}{2}\text{vec}(\widehat{\bfW}_n-\bfW^*\bfP)\nabla^2 f_n(\bfW')\text{vec}(\widehat{\bfW}_n-\bfW^*\bfP)
    \end{aligned}
\end{equation}

\noindent Here $\bfW'$ is on the straight line connecting $\bfW^*\bfP$ and $\widehat{\bfW}_n$. By the fact that $f_n(\widehat{\bfW}_n)\leq f_n(\bfW^*\bfP)$, we have 
\begin{equation}\label{comparision_lm2}
\begin{aligned}
   &\frac{1}{2}\text{vec}(\widehat{\bfW}_n-\bfW^*\bfP)\nabla^2 f_n(\bfW')\text{vec}(\widehat{\bfW}_n-\bfW^*\bfP)\\
   \leq &\Big|\nabla f_n(\bfW^*\bfP)^\top \text{vec}(\widehat{\bfW}_n-\bfW^*\bfP)\Big|
\end{aligned}
\end{equation}
From Lemma \ref{lm: convex} and Lemma \ref{lm: gradient_Bernstein}, we have
\begin{equation}\label{lower_lm2}
\begin{aligned}
&\frac{4}{K^2}\sum_{l=1}^L\lambda_l\frac{\|\bfSg_l^{-1}\|^{-1}}{\eta\tau^K\kappa^2}\rho(\frac{{\bfW^*}^\top\bfmu_l}{\delta_K(\bfW^*)\|\bfSg_l^{-1}\|^{-\frac{1}{2}}}, \\
&\delta_K(\bfW^*)\|\bfSg_l^{-1}\|^{-\frac{1}{2}})||\widehat{\bfW}_n-\bfW^*\bfP||_F^2\\
\leq&\frac{1}{2}\text{vec}(\widehat{\bfW}_n-\bfW^*\bfP)\nabla^2 f_n(\bfW')\text{vec}(\widehat{\bfW}_n-\bfW^*\bfP)
\end{aligned}
\end{equation}
and

\begin{equation}\label{upper_lm2}
    \begin{aligned}
    &\Big|\nabla \tilde{f}_n(\bfW^*\bfP)^\top \text{vec}(\widehat{\bfW}_n-\bfW^*\bfP)\Big|
    \\
    \leq& \|\nabla \tilde{f}_n(\bfW^*\bfP)\|\cdot\|\widehat{\bfW}_n-\bfW^*\bfP\|_F\\
    \leq &(\| \nabla \tilde{f}_n(\bfW^*\bfP)-\nabla \tilde{f}(\bfW^*\bfP)\|+\|\nabla \tilde{f}(\bfW^*\bfP)\|)\\
    &\cdot\|\widehat{\bfW}_n-\bfW^*\bfP\|_F\\
    \leq& O\Big(\sqrt{K \sum_{l=1}^L\lambda_l(\|\bfmu_l\|+\|\bfSg_l^\frac{1}{2}\|)^2}\sqrt{\frac{d\log{n}}{n}}(1+\xi)\Big)\\
    &||\widehat{\bfW}_n-\bfW^*\bfP||_F
    \end{aligned}
\end{equation}

\noindent The second to last step of (\ref{upper_lm2}) comes from the triangle inequality, and the last step follows from the fact $\nabla \bar{f}(\bfW^*\bfP)=0$. Combining (\ref{comparision_lm2}), (\ref{lower_lm2}) and (\ref{upper_lm2}), we have

\begin{equation}
    \begin{aligned}
    &||\widehat{\bfW}_n-\bfW^*\bfP||_F\\
    \lesssim &\frac{K^{\frac{5}{2}}\sqrt{ \sum_{l=1}^L\lambda_l(\|\bfmu_l\|+\|\bfSg_l^\frac{1}{2}\|)^2}(1+\xi)\cdot\sqrt{d\log{n}/n}}{\sum_{l=1}^L\lambda_l\frac{\|\bfSg_l^{-1}\|^{-1}}{\eta\tau^K\kappa^2}\rho(\frac{{\bfW^*}^\top\bfmu_l}{\delta_K(\bfW^*)\|\bfSg_l^{-1}\|^{-\frac{1}{2}}}, \delta_K(\bfW^*)\|\bfSg_l^{-1}\|^{-\frac{1}{2}})}
    \end{aligned}
\end{equation}

\noindent Therefore, we have concluded that there indeed exists a critical point $\widehat{\bfW}$ in $\mathbb{B}(\bfW^*\bfP,r)$. Then we show the linear convergence of Algorithm \ref{gd} as below. By the update rule, we have

\begin{equation}
\begin{aligned}
    &\bfW_{t+1}-\widehat{\bfW}_n\\
    =&\bfW_t-\eta_0(\nabla f_n(\bfW_t)+\frac{1}{n}\sum_{i=1}^n\nu_i)-(\widehat{\bfW}_n-\eta_0\nabla f_n(\widehat{\bfW}_n))\\
    =&\Big(\bfI-\eta_0\int_{0}^1\nabla^2 f_n(\bfW(\gamma))\Big)(\bfW_t-\widehat{\bfW}_n)-\frac{\eta_0}{n}\sum_{i=1}^n \nu_i
\end{aligned}
\end{equation}
where $\bfW(\gamma)=\gamma \widehat{\bfW}_n+(1-\gamma)\bfW_t$ for $\gamma\in(0,1)$. Since $\bfW(\gamma)\in\mathbb{B}(\bfW^*\bfP,r)$, by Lemma \ref{lemma: convexity}, we have
\begin{equation}\label{Hminmax}
    H_{\min}\cdot\bfI\preceq \nabla^2f_n(\bfW(\gamma))\leq H_{\max}\cdot\bfI
\end{equation}
where $H_{\min}=\Omega\Big(\frac{1}{K^2}\sum_{l=1}^L\lambda_l\frac{\|\bfSg_l^{-1}\|^{-1}}{\eta\tau^K\kappa^2}\rho(\frac{{\bfW^*}^\top\bfmu_l}{\delta_K(\bfW^*)\|\bfSg_l^{-1}\|^{-\frac{1}{2}}},\\ \delta_K(\bfW^*)\|\bfSg_l^{-1}\|^{-\frac{1}{2}})\Big)$, $H_{\max}= \sum_{l=1}^L\lambda_l(\|\bfmu_l\|+\|\bfSg_l\|)^2$. Therefore,

\begin{equation}
\begin{aligned}
    &||\bfW_{t+1}-\widehat{\bfW}_n||_F\\
    =&||\bfI-\eta_0\int_{0}^1\nabla^2 f_n(\bfW(\gamma))||\cdot||\bfW_t-\widehat{\bfW}_n||_F+\|\frac{\eta_0}{n}\sum_{i=1}^n \nu_i\|_F\\
    \leq &(1-\eta_0 H_{\min})||\bfW_t-\widehat{\bfW}_n||_F+\|\frac{\eta_0}{n}\sum_{i=1}^n \nu_i\|_F
\end{aligned}
\end{equation}
By setting $\eta_0=\frac{1}{H_{\max}}=O\Big(\frac{1}{\sum_{l=1}^L\lambda_l(\|\bfmu_l\|+\|\bfSg_l\|)^2}\Big)$, we obtain  
\begin{equation}||\widehat{\bfW}_{t+1}-\widehat{\bfW}_n||_F\leq(1-\frac{H_{\min}}{H_{\max}})||\bfW_t-\widehat{\bfW}_n||_F+\frac{\eta_0}{n}\sum_{i=1}^n \|\nu_i\|_F\label{lm2_final}
\end{equation}
Therefore, Algorithm \ref{gd} converges linearly to the local minimizer with an extra statistical error.\\
By Hoeffding's inequality in \cite{V10}, we have
\begin{equation}
\begin{aligned}
    &\mathbb{P}\Big(\frac{1}{n}\sum_{i=1}^n \|\nu_i\|_F\geq \sqrt{\frac{dK\log n}{n}}\xi\Big)\lesssim \exp(-\frac{\xi^2 dK\log n}{dK\xi^2})\\
    \lesssim &d^{-10}
    \end{aligned}
\end{equation}
Therefore, with probability $1-d^{-10}$ we can derive
\begin{equation}
\begin{aligned}
    &||\widehat{\bfW}_{t}-\widehat{\bfW}_n||_F\\
    \leq&(1-\frac{H_{\min}}{H_{\max}})^t||\bfW_0-\widehat{\bfW}_n||_F+\frac{H_{\max}\eta_0}{H_{\min}}\sqrt{\frac{dK \log n}{n}}\xi
\end{aligned}
\end{equation}

\subsection{Proof of Lemma \ref{lemma: tensor bound}}\label{subsec: useful lemms lm3}
We first restate the formal version of Lemma \ref{lemma: tensor bound_informal} in the following.
\begin{customlemma}{3}\label{lemma: tensor bound}(Tensor initialization)
\normalsize For classification model, with $D_6(\Psi)$ defined in Definition \ref{def: D}, we have that if the sample size 
\begin{equation}
n\geq \kappa^8K^4\tau^{12} D_6(\Psi)\cdot d\log^2{d},\label{sp_lm3}
\end{equation}
then the output $\bfW_0\in\mathbb{R}^{d\times K}$ satisfies
\begin{equation}||\bfW_0-\bfW^*\bfP^*||\lesssim \kappa^6 K^3\cdot \tau^6 \sqrt{D_6(\Psi)}\sqrt{\frac{d\log{n}}{n}}||\bfW^*||\end{equation}
with probability at least $1-n^{-\Omega(\delta_1^4)}$ for a specific permutation matrix $\bfP^*\in\mathbb{R}^{K\times K}$.
\end{customlemma}

\subsubsection{Useful lemmas}
Lemmas \ref{lm: p2}, \ref{lm: R3}, \ref{lm: M1}, \ref{lm: bound on subspace estimation}, and \ref{lm: solution to 1st order moment} are needed to prove Lemma \ref{lemma: tensor bound}.

\begin{lemma}\label{lm: p2}\normalsize
 Let $\bfQ_2$ and $\bfQ_3$ follow Definition \ref{def: M}. Let $S$ be a set of i.i.d. samples generated from the mixed Gaussian distribution $\sum_{l=1}^L\lambda_l\mathcal{N}(\bfmu_l,\bfSg_l)$. Let $\widehat{\bfQ}_2$, $\widehat{\bfQ}_3$ be the empirical version of $\bfQ_2$, $\bfQ_3$ using data set $S$, respectively. Then with a probability at least $1-2n^{-\Omega(\delta_1(\bfW^*)^4 d)}$, we have
\begin{equation}||\bfQ_2-\widehat{\bfQ}_2||\lesssim \sqrt{\frac{d\log{n}}{n}}\cdot \delta_1(\bfW^*)^2\cdot\tau^6\sqrt{D_2(\Psi)D_4(\Psi)}\end{equation}
if the mixed Gaussian distribution is not symmetric. We also have
\begin{equation}
\begin{aligned}&||\bfQ_3(\bfI_d,\bfI_d, \boldsymbol{\alpha})-\widehat{\bfQ}_3(\bfI_d,\bfI_d,\boldsymbol{\alpha})||\\\lesssim &\sqrt{\frac{d\log{n}}{n}}\cdot \delta_1(\bfW^*)^2\cdot\tau^6\sqrt{D_2(\Psi)D_4(\Psi)}
\end{aligned}\end{equation}
for any arbitrary vector $\boldsymbol{\alpha}\in\mathbb{R}^d$, if the mixed Gaussian distribution is symmetric.
\end{lemma}

\begin{lemma}\label{lm: R3} \normalsize Let $\bfU\in\mathbb{E}^{d\times K}$ be the orthogonal column span of $\bfW^*$. Let $\boldsymbol{\alpha}$ be a fixed unit vector and $\widehat{\bfU}\in\mathbb{R}^{d\times K}$ denote an orthogonal matrix satisfying $||\bfU\bfU^\top-\widehat{\bfU}\widehat{\bfU}^\top||\leq \frac{1}{4}$. Define $\bfR_3=\bfQ_3(\widehat{\bfU},\widehat{\bfU},\widehat{\bfU})$, where $\bfQ_3$ is defined in Definition \ref{def: M}. Let $\widehat{\bfR}_3$ be the empirical version of $\bfR_3$ using data set $S$, where each sample of $S$ is i.i.d. sampled from the mixed Gaussian distribution   $\sum_{l=1}^L\lambda_l\mathcal{N}(\bfmu_l,\bfSg_l)$. Then with a probability at least $1-n^{-\Omega(\delta^4(\bfW^*))}$, we have
\begin{equation}||\widehat{\bfR}_3-\bfR_3||\lesssim \delta_1(\bfW^*)^2\cdot\big(\tau^{6}\sqrt{D_6(\Psi)}\big)\cdot\sqrt{\frac{\log{n}}{n}}\end{equation}
\end{lemma}
\begin{lemma}\label{lm: M1} \normalsize 
Let $\widehat{\bfQ}_1$ be the empirical version of $\bfQ_1$ using dataset $S$. Then with a probability at least $1-2n^{-\Omega(d)}$, we have
\begin{equation}||\widehat{\bfQ}_1-\bfQ_1||\lesssim \big(\tau^2\sqrt{D_2(\Psi)}\big)\cdot\sqrt{\frac{d\log{n}}{n}}\end{equation}
\end{lemma}

\begin{lemma}\label{lm: bound on subspace estimation}
\normalsize (\cite{ZSJB17}, Lemma E.6) Let $\bfQ_2 $, $\bfQ_3$ be defined in Definition \ref{def: M} and $\widehat{\bfQ}_2$, $\widehat{\bfQ}_3$ be their empirical version, respectively. Let $\bfU\in\mathbb{R}^{d\times K}$ be the column span of $\bfW^*$. Assume $||\bfQ_2-\widehat{\bfQ}_2||\leq\frac{\delta_K(\bfQ_2)}{10}$ for non-symmetric distribution cases and $||\bfQ_3(\bfI_d,\bfI_d,\boldsymbol{\alpha})-\widehat{\bfQ}_3(\bfI_d,\bfI_d,\boldsymbol{\alpha})||\leq\frac{\delta_K(\bfQ_3(\bfI_d,\bfI_d,\boldsymbol{\alpha}))}{10}$ for symmetric distribution cases and any arbitrary vector $\boldsymbol{\alpha}\in\mathbb{R}^d$. Then after $T=O(\log(\frac{1}{\epsilon}))$ iterations, the output of the Tensor Initialization Method \ref{TensorInitialization}, $\widehat{\bfU}$,  will satisfy \begin{equation}||\widehat{\bfU}\widehat{\bfU}^\top-\bfU\bfU^\top||\lesssim\frac{||\widehat{\bfQ}_2-\bfQ_2||}{\delta_K(\bfQ_2)}+\epsilon, \end{equation}
which implies
\begin{equation}||(\bfI-\widehat{\bfU}\widehat{\bfU}^\top)\bfw_i^*||\lesssim(\frac{||\bfQ_2-\widehat{\bfQ}_2||}{\delta_K(\bfQ_2)}+\epsilon)||\bfw_i^*||\end{equation}
if the mixed Gaussian distribution is not symmetric. Similarly, we have
\begin{equation}||\widehat{\bfU}\widehat{\bfU}^\top-\bfU\bfU^\top||\lesssim\frac{||\widehat{\bfQ}_3(\bfI_d,\bfI_d,\boldsymbol{\alpha})-\bfQ_3(\bfI_d,\bfI_d,\boldsymbol{\alpha})||}{\delta_K(\bfQ_3(\bfI_d,\bfI_d,\boldsymbol{\alpha}))}+\epsilon, \end{equation}
which implies
\begin{equation}
\begin{aligned}&||(\bfI-\widehat{\bfU}\widehat{\bfU}^\top)\bfw_i^*||\\
\lesssim&(\frac{||\bfQ_3(\bfI_d,\bfI_d,\boldsymbol{\alpha})-\widehat{\bfQ}_3(\bfI_d,\bfI_d,\boldsymbol{\alpha})||}{\delta_K(\bfQ_3(\bfI_d,\bfI_d,\boldsymbol{\alpha}))}+\epsilon)||\bfw_i^*||
\end{aligned}\end{equation}
if the mixed Gaussian distribution is symmetric.
\end{lemma}
\begin{lemma}\label{lm: solution to 1st order moment}
\normalsize (\cite{ZSJB17}, Lemma E.13) Let $\bfU\in\mathbb{R}^{d\times K}$ be the orthogonal column span of $\bfW^*$. Let $\widehat{\bfU}\in\mathbb{R}^{d\times K}$ be an orthogonal matrix such that $||\bfU\bfU^\top-\widehat{\bfU}\widehat{\bfU}^\top||\lesssim \gamma_1\lesssim \frac{1}{\kappa^2\sqrt{K}}$. For each $i\in[K]$, let $\widehat{\bfv_i}$ denote the vector satisfying $||\widehat{\bfv_i}-\widehat{\bfU}^\top \bar{\bfw_i}^*||\leq \gamma_2\lesssim\frac{1}{\kappa^2\sqrt{K}}$. Let $\bfQ_1$ be defined in Lemma \ref{lm: M1} and $\widehat{\bfQ}_1$ be its empirical version. If $||\bfQ_1-\widehat{\bfQ}_1||\leq \gamma_3||\bfQ_1||\lesssim\frac{1}{4}||\bfQ_1||$, then we have\\
\begin{equation}\Big|||\bfw_i^*||-\widehat{\alpha}_i\Big|\leq (\kappa^4 K^{\frac{3}{2}}(\gamma_1+\gamma_2)+\kappa^2 K^\frac{1}{2}\gamma_3)||\bfw_i^*||\end{equation}
\end{lemma}

\vspace{5mm}
We next show the proof of Lemma \ref{lemma: tensor bound}.

\subsubsection{Proof: }\label{subsec: pf lm3}
By the triangle inequality, we have
\begin{equation}
\begin{aligned}
    &||\bfw_j^*-\widehat{\alpha}_j\widehat{\bfU}\widehat{\bfv}_j||\\
    =&\Big|\Big|\bfw_j^*-||\bfw_j^*||\widehat{\bfU}\widehat{\bfv}_j+||\bfw_j^*||\widehat{\bfU}\widehat{\bfv}_j-\widehat{\alpha}_j\widehat{\bfU}\widehat{\bfv}_j\Big|\Big|\\
    \leq &\Big|\Big|\bfw_j^*-||\bfw_j^*||\widehat{\bfU}\widehat{\bfv}_j\Big|\Big|+\Big|\Big|||\bfw_j^*||\widehat{\bfU}\widehat{\bfv}_j-\widehat{\alpha}_j\widehat{\bfU}\widehat{\bfv}_j\Big|\Big|\\
    \leq &||\bfw_j^*||\Big|\Big|\bar{\bfw_j}^*-\widehat{\bfU}\widehat{\bfv}_j\Big|\Big|+\Big|\Big|||\bfw_j^*||-\widehat{\alpha}_j\Big|\Big|||\widehat{\bfU}\widehat{\bfv}_j||\\
    \leq &||\bfw_j^*||\Big|\Big|\bar{\bfw_j}^*-\widehat{\bfU}\widehat{\bfU}^\top \bar{\bfw}_j^*+\widehat{\bfU}\widehat{\bfU}^\top\bar{\bfw_j}^*-\widehat{\bfU}\widehat{\bfv}_j\Big|\Big|\\
    &+\Big|\Big|||\bfw_j^*||-\widehat{\alpha}_j\Big|\Big|||\widehat{\bfU}\widehat{\bfv}_j||\\
    \leq &\delta_1(\bfW^*)\Big(\Big|\Big|\bar{\bfw_j}^*-\widehat{\bfU}\widehat{\bfU}^\top \bar{\bfw_j}^*\Big|\Big|+\Big|\Big|\widehat{\bfU}^\top\bar{\bfw_j}^*-\widehat{\bfv}_j\Big|\Big|\Big)\\
    &+\Big|\Big|||\bfw_j^*||-\widehat{\alpha}_j\Big|\Big|\\
\end{aligned}
\end{equation}
%\tcr{The second inequality holds from .... The last inequality holds from... }
From Lemma \ref{lm: p2}, Lemma \ref{lm: bound on subspace estimation},  $\delta_K(\bfQ_2)\lesssim \delta_K^2(\bfW^*)$ and $\delta_K(\bfQ_3(\bfI_d,\bfI_d,\boldsymbol{\alpha}))\lesssim \delta_K^2(\bfW^*)$ for any arbitrary vector $\boldsymbol{\alpha}\in\mathbb{R}^d$, we have
\begin{equation}
\begin{aligned}
&\Big|\Big|\bar{\bfw_j}^*-\widehat{\bfU}\widehat{\bfU}^\top \bar{\bfw_j}^*\Big|\Big|\\
\lesssim &\frac{||\bfQ_2-\widehat{\bfQ}_2||}{\delta_K(\bfQ_2)}\lesssim \sqrt{\frac{d\log{n}}{n}}\cdot \frac{\delta_1(\bfW^*)^2}{\delta_K(\bfW^*)^2}\cdot\tau^{6}\sqrt{D_2(\Psi)D_4(\Psi)}\\
= &\sqrt{\frac{d\log{n}}{n}}\cdot \kappa^2\cdot\tau^{6}\sqrt{D_2(\Psi)D_4(\Psi)}\label{part1lm3}
\end{aligned}
\end{equation}
if the mixed Gaussian distribution is not symmetric, and
\begin{equation}
\begin{aligned}
&\Big|\Big|\bar{\bfw_j}^*-\widehat{\bfU}\widehat{\bfU}^\top \bar{\bfw_j}^*\Big|\Big|\lesssim \frac{||\bfQ_3(\bfI_d,\bfI_d,\boldsymbol{\alpha})-\widehat{\bfQ}_3(\bfI_d,\bfI_d,\boldsymbol{\alpha})||}{\delta_K(\bfQ_3(\bfI_d,\bfI_d,\boldsymbol{\alpha}))}\\
= &\sqrt{\frac{d\log{n}}{n}}\cdot \kappa^2\cdot\tau^{6}\sqrt{D_2(\Psi)D_4(\Psi)}\label{part1_lm3}
\end{aligned}
\end{equation}
if the mixed Gaussian distribution is symmetric. Moreover, we have
\begin{equation}
\begin{aligned}
&\Big|\Big|\widehat{\bfU}^\top\bar{\bfw_j}^*-\widehat{\bfv}_j\Big|\Big|\\
\leq &\frac{K^{\frac{3}{2}}}{\delta_K^2(\bfW^*)}||\bfR_3-\widehat{\bfR}_3||\lesssim\kappa^2\cdot\big(\tau^6\sqrt{D_6(\Psi)}\big)\cdot\sqrt{\frac{K^3\log{n}}{n}}\label{part2lm3}
\end{aligned}
\end{equation}
in which the first step is by Theorem 3 in \cite{KCL15}, and the second step is by Lemma \ref{lm: R3}. 
By Lemma \ref{lm: solution to 1st order moment}, we have 
\begin{equation}\Big|\Big|||\bfw_j^*||-\widehat{\alpha}_j\Big|\Big|\leq (\kappa^4K^{\frac{3}{2}}(\gamma_1+\gamma_2)+\kappa^2K^\frac{1}{2}\gamma_3)||\bfW^*||\end{equation}
Therefore, taking the union bound of failure probabilities in Lemmas \ref{lm: p2}, \ref{lm: R3}, and \ref{lm: M1}, %and by $D_2(\Psi)D_4(\Psi)\leq D_6(\Psi)$ from Property \ref{prop: D_ineq}, 
we have that if the sample size $n\geq \kappa^8K^4\tau^{12}D_6(\Psi)\cdot d\log^2{d}$, then the output $\bfW_0\in\mathbb{R}^{d\times K}$ satisfies
\begin{equation}||\bfW_0-\bfW^*||\lesssim \kappa^6 K^3\cdot \tau^6 \sqrt{D_6(\Psi)}\sqrt{\frac{d\log{n}}{n}}||\bfW^*||\end{equation}
with probability at least $1-n^{-\Omega(\delta_1^4(\bfW^*))}$

\subsection{Extension to Multi-Classification}

We only show the analysis of binary classification in the main body of the paper due to the simplicity of presentation and highlight our major conclusions on the group imbalance. We briefly introduce how to extend our analysis on binary classification to multi-classification in this section. The main idea is to define the label as a multi-dimensional vector and apply the analysis for the binary classification case multiple times. Specifically, let $C$ be the number of classes, where $C=2^c$ for a positive integer $c$. The label $\bfy_i$ is a $c$-dimensional vector, and its $j$th entry $y_{i,j}\in\{0,1\}$ for $j\in[c]$ and $i\in[n]$. Such a formulation for the multi-classification problem can be found in \cite{SXK18, ZLOY17}. Then, following the binary setup, data $\bfx_i,y_i$ satisfies 
\begin{equation}
    \mathbb{P}(y_{i,j}=1|\bfx_i)=H_j(\bfW^*,\bfx_i),
\end{equation}
for some unknown ground-truth neural network with unknown weights $\bfW^*$, where 
$H_j(\bfW^*,\bfx_i)$ is the $j$-th entry of $\bfH(\bfW^*,\bfx_i)\in\mathbb{R}^c$ with the parameter $\bfW_j^*\in\mathbb{R}^{d\times K}$. 

The training process is to minimize the empirical risk function with a cross-entropy loss
\begin{equation}
\begin{aligned}
    &\frac{1}{n}\sum_{i=1}^n \sum_{j=1}^c -y_{i,j}\log(H_j(\bfW,\bfx_i))\\
    &-(1-y_{i,j})\log(1-H_j(\bfW,\bfx_i))\\
    :=&\sum_{j=1}^c f_n^{(j)}(\bfW).
\end{aligned}
\end{equation}
Note that $f_n^{(j)}(\bfW)$ has exactly the form as (\ref{eqn:problem}) in our paper for the binary case. Therefore, we can apply the existing theoretical results for 
$f_n^{(j)}(\bfW)$ with all $j\in[c]$, and summing up all the bounds yields the theoretical results for the multi-class case.

We implement experiments on the CelebA dataset for 4-classification. The only change is that we use the combinations of two attributes, “blonde hair” and “pale skin” to generate four classes of data. All other settings are the same. The results are the following.

\begin{figure}[t]
    \centering

        \begin{minipage}{0.4\textwidth}
        \centering
        \includegraphics[width=0.9\textwidth,height=0.55\textwidth]{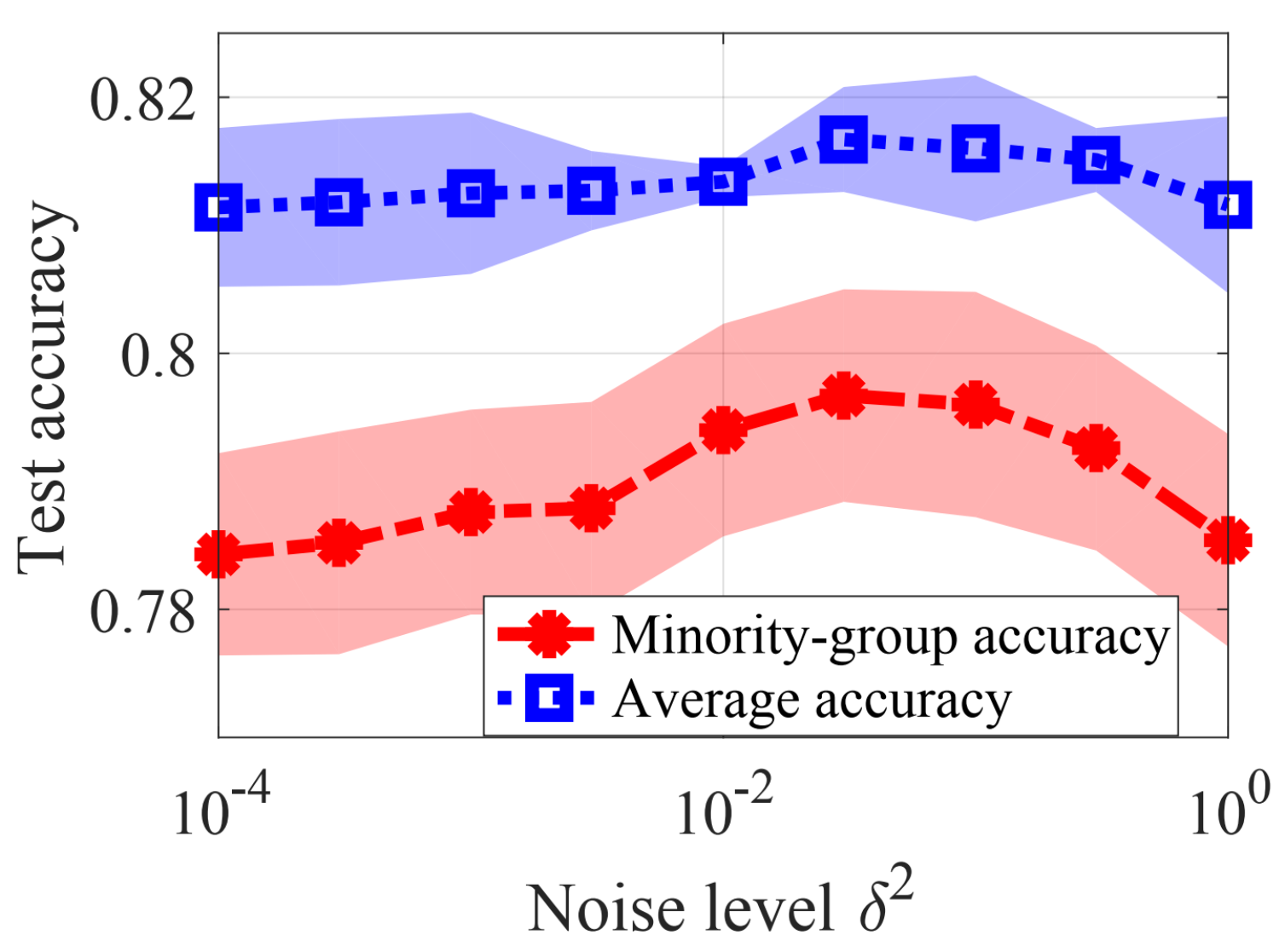}
        \end{minipage}
    
    \vspace{-2mm}
    \caption{Test accuracy against the augmented noise level for 4-classification.}
    \label{figure: 4cla-celebA_noise}
\end{figure}

\begin{figure}[htbp]
    \centering
    \subfigure[]{
        \begin{minipage}{0.4\textwidth}
        \centering
        \includegraphics[width=0.8\textwidth,height=0.55\textwidth]{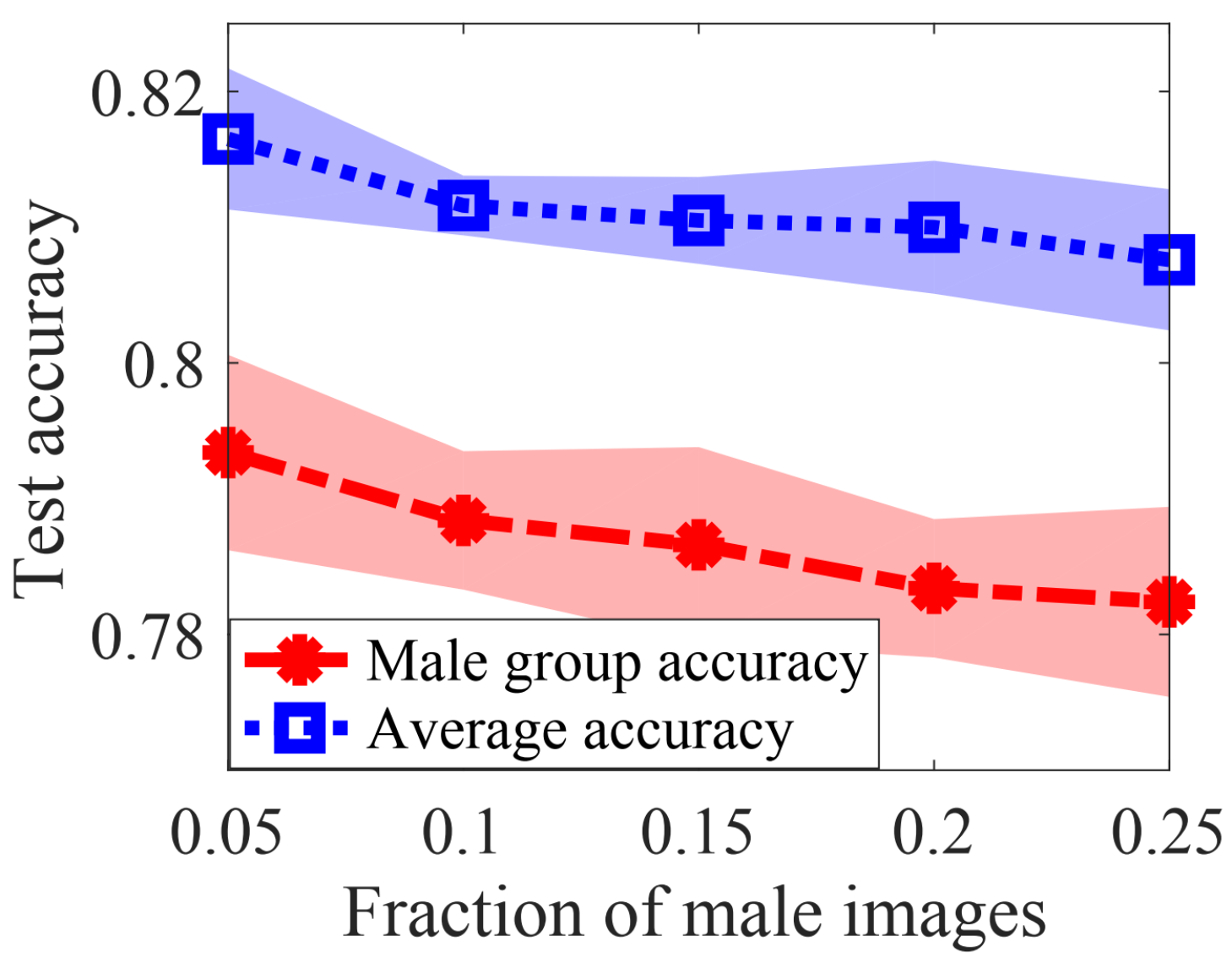}
        \end{minipage}
    }
    ~
    \subfigure[]{
        \begin{minipage}{0.4\textwidth}
        \centering
        \includegraphics[width=0.8\textwidth,height=0.5\textwidth]{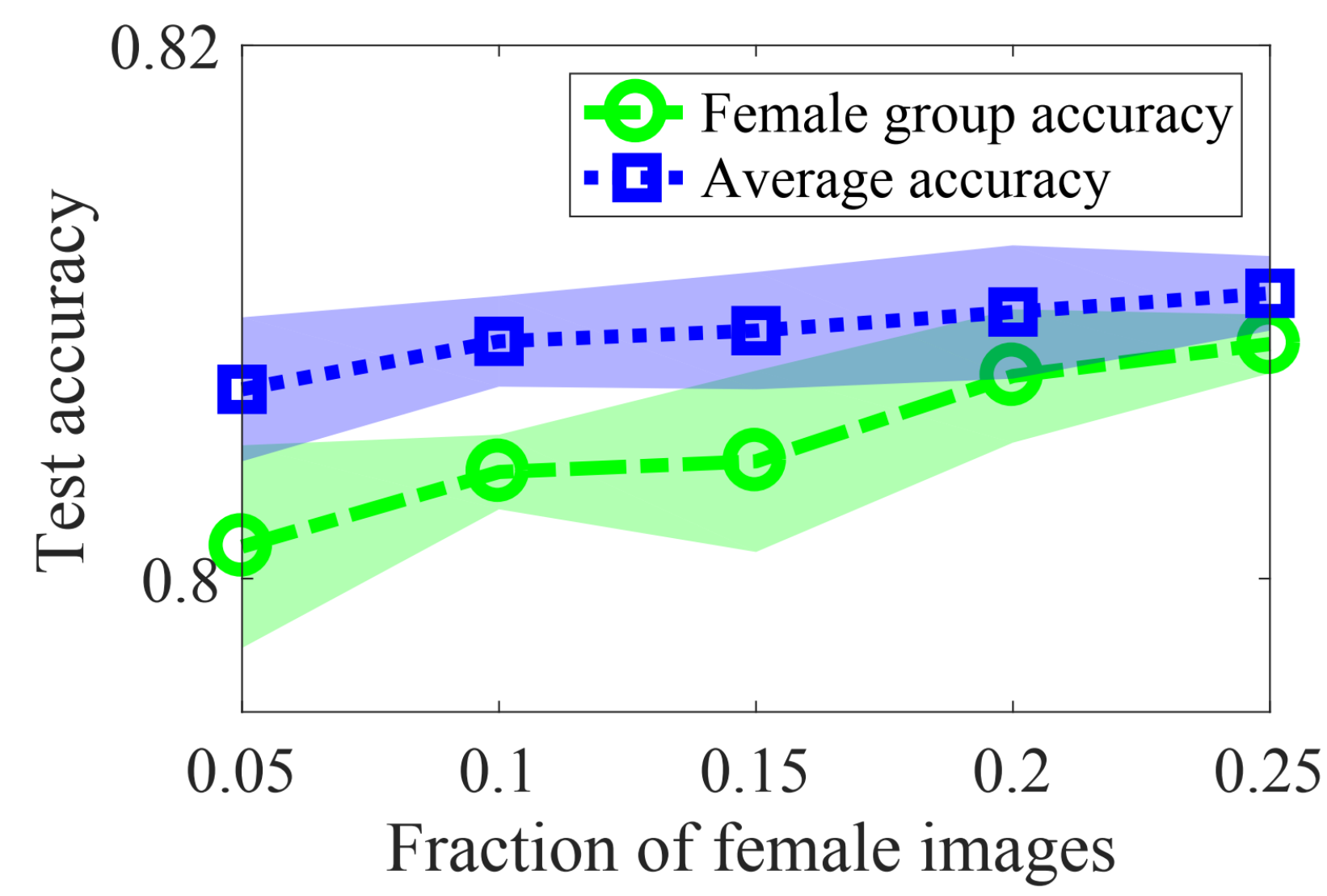}
        \end{minipage}
    }
       \vspace{-2mm}
    \caption{The test accuracy on CelebA dataset has opposite trends when the minority group fraction increases for 4-classification.  (a) Male group is the minority (b) Female group is the minority}\label{figure: 4-cla-male-female}
\end{figure}

One can observe from Figure \ref{figure: 4cla-celebA_noise} that when the noise level $\delta^2$ increases, i.e., when the co-variance of the minority group increases, both the minority-group and average test accuracy increase first and then decrease, coinciding with our insight (P3). In Figure \ref{figure: 4-cla-male-female} (a) and (b), we can see opposite trends if we increase the fraction of the minority group in the training data, with the male being the minority or the female being the minority. Figures \ref{figure: 4cla-celebA_noise} and \ref{figure: 4-cla-male-female} are consistent with our findings in Figures \ref{figure: celebA_noise} and \ref{figure: celebA_group}, respectively.

\iffalse
\begin{table*}[t]
    \centering
    %\vspace{2mm}
    \caption{Test accuracy against the augmented noise level for 4-classification}
    \renewcommand\arraystretch{2}
    \begin{tabular}{p{1.8cm} p{1.2cm} p{1.2cm} p{1.2cm} p{1.2cm} p{1.2cm} p{1.2cm} p{1.2cm} p{1.2cm} p{1.2cm}}
    \hline
 {Noise level $\delta^2$} & $10^{-4}$ & $3\times 10^{-4}$ & $10^{-3}$ & $3\times 10^{-3}$ & $10^{-2}$  & $3\times 10^{-2}$ &  $10^{-1}$ & $3\times 10^{-1}$ & $10^0$ \\
    \hline
 {Minority-group accuracy $(\%)$} & $78.43_{\pm0.99}$ & $78.52_{\pm 1.07}$ & $78.76_{\pm 1.10}$ & $78.79_{\pm 1.13}$ & $79.40_{\pm 1.33}$  & {$\mathbf{79.67}_{\pm1.13}$} &  $79.60_{\pm 1.18}$ & $79.26_{\pm 0.80}$ & $78.54_{\pm 0.83}$ 
    \\
    \hline
     {Average accuracy $(\%)$} & $81.14_{\pm 0.62}$ & $81.18_{\pm0.85}$ & $81.25_{\pm 0.83}$ & $81.27_{\pm 0.31}$ & $81.34_{\pm 0.12}$  & {$\mathbf{81.67}_{\pm0.41}$} &  $81.60_{\pm 0.57}$ & $81.51_{\pm0.25}$ & $81.16_{\pm 0.69}$ 
    \\
    \hline

    \end{tabular}
    \label{tbl:results}
    \vspace{-1mm}
\end{table*}
\fi

\subsection{Discussion about Gaussian Mixture Model (GMM)}\label{subsec: gmm}

%\mw{To show 1.67 is close to 1.8, we need to add another case where the data does not fit GMM well and the corresponding score is very far away from 1.8. }a standard gaussian data can fit GMM well (with other components be 0.01), but GMM data cannot fit standard gaussian well. 

The GMM distribution intuitively means that each data comes from a certain group, which is represented by a certain Gaussian component with mean $\bfmu_l$ and co-variance $\bfSg_l$, $l\in[L]$. The fraction $\lambda_l$ stands for the fraction of group $l\in[L]$. This formulation is motivated by existing works \cite{SRKL20, DYMG23}, which are related to group imbalance in the case of convolutional neural networks. One can see that each data feature follows GMM by Eqn (4) of \cite{SRKL20}. In our setup, we define the data following the GMM for fully connected neural networks, where labels are determined by the mixture of Gaussian input and the ground-truth model. %Our data model is more general in that (a) it can have more components, and (b) the labels do not necessarily depend on the mean of Gaussian components. 

%\mw{Ours is more general than [3]? (a) (b) both do not hold for [3]? Not clear what they mean. }I can remove the last sentence. It is not a very standard comparison. The cited works are for spurious correlation instead of group imbalance. what about now?
We also conduct an experiment on CelebA to show some practical datasets satisfy the GMM model. We select data with two attributes, male and female. We extract features before the fully connected layer of the ResNet 9 model and fit the features to a two-component GMM using the EM algorithm \cite{RW84}. The goodness of fit is measured by the average log-likelihood score as in \cite{ZW12}. We compute the average log-likelihood score of the CelebA dataset as 1.63 bits/dimension. To see that 1.63 bits/dimension reflects a good fitting, we generate synthetic data following the estimated GMM by CelebA and then compute the log-likelihood score of fitting the synthetic data to a two-component GMM. The resulting score is 1.80 bits/dimension for the synthetic two-component GMM data. Therefore, we can see that the quality of fitting CelebA is almost as good as fitting synthetic data generated by a GMM, which indicates that a two-component GMM is a good fitting for the studied practical dataset generated by CelebA.

Since many existing theoretical works \cite{ZSJB17, FCL20, ZWLC20, ZWLC21_sparse, ZWLC21_self} consider the data as standard Gaussian, we also compute the score if we use a single Gaussian to fit the data. The resulting average log-likelihood score is 1.08 bits/dimension, which is evidently smaller than the two-component GMM considered in our manuscript. This shows our GMM can better describe the real data.

Moreover, our GMM assumption goes beyond the state-of-the-art assumption of the standard Gaussian for loss landscape analysis for one-hidden-layer neural networks with convergence guarantees \cite{ZSJB17, ZSD17, ZYWG19, FCL20, ZWLC21_self}.  When generalizing from the standard Gaussian to GMM, we make new technical contributions to analyzing the more complicated and challenging landscape of the risk function because of a mixture of non-zero mean and non-unit standard deviation Gaussians. We characterize the impact of the parameters of the GMM model on the learning convergence and generalization performance. In contrast, other existing theoretical works \cite{ADHL19, ALS19, ALL19, ZCZG20} that consider other input distributions that are more general than the standard Gaussian model do not explicitly quantify the impact of the distribution parameters on the loss landscape and generalization performance.

%research as existing research restricts the input to be standard Gaussian, which makes our theoretical results of additional technical interest. As far as we know, assuming standard Gaussian input \cite{ZSJB17, ZSD17, ZYWG19, FCL20, ZWLC21_self} is the state-of-art practice for loss landscape analysis for one-hidden-layer neural networks with convergence guarantees. Other existing theoretical works \cite{ADHL19, ALS19, ALL19, ZCZG20} that consider a broader input distribution cannot analyze loss landscape and generalization based on the distribution parameters. However, our GMM model generalizes from the standard Gaussian currently used in all existing papers on model estimation and characterizes the impact of the distribution parameters. When generalizing from the standard Gaussian to GMM, we make technical contributions to analyzing the more complicated and challenging landscape of the risk function because of a mixture of non-zero mean and non-standard deviation Gaussians. We characterize the strong local convexity with regard to the GMM parameters, derive concentration bound, and develop tensor initialization using Gaussian mixture variables.}

\subsection{Discussion about $\sigma_{\min}$ and $\tau$}\label{subsec: sigma_min}
In this section, we show that the assumption that $\sigma_{\min}$ is not very close to zero, or equivalently, $\tau=\Theta(1)$, is mild. Even when the real data have singular values very close to zero, they can be approximated by low-rank data without hurting the performance by only keeping a few significant singular values and setting the small ones to zero. %The low-rank approximation does not affect the performance much. 
Thus, every practical dataset can be approximated by a dataset with $\tau=\Theta(1)$ while maintaining the same performance. 
%is usually mild by the low-rank nature in the feature space of real data. In other words, if we use the features with the highest ranks for learning, this assumption becomes valid in practice. 
We verify this by an experiment of binary classification on CelebA \cite{LLWT15}. After training with a ResNet-9, the output feature of each testing image is $256$-dimensional. One can find that the singular value of the covariance matrix of features is close to $0$ except for the top singular values. The feature matrix reconstructed with top singular values can achieve comparable testing accuracy as using all singular values, as shown in Table \ref{tab: lowrank}. One can observe that the feature matrix reconstruction with top-$5$ singular values, which is $2\%$ of all the singular vectors, leads to a test accuracy already close to that using all singular vectors, and the performance gap is smaller than $4.5\%$. We can  compute that $\tau=4.6155=\Theta(1)$ for the feature matrix reconstructed by top $5$ singular values. 

 % if we pick the top $5$ singular values of the covariance of features.}
\begin{table}[h!]
  \begin{center}
        \caption{Testing accuracy with a reconstructed feature matrix using different amounts of singular values (s.v.) }   \label{tab: lowrank}
    \begin{tabular}{l|c|c|c} % <-- Alignments: 1st column left, 2nd middle and 3rd right, with vertical lines in between
 \hline
 \small \small{Reconstruct with } & \small{top $5$ s.v.} & \small{top $25$ s.v.} & \small{all $256$ s.v.}\\
 \hline
 \small{Accuracy} & \small{$84.00\%$}  &
 \small{$85.00\%$} & \small{$88.50\%$}\\
 \hline
    \end{tabular}
  \end{center}
\end{table}


%\bibliographystyle{IEEEtran}
%\bibliography{ref.bib, ref_Hongkang.bib}

\begin{IEEEbiography} [{\includegraphics[width=1.05in,height=1.25in,clip,keepaspectratio]{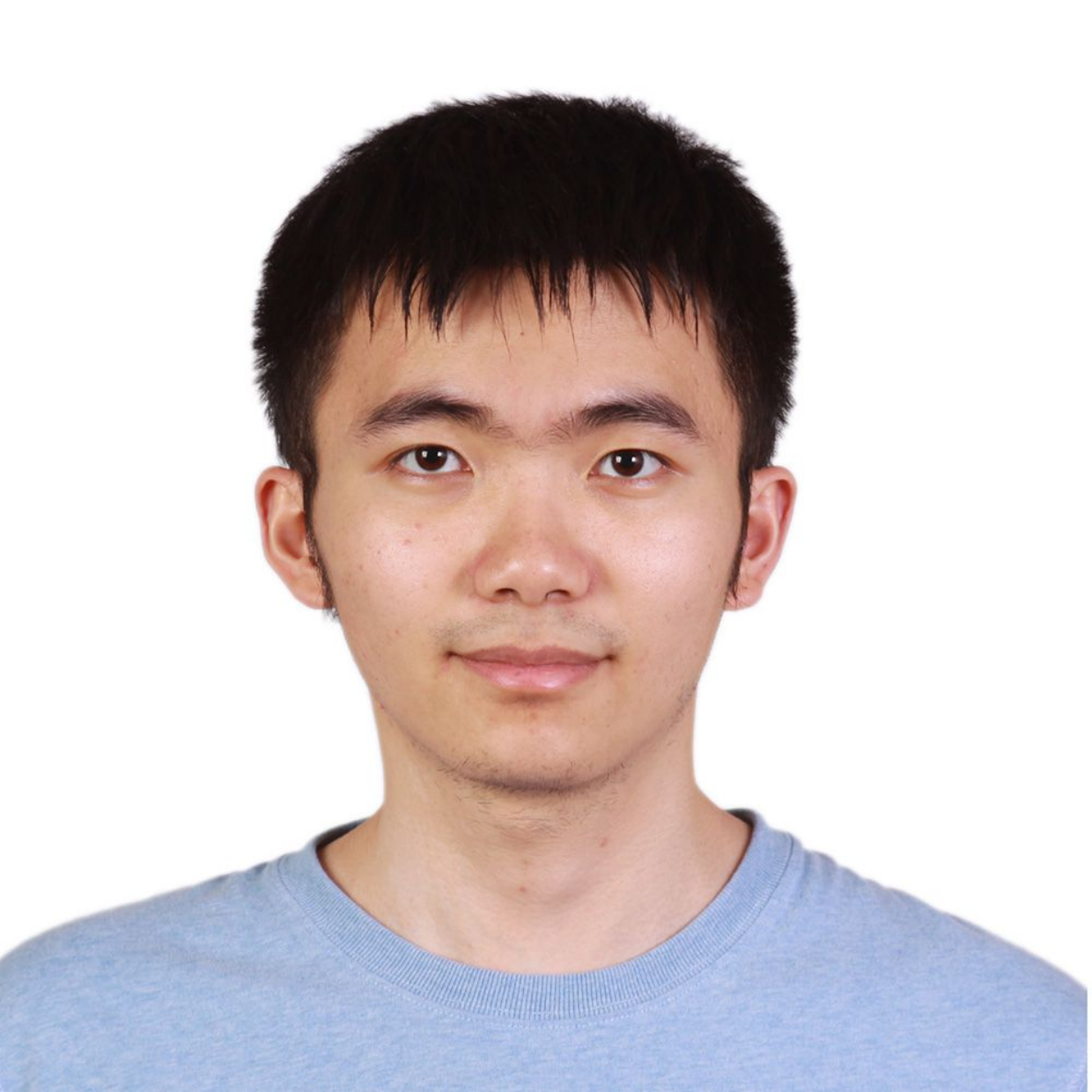}}]    {Hongkang Li} (Student Member, IEEE) received the B.E. degree in the Department of Electronic Engineering and Information Science from the University of Science and Technology of China, Hefei, China, in 2019. He is currently a Ph.D. student in the Department of  Electrical, Computer, and Systems Engineering at Rensselaer Polytechnic Institute, Troy, NY, USA. His research interests include machine learning, deep learning theory, and graph neural network.
\end{IEEEbiography}

\begin{IEEEbiography} [{\includegraphics[width=1.05in,height=1.25in,clip,keepaspectratio]{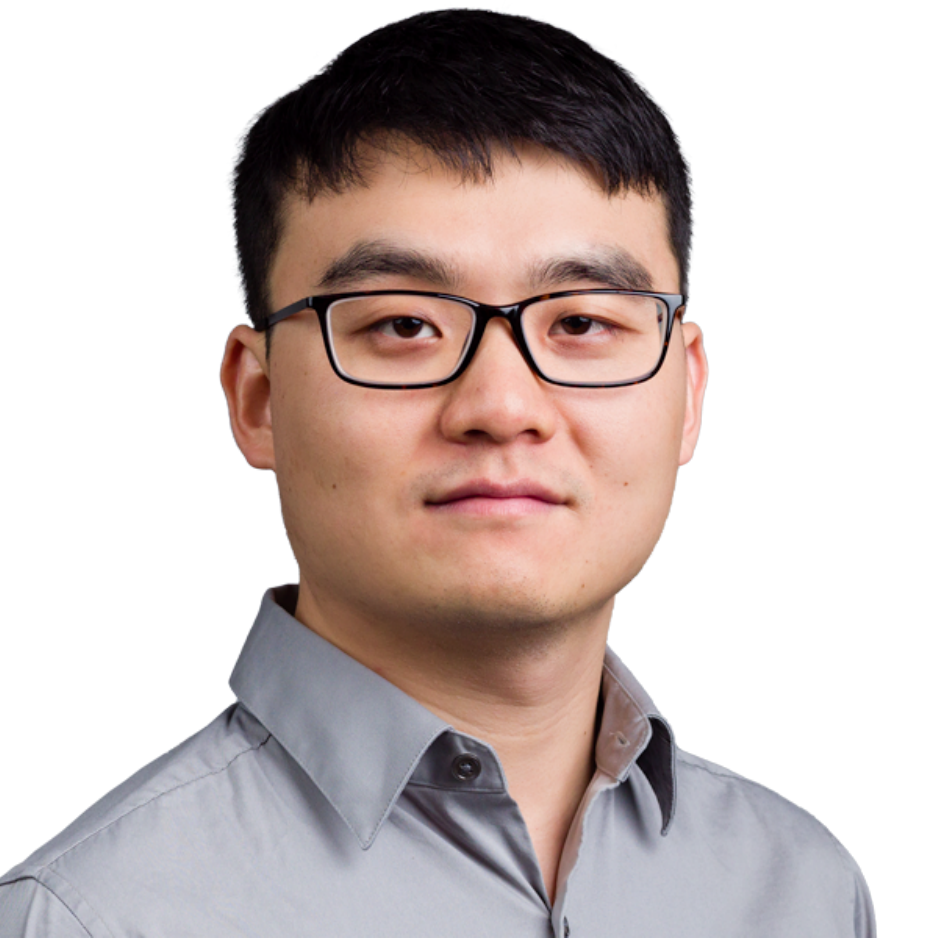}}]    {Shuai Zhang} (Member, IEEE)  received the B.E. degree from the University of Science and Technology of China, Hefei, China, in 2016. He received his Ph.D. degree from Rensselaer Polytechnic Institute, Troy, NY, USA, in 2021. 

He is currently an Assistant Professor in the Department of Data Science at New Jersey Institute of Technology, Newark, NJ, USA. Before that, he was a Postdoctoral Research Associate at Rensselaer Polytechnic Institute in 2022-2023. His research interests span deep learning, optimization, data science, and signal processing, with a particular emphasis on learning theory, algorithmic foundations of machine learning, and the development of efficient and trustworthy AI. 
\end{IEEEbiography}

\begin{IEEEbiography} [{\includegraphics[width=1.05in,height=1.25in,clip,keepaspectratio]{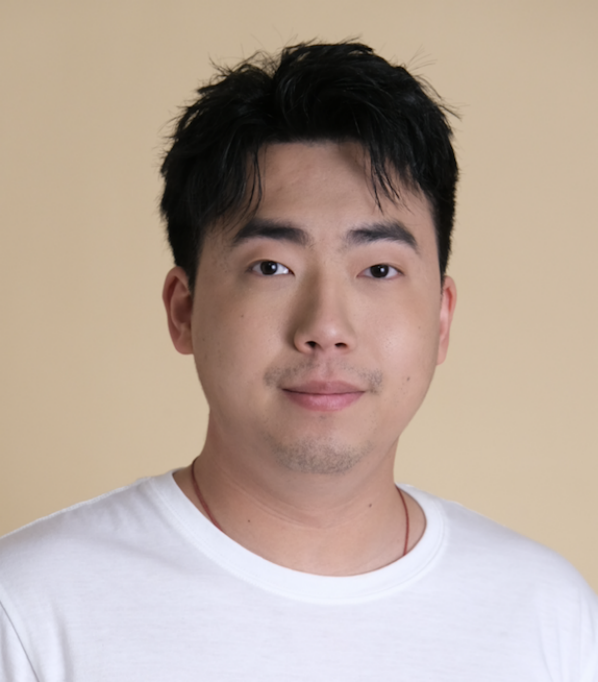}}]    {Yihua Zhang} received the B.E. degree in the School of Mechanical Engineering at Huazhong University of Science and Technology, Wuhan, China, in 2019. He is a Ph.D. student in the Department of Computer Science and Engineering at Michigan State University. His research has been focused on the optimization theory and optimization foundations of various AI applications. In general, his research spans the areas of machine learning (ML)/deep learning (DL), computer vision, and trustworthiness. He has published papers at major ML/AI conferences such as CVPR, NeurIPS,  ICLR, and ICML. He also received the Best Paper Runner-Up Award at the Conference on Uncertainty in Artificial Intelligence (UAI), 2022.
\end{IEEEbiography}

\begin{IEEEbiography}
[{\includegraphics[width=1in,height=1.25in,clip,keepaspectratio]{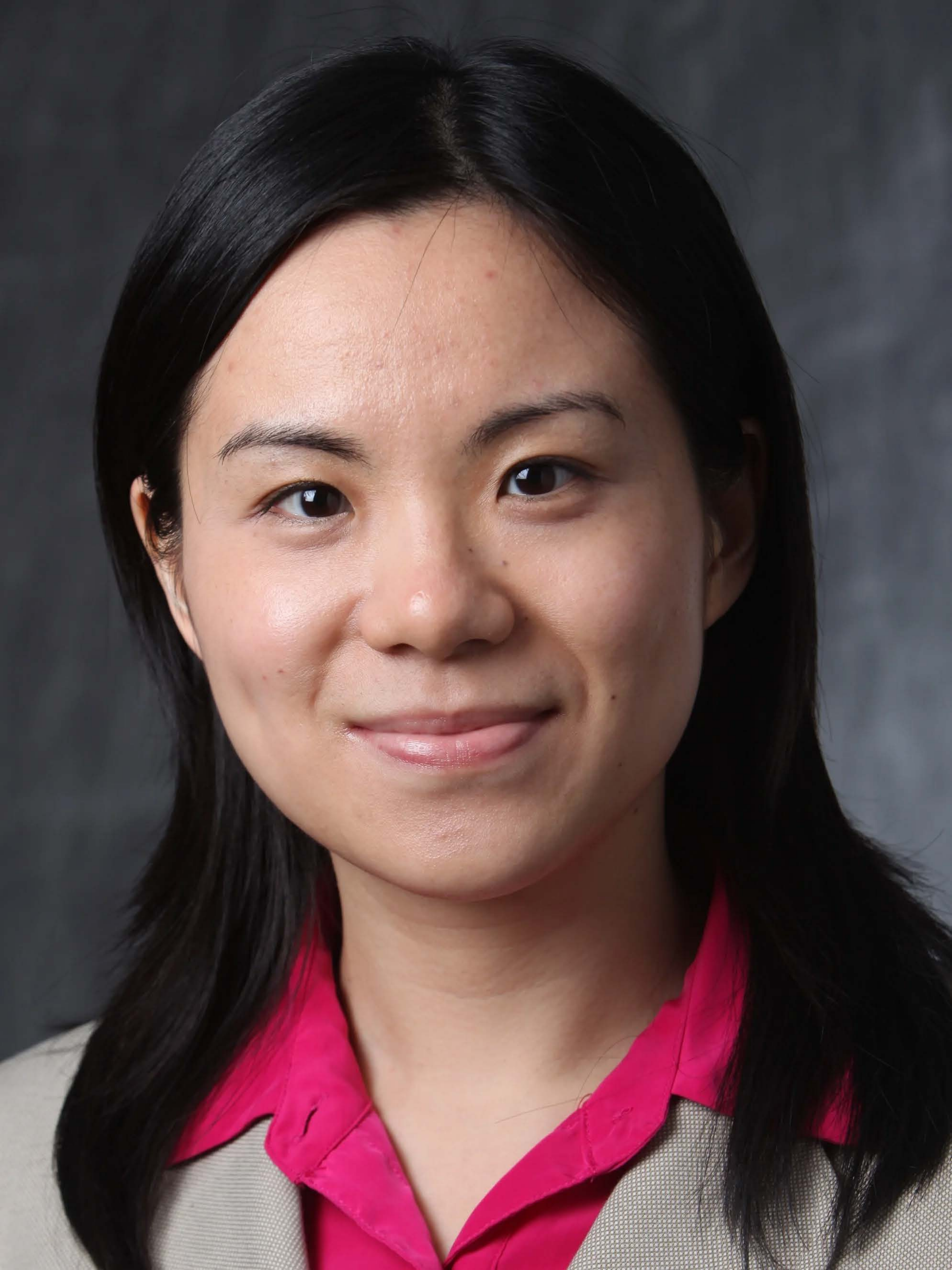}}]
 {Meng Wang} (Senior Member, IEEE) received B.S. and M.S. degrees from Tsinghua University, China, in 2005 and 2007, respectively. She received the Ph.D. degree from Cornell University, Ithaca, NY, USA, in  2012. 
 
 She is an Associate Professor in the Department of  Electrical, Computer, and Systems Engineering at Rensselaer Polytechnic Institute, Troy, NY, USA, where she joined in Dec. 2012. Before that, she was a postdoc scholar at Duke University, Durham, NC, USA. Her research interests include the theory of machine learning and artificial intelligence,  high-dimensional data analytics,  and power systems monitoring. She serves as an Associate Editor for IEEE Transactions on Signal Processing and IEEE Transactions on Smart Grid.
\end{IEEEbiography} 
\vspace{-2mm}

\begin{IEEEbiography}
    [{\includegraphics[width=1in,height=1.25in,clip,keepaspectratio]{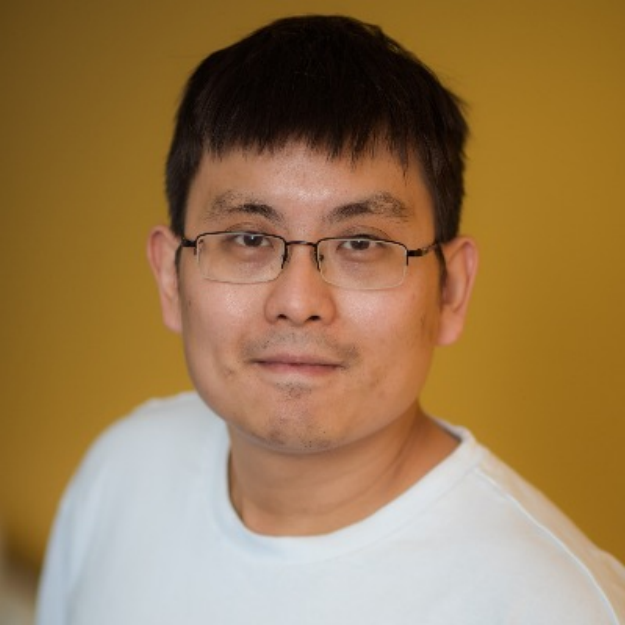}}] {Sijia Liu} (Senior Member, IEEE) received the Ph.D. degree (with All-University Doctoral Prize) in Electrical and Computer Engineering from Syracuse University, NY, USA, in 2016. He was a Postdoctoral Research Fellow at the University of Michigan, Ann Arbor, in 2016-2017, and a Research Staff Member at the MIT-IBM Watson AI Lab in 2018-2020. He is currently an Assistant Professor at the CSE department of Michigan State University, and an Affiliate Professor at the MIT-IBM Watson AI Lab, IBM Research. His research focuses on trustworthy and scalable ML, and optimization theory and methods. He received the Best Student Paper Award at the 42nd IEEE International Conference on Acoustics, Speech and Signal Processing (ICASSP’16), and the Best Paper Runner-Up Award at the 38th Conference on Uncertainty in Artificial Intelligence (UAI’22).
\end{IEEEbiography}
\vspace{-2mm}

\begin{IEEEbiography}
[{\includegraphics[width=1in,height=1.25in,clip,keepaspectratio]{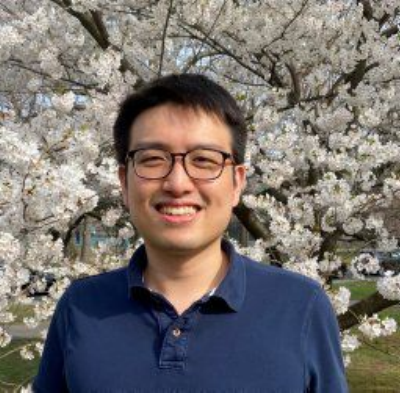}}]
 {Pin-Yu Chen} (Member, IEEE) Dr. Pin-Yu Chen is a principal research scientist at IBM Thomas J. Watson Research Center, Yorktown Heights, NY, USA. He is also the chief scientist of RPI-IBM AI Research Collaboration and PI of ongoing MIT-IBM Watson AI Lab projects. Dr. Chen received his Ph.D. in electrical engineering and computer science from the University of Michigan, Ann Arbor, USA, in 2016. Dr. Chen’s recent research focuses on adversarial machine learning of neural networks for robustness and safety. His long-term research vision is to build trustworthy machine learning systems. He received the IJCAI Computers and Thought Award in 2023. At IBM Research, he received several research accomplishment awards, including IBM Master Inventor, IBM Corporate Technical Award, and IBM Pat Goldberg Memorial Best Paper. His research contributes to IBM open-source libraries including Adversarial Robustness Toolbox (ART 360) and AI Explainability 360 (AIX 360). He has published more than 50 papers related to trustworthy machine learning at major AI and machine learning conferences. He is currently on the editorial board of Transactions on Machine Learning Research and serves as an Area Chair or Senior Program Committee member for NeurIPS, ICML, AAAI, IJCAI, and PAKDD. 
 \end{IEEEbiography}
\vspace{-2mm}

\onecolumn

\setcounter{page}{1}
%\appendix

\begin{center}
\Huge{Supplementary Material}
\end{center}
We begin our Supplementary Material here.

\textbf{Section \ref{sec: appendix_experiment}} provides more experiment results as a supplement of Section \ref{sec: experiment}. 

\textbf{Section \ref{sec: appendix_algorithm}} introduces the algorithm, especially the tensor initialization in detail. 

\textbf{Section \ref{sec: preliminary}} includes some definitions and properties as a preliminary to our proof. 

\textbf{Section \ref{section: thm1}} shows the proof of Theorem \ref{thm1} and Corollary \ref{cor}, followed by \textbf{Section \ref{sec:convexity}, \ref{sec: convergence}, and \ref{sec: tensor bound}} as the proof of three key Lemmas about local convexity, linear convergence and tensor initialization, respectively.

\subsection{More Experiment Results}\label{sec: appendix_experiment}

We present our experiment results%\footnote{codes are available at \url{https://www.dropbox.com/sh/k8d0ssq0ijjsvov/AADMvp0qQZXqF9w-Wp1ev0w9a?dl=0}}
on empirical datasets CelebA \cite{LLWT15} and CIFAR-10 \footnote{Alex Krizhevsky, Vinod Nair, and Geoffrey Hinton. The
CIFAR-10 dataset. \url{www.cs.toronto.edu/˜kriz/cifar.html}}  in this section. To be more specific, we evaluate the impact of the variance levels introduced by different data augmentation methods on the learning performance. We also evaluate the impact of the minority group fraction in the training data on the learning performance. % how the learning performance will be when we try different parameter choices in some data augmentation methods and when different groups play the role of the minority. 
All the experiments are reported in a format of ``mean$\pm2\times$standard deviation'' with a random seed equal to $10$. We implement our experiments on an NVIDIA GeForce RTX 2070 super GPU and a work station with 8 cores of 3.40GHz Intel i7 CPU.

\subsubsection{Tests on CelebA}

\iffalse
Figure \ref{figure: celebA_aug_2} (a), Figure \ref{figure: celebA_lm2_2} (a), and (b) have exactly the same setup as those for Figure \ref{figure: celebA_noise} (b), Figure \ref{figure: celebA_group} (a) and (b), respectively. We repeat these experiments to add error bars of all results. We intend to replace Figure \ref{figure: celebA_noise} (b), \ref{figure: celebA_group} (a) and (b) in the main text with Figure \ref{figure: celebA_aug_2} (a), Figure \ref{figure: celebA_lm2_2} (a) and (b) in the final version. 
\fi

\begin{figure}[htbp]
    \centering
    %\subfigure[]{
     %   \begin{minipage}{0.45\textwidth}
     %   \centering
     %   \includegraphics[width=1\textwidth, %height=0.55\textwidth]{figures/celebA_gaussian_e.png}
      %  \end{minipage}
    %}
    %~

        \centering
        \includegraphics[width=0.6\textwidth]{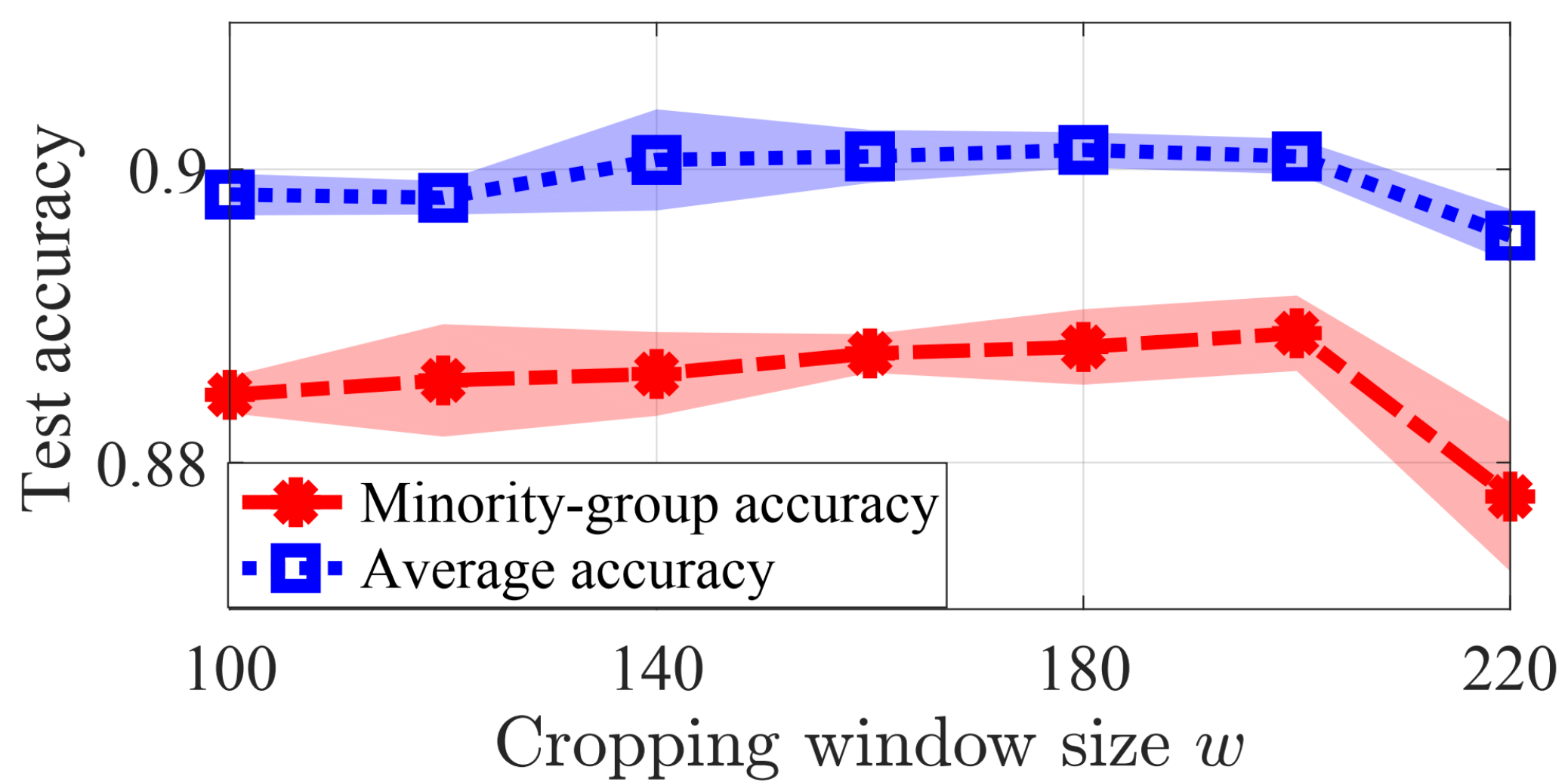}

    \caption{The test accuracy of CelebA dataset with the data augmentation method of cropping.}
    \label{figure: celebA_aug_2}
\end{figure}

In addition to the Gaussian augmentation method in Figure \ref{figure: celebA_noise} (b), we also evaluate the performance of data augmentation by cropping in Figure \ref{figure: celebA_aug_2}.  The setup is exactly the same as that for Gaussian augmentation, expect that we augment the data by cropping instead of adding Gaussian noise. Specifically, to generate an augmented image, we randomly crop an image with a size $w\times w\times 3$ and then resize back to $224\times 224\times 3$. One can observe that the minority-group and average test accuracy first increase and then decrease as $w$ increases, which is in accordance with Insight (P3). 
\iffalse
We implement the same experiment on CelebA. The settings can be found in Section \ref{sec: celebA_main}. We especially add the error bar in Figure \ref{figure: celebA_aug_2} and \ref{figure: celebA_lm2_2} to repeat Figure \ref{figure: celebA_noise} (b), \ref{figure: celebA_group} (a) and (b), respectively. The setting in Figure \ref{figure: celebA_aug_2} (b) is the same as in \ref{figure: celebA_aug_2} (a). We randomly crop the image with a size $w\times w\times 3$ and then resize back to $224\times 224\times 3$. One can observe that the minority-group and average test accuracy increase and then decrease as $w$ increases, which is in accordance with Insight (P3). 
\fi

\subsubsection{Tests on CIFAR-10}

 Group 1 contains  images with attributes ``bird'', ``cat'', ``deer'', ``dog'', ``frog'' and ``horse.''  Group 2 contains ``airplane'' images. In this setting, Group 1 has a larger variance. Because each image in CIFAR-10 only has one attribute, we consider the binary classification setting where all images in Group 1 are labeled as ``animal'' and all images are labeled as ``airplane.'' This is a special scenario that the group label is also the classification label. Note that our results hold for general setups where group labels and classification labels are irrelevant, like our previous results on CelebA. LeNet 5 \cite{LBBH98} is selected to be the learning model.

\begin{figure}[htbp]
    \centering
    \subfigure[]{
        \begin{minipage}{0.45\textwidth}
        \centering
        \includegraphics[width=1\textwidth, height=0.55\textwidth]{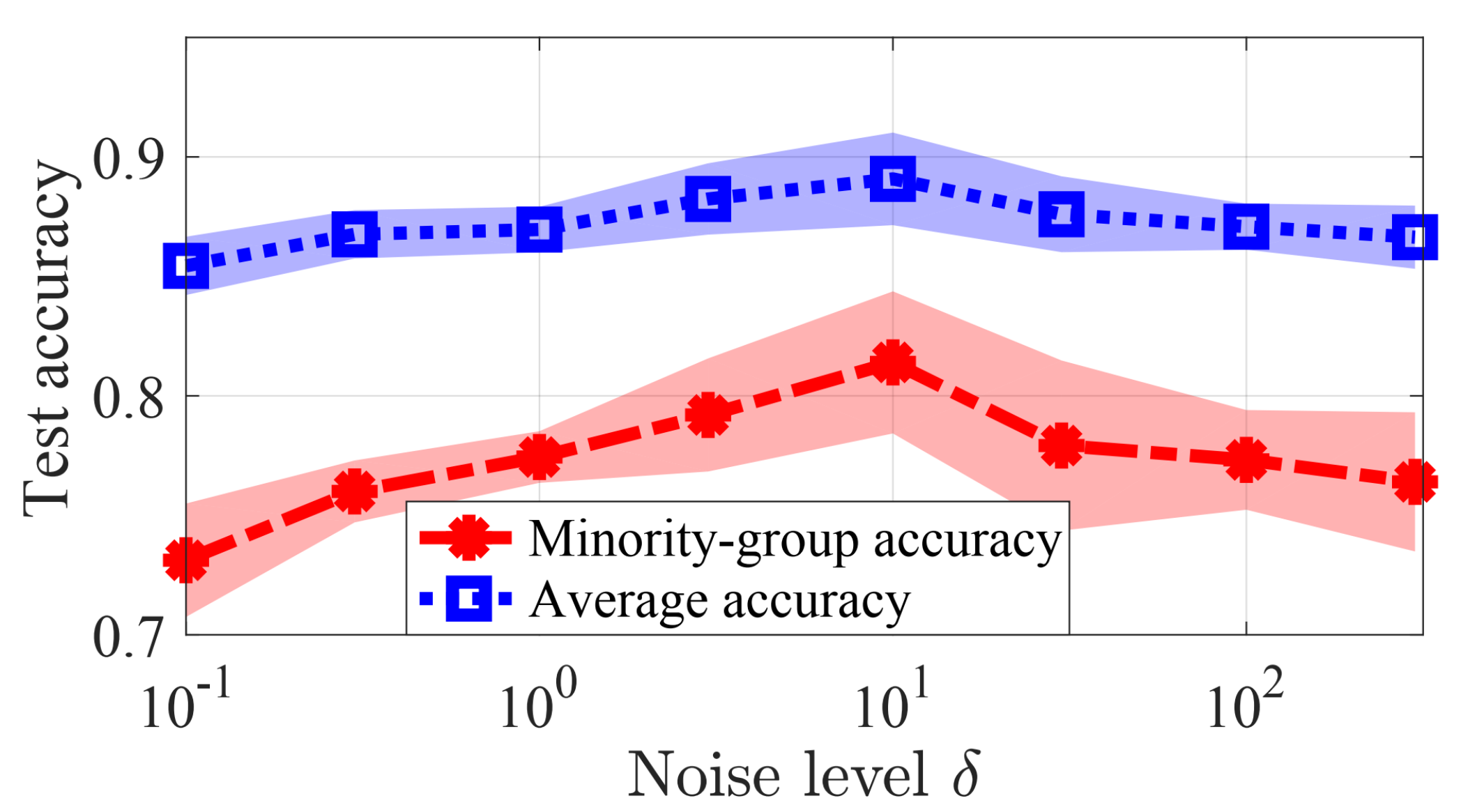}
        \end{minipage}
    }
    ~
    \subfigure[]{
        \begin{minipage}{0.45\textwidth}
        \centering
        \includegraphics[width=1\textwidth,height=0.55\textwidth]{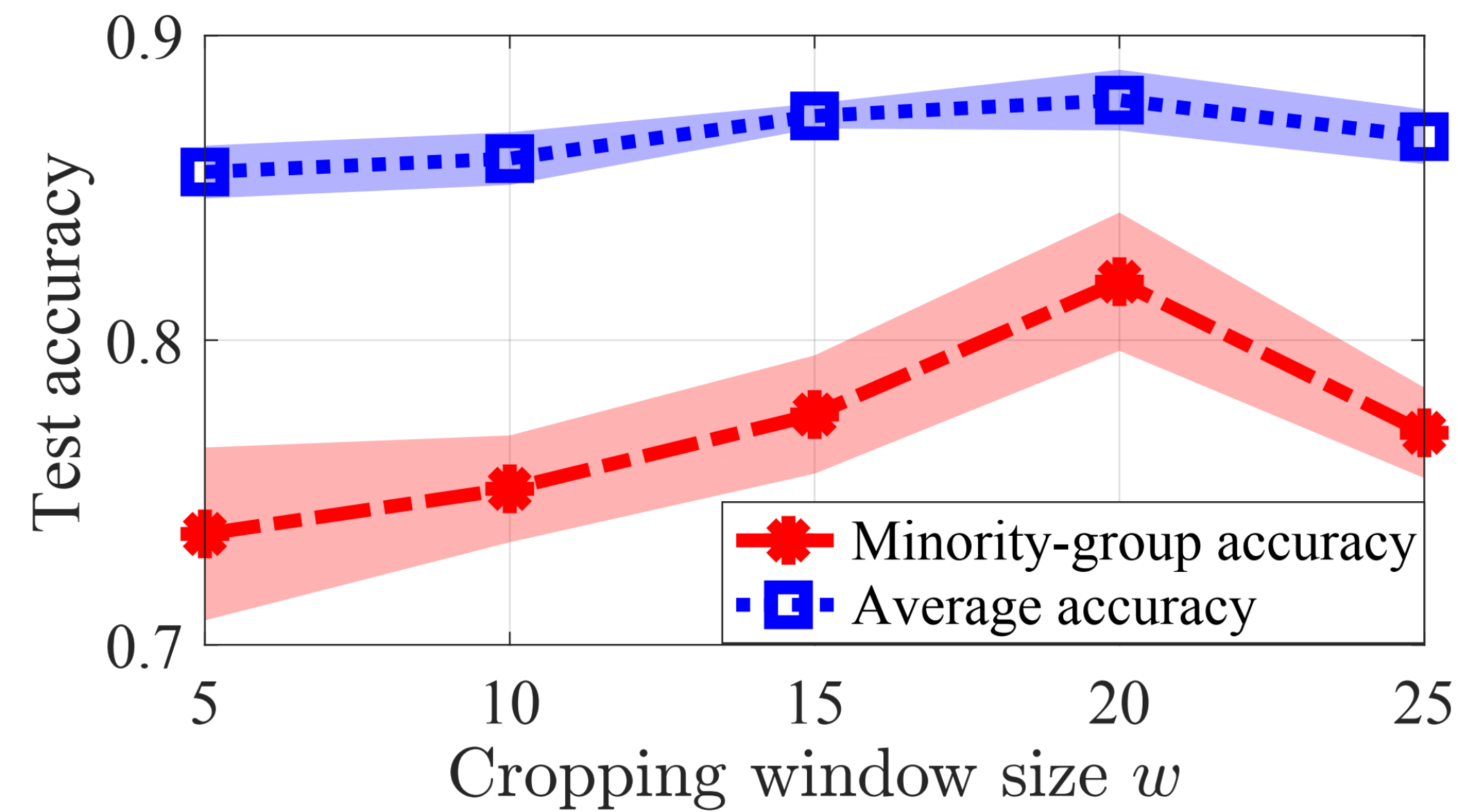}
        \end{minipage}
    }
    \caption{The test accuracy of CIFAR-10 dataset with different data augmentation methods (a) Gaussian noise (b) cropping.}
    \label{figure: cifar_aug}
\end{figure}

%Because each image only has one attribute in
%We generate two groups for the binary classification: one is the class of ``airplane'' and another is the class of ``animal'' by combining ``bird'', ``cat'', ``deer'', ``dog'', ``frog'' and ``horse'' together. Due to limited attributes in CIFAR-10, we denote each group with the same label for one class. LeNet 5 \cite{LBBH98} is selected to be the learning model. 
We first pick $8000$ animal images (majority) and $2000$ airplane images (minority). We select $1000$ out of $2000$ airplane images to implement data augmentation, including both Gaussian augmentation and random cropping. For Gaussian augmentation, we add i.i.d. Gaussian noise drawn from $\mathcal{N}(0,\delta^2)$ to each entry\footnote{In this experiment, the noise is added to the raw image where the pixel value ranges from $0$ to $255$, while in the experiment of CelebA (Figure \ref{figure: celebA_noise} (b)), the noise is added to the image after normalization where the pixel value ranges from $0$ to $1$.}. For random cropping, we randomly crop the image with a certain size $w\times w\times 3$ and then resizing back to $32\times 32 \times 3$. Figure \ref{figure: cifar_aug} shows that when $\delta$ or $w$ increase, i.e., the variance introduced by either augmentation method increases, both the minority-group and average test accuracy increase first and then decrease, which is consistent with our Insight (P3).

\begin{figure}[htbp]
    \centering
    \subfigure[]{
        \begin{minipage}{0.45\textwidth}
        \centering
        \includegraphics[width=1\textwidth, height=0.55\textwidth]{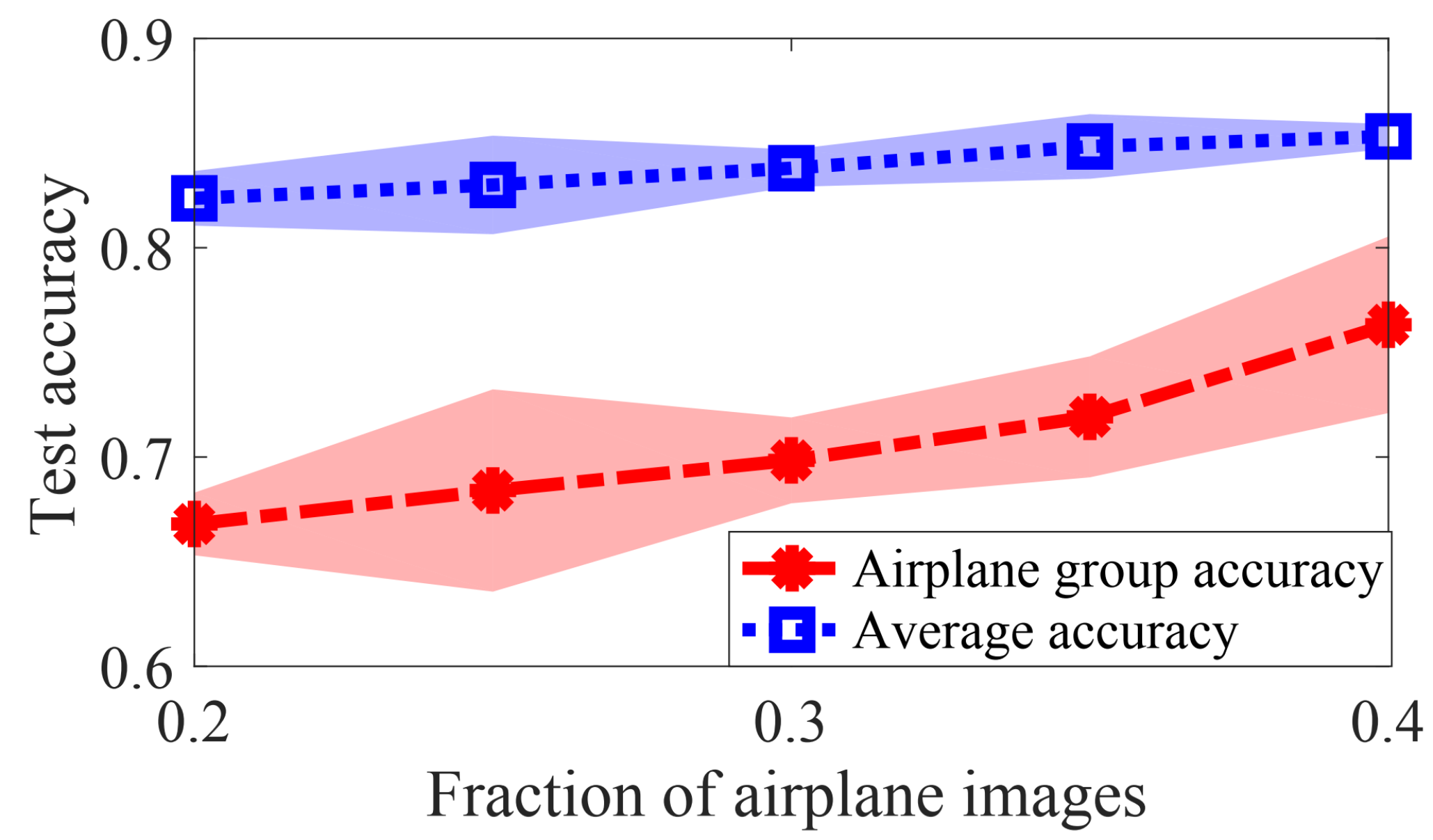}
        \end{minipage}
    }
    ~
    \subfigure[]{
        \begin{minipage}{0.45\textwidth}
        \centering
        \includegraphics[width=1\textwidth,height=0.55\textwidth]{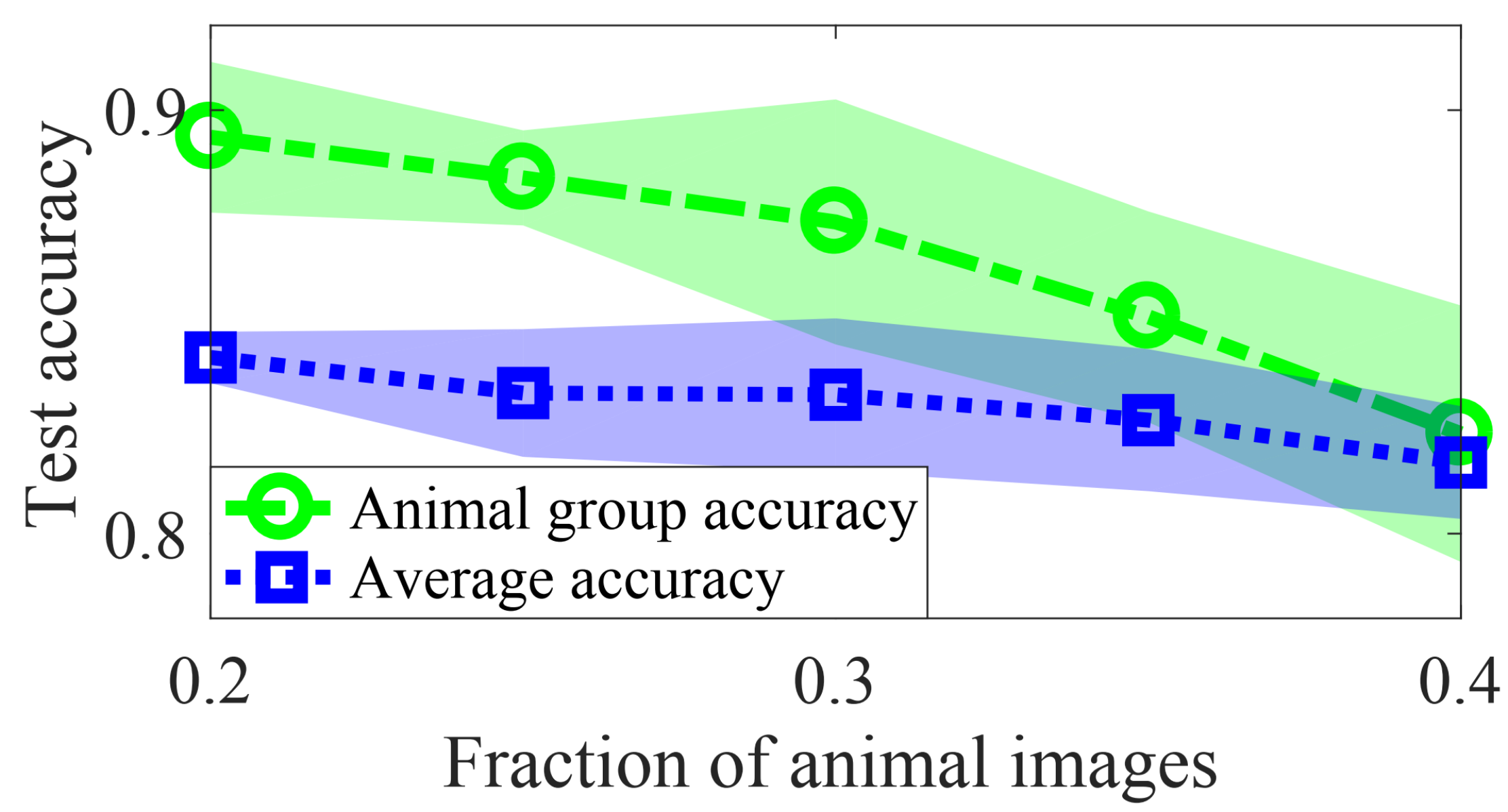}
        \end{minipage}
    }
    \caption{The test accuracy of CIFAR-10 dataset has opposite trends when the minority group fraction increases (a) Airplane group is the minority. (b) Animal group is the minority.}
    \label{figure: cifar_lm2}
\end{figure}

Then we fix the total number of training data to be $5000$ and vary the fractions of the two groups. One can see opposite trends in Figure \ref{figure: cifar_lm2} if we increase the fraction of the minority group with the airplane being the minority and the animal being the minority, which reflects our Insight (P4).


\subsection{Algorithm}\label{sec: appendix_algorithm}
We first introduce new notations to be used in this part and summarize key notions in Table \ref{tbl:notations}.\\
We write $f(x)\lesssim (\gtrsim) g(x)$ if $f(x)\leq (\geq)\Theta(g(x)$. The gradient and the Hessian of a function $f(\bfW)$ are  denoted by $\nabla f(\bfW)$ and $\nabla^2 f(\bfW)$, respectively. $\bfA\succeq0$ means $\bfA$ is a positive semi-definite (PSD) matrix. $\bfA^\frac{1}{2}$ means that $\bfA=(\bfA^\frac{1}{2})^2$. The outer product of  vectors $\bfz_i\in\mathbb{R}^{n_i}$, $i\in[l]$, is defined as $\bfT=\bfz_1\otimes \cdots\otimes \bfz_l\in\mathbb{R}^{n_1\times\cdots\times n_l}$ with $\bfT_{j_1\cdots j_l}=(\bfz_1)_{j_1}\cdots(\bfz_l)_{j_l}$.
Given a tensor $\bfT\in\mathbb{R}^{n_1\times n_2\times n_3}$ and   matrices $\bfA\in\mathbb{R}^{n_1\times d_1}$, $\bfB\in\mathbb{R}^{n_2\times d_2}$, $\bfC\in\mathbb{R}^{n_3\times d_3}$, the $(i_1,i_2, i_3)$-th entry of the tensor $\bfT(\bfA,\bfB, \bfC)$ is given by
\begin{equation}
    \sum_{i_1'}^{n_1}\sum_{i_2'}^{n_2}\sum_{i_3'}^{n_3}\bfT_{i_1',i_2', i_3'}{\bfA}_{i_1',i_1}{\bfB}_{i_2',i_2}{\bfC}_{i_3',i_3}.\label{T(A,B,C)}
\end{equation}

\begin{table}[ht]

  \begin{center}
        \caption{Summary of notations}
        \label{tbl:notations}
\begin{tabularx}{\textwidth}{lX} % <-- Alignments: 1st column left, 2nd middle and 3rd right, with vertical lines in between
 \hline
 \small $\lambda_l,\ \bfmu_l,\ \bfSg_l,\ l\in[L]$ &  The fraction, mean, and covariance of the $l$-th component in the Gaussian mixture distribution, respectively.\\ 
 \hline
 \small $d,\ n,\ K$ & The feature dimension, the number of training samples, and the number of neurons, respectively.\\
 \hline
  \small $\bfW^*,\ \bfW_t$ & {$\bfW^*$ is the ground truth weight. $\bfW_t$ is the updated weight in the $t$-th iteration}.\\
 \hline
  \small $f_n,\ \bar{f},\ \ell$ &  $f_n$ is the empirical risk function. $\bar{f}$ is the average risk or the population risk function. $\ell$ is the cross-entropy loss function.\\
 \hline
 \small $\Psi,\ \sigma_{\max},\ \sigma_{\min},\ \tau$ &  $\Psi$ denotes our Gaussian mixture model $(\lambda_l,\bfmu_l,\bfSg_l,\forall l)$. $\sigma_{\max}=\max_{l\in[L]}\{\|\bfSg_l\|^\frac{1}{2}\}$. $\sigma_{\min}=\min_{l\in[L]}\{\|\bfSg_l^{-1}\|^{-\frac{1}{2}}\}$. $\tau=\sigma_{\max}/\sigma_{\min}$.\\
 \hline
 \small $\delta_i(\bfW^*),\ \eta,\ \kappa,\ i\in[K]$ &  $\delta_i(\bfW^*)$ is the $i$-th largest singular value of $\bfW^*$. $\eta$ and $\kappa$ are two functions of $\bfW^*$.\\
 \hline
 \small $\rho(\bfu,\sigma),\ \Gamma(\Psi),\ D_m(\Psi)$ & These items are functions of the Gaussian mixture distribution $\Psi$ used to develop our Theorem \ref{thm1}.\\
 \hline
 \small $\boldsymbol{\nu}_i,\ \xi$ & $\boldsymbol{\nu}_i$ is the gradient noise. $\xi$ is the upper bound of the noise level. \\
 \hline
 \small $\bfQ_j,\ j=1,2,3$ & $\bfQ_j$'s are tensors used in the initialization. \\
 \hline
 \small $\mathcal{B}(\Psi)$ & A parameter appeared in the sample complexity bound (\ref{final_sp}).\\
 \hline
 \small $v(\Psi),\ q(\Psi)$ & $v(\Psi)$ is the convergence rate (\ref{linear convergence}). $q(\Psi)$ is a parameter in the definition of $v(\Psi)$ (\ref{eqn:v}).\\
 \hline
\small $\mathcal{E}_w(\Psi),\ \mathcal{E},\ \mathcal{E}_l$ & Generalization parameters. $\mathcal{E}_w(\Psi)$ appears in the error bound of the model (\ref{eqn: w_bound}). $\mathcal{E}(\Psi)$ and $\mathcal{E}_l(\Psi)$ are to characterize the average risk (\ref{eqn: f_bound}) and the group-l risk (\ref{eqn: fl_bound}), respectively.\\
\hline

\end{tabularx}
\end{center}

\end{table}

\normalsize The method starts from an initialization $\bfW_0 \in \R^{d\times K}$ computed based on the tensor initialization method (Subroutine \ref{TensorInitialization}) and then updates the iterates $\bfW_t$ using   gradient descent   with the step size $\eta_0$. To model the inaccuracy in computing the gradient, an i.i.d. zero-mean noise $\{\nu_i\}_{i=1}^n\in\mathbb{R}^{d\times K}$ with  bounded magnitude   $ |(\nu_i)_{jk}|\leq\xi$ ($j \in [d], k\in [K]$) for some $\xi \geq 0$ are added in (\ref{eqn:gradient}) when computing the gradient of the loss in (\ref{cross-entropy}).

 \iffalse
\setcounter{algorithm}{0}
\begin{algorithm}

\begin{algorithmic}[1]
\caption{Our proposed learning algorithm}\label{gd}
\STATE{\textbf{Input: }} 
 Training data $\{(\bfx_i,y_i)\}_{i=1}^n$, the step size $\eta_0=O\Big( \big( \sum_{l=1}^L\lambda_l (\|\tilde{\bfmu}_l\|_\infty+\|\bfSg_l^\frac{1}{2}\|)^2 \big)^{-1}\Big)$, the total number of iterations $T$
\STATE{\textbf{Initialization: }}$\bfW_{0}\leftarrow$ Tensor initialization method via Subroutine \ref{TensorInitialization}
\STATE{\textbf{Gradient Descent:}} for $t=0,1,\cdots,T-1$
\vspace{-0.1in}
%\textcolor{red}{$$\bfW_{t+1}=\bfW_t-\eta_0\widetilde{\nabla f_n(\bfW_t)}=\bfW_t-\eta_0\Big(\nabla f_n(\bfW^*)+\frac{1}{n}\sum_{i=1}^n \nu_i\Big)$$}
\begin{equation}\label{eqn:gradient}
\begin{aligned}
\bfW_{t+1}&=\bfW_t-\eta_0 \cdot \frac{1}{n} \sum_{i=1}^n ( \nabla  l(\bfW, \bfx_i, y_i) +\nu_i )\\
&= \bfW_t-\eta_0\Big(\nabla f_n(\bfW)+\frac{1}{n}\sum_{i=1}^n \nu_i\Big)
\end{aligned}
\end{equation}
\vspace{-0.1in}
\STATE{\textbf{Output: }} $\bfW_T$
\end{algorithmic}
\end{algorithm}
\fi

Our tensor initialization  method in Subroutine \ref{TensorInitialization} is extended from  \cite{JSA14} and \cite{ZSJB17}. The idea is to compute quantities ($\bfQ_j$ in (\ref{eqn:mj})) that are tensors of  $\bfw^*_i$ and then apply the tensor decomposition method to estimate  $\bfw^*_i$. Because $\bfQ_j$   can only be estimated from training samples, tensor decomposition does not return $\bfw^*_i$ exactly but provides a close approximation, and this approximation is used as the initialization for Algorithm \ref{gd}.
%based on the training data  to estimate the column space of $\bfW^*$, the direction and magnitude of each $\bfw^*_j$, respectively. 
Because %the method in \cite{ZSJB17} 
the existing method of tensor construction only applies to the standard Gaussian distribution, we exploit the relationship between probability density functions and tensor expressions developed in  \cite{JSA14} to design tensors suitable for the Gaussian mixture model. Formally, 

%To describe our initialization method, we need the following definitions. 

\begin{definition}\label{def: M}
 For $j=1,2,3$, we define 
%Define $\widetilde{\otimes}$, $S_m(\bfx)$, $\boldsymbol{M}_1$, $\boldsymbol{M}_2$, $\boldsymbol{M}_3$ as follows:\\
%For $\bfx\sim\sum_{l=1}^L\lambda_l\mathcal{N}(\bfmu_l,\sigma_l^2\bfI_d)$, we define
\begin{equation}\label{eqn:mj}
    \bfQ_j:=\mathbb{E}_{\bfx\sim\sum_{l=1}^L\lambda_l\mathcal{N}(\bfmu_l,\bfSg_l)}[y\cdot (-1)^j p^{-1}(\bfx)\nabla^{(j)}p(\bfx)], 
 %   \bfM_j:=\mathbb{E}_{\bfx\sim\sum_{l=1}^L\lambda_l\mathcal{N}(\bfmu_l,\sigma_l^2\bfI)}[y\cdot (-1)^j\frac{\nabla^{(m)}p(\bfx)}{p(\bfx)}],\ j=1,2,3  
\end{equation}
%\tcr{expectation over what? after the expectation, still depends on x?}

%\end{definition}
%
%We  follow the conventional definition (see [\cite{ZSJB17}]) of tensor operators as follows. For a tensor $\bfT\in\mathbb{R}^{n_1\times n_2\times n_3}$ and three matrices $\bfA\in\mathbb{R}^{n_1\times d_1}$, $\bfB\in\mathbb{R}^{n_2\times d_2}$, $\bfC\in\mathbb{R}^{n_3\times d_3}$, the $(i_1,i_2, i_3)$-th entry of the tensor $\bfT(\bfA,\bfB, \bfC)$ is given by
%\begin{equation}
 %   \sum_{i_1'}^{n_1}\sum_{i_2'}^{n_2}\sum_{i_3'}^{n_3}\bfT_{i_1',i_2', i_3'}{\bfA}_{i_1',i_1}{\bfB}_{i_2',i_2}{\bfC}_{i_3',i_3}.\label{T(A,B,C)}
%\end{equation}
%With this definition, we can define the following useful quantities.
%\begin{definition}\label{def: P}
where $p(\bfx)$, the probability density function of GMM is defined as

\begin{equation}\label{eqn:p}
 p(\boldsymbol{x})=\sum_{l=1}^L\lambda_l(2\pi)^{-\frac{d}{2}}|\bfSg_l|^{-\frac{1}{2}}\exp\big(-\frac{1}{2}(\boldsymbol{x}-\boldsymbol{\mu}_l)\bfSg_l^{-1}(\boldsymbol{x}-\boldsymbol{\mu}_l)\big)
 \end{equation}
 If the Gaussian mixture model is symmetric, the symmetric distribution can be written as
\be
{\bfx\sim \left\{
    \begin{array}{ll}
     \sum\limits_{l=1}^{\frac{L}{2}}\lambda_l\big(\mathcal{N}(\bfmu_l,\bfSg_l)+\mathcal{N}(-\bfmu_l,\bfSg_l)\big)  & L \textrm{ is even} \\
         \lambda_1\mathcal{N}(\boldsymbol{0},\bfSg_1)+\sum\limits_{l=2}^{\frac{L-1}{2}}\lambda_l\big(\mathcal{N}(\bfmu_l,\bfSg_l)+\mathcal{N}(-\bfmu_l,\bfSg_l)\big) & L \textrm{ is odd}
    \end{array}\right.}\label{symmetric_GMM}
\ee
\end{definition}

$\bfQ_j$ is a $j$th-order tensor of $\bfw^*_i$, e.g.,  $\bfQ_3=\frac{1}{K}\sum_{i=1}^K\mathbb{E}_{\bfx\sim\sum_{l=1}^L\lambda_l\mathcal{N}(\bfmu_l,\bfSg_l)}[\phi'''({\bfw_i^*}^\top\bfx)]{\bfw_i^*}^{\otimes 3}$.
These quantifies cannot be directly computed from (\ref{eqn:mj})  but can be estimated 
%Note that $\bfM_i$ ($i=1,2,3$) and $\bfP_2$ depend on $\bfW^*$ through having $y$ in the definition %. Therefore, they 
%and cannot be directly computed. % from the definition if $\bfW^*$ is unknown. 
%We estimate these values
by sample means, denoted by $\widehat{\bfQ}_j$ ($j=1,2,3$), %and $\widehat{\bfP}_2$, 
from samples $\{\bfx_i, y_i \}_{i=1}^n$.  The following assumption    guarantees  that  % $\bfM_i$ ($i=1,2,3$) and
%$\bfP_2$ are all nonzero, and 
these tensors are nonzero and  can thus % Then these quantities can
be
leveraged to estimate $\bfW^*$.  % one can estimate the estimators still contain the information of the direction and the magnitude of $\bfw_i^*$. 

\begin{assumption}\label{assumption1}
The Gaussian Mixture Model in (\ref{symmetric_GMM}) satisfies the following conditions: 
%(1) $\bfM_1\neq0$, where $\bfM_1$ is defined in Definition \ref{def: M}\\ %\tcr{Is M_1 a function of x and/or $\alpha$?}{\color{blue}no, it is defined in definition 1}
\begin{enumerate}
    \item  $\bfQ_1$ and $\bfQ_3$ are nonzero.
\item If the distribution is not symmetric, then $\bfQ_2$ is nonzero.
\end{enumerate}
\end{assumption}

%By Theorem 6 in \cite{JSA14}, one can check that  Assumption \ref{assumption1} guarantees that $\bfM_2$, $\bfM_3$, and $\bfP_2$ are nonzero. Theorem 6 in \cite{JSA14} also shows that $\bfM_1$ is nonzero. 
  Assumption \ref{assumption1} is a very mild assumption\footnote{By mild, we mean   given $L$, if Assumption 1 is not met for some %In any neighborhood of the distribution parameter
$\Psi_0$, there   exists  an infinite number of $\Psi'$   in any neighborhood of  $\Psi_0$ such that Assumption 1 holds for $\Psi'$,}. % and holds naturally unless the parameters of the Gaussian Mixture model is specially designed.
Moreover, as indicated in \cite{JSA14}, in the rare case that some quantities $\bfQ_i$ ($i=1,2,3$) are zero, one can construct higher-order tensors in a similar way as in Definition \ref{def: M} and then estimate $\bfW^*$ from higher-order tensors.

Subroutine \ref{TensorInitialization}
% \ref{TensorInitialization} 
describes the tensor initialization method, which estimates the direction and magnitude of $\bfw_j^*, j\in[K]$, separately. The direction vectors are denoted as $\boldsymbol{\bar{w}}_j^*=\bfw_j^*/\|\bfw_j^*\|$ and the  magnitude $\|\bfw_j^*\|$ is denoted as $z_j$.
Lines 2-6 estimate  the subspace  $\widehat{\bfU}$ spanned by $\{\bfw_1^*,\cdots,\bfw_K^*\}$ using $\widehat{\bfQ}_2$ or, in the case that $\bfQ_2=0$, a second-order tensor projected by  $\widehat{\bfQ}_3$.  Lines 7-8 estimate   $\boldsymbol{\bar{w}}_j^*$ by employing the KCL algorithm \cite{KCL15}. Lines 9-10 estimate  the magnitude $z_j$. Finally, the returned estimation of $\bfW^*$ is used as an initialization $\bfW_0$ for Algorithm \ref{gd}. The computational complexity of Subroutine \ref{TensorInitialization}  is $O(Knd)$ based on similar calculations as those in \cite{ZSJB17}.

\floatname{algorithm}{Subroutine}
\setcounter{algorithm}{0}
\begin{algorithm}
\caption{Tensor Initialization Method}\label{TensorInitialization}
\begin{algorithmic}[1]
\STATE{\textbf{Input: }} Partition $n$ pairs of data $\{(\bfx_i,y_i)\}_{i=1}^n$ into three disjoint subsets $\mathcal{D}_1$, $\mathcal{D}_2$, $\mathcal{D}_3$
%\STATE{\textbf{Output: }}
\IF{the Gaussian Mixture distribution is not symmetric}
\STATE Compute $\widehat{\bfQ}_2$ using $\mathcal{D}_1$. Estimate the subspace $\widehat{\bfU}$   by orthogonalizing the eigenvectors with respect to the $K$ largest eigenvalues of  $\widehat{\bfQ}_2$
\ELSE \STATE Pick an arbitrary vector  $\boldsymbol{\alpha} \in \R^{d}$, and use $\mathcal{D}_1$ to compute $\widehat{\bfQ}_3(\bfI_d,\bfI_d,\boldsymbol{\alpha})$.  Estimate   $\widehat{\bfU}$   by orthogonalizing the eigenvectors with respect to the $K$ largest eigenvalues of   $\widehat{\bfQ}_3(\bfI_d,\bfI_d,\boldsymbol{\alpha})$.
\ENDIF

%\STATE \tcr{If the Gaussian Mixture distribution is symmetrix as in (\ref{symmetric_GMM}), estimate the subspace $\widehat{\bfU}$ spanned by $\{\bfw_1^*,\cdots,\bfw_K^*\}$ by orthogonalizing the eigenvectors with respect to the K largest eigenvalues of $\widehat{\bfQ}_3(\bfI_d,\bfI_d,\boldsymbol{\alpha})$ (computed by $\mathcal{D}_1$), where $\boldsymbol{\alpha}$ is a randomly picked vector. Otherwise, estimate the subspace $\widehat{\bfU}$ by orthogonalizing the eigenvectors with respect to the K largest eigenvalues of $\widehat{\bfQ}_2$ (computed by $\mathcal{D}_1$) instead.}
\STATE Compute $\widehat{\bfR}_3=\widehat{\bfQ}_3(\widehat{\bfU},\widehat{\bfU},\widehat{\bfU})$ from data set $\mathcal{D}_2$
\STATE Employ the KCL algorithm  to compute vectors $\{\hat{\bfv}_i\}_{i\in[K]}$, which are the estimates of $\{\widehat{\bfU}^\top\boldsymbol{\bar{w}}_i^*\}_{i=1}^K$. Then the direction vectors $\{\boldsymbol{\bar{w}}_i^*\}_{i=1}^K$ can be approximated by $\{\widehat{\bfU}\hat{\bfv}_i\}_{i=1}^K$.
\STATE Compute $\widehat{\bfQ}_1$ from data set $\mathcal{D}_3$.
\STATE Estimate the magnitude $\widehat{\bfz}$ by solving the optimization problem
\begin{equation}
    \widehat{\boldsymbol{z}}= \arg\min_{\boldsymbol{\alpha}\in\mathbb{R}^K}\frac{1}{2}\|\widehat{\bfQ}_1-\sum_{j=1}^K \alpha_j\boldsymbol{\bar{w}}_j^*\|^2
\end{equation}

\STATE \textbf{Return:}
Use $\hat{z}_j\widehat{\bfU}\hat{\bfv}_j$ as the $j$th column of $\bfW_0$,  $j\in [K]$.
\end{algorithmic}
\end{algorithm}

\subsubsection{Numerical Evaluation of Tensor Initialization}\label{sec: tensor_compare}
%\paragraph{Tensor initialization}

\begin{figure}[h]
\centering
%\begin{minipage}{0.90\linewidth}
 \centering
\includegraphics[width=0.7\linewidth]{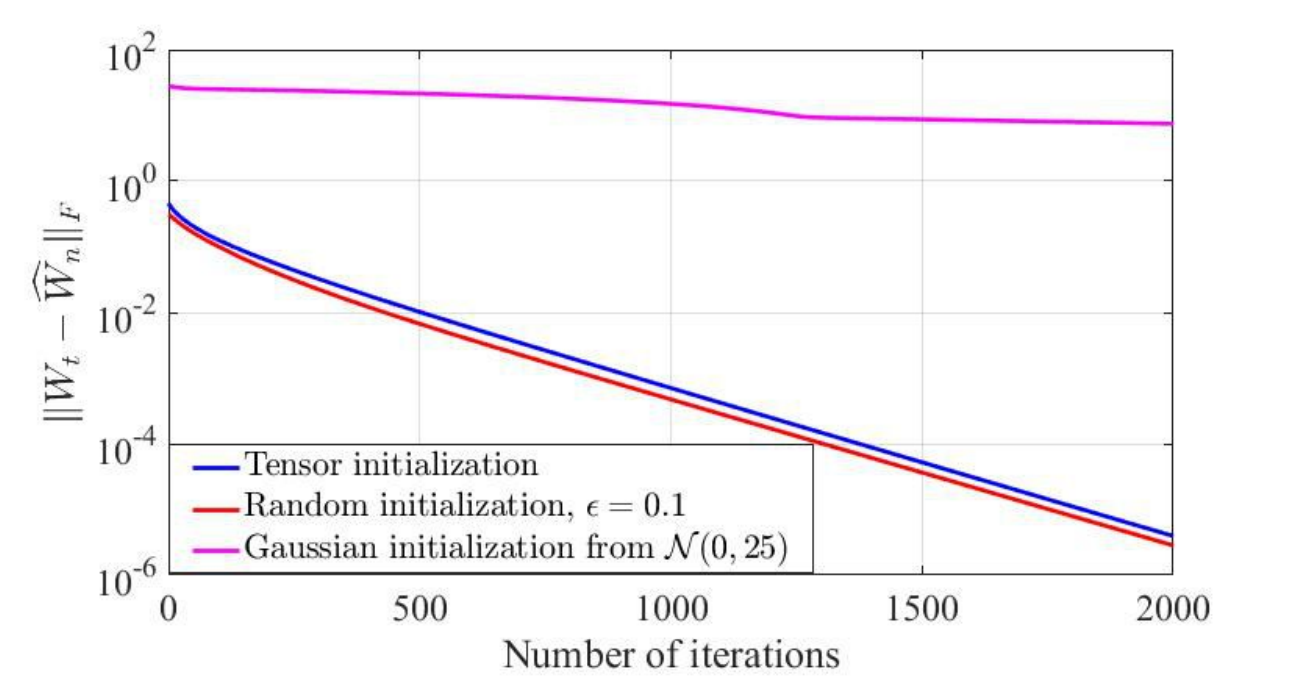}
\caption{Comparison  between tensor initialization, a random initialization near $\bfW^*$, and an arbitrary random initialization} % against $1/K^2$}
\label{fig: tensor}
\end{figure}

Figure~\ref{fig: tensor} shows the accuracy of the returned model by Algorithm \ref{gd}. Here $n=2\times 10^5$, $d=50$, $K=2$, $\lambda_1=\lambda_2=0.5$, $\bfmu_1=-0.3\cdot\boldsymbol{1}$ and $\bfmu_2=\boldsymbol{0}$. We compare the tensor initialization with a random initialization in a local region $\{\bfW\in\mathbb{R}^{d\times K}:||\bfW-\bfW^*||_F\leq \epsilon\}$. Each entry of $\bfW^*$ is selected from $[-0.1,0.1]$ uniformly. Tensor initialization in Subroutine \ref{TensorInitialization} returns an initial point close to one permutation of $\bfW^*$,  with a relative error of $0.65$. 
If the random initialization is also close to $\bfW^*$, 
e.g., $\epsilon=0.1$, then the gradient descent algorithm  converges to a critical point from both initializations, and the linear convergence rate is the same. We also test a random initialization with each entry drawn from %If the random initialization is a matrix with each entry belonging to 
$\mathcal{N}(0,25)$. The initialization is sufficiently far from $\bfW^*$, and the algorithm does not converge.  On a MacBook Pro with Intel(R) Core(TM) i5-7360U CPU at 2.30GHz and MATLAB 2017a, it takes 5.52 seconds to compute the tensor initialization.  Thus, to reduce the computational time, we consider a random initialization with $\epsilon=0.1$ in the   experiments instead of computing tensor initialization.

\subsection{Preliminaries of the Main Proof}\label{sec: preliminary}
In this section, we introduce some  \textbf{definitions} and  \textbf{properties} that will be used to prove the main results.

First, we define the sub-Gaussian random variable and sub-Gaussian norm.
\begin{definition}\label{def: sub-Gaussian}
We say $X$ is a sub-Gaussian random variable with sub-Gaussian norm $K>0$, if $(\mathbb{E}|X|^p)^{\frac{1}{p}}\leq K\sqrt{p}$ for all $p\geq 1$. In addition, the sub-Gaussian norm of X, denoted $\|X\|_{\psi_2}$, is defined as $\|X\|_{\psi_2}=\sup_{p\geq 1}p^{-\frac{1}{2}}(\mathbb{E}|X|^p)^{\frac{1}{p}}$.
\end{definition}

Then we  define  the following three  quantities.  %$\rho(\bfmu,\sigma)$, 
%$\Gamma(\boldsymbol{\lambda},\bfM,\boldsymbol{\sigma},\bfW^*)$ and $D_m(\boldsymbol{\lambda},\bfM,\boldsymbol{\sigma})$, that reflect  the influence of the activation function $\phi(z)$ and   the Gaussian mixture model. Note that
$\rho(\bfmu,\sigma)$ is motivated by the $\rho$ parameter for the standard Gaussian distribution in \cite{ZSJB17}, and we generalize it to a Gaussian with an arbitrary mean and variance.  We define the new quantities $\Gamma(\Psi)$ and $D_m(\Psi)$ for the Gaussian mixture model.  

\iffalse
\begin{definition}\label{def: rho}
($\rho$-function). Let $\bfz\sim\mathcal{N}(\bfu,\bfI_d)\in\mathbb{R}^d$. Define $\alpha_q(i,\bfu,\sigma)=\mathbb{E}_{z_i\sim\mathcal{N}(u_i,1)}[\phi'(\sigma\cdot z_i)z_i^q]$ and $\beta_q(i,\bfu,\sigma)=\mathbb{E}_{z_i\sim\mathcal{N}(u_i,1)}[\phi'^2(\sigma\cdot z_i)z_i^q]$, $\forall\ q\in\{0,1,2\}$, where $z_i$ and $u_i$ is the $i$-th entry of $\bfz$ and $\bfu$, respectively. Define $\rho(\bfu,\sigma)$ as
\begin{equation}
\rho(\bfu,\sigma)=\min_{i,j\in[d],j\neq i}\{(u_{j}^2+1)(\beta_0(i,\bfu,\sigma)-\alpha_0(i,\bfu,\sigma)^2), 
\beta_2(i,\bfu,\sigma)-\frac{\alpha_2(i,\bfu,\sigma)^2}{u_i^2+1}\}\label{rho_function}
\end{equation}
\end{definition}
\fi

\begin{definition}\label{def: Gamma}
($\Gamma$-function). With (\ref{rho_function}) and $\kappa$, $\eta$ defined in Section \ref{sec:formulation}, we define
\begin{equation}
    \Gamma(\Psi)=\sum_{l=1}^L\frac{\lambda_l}{\tau^K\kappa^2\eta}\frac{\|\bfSg_l^{-1}\|^{-1}}{\sigma_{\max}^2}\rho(\frac{{\bfW^*}^\top\bfmu_l}{\delta_K(\bfW^*)\|\bfSg_l^{-1}\|^{-\frac{1}{2}}},\delta_K(\bfW^*)\|\bfSg_l^{-1}\|^{-\frac{1}{2}})
\end{equation}
\end{definition}

\iffalse
\begin{definition}\label{def: D}
(D-function). Given the Gaussian Mixture Model and any positive integer $m$,   define $D_m(\Psi)$ as
\begin{equation}D_m(\Psi)=\sum_{l=1}^L\lambda_l(\frac{\|{\bfmu}_l\|}{\|\bfSg_l^{-1}\|^{-\frac{1}{2}}}+1)^m,
\end{equation}
\end{definition}
\fi

%$\rho$-function is defined to compute the lower bound of the Hessian of the population risk with Gaussian input. 
$\Gamma$ function is the weighted sum of $\rho$-function under mixture Gaussian distribution. This function is positive and upper bounded by a small value. $\Gamma$ goes to zero if all $\|\bfmu_l\|$ or all $\sigma_l$ goes to infinity. 
%$D$-function is a normalized parameter for the means and variances. It is lower bounded by 1. $D$-function is an increasing function of $\|\bfmu_l\|$ and a decreasing function of $\sigma_l$.

\begin{property}\label{prop: column space}
\normalsize Given $\bfW^*=\bfU\bfV\in\mathbb{R}^{d\times k}$, where $\bfU\in\mathbb{R}^{d\times K}$ is the orthogonal basis of $\bfW^*$. For any $\bfmu\in\mathbb{R}^d$, we can find an orthogonal decomposition of $\bfmu$ based on the colomn space of $\bfW^*$, i.e. $\bfmu=\bfmu_{\bfU}+\bfmu_{\bfU_\perp}$. If we consider the recovery problem of FCN with a dataset of Gaussian Mixture Model, in which $\bfx_i\sim\mathcal{N}(\bfmu_h,\bfSg_h)$ for some $h\in[L]$, the problem is equivalent to the problem of FCN with $\bfx_i\sim\mathcal{N}({\bfmu_{\bfU}}_h,\bfSg_h)$. Hence, we can assume without loss of generality that $\bfmu_l$ belongs to the column space of $\bfW^*$ for all $l \in[L]$.
\end{property}
\noindent \textbf{Proof: }\\
From (\ref{cla_model}) and (\ref{cross-entropy}), the recovery problem can be formulated as $$\min_{\bfW^*}\ {g({\bfW^*}^\top \bfx_i, y_i)}$$
For any $\bfx_i\sim\mathcal{N}(\bfmu_h,\bfSg_h)$, $\bfx_i$ can be written as $$\bfx_i=\bfz+\bfmu_h$$
where $\bfz\sim\mathcal{N}(\bfzero,\bfSg_h)$. Therefore,\\
$${\bfW^*}^\top \bfx_i={\bfW^*}^\top (\bfz+\bfmu_h)={\bfW^*}^\top (\bfz+{\bfmu_{\bfU}}_h+{\bfmu_{\bfU_\perp}}_h)={\bfW^*}^\top (\bfz+{\bfmu_{\bfU}}_h)$$
The final step is because ${\bfW^*}^\top \bfmu_{\bfU_\perp}=\bfzero$. So the problem is equivalent to the recovery problem of FCN with $\bfx_i\sim\mathcal{N}({\bfmu_{\bfU}}_h,\bfSg_h)$.\\

Recall that the gradient noise $\boldsymbol{\nu}_i\in\mathbb{R}^{d\times K}$ is zero-mean, and each of its entry is upper bounded by $\xi>0$. 
\begin{property}\label{prop: sub-Gaussian}
We have that $\|\boldsymbol{\nu}_i\|_F$ is a sub-Gaussian random variable with its sub-Gaussian norm bounded bu $\xi\sqrt{dK}$.
\end{property}
\noindent \textbf{Proof:}\\
\begin{equation}
    (\mathbb{E}\|\boldsymbol{\nu}_i\|_F^p)^{\frac{1}{p}}\leq (\mathbb{E}|\sqrt{dK}\xi|^p)^{\frac{1}{p}}\leq \xi\sqrt{dK}
\end{equation}

We state some general properties of the $\rho$ function defined in Definition \ref{def: rho} in the following. 
\begin{property}\label{prop: rho_property}
  $\rho(\bfu,\sigma)$ in Definition \ref{def: rho}  satisfies the following properties,
  \begin{enumerate}

\item (\textbf{Positive}) $\rho(\bfu,\sigma)>0$ for any $\bfu\in\mathbb{R}^d$ and $\sigma\neq0$. 
\item (\textbf{Finite limit point for zero mean}) $\rho(\bfu,\sigma)$ converges to a positive value function of $\sigma$ as $u_i$ goes to 0, i.e. $\lim_{u_i\rightarrow0}\rho(\bfu,\sigma):=\mathcal{C}_m(\sigma)$.
\item (\textbf{Finite limit point for zero variance}) When all $u_i\neq0\ (i\in[d])$, $\rho(\frac{\bfu}{\sigma},\sigma)$ converges to a strictly positive real function of $\bfu$ as $\sigma$ goes to 0, i.e. $\lim_{\sigma\rightarrow0}\rho(\frac{\bfu}{\sigma},\sigma):=\mathcal{C}_s(\bfu)$. When $u_i=0$ for some $i\in[d]$, $\lim_{\sigma\rightarrow 0}\rho(\frac{\bfu}{\sigma},\sigma)=0$.
\item (\textbf{Lower bound function of the mean}) When everything else except $|u_i|$ is fixed, $\rho(\frac{{\bfW^*}^\top\bfu}{\sigma\delta_K(\bfW^*)},\sigma\delta_K(\bfW^*))$ is lower bounded by a strictly positive real function, $\mathcal{L}_m(\frac{(\bfLM{\bfW^*})^\top\bfLM\bfu}{\sigma\delta_K(\bfW^*)}, \sigma\delta_K(\bfW^*))$, which is monotonically decreasing as $|u_i|$ increases. 
\item (\textbf{Lower bound function of the variance}) When everything else except $\sigma$ is fixed, $\rho(\frac{{\bfW^*}^\top\bfu}{\sigma\delta_K(\bfW^*)},\sigma\delta_K(\bfW^*))$ is lower bounded by a strictly positive real function, $\mathcal{L}_s(\frac{{\bfW^*}^\top\bfu}{\sigma\delta_K(\bfW^*)},\sigma\delta_K(\bfW^*))$, which satisfies the following conditions: (a) there exists $\zeta_{s'}>0$, such that $\sigma^{-1}\mathcal{L}_s(\frac{{\bfW^*}^\top\bfu}{\sigma\delta_K(\bfW^*)},\sigma\delta_K(\bfW^*))$ is an increasing function of $\sigma$ when $\sigma\in(0,\zeta_{s'})$; (b) there exists $\zeta_s>0$ such  that $\mathcal{L}_s(\frac{{\bfW^*}^\top\bfu}{\sigma\delta_K(\bfW^*)},\sigma\delta_K(\bfW^*))$ is a decreasing function of $\sigma$ when $\sigma\in(\zeta_s,+\infty)$.
  \end{enumerate}

\end{property}
\noindent \textbf{Proof:}\\
(1) From Cauchy Schwarz's inequality, we have
\begin{equation}
    \mathbb{E}_{z_i\sim\mathcal{N}(u_i,1)}[\phi'(\sigma\cdot z_i)]\leq \sqrt{\mathbb{E}_{z_i\sim\mathcal{N}(u_i,1)}[\phi'^2(\sigma\cdot z_i)]}\label{rho_ineq1}
\end{equation}
\begin{equation}
\begin{aligned}
    \mathbb{E}_{z_i\sim\mathcal{N}(u_i,1)}[\phi'(\sigma\cdot z_i)z_i\cdot z_i]&\leq \sqrt{\mathbb{E}_{z_i\sim\mathcal{N}(u_i,1)}[\phi'^2(\sigma\cdot z_i)z_i^2]}\cdot\sqrt{\mathbb{E}_{z_i\sim\mathcal{N}(u_i,1)}[z_i^2]}\\
    &=\sqrt{\mathbb{E}_{z_i\sim\mathcal{N}(u_i,1)}[\phi'^2(\sigma\cdot z_i)z_i^2]}\cdot\sqrt{u_i^2+1}\label{rho_ineq2}
\end{aligned}
\end{equation}
The equalities of the (\ref{rho_ineq1}) and (\ref{rho_ineq2}) hold if and only if $\phi'$ is a constant function. Since that $\phi$ is the sigmoid function, the equalities of (\ref{rho_ineq1}) and (\ref{rho_ineq2}) cannot hold. \\
By the definition of $\rho(\bfu,\sigma)$ in Definition \ref{def: rho}, we have 
\begin{equation}
    \beta_0(i,\bfu,\sigma)-\alpha_0^2(i,\bfu,\sigma)>0, 
\end{equation}
\begin{equation}
    \beta_2(i,\bfu,\sigma)-\frac{\alpha_2^2(i,\bfu,\sigma)}{u_i^2+1}>0.
\end{equation}
Therefore,
\begin{equation}
    \rho(\bfu,\sigma)>0\label{rho_positive}
\end{equation}
(2) We can derive that
\begin{equation}
\begin{aligned}
    &\lim_{u_i\rightarrow0}(\frac{u_j^2}{\sigma^2}+1)\big(\beta_0(i,\bfu,\sigma)-\alpha_0^2(i,\bfu,\sigma)\big)\\
    =&\lim_{u_i\rightarrow0}(\frac{u_j^2}{\sigma^2}+1)\big(\int_{-\infty}^\infty\phi'^2(\sigma \cdot z_i)(2\pi)^{-\frac{1}{2}}\exp(-\frac{\|z_i-u_i\|^2}{2})dz_i\\
    &-(\int_{-\infty}^\infty\phi'(\sigma \cdot z_i)(2\pi)^{-\frac{1}{2}}\exp(-\frac{\|z_i-u_i\|^2}{2})dz_i)^2\big)\\
    =& (\frac{u_j^2}{\sigma^2}+1)\big(\int_{-\infty}^\infty\phi'^2(\sigma \cdot z_i)(2\pi)^{-\frac{1}{2}}\exp(-\frac{\|z_i\|^2}{2})dz_i
    -(\int_{-\infty}^\infty\phi'(\sigma \cdot z_i)(2\pi)^{-\frac{1}{2}}\exp(-\frac{\|z_i\|^2}{2})dz_i)^2\big),\label{rho1_u}
\end{aligned}
\end{equation}
where the first step is by Definition \ref{def: rho}, and the second step comes from the limit laws. Similarly, we also have
\begin{equation}
\begin{aligned}
    &\lim_{u_i\rightarrow0}\big(\beta_2(i,\bfu,\sigma)-\frac{1}{u_i^2+1}\alpha_2^2(i,\bfu,\sigma)\big)\\
    =&\lim_{u_i\rightarrow0} \int_{-\infty}^\infty\phi'^2(\sigma \cdot z_i)z_i^2(2\pi)^{-\frac{1}{2}}\exp(-\frac{\|z_i-u_i\|^2}{2})dz_i\\
    &-(\frac{1}{u_i^2+1}\int_{-\infty}^\infty\phi'(\sigma \cdot z_i)z_i^2(2\pi)^{-\frac{1}{2}}\exp(-\frac{\|z_i-u_i\|^2}{2})dz_i)^2\\
    =& \int_{-\infty}^\infty\phi'^2(\sigma \cdot z_i)z_i^2(2\pi)^{-\frac{1}{2}}\exp(-\frac{\|z_i\|^2}{2})dz_i
    -(\int_{-\infty}^\infty\phi'(\sigma \cdot z_i)z_i^2(2\pi)^{-\frac{1}{2}}\exp(-\frac{\|z_i\|^2}{2})dz_i)^2\label{rho2_u}
\end{aligned}
\end{equation}\\
Since that (\ref{rho1_u}) and (\ref{rho2_u}) are positive due to Jensen's inequality, we can derive that $\rho(\bfu,\sigma)$ converges to a positive value function of $\sigma$ as $u_i$ goes to 0, i.e. 
\begin{equation}
    \lim_{u\rightarrow0}\rho(\bfu,\sigma):=\mathcal{C}_m(\sigma)
\end{equation}\\
(3) When all $u_i\neq0\ (i\in[d])$, 
\begin{equation}
\begin{aligned}
    &\lim_{\sigma\rightarrow0}\big(\beta_2(i,\frac{\bfu}{\sigma},\sigma)-\frac{1}{\frac{u_i^2}{\sigma^2}+1}\alpha_2^2(i,\frac{\bfu}{\sigma},\sigma)\big)\\
    =&\lim_{\sigma\rightarrow0} \int_{-\infty}^\infty\phi'^2(\sigma \cdot z_i)z_i^2(2\pi)^{-\frac{1}{2}}\exp(-\frac{\|z_i-\frac{u_i}{\sigma}\|^2}{2})dz_i\\
    &-\frac{1}{\frac{u_i^2}{\sigma^2}+1}\big(\int_{-\infty}^\infty\phi'(\sigma \cdot z_i)z_i^2(2\pi)^{-\frac{1}{2}}\exp(-\frac{\|z_i-\frac{u_i}{\sigma}\|^2}{2})dz_i\big)^2\\
    =&\lim_{\sigma\rightarrow0} \int_{-\infty}^\infty\phi'^2(u_i \cdot x_i)\frac{u_i^2}{\sigma^2}x_i^2(2\pi\frac{\sigma^2}{u_i^2})^{-\frac{1}{2}}\exp(-\frac{\|x_i-1\|^2}{2\frac{\sigma^2}{u_i^2}})dx_i\\
    &-\frac{1}{\frac{u_i^2}{\sigma^2}+1}\big(\int_{-\infty}^\infty\phi'(u_i \cdot x_i)\frac{u_i^2}{\sigma^2}x_i^2(2\pi\frac{\sigma^2}{u_i^2})^{-\frac{1}{2}}\exp(-\frac{\|x_i-1\|^2}{2\frac{\sigma^2}{u_i^2}})dx_i\big)^2\ \ \ \ \ \ \ \ \ z_i=\frac{u_i}{\sigma}x_i\\
    =& \lim_{\sigma\rightarrow0}\phi'^2(u_i)\frac{u_i^2}{\sigma^2}-\frac{1}{\frac{u_i^2}{\sigma^2}+1}(\phi'(u_i)\frac{u_i^2}{\sigma^2})^2\\
    =& \lim_{\sigma\rightarrow0}\phi'^2(u_i)\frac{u_i^2}{\sigma^2}\big(1-\frac{\frac{u_i^2}{\sigma^2}}{1+\frac{u_i^2}{\sigma^2}}\big)\\
    =&\lim_{\sigma\rightarrow0}\phi'^2(u_i)\frac{1}{1+\frac{\sigma^2}{u_i^2}}\\
    =&\phi'^2(u_i)\label{rho2_sigma}
\end{aligned}
\end{equation}\\
The first step of (\ref{rho2_sigma}) comes from Definition \ref{def: rho}. The second step and the last three steps are derived from some basic mathematical computation and the limit laws. The third step of (\ref{rho2_sigma}) is by the fact that the Gaussian distribution goes to a Dirac delta function when $\sigma$ goes to $0$. Then the integral will take the value when $x_i=1$. Similarly, we can obtain the following\\
\begin{equation}
\begin{aligned}
    &\lim_{\sigma\rightarrow0}\big(\beta_0(i,\frac{\bfu}{\sigma},\sigma)-\alpha_0^2(i,\frac{\bfu}{\sigma},\sigma)\big)\\
    =&\lim_{\sigma\rightarrow0} \int_{-\infty}^\infty\phi'^2(\sigma \cdot z_i)(2\pi)^{-\frac{1}{2}}\exp(-\frac{\|z_i-\frac{u_i}{\sigma}\|^2}{2})dz_i\\
    &-\big(\int_{-\infty}^\infty\phi'(\sigma \cdot z_i)(2\pi)^{-\frac{1}{2}}\exp(-\frac{\|z_i-\frac{u_i}{\sigma}\|^2}{2})dz_i\big)^2\\
    =& \phi'^2(u_i)-\phi'^2(u_i)=0\label{rho1_sigma_1}
\end{aligned}
\end{equation}
\begin{equation}
    \begin{aligned}
        &\lim_{\sigma\rightarrow0}\Big(\frac{\partial}{\partial\sigma}\big(\beta_0(i,\frac{\bfu}{\sigma},\sigma)-\alpha_0^2(i,\frac{\bfu}{\sigma},\sigma)\big)\Big)\\
        =& \lim_{\sigma\rightarrow0}\Big(\frac{\partial}{\partial\sigma} \Big(\int_{-\infty}^\infty\phi'^2(x_i)(2\pi\sigma^2)^{-\frac{1}{2}}\exp(-\frac{\|x_i-u_i\|^2}{2\sigma^2})dx_i\\
        &-\big(\int_{-\infty}^\infty\phi'(x_i)(2\pi\sigma^2)^{-\frac{1}{2}}\exp(-\frac{\|x_i-u_i\|^2}{2\sigma^2})dx_i\big)^2\Big)\Big)\ \ \ \ \ \ \ \ x_i=\sigma\cdot z_i\\
        =&  \lim_{\sigma\rightarrow0}\Big(\int_{-\infty}^\infty\phi'^2(x_i)(2\pi\sigma^2)^{-\frac{1}{2}}\exp(-\frac{\|x_i-u_i\|^2}{2\sigma^2})(-\sigma^{-1}+\|x_i-u_i\|^2\sigma^{-2})dx_i\\
        &-2\big(\int_{-\infty}^\infty\phi'(x_i)(2\pi\sigma^2)^{-\frac{1}{2}}\exp(-\frac{\|x_i-u_i\|^2}{2\sigma^2})dx_i\big)\\
        &\cdot\int_{-\infty}^\infty\phi'(x_i)(2\pi\sigma^2)^{-\frac{1}{2}}\exp(-\frac{\|x_i-u_i\|^2}{2\sigma^2})(-\sigma^{-1}+\|x_i-u_i\|^2\sigma^{-2})dx_i\Big)\\
        =& \lim_{\sigma\rightarrow0}\Big( \frac{\phi'^2(u_i)}{-\sigma}-2\phi'(u_i)\frac{\phi'(u_i)}{-\sigma}\Big)\\
        =&\lim_{\sigma\rightarrow0}\frac{\phi'^2(u_i)}{\sigma}=+\infty
    \end{aligned}\label{rho1_sigma_2}
\end{equation}
Therefore, by L'Hopital's rule and (\ref{rho1_sigma_1}), (\ref{rho1_sigma_2}), we have 
\begin{equation}
    \begin{aligned}
        &\lim_{\sigma\rightarrow0}(\frac{u_j^2}{\sigma^2}+1)(\beta_0(i,\frac{\bfu}{\sigma},\sigma)-\alpha_0(i,\frac{\bfu}{\sigma},\sigma))\\
        =&\lim_{\sigma\rightarrow0}\frac{u_i^2}{2\sigma}\frac{\partial}{\partial\sigma}(\beta_0(i,\frac{\bfu}{\sigma},\sigma)-\alpha_0(i,\frac{\bfu}{\sigma},\sigma))\\
        =&+\infty\label{rho1_sigma}
    \end{aligned}
\end{equation}
Combining (\ref{rho1_sigma}) and (\ref{rho2_sigma}), we can derive that $\rho(\frac{\bfu}{\sigma},\sigma)$ converges to a positive value function of $\bfu$ as $\sigma$ goes to 0, i.e.
\begin{equation}
    \lim_{\sigma\rightarrow0}\rho(\frac{\bfu}{\sigma},\sigma):=\mathcal{C}_s(\bfu).
\end{equation}
When $u_i=0$ for some $i\in[d]$, $\lim_{\sigma\rightarrow 0}(\frac{u_i^2}{\sigma^2}+1)(\beta_0(j,\frac{\bfu}{\sigma},\sigma)-\alpha^2(j,\frac{\bfu}{\sigma},\sigma))=0$ by (\ref{rho1_sigma_1}). Then from the Definition \ref{def: rho}, we have 
\begin{equation}
    \lim_{\sigma\rightarrow0}\rho(\frac{\bfu}{\sigma},\sigma)=0
\end{equation}
(4) We can define $\mathcal{L}_m(\frac{(\bfLM\bfW^*)^\top\bfLM\bfu}{\sigma\delta_K(\bfW^*)},\sigma\delta_K(\bfW^*))$ as
\begin{equation}
    \mathcal{L}_m(\frac{(\bfLM\bfW^*)^\top\bfLM\bfu}{\sigma\delta_K(\bfW^*)},\sigma\delta_K(\bfW^*))=\min_{v_i\in[0,u_i]}\Big\{\rho(\frac{(\bfLM\bfW^*)^\top\bfLM\bfv}{\sigma_l\delta_K(\bfW^*)},\sigma\delta_K(\bfW^*)): v_j=u_j\text{ for all }j\neq i\Big\}
\end{equation}
Then by this definition, we have
\begin{equation}
    0<\mathcal{L}_m(\frac{(\bfLM\bfW^*)^\top\bfLM\bfu}{\sigma\delta_K(\bfW^*)},\sigma\delta_K(\bfW^*))\leq \rho(\frac{(\bfLM\bfW^*)^\top\bfLM\bfu}{\sigma_l\delta_K(\bfW^*)},\sigma\delta_K(\bfW^*))
\end{equation}
Meanwhile, for any $0\leq u_i'\leq u_i^*$, since that $[0,u_i']\subset[0,u_i^*]$, we can obtain
\begin{equation}
\mathcal{L}_m(\frac{(\bfLM\bfW^*)^\top\bfLM\bfu}{\sigma\delta_K(\bfW^*)},\sigma\delta_K(\bfW^*))|_{u_i=u_i'}\geq \mathcal{L}_m(\frac{(\bfLM\bfW^*)^\top\bfLM\bfu}{\sigma\delta_K(\bfW^*)},\sigma\delta_K(\bfW^*))|_{u_i=u_i^*}\label{Lm}
\end{equation}
Hence, $\mathcal{L}_m(\frac{(\bfLM\bfW^*)^\top\bfLM\bfu}{\sigma\delta_K(\bfW^*)},\sigma\delta_K(\bfW^*))$ is a strictly positive real function which is monotonically decreasing.\\
(5) Therefore, we only need to show the condition (a).\\
When $({\bfW^*}^\top\bfu)_i\neq0$ for all $i\in[K]$, \begin{equation}\lim_{\sigma\rightarrow 0}\rho(\frac{{\bfW^*}^\top\bfu}{\sigma\delta_K(\bfW^*)},\sigma\delta_K(\bfW^*))=\mathcal{C}_s(\bfu)>0.\end{equation} Therefore, there exists $\zeta_s>0$, such that when $0<\sigma<\zeta_s$, \begin{equation}\rho(\frac{{\bfW^*}^\top\bfu}{\sigma\delta_K(\bfW^*)},\sigma\delta_K(\bfW^*))>\frac{\mathcal{C}_s({\bfW^*}^\top\bfu)}{2}.\end{equation} Then we can define \begin{equation}
    \mathcal{L}_s(\frac{{\bfW^*}^\top\bfu}{\sigma\delta_K(\bfW^*)},\sigma\delta_K(\bfW^*)):=\frac{\mathcal{C}_s({\bfW^*}^\top\bfu)}{2\zeta_s}\sigma^2
\end{equation} such that $\sigma^{-1}\mathcal{L}_s(\frac{{\bfW^*}^\top\bfu}{\sigma\delta_K(\bfW^*)},\sigma\delta_K(\bfW^*))$ is an increasing function of $\sigma$ below $\rho(\frac{{\bfW^*}^\top\bfu}{\sigma\delta_K(\bfW^*)},\sigma\delta_K(\bfW^*))$.\\
When $({\bfW^*}^\top\bfu)_i=0$ for some $i\in[K]$, then \begin{equation}\lim_{\sigma\rightarrow0}\rho(\frac{{\bfW^*}^\top\bfu}{\sigma\delta_K(\bfW^*)},\sigma\delta_K(\bfW^*))=0.\end{equation} We can define
\begin{equation}
    \mathcal{L}_s(\frac{{\bfW^*}^\top\bfu}{\sigma\delta_K(\bfW^*)},\sigma\delta_K(\bfW^*))=\sigma\cdot\min_{v_i\in[u_i,\zeta_{s'}]}\Big\{\rho(\frac{{\bfW^*}^\top\bfv}{\sigma\delta_K(\bfW^*)},\sigma\delta_K(\bfW^*)):  v_j\neq u_j\text{ for all }j\neq i\Big\}\label{Ls}
\end{equation}
Then,  
\begin{equation}
    \sigma^{-1}\mathcal{L}_s(\frac{{\bfW^*}^\top\bfu}{\sigma\delta_K(\bfW^*)},\sigma\delta_K(\bfW^*))=\min_{v_i\in[u_i,\zeta_{s'}]}\Big\{\rho(\frac{{\bfW^*}^\top\bfv}{\sigma\delta_K(\bfW^*)},\sigma\delta_K(\bfW^*)):  v_j=u_j\text{ for all }j\neq i\Big\}
\end{equation}
For any $0\leq u_i'\leq u_i^*< \zeta_{s'}$, since that $[u_i^*,\zeta_{s'}]\subset[u_i',\zeta_{s'}]$, we can obtain
\begin{equation}
\sigma^{-1}\mathcal{L}_s(\frac{{\bfW^*}^\top\bfu}{\sigma\delta_K(\bfW^*)},\sigma\delta_K(\bfW^*))|_{u_i=u_i'}\leq \sigma^{-1}\mathcal{L}_s(\frac{{\bfW^*}^\top\bfu}{\sigma\delta_K(\bfW^*)},\sigma\delta_K(\bfW^*))|_{u_i=u_i^*}
\end{equation}
Therefore, we can derive that $\sigma^{-1}\mathcal{L}_s(\frac{{\bfW^*}^\top\bfu}{\sigma\delta_K(\bfW^*)},\sigma\delta_K(\bfW^*))$ is monotonically increasing. Following the steps in (4), we can have that $\sigma^{-1}\mathcal{L}_s(\frac{{\bfW^*}^\top\bfu}{\sigma\delta_K(\bfW^*)},\sigma\delta_K(\bfW^*))$ is a strictly positive real function which is upper bounded by $\rho(\frac{{\bfW^*}^\top\bfu}{\sigma\delta_K(\bfW^*)},\sigma\delta_K(\bfW^*))$.\\
In conclusion, condition (a) is proved.\\
For condition (b), since that $\zeta_s>0$, $\rho(\frac{{\bfW^*}^\top\bfu}{\sigma_l\delta_K(\bfW^*)},\sigma\delta_K(\bfW^*))$ is continuous and positive, we can obtain
\begin{equation}
    \rho(\frac{{\bfW^*}^\top\bfv}{\sigma\delta_K(\bfW^*)},\sigma\delta_K(\bfW^*))\Big|_{\sigma=\zeta_s}>0
\end{equation}
Then condition (b) can be easily proved as in (4). 

We then characterize the order of the $\rho$ function in different cases as follows. 
\begin{property}\label{prop: rho_simple}
  To specify the order with regard to the distribution parameters, $\rho(\bfu,\sigma)$ in Definition \ref{def: rho}  satisfies the following properties,
\begin{enumerate}
    \item (\textbf{Small variance}) $\lim_{\sigma\rightarrow 0^+}\rho(\bfu,\sigma)=\Theta(\sigma^4)$.
    \item (\textbf{Large variance}) For any $\epsilon>0$, $\lim_{\sigma\rightarrow \infty}\rho(\bfu,\sigma)\geq\Theta(\frac{1}{\sigma^{3+\epsilon}})$.
    \item (\textbf{Large mean}) For any $\epsilon>0$, $\lim_{\mu\rightarrow \infty}\rho(\bfu,\sigma)\geq\Theta(e^{-\frac{\|\bfu\|^2}{2}})\frac{1}{\|\bfu\|^{3+\epsilon}}$.
\end{enumerate}
\end{property}
\noindent\textbf{Proof:}\\
(1) 
\begin{equation}
    \begin{aligned}
    &\beta_0(i,\bfu,\sigma)-\alpha_0(i,\bfu,\sigma)^2\\
    =&\mathbb{E}_{z\sim\mathcal{N}(\mu,1)}[{\phi'}^2(\sigma\cdot z)]-(\mathbb{E}_{z\sim\mathcal{N}(\mu,1)}[{\phi'}(\sigma\cdot z)])^2\\
    =&\int_{-\infty}^\infty {\phi'}^2(\sigma\cdot z)\frac{1}{\sqrt{2\pi}}e^{-\frac{(z-\mu)^2}{2}}dz-(\int_{-\infty}^\infty {\phi'}(\sigma\cdot z)\frac{1}{\sqrt{2\pi}}e^{-\frac{(z-\mu)^2}{2}}dz)^2\\
    =&\int_{-\infty}^\infty (\frac{1}{4}-\frac{ t^2}{16}+\frac{ t^4}{96}\cdots)^2\frac{1}{\sqrt{2\pi}\sigma}e^{-\frac{(t-\mu\sigma)^2}{2\sigma^2}}dt\\
    &-(\int_{-\infty}^\infty (\frac{1}{4}-\frac{ t^2}{16}+\frac{ t^4}{96}+\cdots)\frac{1}{\sqrt{2\pi}\sigma}e^{-\frac{(t-\mu\sigma)^2}{2\sigma^2}}dt)^2\\
    =& (\frac{1}{16}-\frac{1}{32}(\mu^2\sigma^2+\sigma^2)+\frac{7}{768}(3\sigma^4+6\mu^2\sigma^4+\mu^4\sigma^4)+\cdots)\\
    &-(\frac{1}{4}-\frac{\mu^2\sigma^2+\sigma^2}{16}+\frac{3\sigma^4+6\mu^2\sigma^4+\mu^4\sigma^4}{192}+\cdots)^2\\
    =& \frac{1}{128}\sigma^4+\frac{\mu^2\sigma^4}{64}+o(\sigma^4), \ \ \ \ \text{as }\sigma\rightarrow 0^+.
    \end{aligned}\label{small_sigma_beta0-alpha0}
\end{equation}
The first step of (\ref{small_sigma_beta0-alpha0}) is by Definition \ref{def: rho}. The second step and the last steps come from some basic mathematical computation. The third step is from Taylor expansion. Hence,
\begin{equation}
    \lim_{\sigma\rightarrow 0^+}(\beta_0(i,\bfu,\sigma)-\alpha_0(i,\bfu,\sigma)^2)=\frac{1}{128}\sigma^4+\frac{\mu^2\sigma^4}{64}+o(\sigma^4)
\end{equation}
Similarly, we can obtain
\begin{equation}
    \begin{aligned}
    &\beta_2(i,\bfu,\sigma)-\frac{\alpha_2(i,\bfu,\sigma)^2}{\mu^2+1}\\
    =&\mathbb{E}_{z\sim\mathcal{N}(0,1)}[{\phi'}^2(\sigma\cdot z)z^2]-\frac{(\mathbb{E}_{z\sim\mathcal{N}(0,1)}[{\phi'}(\sigma\cdot z)z^2])^2}{\mu^2+1}\\
    =&\int_{-\infty}^\infty {\phi'}^2(\sigma\cdot z)z^2\frac{1}{\sqrt{2\pi}}e^{-\frac{(z-\mu)^2}{2}}dz-\frac{1}{\mu^2+1}(\int_{-\infty}^\infty {\phi'}(\sigma\cdot z)z^2\frac{1}{\sqrt{2\pi}}e^{-\frac{(z-\mu)^2}{2}}dz)^2\\
    =&\int_{-\infty}^\infty (\frac{t}{4\sigma}-\frac{ t^3}{16\sigma}+\frac{ t^5}{96\sigma}\cdots)^2\frac{1}{\sqrt{2\pi}\sigma}e^{-\frac{(t-\mu\sigma)^2}{2\sigma^2}}dt\\
    &-\frac{1}{\mu^2+1}(\int_{-\infty}^\infty (\frac{t^2}{4\sigma^2}-\frac{ t^4}{16\sigma^2}+\frac{ t^6}{96\sigma^2}+\cdots)\frac{1}{\sqrt{2\pi}\sigma}e^{-\frac{(t-\mu\sigma)^2}{2\sigma^2}}dt)^2\\
    =& (\frac{1+\mu^2}{16}-\frac{3\sigma^2+6\mu^2\sigma^2+\mu^4\sigma^2}{32}+\cdots)\\
    &-\frac{1}{\mu^2+1}(\frac{1+\mu^2}{4}-\frac{15\sigma^2+45\mu^2\sigma^2+15\mu^4\sigma^2+\mu^6\sigma^2}{32}+\cdots)^2\\
    =&\frac{9}{64}\sigma^2+\frac{33}{64}\mu^2\sigma^2+\frac{13}{64}\mu^4\sigma^2+\frac{1}{64}\mu^6\sigma^2+o(\sigma^2), \ \ \ \ \text{as }\sigma\rightarrow 0^+
    \end{aligned}
\end{equation}
Hence,
\begin{equation}
    \lim_{\sigma\rightarrow 0^+}(\beta_2(i,\bfu,\sigma)-\frac{\alpha_2(i,\bfu,\sigma)^2}{\mu^2+1})=\frac{9}{64}\sigma^2+o(\sigma^2)
\end{equation}
Therefore, 
\begin{equation}
    \lim_{\sigma\rightarrow 0^+}\rho(\bfu,\sigma)=\min_{j\in[d], u_j\neq \mu}\{(u_j^2+1)\}\frac{1}{128}\sigma^4
\end{equation}
(2) 
Note that by some basic mathematical derivation,
\begin{equation}
\begin{aligned}
    \int_{-\infty}^\infty {\phi'}^2(\sigma\cdot z)\frac{1}{\sqrt{2\pi}}e^{-\frac{(z-\mu)^2}{2}}dz&=\int_{-\infty}^\infty \frac{1}{(e^{\sigma\cdot z}+e^{-\sigma\cdot z}+2)^2}\frac{1}{\sqrt{2\pi}}e^{-\frac{(z-\mu)^2}{2}}dz\\
    &\geq 2\int_{0}^\infty \frac{1}{16e^{2\sigma\cdot z}}\frac{1}{\sqrt{2\pi}}e^{-\frac{(z+|\mu|)^2}{2}}dz\\
    &= \frac{1}{8}e^{2|\mu|\sigma+2\sigma^2}\int_{0}^\infty \frac{1}{\sqrt{2\pi}}e^{-\frac{(z+2\sigma)^2}{2}}dz\\
    &= \frac{1}{8\sqrt{2\pi}}e^{2|\mu|\sigma+2\sigma^2}\int_{|\mu|+2\sigma}^\infty e^{-\frac{t^2}{2}}dt\\
\end{aligned}\label{beta0_special}
\end{equation}
We then provide the following Claim with its proof to give a lower bound for (\ref{beta0_special}).\\
\textbf{Claim}: $\int_{|\mu|+2\sigma}^\infty e^{-\frac{t^2}{2}}dt> e^{-2|\mu|\sigma-2\sigma^2-k_1\log \sigma}$ for $k_1>1$.

\textbf{Proof:} Let \begin{equation}f(\sigma)=\int_{|\mu|+2\sigma}^\infty e^{-\frac{t^2}{2}}dt-e^{-2|\mu|\sigma-2\sigma^2-k_1\log \sigma}.\end{equation} Then, \begin{equation}f'(\sigma)=e^{-2\sigma^2}((2|\mu|+4\sigma+\frac{k_1}{\sigma})\sigma^{-k_1}-2e^{-\frac{1}{2}\mu^2}).\end{equation} It can be easily verified that for a given $|\mu|\geq0$, $f'(\sigma)<0$ when $\sigma$ is large enough if $k_1>1$. Combining that $\lim_{\sigma\rightarrow\infty} f(\sigma)=0$, we have $f(\sigma)>0$ when $\sigma$ is large enough by showing the contradiction in the following: \\
Suppose there is a strictly increasing function $f(x)>0$ with $\lim_{x\rightarrow \infty}f(x)=0$ when $x$ is large enough. Then there exists $x_0>0$ such that for any $\epsilon>0$, $f(x)<\epsilon$ for $x>x_0$. Pick $\epsilon=f(x_0)>0$, then for $x_1>x_0$, $f(x_1)>f(x_0)=\epsilon$. Contradiction!\\
Similarly, we also have \begin{equation}
\begin{aligned}
    \int_{-\infty}^\infty {\phi'}(\sigma\cdot z)\frac{1}{\sqrt{2\pi}}e^{-\frac{z^2}{2}}dz&=\int_{-\infty}^\infty \frac{1}{e^{\sigma\cdot z}+e^{-\sigma\cdot z}+2}\frac{1}{\sqrt{2\pi}}e^{-\frac{(z-\mu)^2}{2}}dz\\
    &\leq 2\int_{0}^\infty \frac{1}{e^{\sigma\cdot z}}\frac{1}{\sqrt{2\pi}}e^{-\frac{(z-\mu)^2}{2}}dz\\
    &= e^{|\mu|\sigma+\frac{1}{2}\sigma^2}\int_{0}^\infty \frac{2}{\sqrt{2\pi}}e^{-\frac{(z+|\mu|+\sigma)^2}{2}}dz\\
    &= \frac{2}{\sqrt{2\pi}}e^{|\mu|\sigma+\frac{1}{2}\sigma^2}\int_{|\mu|+\sigma}^\infty e^{-\frac{t^2}{2}}dt,\\
\end{aligned}\label{alpha0_special}
\end{equation}
and the \textbf{Claim}: $\int_{|\mu|+\sigma}^\infty e^{-\frac{t^2}{2}}dt< e^{-|\mu|\sigma-\frac{1}{2}\sigma^2-k_2\log \sigma}$ for $k_2\leq1$ to give an upper bound for (\ref{alpha0_special}).

Therefore, combining (\ref{beta0_special}, \ref{alpha0_special}) and two claims, we have that for any $\epsilon>0$, 
\begin{equation}
    \beta_0(i,\bfu,\sigma)-\alpha_0(i,\bfu,\sigma)^2\geq \frac{1}{8\sqrt{2\pi}}\frac{1}{\sigma^{k_1}}-\frac{1}{2\pi}\frac{1}{\sigma^{2k_2}}\gtrsim \frac{1}{\sigma^{1+\epsilon}}\label{rho_2_bound1}
\end{equation}
(The above inequality holds for any $2k_2>k_1$ where $k_1>1$ and $k_2\leq1$.)\\
Similarly,
\begin{equation}
\begin{aligned}
    \int_{-\infty}^\infty {\phi'}^2(\sigma\cdot z)z^2\frac{1}{\sqrt{2\pi}}e^{-\frac{z^2}{2}}dz&=\int_{-\infty}^\infty \frac{z^2}{(e^{\sigma\cdot z}+e^{-\sigma\cdot z}+2)^2}\frac{1}{\sqrt{2\pi}}e^{-\frac{(z-\mu)^2}{2}}dz\\
    &\geq 2\int_{0}^\infty \frac{z^2}{16e^{2\sigma\cdot z}}\frac{1}{\sqrt{2\pi}}e^{-\frac{(z+|\mu|)^2}{2}}dz\\
    &= \frac{1}{8\sqrt{2\pi}}e^{|\mu|\sigma+2\sigma^2}\int_{2|\mu|+2\sigma}^\infty (t-2\sigma)^2 e^{-\frac{t^2}{2}}dt\\
\end{aligned}
\end{equation}
\textbf{Claim}: $\int_{|\mu|+2\sigma}^\infty (t-2\sigma)^2 e^{-\frac{t^2}{2}}dt\geq e^{-2|\mu|\sigma-2\sigma^2-k_1\log \sigma}$ if $k_1>3$.\\
\textbf{Proof}: Let \begin{equation}f(\sigma)=\int_{|\mu|+2\sigma}^\infty (t-2\sigma)^2 e^{-\frac{t^2}{2}}dt- e^{-2|\mu|\sigma-2\sigma^2-k_1\log \sigma}.\end{equation}
\begin{equation}f'(\sigma)=8\sigma\int_{|\mu|+2\sigma}^\infty e^{-\frac{t^2}{2}}dt+e^{-2|\mu|\sigma-2\sigma^2}(4\sigma^{1-k_1}+k_1\sigma^{-1-k_1}+2|\mu|\sigma^{-k_1}-4e^{-\frac{1}{2}\mu^2}).\end{equation} We need $f'(\sigma)<0$ when $\sigma$ is large enough.  Since that $f'(\sigma)\rightarrow 0, f''(\sigma)\rightarrow0$ when $\sigma$ is large, we need $f''(\sigma)>0$ and $f'''(\sigma)<0$ recursively. Hence, \begin{equation}
\begin{aligned}f'''(\sigma)=&e^{-2|\mu|\sigma-2\sigma^2}(64\sigma^{3-k_1}+96\mu\sigma^{2-k_  1}+16(3k_1-3+\mu^2)\sigma^{1-k_1}+8\mu(-\mu^2-3+6k_1)\sigma^{-k_1}\\
&+4k_1(3k_1+\mu^2)\sigma^{-1-k_1}+2k_1(1+k_1)(\mu+2)\sigma^{-2-k_1}\\
&+k_1(1+k_1)(2+k_1)\sigma^{-3-k_1}-16e^{-\frac{1}{2}\mu^2})<0\end{aligned}\end{equation} requires $k_1>3$.\\
Similarly, we have
\begin{equation}
\begin{aligned}
    \int_{-\infty}^\infty {\phi'}(\sigma\cdot z)z^2\frac{1}{\sqrt{2\pi}}e^{-\frac{z^2}{2}}dz
    &\leq 2\int_{0}^\infty \frac{1}{e^{\sigma\cdot z}}\frac{1}{\sqrt{2\pi}}z^2e^{-\frac{z^2}{2}}dz
    = \frac{2}{\sqrt{2\pi}}e^{\frac{1}{2}\sigma^2}\int_{\sigma}^\infty (t-\sigma)^2e^{-\frac{t^2}{2}}dt\\
\end{aligned}
\end{equation}
and the \textbf{Claim}: $\int_{\sigma}^\infty (t-\sigma)^2 e^{-\frac{t^2}{2}}dt< e^{-\frac{\sigma^2}{2}-k_2\log \sigma}$. Hence,

\begin{equation}
    \beta_2(i,\bfu,\sigma)-\frac{\alpha_2(i,\bfu,\sigma)^2}{\mu^2+1}\geq \frac{1}{8\sqrt{2\pi}}\frac{1}{\sigma^{k_1}}-\frac{2}{\pi(\mu^2+1)}\frac{1}{\sigma^{2k_2}}\gtrsim \frac{1}{\sigma^{3.1}}\label{rho_2_bound2}
\end{equation}
(The above inequality holds for any $2k_2>k_1$ where $k_1>3$ and $k_2<3$.)\\
Therefore, by combining (\ref{rho_2_bound1}) and (\ref{rho_2_bound2}), for any $\epsilon>0$
\begin{equation}
    \lim_{\sigma\rightarrow \infty}\rho(\bfu,\sigma)\geq\Theta(\frac{1}{\sigma^{3+\epsilon}}).
\end{equation}
(3)
Let $\sigma$ be fixed. For any $\epsilon>0$, following the steps in (2), we can obtain
\begin{equation}
\begin{aligned}
    \int_{-\infty}^\infty {\phi'}^2( \sigma\cdot z)\frac{1}{\sqrt{2\pi}}e^{-\frac{(z-\mu)^2}{2}}dz&=\int_{-\infty}^\infty \frac{1}{(e^{\sigma\cdot z}+e^{-\sigma\cdot  z}+2)^2}\frac{1}{\sqrt{2\pi}}e^{-\frac{(z-\mu)^2}{2}}dz\\
    &\geq 2\int_{0}^\infty \frac{1}{16e^{2 \sigma\cdot z}}\frac{1}{\sqrt{2\pi}}e^{-\frac{(z+|\mu|)^2}{2}}dz\\
    &= \frac{1}{8\sqrt{2\pi}}e^{2|\mu|\sigma+2\sigma^2}\int_{|\mu|+2\sigma}^\infty e^{-\frac{t^2}{2}}dt\\
    &\geq \frac{1}{8\sqrt{2\pi}}e^{-\frac{\mu^2}{2}}\frac{1}{\mu^{1+\epsilon}}
\end{aligned}
\end{equation}
\begin{equation}
\begin{aligned}
    \int_{-\infty}^\infty {\phi'}( \sigma\cdot z)\frac{1}{\sqrt{2\pi}}e^{-\frac{(z-\mu)^2}{2}}dz&=\int_{-\infty}^\infty \frac{1}{e^{\sigma\cdot z}+e^{-\sigma\cdot z}+2}\frac{1}{\sqrt{2\pi}}e^{-\frac{(z-\mu)^2}{2}}dz\\
    &\leq 2\int_{0}^\infty \frac{1}{e^{\sigma\cdot z}}\frac{1}{\sqrt{2\pi}}e^{-\frac{(z-\mu)^2}{2}}dz\\
    &= \frac{2}{\sqrt{2\pi}}e^{-\frac{\mu^2}{2}}\frac{1}{\mu^{1-\epsilon}}\\
\end{aligned}
\end{equation}
Similarly, 
\begin{equation}
\begin{aligned}
    \int_{-\infty}^\infty {\phi'}^2( \sigma\cdot z)z^2\frac{1}{\sqrt{2\pi}}e^{-\frac{(z-\mu)^2}{2}}dz\geq \frac{1}{8\sqrt{2\pi}}e^{-\frac{\mu^2}{2}}\frac{1}{\mu^{3+\epsilon}}\\
\end{aligned}
\end{equation}
\begin{equation}
\begin{aligned}
    \int_{-\infty}^\infty {\phi'}( \sigma\cdot z)z^2\frac{1}{\sqrt{2\pi}}e^{-\frac{(z-\mu)^2}{2}}dz\leq \frac{2}{\sqrt{2\pi}}e^{-\frac{\mu^2}{2}}\frac{1}{\mu^{3-\epsilon}}\\
\end{aligned}
\end{equation}
We can conclude that $\lim_{\mu\rightarrow \infty}\rho(\bfu,\sigma)\geq\Theta(e^{-\frac{\|\bfu\|^2}{2}})\frac{1}{\|\bfu\|^{3+\epsilon}}$.

\begin{property}
\label{prop: even}
\normalsize If a function $f(\bfx)$ is an even function, then 
\begin{equation}\mathbb{E}_{\bfx\sim\mathcal{N}(\bfmu,\bfSg)}[f(\bfx)]=\mathbb{E}_{\bfx\sim\frac{1}{2}\mathcal{N}(\bfmu,\bfSg)+\frac{1}{2}\mathcal{N}(-\bfmu,\bfSg)}[f(\bfx)]
\end{equation}
\end{property}
\noindent \textbf{Proof: }\\
Denote \begin{equation}g(\bfx)=f(\bfx)(2\pi|\bfSg|^2)^{-\frac{d}{2}}\exp(-\frac{1}{2}(\bfx-\bfmu)\bfSg^{-1}(\bfx-\bfmu))
\end{equation}
By some basic mathematical computation, 
\begin{equation}
\begin{aligned}
    \mathbb{E}_{\bfx\sim\mathcal{N}(\bfmu,\bfSg)}[f(\bfx)]&=\int_{\bfx\in\mathbb{R}^d}g(\bfx)d\bfx=\int_{-\infty}^\infty\cdots\int_{-\infty}^\infty g(x_1,\cdots,x_d)dx_1\cdots dx_d\\
    &=\int_{-\infty}^\infty\cdots\int_{-\infty}^\infty\int_{\infty}^{-\infty} g(x_1,x_2,\cdots,x_d)d(-x_1)dx_2\cdots dx_d\\
    &=\int_{-\infty}^\infty\cdots\int_{-\infty}^\infty g(-x_1,x_2\cdots,x_d)dx_1dx_2\cdots dx_d\\
    &=\int_{\bfx\in\mathbb{R}^d}g(-\bfx)d\bfx\\
    &=\int_{\bfx\in\mathbb{R}^d}f(\bfx)(2\pi|\bfSg|^2)^{-\frac{d}{2}}\exp(-\frac{1}{2}(\bfx+\bfmu)\bfSg^{-1}(\bfx+\bfmu))\\
    &=\mathbb{E}_{\bfx\sim\mathcal{N}(-\bfmu,\bfSg)}[f(\bfx)]
\end{aligned}
\end{equation}
Therefore, we have \begin{equation}\mathbb{E}_{\bfx\sim\mathcal{N}(\bfmu,\bfSg)}[f(\bfx)]=\mathbb{E}_{\bfx\sim\frac{1}{2}\mathcal{N}(\bfmu,\bfSg)+\frac{1}{2}\mathcal{N}(-\bfmu,\bfSg)}[f(\bfx)]\end{equation}

\begin{property}\label{prop: bound_simple} 
\normalsize Under Gaussian Mixture Model $\bfx\sim\sum_{l=1}^L\lambda_l\mathcal{N}(\bfmu_l, \bfSg_l)$ where $\bfSg_l=\text{diag}(\sigma_{l1}^2,\cdots,\sigma_{ld}^2)$, we have the following upper bound.
\begin{equation}\mathbb{E}_{\bfx\sim\sum_{l=1}^L\lambda_l\mathcal{N}(\bfmu_l,\bfSg_l)}[(\bfu^\top \bfx)^{2t}]\leq (2t-1)!!||\bfu||^{2t}\sum_{l=1}^L \lambda_l(||\bfmu_l||+\|\bfSg_l^\frac{1}{2}\|)^{2t}\end{equation}
\end{property}
\noindent \textbf{Proof: }\\
Note that
\begin{equation}
    \mathbb{E}_{\bfx\sim\sum_{l=1}^L\lambda_l\mathcal{N}(\bfmu_l,\bfSg_l)}[(\bfu^\top \bfx)^{2t}]=\sum_{l=1}^L \lambda_l\mathbb{E}_{\bfx\sim\mathcal{N}(\bfmu_l,\bfSg_l)}[(\bfu^\top\bfx)^{2t}]=\sum_{l=1}^L \lambda_l\mathbb{E}_{y\sim\mathcal{N}(\bfu^\top\bfmu_l,\bfu^\top\bfSg_l\bfu)}[y^{2t}],
\end{equation}
where the last step is by that $\bfu^\top\bfx\sim\mathcal{N}(\bfu^\top\bfmu,\bfu^\top\bfSg_l\bfu)$ for $\bfx\sim\mathcal{N}(\bfmu_l,\bfSg_l)$. By some basic mathematical computation, we know
\begin{equation}
    \begin{aligned}
    &\mathbb{E}_{y\sim\mathcal{N}(\bfu^\top\bfmu_l, \bfu^\top\bfSg_l\bfu)}[y^{2t}]\\
   =&\int_{-\infty}^\infty (y-\bfu^\top\bfmu_l+\bfu^\top\bfmu_l)^{2t}\frac{1}{\sqrt{2\pi\bfu^\top\bfSg_l\bfu}}e^{-\frac{(y-\bfu^\top\bfmu_l)^2}{2 \bfu^\top\bfSg_l\bfu}}dy\\
   =&\int_{-\infty}^\infty \sum_{p=0}^{2t}\binom{2t}{p}(\bfu^\top\bfmu_l)^{2t-p}(y-\bfu^\top\bfmu_l)^{p}\frac{1}{\sqrt{2\pi\bfu^\top\bfSg_l\bfu}}e^{-\frac{(y-\bfu^\top\bfmu_l)^2}{2 \bfu^\top\bfSg_l\bfu}}dy\\
   =&\sum_{p=0}^{2t}\binom{2t}{p}(\bfu^\top\bfmu_l)^{2t-p}\cdot\left\{
    \begin{array}{rcl}
    0\ \ \ \ \ \ \ \ ,& p\text{ is odd}\\
    (p-1)!!(\bfu^\top\bfSg_l\bfu)^\frac{p}{2}, & p\text{ is even}
    \end{array}\right.\\
    \leq &\sum_{p=0}^{2t}\binom{2t}{p}|\bfu^\top\bfmu_l|^{2t-p}
    (p-1)!!|\bfu^\top\bfSg_l\bfu|^\frac{p}{2}\\
    \leq & (2t-1)!! (|\bfu^\top\bfmu_l|+|\bfu^\top\bfSg_l\bfu|^\frac{1}{2})^{2t}\\
    \leq & (2t-1)!!\|\bfu\|^{2t}(\|\bfmu_l\|+\|\bfSg\|^\frac{1}{2})^{2t},\\
    \end{aligned}
\end{equation}
where the second step is by the Binomial theorem. Hence,
\begin{equation}\mathbb{E}_{\bfx\sim\sum_{l=1}^L\lambda_l\mathcal{N}(\bfmu_l,\bfSg_l)}[(\bfu^\top \bfx)^{2t}]\leq (2t-1)!!||\bfu||^{2t}\sum_{l=1}^L \lambda_l(||\bfmu_l||+\|\bfSg_l^\frac{1}{2}\|)^{2t}\end{equation}

\begin{property}\label{prop: bound2} 
\normalsize With the Gaussian Mixture Model, we have
\begin{equation}\mathbb{E}_{\bfx\sim\sum_{l=1}^L\lambda_l\mathcal{N}(\bfmu_l,\bfSg_l)}[||\bfx||^{2t}]\leq d^t(2t-1)!!\sum_{l=1}^L\lambda_l(\|\bfmu_l\|+\|\bfSg_l^{\frac{1}{2}}\|)^{2t}\end{equation}
\end{property}
\noindent \textbf{Proof:}\\
\begin{equation}
\begin{aligned}
    &\mathbb{E}_{\bfx\sim\sum_{l=1}^L\lambda_l\mathcal{N}(\bfmu_l,\bfSg_l)}[||\bfx||_2^{2t}]\\
    =&\mathbb{E}_{\bfx\sim\sum_{l=1}^L\lambda_l\mathcal{N}(\bfmu_l,\bfSg_l)}[(\sum_{i=1}^d x_i^2)^t]\\
    =& \mathbb{E}_{\bfx\sim\sum_{l=1}^L\lambda_l\mathcal{N}(\bfmu_l,\bfSg_l)}[d^t(\sum_{i=1}^d \frac{x_i^2}{d})^t]\\
    \leq& \mathbb{E}_{\bfx\sim\sum_{l=1}^L\lambda_l\mathcal{N}(\bfmu_l,\bfSg_l)}[d^t\sum_{i=1}^d\frac{x_i^{2t}}{d}] \\
    =&d^{t-1}\sum_{i=1}^d\sum_{j=1}^{L}\int_{-\infty}^\infty (x_i-\mu_{ji}+\mu_{ji})^{2t}\lambda_j\frac{1}{\sqrt{2\pi}\sigma_{ji}}\exp(-\frac{(x_i-\mu_{ji})^2}{2\sigma_{ji}^2})dx_i\\
    =&d^{t-1}\sum_{i=1}^d\sum_{j=1}^{L}\sum_{k=1}^{2t}\binom{2t}{k}\lambda_j|\mu_{ji}|^{2t-k}\cdot
    \left\{
    \begin{array}{rcl}
    0\ \ \ \ \ \ \ \ ,& k\text{ is odd}\\
    (k-1)!!\sigma_{ji}^k, &k\text{ is even}
    \end{array}\right.\\
    \leq& d^{t-1}\sum_{i=1}^d\sum_{j=1}^{L}\sum_{k=1}^{2t}\binom{2t}{k}\lambda_j|\mu_{ji}|^{2t-k}\sigma_j^k\cdot(2t-1)!!\\
    =&d^{t-1}\sum_{i=1}^d\sum_{j=1}^L\lambda_j(|\mu_{ji}|+\sigma_{ji})^{2t}(2t-1)!!\\
    \leq& d^t(2t-1)!!\sum_{l=1}^L\lambda_l(\|\bfmu\|+\|\bfSg_l^{\frac{1}{2}}\|)^{2t}
\end{aligned}
\end{equation}
In the 3rd step, we apply Jensen inequality because $f(x)=x^t$ is convex when $x\geq0$ and $t\geq1$. In the 4th step we apply the Binomial theorem and the result of k-order central moment of Gaussian variable.\\

\begin{property}\label{prop: bound} 
\normalsize Under the Gaussian Mixture Model $\bfx\sim\sum_{l=1}^L\lambda_l\mathcal{N}(\bfmu_l, \bfSg_l)$ where $\bfSg_l=\bfLM_l^\top\bfD_l\bfLM_l$, we have the following upper bound.
\begin{equation}\mathbb{E}_{\bfx\sim\sum_{l=1}^L\lambda_l\mathcal{N}(\bfmu_l,\bfSg_l)}[(\bfu^\top \bfx)^{2t}]\leq (2t-1)!!||\bfu||^{2t}\sum_{l=1}^L \lambda_l(||\bfmu_l||+\|\bfSg_l^\frac{1}{2}\|)^{2t}\end{equation}
\end{property}
\noindent \textbf{Proof:}\\
If $\bfx\sim\mathcal{N}(\bfmu_l,\bfSg_l)$, then $\bfu^\top\bfx\sim\mathcal{N}(\bfu^\top\bfmu_l,\bfu^\top\bfSg_l\bfu)=\mathcal{N}((\bfLM_l\bfu)^\top \bfLM_l\bfmu_l,(\bfLM_l\bfu)^\top\bfD_l(\bfLM_l\bfu))$. 
By Property \ref{prop: bound_simple}, we have
\begin{equation}
\mathbb{E}_{\bfx\sim\mathcal{N}(\bfmu_l,\bfSg_l)}[(\bfu^\top\bfx)^{2t}]\leq (2t-1)!!\|\bfu\|^{2t}(\|\bfmu_l\|+\|\bfSg_l^\frac{1}{2}\|)^{2t}
\end{equation}
Then we can derive the final result.\\

\begin{property}
\label{prop: fx} \normalsize The population risk function  $\bar{f}(\bfW)$ is defined as 
\begin{equation}
\begin{aligned}
    &\bar{f}(\bfW)=\mathbb{E}_{\bfx\sim\sum_{l=1}^L\lambda_l\mathcal{N}(\bfmu_l,\bfSg_l)}[f_n(\bfW)]\\
    =&\mathbb{E}_{\bfx\sim\sum_{l=1}^L\lambda_l\mathcal{N}(\bfmu_l,\bfSg_l)}\Big[\frac{1}{n}\sum_{i=1}^n\ell(\bfW;\bfx_i,y_i)\Big]\\
    =&\mathbb{E}_{\bfx\sim\sum_{l=1}^L\lambda_l\mathcal{N}(\bfmu_l,\bfSg_l)}[\ell(\bfW;\bfx_i,y_i)]\label{population_loss}
\end{aligned}
\end{equation}

For any permutation matrix $\bfP$, where $\{\pi(j)\}_{j=1}^K$ is the indices permuted by $\bfP$, we have
\begin{equation}
\begin{aligned}
    H(\bfW\bfP,\bfx)&=\frac{1}{K}\sum_{\pi^*(j)}\phi({\bfw_{\pi(j)}}^\top\bfx)\\
    &=\frac{1}{K}\sum_{j=1}^K\phi({\bfw_j}^\top\bfx)\\
    &=H(\bfW,\bfx)
\end{aligned}
\end{equation}
Therefore,
\begin{equation}
    \bar{f}(\bfW)=\bar{f}(\bfW\bfP)
\end{equation}
Based on (\ref{cla_model}) and (\ref{cross-entropy}), we can derive its  gradient and Hessian as follows.
\begin{equation}
    \frac{\partial \ell(\bfW;\bfx,y)}{\partial \bfw_j}= -\frac{1}{K}\frac{y-H(\bfW)}{H(\bfW)(1-H(\bfW))}\phi'(\bfw_j^\top\bfx)\bfx=\zeta(\bfW)\cdot\bfx\label{gradient}
\end{equation}
\begin{equation}
    \frac{\partial^2 \ell(\bfW;\bfx,y)}{\partial\bfw_j\partial\bfw_l}=\xi_{j,l}\cdot\bfx\bfx^\top\label{Hessian}
\end{equation}
\begin{equation}
\begin{aligned}
\xi_{j,l}(\bfW)=\left\{
    \begin{array}{rcl}
    \frac{1}{K^2}\phi'(\bfw_j^\top \bfx)\phi'(\bfw_l^\top x)\frac{H(\bfW)^2+y-2y\cdot H(\bfW)}{H^2(\bfW)(1-H(\bfW))^2},\ \ \ \ \ \ \ \ \ \ \ \ \ \ \ \ \ \ \ \ & j\neq l\\
    \frac{1}{K^2}\phi'(\bfw_j^\top \bfx)\phi'(\bfw_l^\top \bfx)\frac{H(\bfW)^2+y-2y\cdot H(\bfW)}{H^2(\bfW)(1-H(\bfW))^2}-\frac{1}{K}\phi''(\bfw_j^\top \bfx)\frac{y-H(\bfW)}{H(\bfW)(1-H(\bfW))}, &j=l
    \end{array}\right.\label{xi}
\end{aligned}
\end{equation}\\
\end{property}

\begin{property}\label{prop: D_ineq}
With $D_m(\Psi$ defined in definition \ref{def: D}, we have
\begin{equation}
(i)\ D_m(\Psi)D_{2m}(\Psi)\leq D_{3m}(\Psi)\label{ineq_D1}
\end{equation}
\begin{equation}
    (ii)\ \big(D_m(\Psi)\big)^2\leq D_{2m}(\Psi)\label{ineq_D2}
\end{equation}
\end{property}
\noindent\textbf{Proof: }\\
To prove (\ref{ineq_D1}), we can first compare the terms $\sum_{i=1}^L \lambda_i a_i \sum_{i=1}^L \lambda_i a_i^2$ and $\sum_{i=1}^L \lambda_i a_i^3$, where $a_i\geq 1,\ i\in[L]$ and $\sum_{i=1}^L\lambda_i=1$.
\begin{equation}
    \begin{aligned}
        \sum_{i=1}^L \lambda_i a_i^3-\sum_{i=1}^L \lambda_i a_i \sum_{i=1}^L \lambda_i a_i^2 &=\sum_{i=1}^L \lambda_i a_i\cdot \big( a_i^2-\sum_{j=1}^L\lambda_j a_j^2\big)\\
        &=\sum_{i=1}^L \lambda_i a_i\cdot \big( (1-\lambda_i)a_i^2-\sum_{1\leq j\leq L, j\neq i}\lambda_j a_j^2\big)\\
        &=\sum_{i=1}^L \lambda_i a_i\cdot \big( \sum_{1\leq j\leq L, j\neq i}\lambda_j a_i^2-\sum_{1\leq j\leq L, j\neq i}\lambda_j a_j^2\big)\\
        &=\sum_{i=1}^L \lambda_i a_i\cdot \big(\sum_{1\leq j\leq L, j\neq i}\lambda_j (a_i^2-a_j^2)\big)\\
        &=\sum_{1\leq i,j\leq L, i\neq j}\big(\lambda_i\lambda_j a_i(a_i^2-a_j^2)+\lambda_i\lambda_j a_j (a_j^2-a_i^2)\big)\\
        &=\sum_{1\leq i,j\leq L, i\neq j} \lambda_i\lambda_j (a_i-a_j)^2(a_i+a_j)\geq 0
    \end{aligned}\label{D1_cmp1_derivation}
\end{equation}
The second to last step is because we can find the pairwise terms $\lambda_i a_i\cdot \lambda_j (a_i^2-a_j^2)$ and $\lambda_j a_j\cdot \lambda_i(a_j^2-a_i^2)$ in the summation that can be putted together. From (\ref{D1_cmp1_derivation}), we can obtain
\begin{equation}
    \sum_{i=1}^L\lambda_i a_i \sum_{i=1}^L\lambda_i a_i^2\leq \sum_{i=1}^L \lambda_i a_i^3\label{D1_cmp1}
\end{equation}
Combining (\ref{D1_cmp1}) and the definition of $D_m(\Psi)$ in (\ref{def: D}), we can derive (\ref{ineq_D1}).\\
Similarly, to prove (\ref{ineq_D2}), we can first compare the terms $(\sum_{i=1}^L\lambda_i a_i)^2$ and $\sum_{i=1}^L\lambda_i a_i^2$, where $a_i\geq 1,\ i\in[L]$ and $\sum_{i=1}^L\lambda_i=1$.
\begin{equation}
    \begin{aligned}
        \sum_{i=1}^L \lambda_i a_i^2-(\sum_{i=1}^L \lambda_i a_i)^2 &=\sum_{i=1}^L \lambda_i a_i\cdot \big( a_i-\sum_{j=1}^L\lambda_j a_j\big)\\
        &=\sum_{i=1}^L \lambda_i a_i\cdot \big( (1-\lambda_i)a_i-\sum_{1\leq j\leq L, j\neq i}\lambda_j a_j\big)\\
        &=\sum_{i=1}^L \lambda_i a_i\cdot \big( \sum_{1\leq j\leq L, j\neq i}\lambda_j a_i-\sum_{1\leq j\leq L, j\neq i}\lambda_j a_j\big)\\
        &=\sum_{i=1}^L \lambda_i a_i\cdot \big(\sum_{1\leq j\leq L, j\neq i}\lambda_j (a_i-a_j)\big)\\
        &=\sum_{1\leq i,j\leq L, i\neq j}\big(\lambda_i\lambda_j a_i(a_i-a_j)+\lambda_i\lambda_j a_j (a_j-a_i)\big)\\
        &=\sum_{1\leq i,j\leq L, i\neq j} \lambda_i\lambda_j (a_i-a_j)^2\geq 0
    \end{aligned}\label{D2_cmp2_derivation}
\end{equation}
The derivation of (\ref{D2_cmp2_derivation}) is close to (\ref{D1_cmp1_derivation}). By (\ref{D2_cmp2_derivation}) we have
\begin{equation}
    (\sum_{i=1}^L\lambda_i a_i )^2\leq \sum_{i=1}^L \lambda_i a_i^2\label{D2_cmp2}
\end{equation}
Combining (\ref{D2_cmp2}) and the definition of $D_m(\Psi)$ in (\ref{def: D}), we can derive (\ref{ineq_D2}).\\

\subsection{Proof of Theorem \ref{thm1} and Corollary \ref{cor}}\label{section: thm1}

 Theorem \ref{thm1} is built upon \textbf{three lemmas}. 
 
 \textbf{Lemma \ref{lemma: convexity}} shows that with $O(dK^5\log^2{d})$ samples, the empirical risk function is strongly convex in the neighborhood of   $\bfW^*$.
 
 \textbf{Lemma \ref{lemma: convergence}} shows that if initialized in the convex region, the gradient descent algorithm    converges linearly to a critical point $\widehat{\bfW}_n$, which is close to  $\bfW^*$.  
 
 \textbf{Lemma \ref{lemma: tensor bound}} shows that the Tensor Initialization Method in Subroutine \ref{TensorInitialization} initializes $\bfW_0\in\mathbb{R}^{d\times K}$ in the local convex region. Theorem 1 follows naturally by combining these three lemmas. 
  
  This proving approach is built upon those in \cite{FCL20}. One of our major technical contribution is extending Lemmas \ref{lemma: convexity} and
\ref{lemma: convergence} to the Gaussian mixture model, while the results in \cite{FCL20} only apply to Standard Gaussian models. The second major contribution is a new tensor initialization method for Gaussian mixture model such that the initial point is in the convex region (see Lemma \ref{lemma: tensor bound}). Both contributions require the development of new tools, and our analyses are much more involved than those for the standard Gaussian due to the complexity introduced by the Gaussian mixture model.

%We first introduce two notations used in the following lemmas. 
To present these lemmas, the Euclidean ball  $\mathbb{B}(\bfW^*\bfP^*,r)$ is used to denote the neighborhood of $\bfW^*\bfP^*$, where $r$ is the radius of the ball.
\begin{equation}\mathbb{B}(\bfW^*\bfP^*,r)=\{\bfW\in\mathbb{R}^{d\times K}:||\bfW-\bfW^*\bfP^*||_F\leq r\}\end{equation}

 The radius of the convex region is
\begin{equation}\label{radius}
 r:=\Theta\Big(\frac{C_3\epsilon_0\cdot\sum_{l=1}^L\lambda_l\frac{\|\bfSg_l^{-1}\|^{-1}}{\eta\tau^K\kappa^2}\rho(\frac{{\bfW^*}^\top\bfmu_l}{\delta_K(\bfW^*)\|\bfSg_l^{-1}\|^{-\frac{1}{2}}}, \delta_K(\bfW^*)\|\bfSg_l^{-1}\|^{-\frac{1}{2}})}{K^{\frac{7}{2}}\Big(\sum_{l=1}^L\lambda_l(\|\bfmu_l\|+\|\bfSg_l^\frac{1}{2}\|)^4\sum_{l=1}^L\lambda_l(\|\bfmu_l\|+\|\bfSg_l^\frac{1}{2}\|)^8\Big)^{\frac{1}{4}}}\Big)
\end{equation}
with some constant $C_3>0$.

\noindent \textbf{Proof of Theorem \ref{thm1}}\\
From Lemma \ref{lemma: convergence} and Lemma \ref{lemma: tensor bound}, we know  that if $n$ is sufficiently large such that the initialization  $\bfW_0$ by the tensor method is in the region $\mathbb{B}(\bfW^*\bfP,r)$, then the gradient descent method converges to a critical point $\widehat{\bfW}_n$ that is sufficiently close to $\bfW^*$. To achieve that, one sufficient condition is
%We require that the initialized point is located in the strong convex region. Hence,
\begin{equation}
\begin{aligned}
||\bfW_0-\bfW^*\bfP^*||_F&\leq \sqrt{K}||\bfW_0-\bfW^*\bfP^*||\leq \kappa^6 K^{\frac{7}{2}}\cdot \tau^6 \sqrt{D_6(\Psi)}\sqrt{\frac{d\log{n}}{n}}||\bfW^*\bfP||\\
&\leq \frac{C_3\epsilon_0\Gamma(\Psi)\sigma_{\max}^2}{K^{\frac{7}{2}}\Big(\sum_{l=1}^L\lambda_l(\|\bfmu_l\|+\|\bfSg_l^\frac{1}{2}\|)^4\sum_{l=1}^L\lambda_l(\|\bfmu_l\|+\|\bfSg_l^\frac{1}{2}\|)^8\Big)^{\frac{1}{4}}}
\end{aligned}
\end{equation}
where the first inequality follows from $||\bfW||_F\leq \sqrt{K}||\bfW||$ for $\bfW\in\mathbb{R}^{d\times K}$, the second inequality comes from Lemma \ref{lemma: tensor bound}, and the third inequality comes from the requirement to be    in the region $\mathbb{B}(\bfW^*\bfP,r)$. % \textbf{Lemma }\ref{lemma: convergence}.
That is equivalent to the following condition
\begin{equation}
\begin{aligned}
n\geq &C_0\epsilon_0^{-2}\cdot\tau^{12}\kappa^{12} K^{14}\Big(\sum_{l=1}^L\lambda_l(\|\bfmu_l\|+\|\bfSg_l^\frac{1}{2}\|)^4\sum_{l=1}^L\lambda_l(\|\bfmu_l\|+\|\bfSg_l^\frac{1}{2}\|)^8\Big)^{\frac{1}{2}}\\
&\cdot(\delta_1(\bfW^*))^2 D_6(\Psi)
\Gamma(\Psi)^{-2}\sigma_{\max}^{-4}\cdot d\log^2{d}\label{fn_sp0}
\end{aligned}
\end{equation}
where $C_0=\max\{C_4,C_3^{-2}\}$. By Definition \ref{def: D}, we can obtain
\begin{equation}
    \Big(\sum_{l=1}^L\lambda_l(\|\bfmu_l\|+\|\bfSg_l^\frac{1}{2}\|)^4\sum_{l=1}^L\lambda_l(\|\bfmu_l\|+\|\bfSg_l^\frac{1}{2}\|)^8\Big)^{\frac{1}{2}}\leq \sqrt{D_4(\Psi)D_8(\Psi)}\sigma_{\max}^6\label{D_def_ineq}
\end{equation}
From Property \ref{prop: D_ineq}, we have that 
\begin{equation}
\begin{aligned}
    &\sqrt{D_4(\Psi)D_8(\Psi)}D_6(\Psi)\\
    \leq& \sqrt{D_{12}(\Psi)}\sqrt{D_{12}(\Psi)}=D_{12}(\Psi)\label{cmp_sp}
\end{aligned}
\end{equation}
Plugging (\ref{D_def_ineq}), (\ref{cmp_sp}) into (\ref{fn_sp0}), we have 
\begin{equation}
n\geq C_0\epsilon_0^{-2}\cdot\kappa^{12} K^{14}(\sigma_{\max}\delta_1(\bfW^*))^2\tau^{12}\Gamma(\Psi)^{-2}D_{12}(\Psi)\cdot d\log^2{d}\label{fn_sp}
\end{equation}
Considering   the requirements on the sample complexity  in (\ref{sp_1}), (\ref{sp_lm3}), and (\ref{fn_sp}),  (\ref{fn_sp}) shows a sufficient number of samples. Taking the union bound of all the failure probabilities in Lemma \ref{lemma: convexity}, and \ref{lemma: tensor bound}, (\ref{fn_sp}) holds with probability $1-d^{-10}$.\\
By Property \ref{prop: rho_property}.4, $\rho(\frac{{\bfW^*}^\top\bfmu_l}{\delta_K(\bfW^*)\|\bfSg_l^{-1}\|^{-\frac{1}{2}}},\delta_K(\bfW^*)\|\bfSg_l^{-1}\|^{-\frac{1}{2}})$ can be lower bounded by positive and monotonically decreasing functions $\mathcal{L}_m(\frac{(\bfLM_l{\bfW^*})^\top\tilde{\bfmu}_l}{\delta_K(\bfW^*)\|\bfSg_l^{-1}\|^{-\frac{1}{2}}},\delta_K(\bfW^*)\|\bfSg_l^{-1}\|^{-\frac{1}{2}})$ when everything else except $|\tilde{\bfmu}_{l(i)}|$ is fixed, or $\mathcal{L}_s(\frac{{\bfW^*}^\top\bfmu_l}{\delta_K(\bfW^*)\|\bfSg_l^{-1}\|^{-\frac{1}{2}}},\delta_K(\bfW^*)\|\bfSg_l^{-1}\|^{-\frac{1}{2}})$ when everything else except $\|\bfSg_l^\frac{1}{2}\|$ is fixed. Then, by replacing the lower bound of $\rho(\frac{{\bfW^*}^\top\bfmu_l}{\delta_K(\bfW^*)\|\bfSg_l^{-1}\|^{-\frac{1}{2}}},\delta_K(\bfW^*)\|\bfSg_l^{-1}\|^{-\frac{1}{2}})$ with these two functions in $\Gamma(\Psi)$, we can have an upper bound of $(\sigma_{\max}\delta_1(\bfW^*))^2\tau^{12}\Gamma(\Psi)^{-2}D_{12}(\Psi)$, denoted as $\mathcal{B}(\Psi)$.\\
To be more specific, when everything else except $|\tilde{\bfmu}_{l(i)}|$ is fixed, $\mathcal{L}_m(\frac{(\bfLM_l{\bfW^*})^\top\tilde{\bfmu}_l}{\delta_K(\bfW^*)\|\bfSg_l^{-1}\|^{-\frac{1}{2}}},\delta_K(\bfW^*)\|\bfSg_l^{-1}\|^{-\frac{1}{2}})$ is plugged in $\mathcal{B}(\Psi)$. Then since that $D_{12}(\Psi)$ and $\mathcal{L}_m(\frac{(\bfLM_l{\bfW^*})^\top\tilde{\bfmu}_l}{\delta_K(\bfW^*)\|\bfSg_l^{-1}\|^{-\frac{1}{2}}},\delta_K(\bfW^*)\|\bfSg_l^{-1}\|^{-\frac{1}{2}})$ are both increasing function of $|\tilde{\bfmu}_{l(i)}|$, $\mathcal{B}(\Psi)$ is an increasing function of $|\tilde{\bfmu}_{l(i)}|$.\\
When everything else except $\|\bfSg_l^\frac{1}{2}\|$ is fixed, if $\|\bfSg_l^\frac{1}{2}\|=\sigma_{\max}>\zeta_s$, then $\sigma_{\max}^2\tau^{12}D_{12}(\Psi)$ is an increasing function of $\|\bfSg_l^\frac{1}{2}\|$. Since that $\mathcal{L}_s(\frac{{\bfW^*}^\top\bfmu_l}{\delta_K(\bfW^*)\|\bfSg_l^{-1}\|^{-\frac{1}{2}}},\delta_K(\bfW^*)\|\bfSg_l^{-1}\|^{-\frac{1}{2}})$ is a decreasing function, $\mathcal{L}_s(\frac{{\bfW^*}^\top\bfmu_l}{\delta_K(\bfW^*)\|\bfSg_l^{-1}\|^{-\frac{1}{2}}},\delta_K(\bfW^*)\|\bfSg_l^{-1}\|^{-\frac{1}{2}})^{-2}$ is an increasing function of $\|\bfSg_l^\frac{1}{2}\|$. Hence, $\mathcal{B}(\Psi)$ is an increasing function of $\|\bfSg_l^\frac{1}{2}\|$. Moreover, when all $\|\bfSg_l^\frac{1}{2}\|<\zeta_{s'}$ and go to $0$, two decreasing functions of $\|\bfSg_l^\frac{1}{2}\|$,  $\sigma^2_{\max}\mathcal{L}_s(\frac{{\bfW^*}^\top\bfmu_l}{\delta_K(\bfW^*)\|\bfSg_l^{-1}\|^{-\frac{1}{2}}},\delta_K(\bfW^*)\|\bfSg_l^{-1}\|^{-\frac{1}{2}})^{-2}$ and $D_{12}(\Psi)$ will be the dominant term of $\mathcal{B}(\Psi)$. Therefore, $\mathcal{B}(\Psi)$ increases to infinity as all $\|\bfSg_l^\frac{1}{2}\|$'s go to $0$. In sum, we can define a universe $\mathcal{B}(\Psi)$ as:
\begin{equation}
\begin{aligned}
    &\mathcal{B}(\Psi)\\
    =&\begin{cases}&(\sigma_{\max}\delta_1(\bfW^*))^2\tau^{12}\Big(\sum_{l=1}^L\frac{\lambda_l\|\bfSg_l^{-1}\|^{-1}}{\eta\sigma_{\max}^2}\mathcal{L}_m(\frac{(\bfLM_l{\bfW^*})^\top\tilde{\bfmu}_l}{\delta_K(\bfW^*)\|\bfSg_l^{-1}\|^{-\frac{1}{2}}},\delta_K(\bfW^*)\|\bfSg_l^{-1}\|^{-\frac{1}{2}})\Big)^{-2}\\
    &\cdot D_{12}(\Psi), \text{if }\bfS\text{ is fixed}\\
    &(\sigma_{\max}\delta_1(\bfW^*))^2\tau^{12}\Big(\sum_{l=1}^L\frac{\lambda_l\|\bfSg_l^{-1}\|^{-1}}{\eta\sigma_{\max}^2}\mathcal{L}_s(\frac{(\bfLM_l{\bfW^*})^\top\tilde{\bfmu}_l}{\delta_K(\bfW^*)\|\bfSg_l^{-1}\|^{-\frac{1}{2}}},\delta_K(\bfW^*)\|\bfSg_l^{-1}\|^{-\frac{1}{2}})\Big)^{-2}\\
    &\cdot D_{12}(\Psi), \text{if }\bf\bfM\text{ is fixed}\\
    &(\sigma_{\max}\delta_1(\bfW^*))^2\tau^{12}\Big(\sum_{l=1}^L\frac{\lambda_l\|\bfSg_l^{-1}\|^{-1}}{\eta\sigma_{\max}^2}\rho(\frac{(\bfLM_l{\bfW^*})^\top\tilde{\bfmu}_l}{\delta_K(\bfW^*)\|\bfSg_l^{-1}\|^{-\frac{1}{2}}},\delta_K(\bfW^*)\|\bfSg_l^{-1}\|^{-\frac{1}{2}})\Big)^{-2}\\
    &\cdot D_{12}(\Psi), \text{otherwise}
    \end{cases}
\end{aligned}\label{B_closedform}
\end{equation}
where $\mathcal{L}_m, \mathcal{L}_s$ and $D_{12}$ are defined in (\ref{Lm}), (\ref{Ls}) and Definition \ref{def: D}, respectively.\\
Hence, we have
\begin{equation}
    n\geq poly(\epsilon_0^{-1}, \kappa, \eta, \tau K)\mathcal{B}(\Psi)\cdot d\log^2 d\label{modified_sp}
\end{equation}
Similarly, by replacing $\rho(\frac{{\bfW^*}^\top\bfmu_l}{\delta_K(\bfW^*)\|\bfSg_l^{-1}\|^{-\frac{1}{2}}},\delta_K(\bfW^*)\|\bfSg_l^{-1}\|^{-\frac{1}{2}})$ with $\mathcal{L}_m(\frac{(\bfLM_l{\bfW^*})^\top\tilde{\bfmu}_l}{\delta_K(\bfW^*)\|\bfSg_l^{-1}\|^{-\frac{1}{2}}},\delta_K(\bfW^*)\|\bfSg_l^{-1}\|^{-\frac{1}{2}})$ when everything else except $|\tilde{\bfmu}_{l(i)}|$ is fixed, or $\mathcal{L}_s(\frac{{\bfW^*}^\top\bfmu_l}{\delta_K(\bfW^*)\|\bfSg_l^{-1}\|^{-\frac{1}{2}}},\delta_K(\bfW^*)\|\bfSg_l^{-1}\|^{-\frac{1}{2}})$ (or $\|\bfSg_l^{-1}\|\mathcal{L}_s(\frac{{\bfW^*}^\top\bfmu_l}{\delta_K(\bfW^*)\|\bfSg_l^{-1}\|^{-\frac{1}{2}}},\delta_K(\bfW^*)\|\bfSg_l^{-1}\|^{-\frac{1}{2}})$ for $\|\bfSg_l^{-1}\|^{-1}\geq 1$) when everything else except $\|\bfSg_l^\frac{1}{2}\|$ is fixed, (\ref{convergence_rate_initial}) can also be transferred to another feasible upper bound. We denote the modified version of the convergence rate as $v=1-K^{-2}q(\Psi)$. Since that $q(\Psi)$ is a ratio between the smallest and the largest singular value of $\nabla ^2 \bar{f}(\bfW^*)$, we have $q(\Psi)\in(0,1)$. Hence, we can obtain $1-K^{-2}q(\Psi)\in(0,1)$ by $K\geq1$. When everything else except $|\tilde{\bfmu}_{l(i)}|$ is fixed, since that $\mathcal{L}_m(\frac{(\bfLM_l{\bfW^*})^\top\tilde{\bfmu}_l}{\delta_K(\bfW^*)\|\bfSg_l^{-1}\|^{-\frac{1}{2}}},\delta_K(\bfW^*)\|\bfSg_l^{-1}\|^{-\frac{1}{2}})$ is monotonically decreasing and $\sum_{l=1}^L\lambda(\|\bfmu_l\|+\|\bfSg_l^\frac{1}{2}\|)^2$ is increasing as $|\tilde{\bfmu}_{l(i)}|$ increases, $v$ is an increasing function of $|\tilde{\bfmu}_{l(i)}|$ to $1$. Similarly, when everything else except $\|\bfSg_l^\frac{1}{2}\|$ is fixed where $\|\bfSg_l^\frac{1}{2}\|\geq\max\{1,\zeta_s\}$, $\frac{1}{\sum_{l=1}^L\lambda_l(\|\bfmu_l\|+\|\bfSg_l^\frac{1}{2}\|)^2}$ decreases to $0$ as $\|\bfSg_l\|$ increases. We replace $\rho(\frac{{\bfW^*}^\top\bfmu_l}{\delta_K(\bfW^*)\|\bfSg_l^{-1}\|^{-\frac{1}{2}}},\delta_K(\bfW^*)\|\bfSg_l^{-1}\|^{-\frac{1}{2}})$ by $\|\bfSg_l^{-1}\|\mathcal{L}_s(\frac{{\bfW^*}^\top\bfmu_l}{\delta_K(\bfW^*)\|\bfSg_l^{-1}\|^{-\frac{1}{2}}},\delta_K(\bfW^*)\|\bfSg_l^{-1}\|^{-\frac{1}{2}})$ and then \begin{equation}\begin{aligned}&\|\bfSg_l^{-1}\|^{-1}\cdot \|\bfSg_l^{-1}\|\mathcal{L}_s(\frac{{\bfW^*}^\top\bfmu_l}{\delta_K(\bfW^*)\|\bfSg_l^{-1}\|^{-\frac{1}{2}}},\delta_K(\bfW^*)\|\bfSg_l^{-1}\|^{-\frac{1}{2}})\\=&\mathcal{L}_s(\frac{{\bfW^*}^\top\bfmu_l}{\delta_K(\bfW^*)\|\bfSg_l^{-1}\|^{-\frac{1}{2}}},\delta_K(\bfW^*)\|\bfSg_l^{-1}\|^{-\frac{1}{2}})\end{aligned}\end{equation}
is an decreasing function less than $\rho(\frac{{\bfW^*}^\top\bfmu_l}{\delta_K(\bfW^*)\|\bfSg_l^{-1}\|^{-\frac{1}{2}}},\delta_K(\bfW^*)\|\bfSg_l^{-1}\|^{-\frac{1}{2}})$. Therefore, $v$ is an increasing function of $\|\bfSg_l^\frac{1}{2}\|$ to $1$ when $\|\bfSg_l^\frac{1}{2}\|\geq\max\{1,\zeta_s\}$. When everything else except all $\|\bfSg_l^\frac{1}{2}\|\leq \zeta_{s'}$'s go to $0$, all $\mathcal{L}_s(\frac{{\bfW^*}^\top\bfmu_l}{\delta_K(\bfW^*)\|\bfSg_l^{-1}\|^{-\frac{1}{2}}},\delta_K(\bfW^*)\|\bfSg_l^{-1}\|^{-\frac{1}{2}}$'s will decrease and  all $\frac{\|\bfSg_l^{-1}\|^{-1}}{\sum_{l=1}^L\lambda_l(\|\bfmu_l\|_\infty+\|\bfSg_l^\frac{1}{2}\|)^2}$'s will decrease to $0$. Therefore, $v$ increases to $1$.\\
$q(\Psi)$ can then be defined as
\begin{equation}
\begin{aligned}
    &q(\Psi)\\
    =&
    \begin{cases}&\Omega\big(\frac{\sum_{l=1}^L\lambda_l\frac{\|\bfSg_l^{-1}\|^{-1}}{\eta \tau^K\kappa^2}\mathcal{L}_m(\frac{({\bfLM_l\bfW^*})^\top\tilde{\bfmu_l}}{\delta_K(\bfW^*)\|\bfSg_l^{-1}\|^{-\frac{1}{2}}},\delta_K(\bfW^*)\|\bfSg_l^{-1}\|^{-\frac{1}{2}})}{\sum_{l=1}^L\lambda_l(\|\bfmu_l\|+\|\bfSg_l^\frac{1}{2}\|)^2}\big)),\\ &\text{if }\bfS\text{ is fixed}\\
    &\Omega\big(\frac{\sum_{l=1}^L\lambda_l\frac{\|\bfSg_l^{-1}\|^{-1}}{\eta \tau^K\kappa^2}\mathcal{L}_s(\frac{{\bfW^*}^\top\bfmu_l}{\delta_K(\bfW^*)\|\bfSg_l^{-1}\|^{-\frac{1}{2}}},\delta_K(\bfW^*)\|\bfSg_l^{-1}\|^{-\frac{1}{2}})}{\sum_{l=1}^L\lambda_l(\|\bfmu_l\|+\|\bfSg_l^\frac{1}{2}\|)^2}\big),\\ &\text{if }\bfM\text{ is fixed and all }\|\bfSg_l^\frac{1}{2}\|\leq \zeta_{s'}\\
    &\Omega\big(\frac{\lambda_l\frac{1}{\eta \tau^K\kappa^2}\mathcal{L}_s(\frac{{\bfW^*}^\top\bfmu_i}{\delta_K(\bfW^*)\|\bfSg_i^{-1}\|^{-\frac{1}{2}}},\delta_K(\bfW^*)\|\bfSg_i^{-1}\|^{-\frac{1}{2}})+\sum_{l\neq i}r(\lambda_l,\bfmu_l,\bfSg_l,\bfW^*)}{\sum_{l=1}^L\lambda_l(\|\bfmu_l\|+\|\bfSg_l^\frac{1}{2}\|)^2}\big),\\ &\text{if }\bfM\text{ is fixed and one }\|\bfSg_i^\frac{1}{2}\|\geq\max\{1,\zeta_s\}\\
    &\Omega\big(\frac{\sum_{l=1}^L\lambda_l\frac{\|\bfSg_l^{-1}\|^{-1}}{\eta \tau^K\kappa^2}\rho(\frac{{\bfW^*}^\top\bfmu_l}{\delta_K(\bfW^*)\|\bfSg_l^{-1}\|^{-\frac{1}{2}}},\delta_K(\bfW^*)\|\bfSg_l^{-1}\|^{-\frac{1}{2}})}{\sum_{l=1}^L\lambda_l(\|\bfmu_l\|+\|\bfSg_l^\frac{1}{2}\|)^2}\big),\\ &\text{otherwise}\label{q_closedform}
    \end{cases}.
\end{aligned}
\end{equation}
where $r(\lambda_l,\bfmu_l,\bfSg_l,\bfW^*)=\lambda_l\frac{\|\bfSg_l^{-1}\|^{-1}}{\eta\tau^K\kappa^2}\rho(\frac{{\bfW^*}^\top\bfmu_l}{\delta_K(\bfW^*)\|\bfSg_l^{-1}\|^{-\frac{1}{2}}},\delta_K(\bfW^*)\|\bfSg_l^{-1}\|^{-\frac{1}{2}})$. Note that here the $\rho(\cdot)$ function is defined in Definition \ref{def: rho}. $\mathcal{L}_m(\cdot)$ and $\mathcal{L}_s(\cdot)$ are defined in (\ref{Lm}) and (\ref{Ls}), respectively.\\
The bound of $\|\widehat{\bfW}_n-\bfW^*\bfP\|_F$ is directly from (\ref{critical_groundtruth}). We can derive that
\begin{equation}
    \mathcal{E}_w(\Psi)=O(\frac{\sqrt{ \sum_{j=1}^L\lambda_l(\|\bfmu_j\|+\|\bfSg_j^\frac{1}{2}\|)^2}}{\sum_{j=1}^L\lambda_l\|\bfSg_j^{-1}\|^{-1}\rho(\frac{{\bfW^*}^\top\bfmu_j}{\delta_K(\bfW^*)\|\bfSg_j^{-1}\|^{-\frac{1}{2}}}, \delta_K(\bfW^*)\|\bfSg_j^{-1}\|^{-\frac{1}{2}})})\label{E_w_closedform}
\end{equation}
\begin{equation}
    \mathcal{E}(\Psi)=O(\frac{ \sum_{j=1}^L\lambda_l(\|\bfmu_j\|+\|\bfSg_j^\frac{1}{2}\|)^2}{\sum_{j=1}^L\lambda_l\|\bfSg_j^{-1}\|^{-1}\rho(\frac{{\bfW^*}^\top\bfmu_j}{\delta_K(\bfW^*)\|\bfSg_j^{-1}\|^{-\frac{1}{2}}}, \delta_K(\bfW^*)\|\bfSg_j^{-1}\|^{-\frac{1}{2}})})\label{E_closedform}
\end{equation}
\begin{equation}
    \mathcal{E}_l(\Psi)=O(\frac{\sqrt{ \sum_{j=1}^L\lambda_l(\|\bfmu_j\|+\|\bfSg_j^\frac{1}{2}\|)^2}(\|\bfmu_l\|+\|\bfSg_l\|^\frac{1}{2})}{\sum_{j=1}^L\lambda_l\|\bfSg_j^{-1}\|^{-1}\rho(\frac{{\bfW^*}^\top\bfmu_j}{\delta_K(\bfW^*)\|\bfSg_j^{-1}\|^{-\frac{1}{2}}}, \delta_K(\bfW^*)\|\bfSg_j^{-1}\|^{-\frac{1}{2}})})\label{E_l_closedform}
\end{equation}
The discussion of the monotonicity of $\mathcal{E}_w(\Psi)$,  $\mathcal{E}(\Psi)$ and $\mathcal{E}_l(\Psi)$ can follow the analysis of $q(\Psi)$. \\
We finish our proof of Theorem \ref{thm1} here. The parameters $\mathcal{B}(\Psi),\ q(\Psi),\ \mathcal{E}_w(\Psi),\ \mathcal{E}(\Psi),\ \text{and }\mathcal{E}_l(\Psi)$ can be found in (\ref{B_closedform}), (\ref{q_closedform}), (\ref{E_w_closedform}), (\ref{E_closedform}), and (\ref{E_l_closedform}), respectively.\\\\
\textbf{Proof of Corollary \ref{cor}}:\\
The monotonicity analysis has been included in the proof of Theorem \ref{thm1}. In this part, we specify our proof for the results in Table \ref{tbl:results}. For simplicity, we denote $\rho_l=\rho(\frac{{\bfW^*}^\top\bfmu_l}{\delta_K(\bfW^*)\|\bfSg_l^{-1}\|^{-\frac{1}{2}}},\delta_K(\bfW^*)\|\bfSg_l^{-1}\|^{-\frac{1}{2}})$.\\
When everything else except $\|\bfSg_l\|^\frac{1}{2}$ is fixed, if $\|\bfSg_l\|=o(1)$, by some basic mathematical computation, then we have
\begin{equation}
    \begin{aligned}
        n_{sc}=&C_0\epsilon_0^{-2}\cdot\eta^2\tau^{12}\kappa^{16} K^{14}\Big(\sum_{l=1}^L\lambda_l(\|\tilde{\bfmu}_l\|+\|\bfSg_l^\frac{1}{2}\|)^4\sum_{l=1}^L\lambda_l(\|\tilde{\bfmu}_l\|+\|\bfSg_l^\frac{1}{2}\|)^8\Big)^{\frac{1}{2}}(\delta_1(\bfW^*))^2 D_6(\Psi)\\
&\cdot(\frac{1}{\sum_{l=1}^L \lambda_l \|\bfSg_l^{-1}\|^{-1}\rho_l})^2\cdot d\log^2{d}\\
\lesssim &\text{poly}(\epsilon_0^{-1}, \eta, \tau, \kappa, K, \delta_1(\bfW^*))\cdot d\log^2 d\cdot  O(\lambda_L\frac{1}{\|\bfSg_L^\frac{1}{2}\|^{6}})
    \end{aligned}
\end{equation}
\begin{equation}
    \begin{aligned}
        v(\Psi)&=1-\frac{\sum_{l=1}^L \lambda_l\frac{\|\bfSg_l^{-1}\|^{-1}}{\eta \kappa^2}\rho_l}{K^2(\sum_{l=1}^L \lambda_l(\|\bfmu_l\|+\|\bfSg_l^\frac{1}{2}\|)^2)}\\
        &\leq 1-\frac{\lambda_l}{K^2\eta\kappa^2\tau^K}\Theta(\|\bfSg_l\|^3)
    \end{aligned}
\end{equation}
\begin{equation}
    \begin{aligned}
        \|\widehat{\bfW}_n-\bfW^*\bfP^*\|&\leq O(\frac{K^{\frac{5}{2}}\sqrt{ \sum_{l=1}^L\lambda_l(\|\bfmu_l\|+\|\bfSg_l^\frac{1}{2}\|)^2}(1+\xi)}{\sum_{l=1}^L\lambda_l\frac{\|\bfSg_l^{-1}\|^{-1}}{\eta\tau^K\kappa^2}\rho(\frac{{\bfW^*}^\top\bfmu_l}{\delta_K(\bfW^*)\|\bfSg_l^{-1}\|^{-\frac{1}{2}}}, \delta_K(\bfW^*)\|\bfSg_l^{-1}\|^{-\frac{1}{2}})}\sqrt{\frac{d\log{n}}{n}})\\
        &\lesssim \text{poly}(\eta, \kappa, \tau, \delta_K(\bfW^*))\sqrt{\frac{d\log n}{n}} K^2(1+\xi)\cdot O(1-\|\bfSg_l\|^3)
    \end{aligned}
\end{equation}
\begin{equation}
    \begin{aligned}
        \bar{f}_l(\bfW_t)&=\bar{f}_l(\bfW_t)-\bar{f}_l(\bfW^*)\\
    &\leq \mathbb{E}\Big[\sum_{k=1}^K \frac{\partial (\bar{f}_l(\bfW_t)}{\partial \tilde{\bfw_k}})^\top(\bfw_{t(k)}-\bfw_k^*)\Big]\\
    &\leq \|\bfW_t-\bfW^*\bfP^*\|(\|\bfmu_l\|+\|\bfSg_l\|^\frac{1}{2})\\
    &\lesssim O\Big(\frac{\sum_{j=1}^L \sqrt{\lambda_l}(\|\bfmu_j\|+\|\bfSg_j\|^\frac{1}{2})}{\sum_{j=1}^L\lambda_j\|\bfSg_j^{-1}\|^{-1}\rho_j}(\|\bfmu_j\|+\|\bfSg_j\|^\frac{1}{2})\cdot\sqrt{\frac{d\log n}{n}}\eta\kappa^2 K^2(1+\xi)\Big)\\
    &\lesssim \text{poly}(\eta, \kappa, \tau, \delta_K(\bfW^*))\sqrt{\frac{d\log n}{n}} K^2(1+\xi)\cdot O(\frac{1}{1+\|\bfSg_l\|^3})\\
    &\lesssim \text{poly}(\eta, \kappa, \tau, \delta_K(\bfW^*))\sqrt{\frac{d\log n}{n}} K^2(1+\xi)\cdot O(1)-\Theta(\|\bfSg_l\|^3),\label{f_l_bound}
    \end{aligned}
\end{equation}
The first inequality of (\ref{f_l_bound}) is by the Mean Value Theorem. The second inequality of (\ref{f_l_bound}) is from Property \ref{prop: bound}, and the third inequality is derived from (\ref{critical_groundtruth}, \ref{convergence_rate_initial}). The last inequality is obtained by the condition that $\|\bfSg_l\|=o(1)$. We can similarly have
\begin{equation}
\begin{aligned}
    \bar{f}(\bfW_t)&\leq \mathbb{E}\Big[\sum_{k=1}^K \frac{\partial (\bar{f}(\bfW_t)}{\partial \tilde{\bfw_k}})^\top(\bfw_{t(k)}-\bfw_k^*)\Big]\\
    &\lesssim \text{poly}(\eta, \kappa, \tau, \delta_K(\bfW^*))\sqrt{\frac{d\log n}{n}} K^2(1+\xi)\cdot O(\frac{1}{1+\|\bfSg_l\|^3})\\
    &\lesssim \text{poly}(\eta, \kappa, \tau, \delta_K(\bfW^*))\sqrt{\frac{d\log n}{n}} K^2(1+\xi)\cdot O(1)-\Theta(\|\bfSg_l\|^3)
    \end{aligned}
\end{equation}
If $\|\bfSg_l\|^\frac{1}{2}=\Omega(1)$, we have
\begin{equation}
    n_{sc}\lesssim \text{poly}(\epsilon_0^{-1}, \eta, \tau, \kappa, K, \delta_1(\bfW^*))\cdot d\log^2 d\cdot O(\|\bfSg_l\|^3)
\end{equation}
\begin{equation}
    v(\Psi)\leq 1-\frac{1}{ K^2\tau^K \eta\kappa^2}\Theta(\frac{1}{1+\|\bfSg_l\|})
\end{equation}
\begin{equation}
    \|\widehat{\bfW}_n-\bfW^*\bfP^*\|_F\lesssim \text{poly}(\eta, \tau, \kappa, \delta_K(\bfW^*))\sqrt{\frac{d\log n}{n}} K^\frac{5}{2}(1+\xi)\cdot\sqrt{\|\bfSg_l\|}
\end{equation}
\begin{equation}
    \bar{f}_l(\bfW_t)\lesssim \text{poly}(\eta, \tau, \kappa, \delta_K(\bfW^*))\sqrt{\frac{d\log n}{n}} K^2(1+\xi)\cdot\|\bfSg_l\|
\end{equation}
\begin{equation}
    \bar{f}(\bfW_t)\lesssim \text{poly}(\eta, \tau, \kappa, \delta_K(\bfW^*))\sqrt{\frac{d\log n}{n}} K^2(1+\xi)\cdot\|\bfSg_l\|
\end{equation}
When everything is fixed except $\|\bfmu_l\|$, by combining (\ref{sp_1}) and (\ref{fn_sp0}), we have
\begin{equation}
    n_{sc}\lesssim \text{poly}(\epsilon_0^{-1}, \eta, \tau, \kappa, K, \delta_1(\bfW^*))\cdot d\log^2 d\cdot \begin{cases}O(\|\bfmu_l\|^4), &\text{if }\|\bfmu_l\|\leq 1\\
    O(\|\bfmu_l\|^{12}), &\text{if }\|\bfmu_l\|\geq1\end{cases}
\end{equation}
\begin{equation}
    v(\Psi)\leq 1-\frac{1}{ K^2\tau^K \eta\kappa^2}\Theta(\frac{1}{1+\|\bfmu_l\|^2})
\end{equation}
\begin{equation}
    \|\widehat{\bfW}_n-\bfW^*\bfP^*\|_F\lesssim \text{poly}(\eta, \tau, \kappa, \delta_K(\bfW^*))\sqrt{\frac{d\log n}{n}} K^\frac{5}{2}(1+\xi)\cdot(1+\|\bfmu_l\|)
\end{equation}
\begin{equation}
    \bar{f}_l(\bfW_t)\lesssim \text{poly}(\eta, \tau, \kappa, \delta_K(\bfW^*))\sqrt{\frac{d\log n}{n}} K^2(1+\xi)\cdot(1+\|\bfmu_l\|^2)
\end{equation}
\begin{equation}
    \bar{f}(\bfW_t)\lesssim \text{poly}(\eta, \tau, \kappa, \delta_K(\bfW^*))\sqrt{\frac{d\log n}{n}} K^2(1+\xi)\cdot(1+\|\bfmu_l\|^2)
\end{equation}
When everything else is fixed except $\lambda_1, \lambda_2,\cdots,\lambda_L$, where $\|\bfSg_j\|=\Omega(1),\ j\in[L]$ and $\|\bfmu_j\|=\|\bfmu_i\|,\ i,j\in[L]$, if $\|\bfSg_l\|\leq \|\bfSg_j\|, \ j\in[L]$, we have
\begin{equation}
\begin{aligned}
    n_{sc}\lesssim &\text{poly}(\epsilon_0^{-1},\eta,\kappa, K, \delta_1(\bfW^*))\cdot d\log^2 d \cdot \frac{(a_1\lambda_l^2+a_2\lambda_l^\frac{3}{2}+a_3\lambda_l+a_4\lambda_l^\frac{1}{2}+a_5)}{(\sum_{j=1}^L\lambda_j\rho_j)^2}\\
    \leq & \text{poly}(\epsilon_0^{-1},\eta,\kappa, K, \delta_1(\bfW^*))\cdot d\log^2 d \cdot \frac{a_5}{(\sum_{j=1}^L\lambda_j\rho_j)^2}\\
    \lesssim &\text{poly}(\epsilon_0^{-1},\eta,\kappa, K, \delta_1(\bfW^*))\cdot d\log^2 d \cdot O((1+\lambda_l)^{-2})\label{n_sc_smallsigma}
\end{aligned}
\end{equation}
where $a_1=(\|\bfmu_l\|+\|\bfSg_l\|^\frac{1}{2})^{12}/\|\bfSg_l\|^3$, $a_2=(\|\bfmu_l\|+\|\bfSg_l^\frac{1}{2}\|)^8(\sum_{j\neq l}\lambda_j(\|\bfmu_j\|+\|\bfSg_j\|^\frac{1}{2})^8)^\frac{1}{2}/\|\bfSg_l\|^3$, $a_3=(\|\bfmu_l\|/\|\bfSg_l\|^\frac{1}{2}+1)^6(\sum_{j\neq l}\lambda_j(\|\bfmu_j\|+\|\bfSg_j\|^\frac{1}{2})^4\sum_{j\neq l}\lambda_j(\|\bfmu_j\|+\|\bfSg_j\|^\frac{1}{2})^8)^\frac{1}{2}+(\|\bfmu_l\|+\|\bfSg_l\|^\frac{1}{2})^6\sum_{j\neq l}\lambda_j(\|\bfmu_j\|/\|\bfSg_j\|^\frac{1}{2}+1)^6$, $a_4=\sum_{j\neq l}\lambda_j(\|\bfmu_j\|/\|\bfSg_j\|^\frac{1}{2}+1)^6(\|\bfmu_l\|+\|\bfSg_l\|^\frac{1}{2})^2(\sum_{j\neq l}\lambda_j(\|\bfmu_j\|+\|\bfSg_j\|^\frac{1}{2})^8)^\frac{1}{2}$, $a_5=(\sum_{j\neq l}\lambda_j(\|\bfmu_j\|+\|\bfSg_j\|^\frac{1}{2})^4\sum_{j\neq l}\lambda_j(\|\bfmu_j\|+\|\bfSg_j\|^\frac{1}{2})^8)^\frac{1}{2}\cdot \sum_{j\neq l}\lambda_j(\|\bfmu_j\|/\|\bfSg_j\|^\frac{1}{2}+1)^6$. The second step of (\ref{n_sc_smallsigma}) is by $a_i= O(a_{5})$, $i=1,2,3,4$. 
\begin{equation}
    v\leq \frac{1}{K^2\eta\tau^K\kappa^2}\Theta(\frac{1}{1+\lambda_l})
\end{equation}
\begin{equation}
    \|\widehat{\bfW}_n-\bfW^*\bfP\|_F\leq \text{poly}(\eta,\kappa,, \tau,\delta_1(\bfW^*))\cdot\sqrt{\frac{d\log n}{n}}K^\frac{5}{2}(1+\xi)\cdot O(\frac{1}{1+\sqrt{\lambda_l}})
\end{equation}
\begin{equation}
   \bar{f}_l(\bfW_t)\leq \text{poly}(\eta,\kappa,, \tau,\delta_1(\bfW^*))\cdot\sqrt{\frac{d\log n}{n}}K^2(1+\xi)\cdot O(\frac{1}{1+\sqrt{\lambda_l}})
\end{equation}
\begin{equation}
   \bar{f}(\bfW_t)\leq \text{poly}(\eta,\kappa,, \tau,\delta_1(\bfW^*))\cdot\sqrt{\frac{d\log n}{n}}K^2(1+\xi)\cdot O(\frac{1}{1+\lambda_l})
\end{equation}

If $\|\bfSg_l\|\geq \|\bfSg_j\|,\ j\in[L]$, we can similarly derive that
\begin{equation}
\begin{aligned}
    n_{sc}\lesssim &\text{poly}(\epsilon_0^{-1},\eta,\kappa, K, \delta_1(\bfW^*))\cdot d\log^2 d \cdot \frac{(a_1\lambda_l^2+a_2\lambda_l^\frac{3}{2}+a_3\lambda_l+a_4\lambda_l^\frac{1}{2}+a_5)}{(\sum_{j=1}^L\lambda_j\rho_j)^2}\\
    \lesssim &\text{poly}(\epsilon_0^{-1},\eta,\kappa, K, \delta_1(\bfW^*))\cdot d\log^2 d \cdot (O(1)-\Theta((1+\lambda_l)^{-2}))
\end{aligned}
\end{equation}
\begin{equation}
    v\leq 1-\frac{1}{K^2\eta\tau^K\kappa^2} \Theta(\frac{1}{1+\lambda_l})
\end{equation}
\begin{equation}
    \|\widehat{\bfW}_n-\bfW^*\bfP\|_F\leq \text{poly}(\eta,\kappa,, \tau,\delta_1(\bfW^*))\cdot\sqrt{\frac{d\log n}{n}}K^\frac{5}{2}(1+\xi)\cdot O(1+\sqrt{\lambda_l})
\end{equation}
\begin{equation}
   \bar{f}_l(\bfW_t)\leq \text{poly}(\eta,\kappa,, \tau,\delta_1(\bfW^*))\cdot\sqrt{\frac{d\log n}{n}}K^2(1+\xi)\cdot O(1+\sqrt{\lambda_l})
\end{equation}
\begin{equation}
   \bar{f}(\bfW_t)\leq \text{poly}(\eta,\kappa,, \tau,\delta_1(\bfW^*))\cdot\sqrt{\frac{d\log n}{n}}K^2(1+\xi)\cdot (O(1)-\frac{\Theta(1)}{1+\lambda_l})
\end{equation}

%Compared with all the sample complexity, we adopt (\ref{sample}) as the final one for the theorem. Selecting the least probability, we will have the algorithm performance in \textbf{Lemma }\ref{lemma: convergence} with probability at least $1-d^{-10}$.\\\\

\subsection{Proof of Lemma \ref{lemma: convexity} and its supportive lemmas} \label{sec:convexity}

We first %state some important lemmas used in proof in Section \ref{subsec: useful lemmas Lm1} and 
describe the proof of Lemma \ref{lemma: convexity} in Section \ref{subsecc: pf lm1}. The proofs of the supportive lemmas are provided in Section \ref{subsec: pf lm_rho} to \ref{subsec: pf lm_bernstein} in sequence.  
The proof idea mainly follows from \cite{FCL20}.  Lemma \ref{lm: smoothness} shows the Hessian $\nabla^2\bar{f}(\bfW)$ of the population risk function is smooth. Lemma \ref{lm: convex} illustrates that $\nabla^2\bar{f}(\bfW)$ is strongly convex in the neighborhood around $\bfmu^*$. Lemma \ref{lm: bernstein} shows the Hessian of the empirical risk function $\nabla^2 f_n(\bfW^*)$ is close to its population risk $\nabla^2 \bar{f}(\bfW^*)$ in the local convex region. Summing up these three lemmas, we can   derive the proof of Lemma \ref{lemma: convexity}. Lemma \ref{lm: rho} is used in the proof of Lemma \ref{lm: convex}. Lemma \ref{lm: fraction bound} is used in the proof of Lemma \ref{lm: bernstein}. \\

The analysis of the Hessian matrix of the population loss in \cite{FCL20} and \cite{ZSJB17} can not be extended to the Gaussian mixture model. To solve this problem, we develop new tools using some good properties of symmetric distribution and even function. Our approach can also be applied to other activations like tanh or erf. Moreover, if we directly apply the existing matrix concentration inequalities in these works in bounding the error between the empirical loss and the population loss, the resulting sample complexity bound is loose and cannot reflect the influence of each component of the Gaussian mixture distribution. We develop a new version of Bernstein’s inequality (see (\ref{entire_bernstein})) so that the final bound is $O(d\log^2 d)$.

\cite{MBM16} showed that the landscape of the empirical risk is close to that of the population risk when the number of samples is sufficiently large for the special case that $K=1$. %Compared with the conclusion in Theorem 1 and Theorem 8 for Gaussian mixture input in (\cite{MBM16}), 
Focusing on Gaussian mixture models,  our result explicitly shows how the parameters of the input distribution, including the proportion, mean and, variance of each component will affect the error bound between the empirical loss and the population loss in Lemma \ref{lm: bernstein}.

\subsubsection{Proof of Lemma \ref{lm: rho}}\label{subsec: pf lm_rho}
\noindent Following the proof idea in Lemma D.4 of \cite{ZSJB17}, we have
\begin{equation}\mathbb{E}_{\bfx\sim\frac{1}{2}\mathcal{N}(\bfmu,\bfI_d)+\frac{1}{2}\mathcal{N}(-\bfmu,\bfI_d)}\Big[(\sum_{i=1}^k \bfr_i^\top \bfx\cdot\phi'(\sigma\cdot x_i))^2\Big]=A_0+B_0
\end{equation}
\begin{equation}A_0=\mathbb{E}_{\bfx\sim\frac{1}{2}\mathcal{N}(\bfmu,\bfI_d)+\frac{1}{2}\mathcal{N}(-\bfmu,\bfI_d)}\Big(\sum_{i=1}^k \bfr_i^\top\bfx \cdot\phi'^2(\sigma\cdot x_i)\cdot \bfx\bfx^\top \bfr_i\Big)\end{equation}
\begin{equation}B_0=\mathbb{E}_{\bfx\sim\frac{1}{2}\mathcal{N}(\bfmu,\bfI_d)+\frac{1}{2}\mathcal{N}(-\bfmu,\bfI_d)}\Big(\sum_{i\neq l} \bfr_i^\top \phi'(\sigma\cdot x_i)\phi'(\sigma\cdot x_l)\cdot \bfx\bfx^\top \bfr_l\Big)\end{equation}
In $A_0$, we know that $\mathbb{E}_{\bfx\sim\frac{1}{2}\mathcal{N}(\bfmu,\bfI_d)+\frac{1}{2}\mathcal{N}(-\bfmu,\bfI_d)} x_j=0$. Therefore, by some basic mathematical computation, 
\begin{equation}
\begin{aligned}
A_0&=\sum_{i=1}^k \mathbb{E}_{\bfx\sim\frac{1}{2}\mathcal{N}(\bfmu,\bfI_d)+\frac{1}{2}\mathcal{N}(-\bfmu,\bfI_d)}\Big[ \bfr_i^\top\Big(\phi'^2(\sigma\cdot x_i)\Big(x_i^2\bfe_i\bfe_i^\top+\sum_{j\neq i}x_ix_j(\bfe_i\bfe_j^\top \\
&\ \ +\bfe_j \bfe_i^\top )+\sum_{j\neq i}\sum_{l\neq i}x_j x_l \bfe_j \bfe_l^\top\Big)\Big)\bfr_i\Big]\\
&=\sum_{i=1}^k \mathbb{E}_{\bfx\sim\frac{1}{2}\mathcal{N}(\bfmu,\bfI_d)+\frac{1}{2}\mathcal{N}(-\bfmu,\bfI_d)}\Big[ \bfr_i^\top\Big(\phi'^2(\sigma\cdot x_i)\Big(x_i^2\bfe_i\bfe_i^\top+\sum_{j\neq i}x_j^2 \bfe_j \bfe_j^\top\Big)\Big)\bfr_i\Big]\\
&=\sum_{i=1}^k \Big[ \mathbb{E}_{\bfx\sim\frac{1}{2}\mathcal{N}(\bfmu,\bfI_d)+\frac{1}{2}\mathcal{N}(-\bfmu,\bfI_d)}[\phi'^2(\sigma\cdot x_i)x_i^2]\bfr_i^\top \bfe_i\bfe_i^\top \bfr_i\\
&\ \ \ +\sum_{j\neq i}\mathbb{E}_{\bfx\sim\frac{1}{2}\mathcal{N}(\bfmu,\bfI_d)+\frac{1}{2}\mathcal{N}(-\bfmu,\bfI_d)}[x_j^2]\mathbb{E}_{\bfx\sim\frac{1}{2}\mathcal{N}(\bfmu,\bfI)+\frac{1}{2}\mathcal{N}(-\bfmu,\bfI)}[\phi'^2(\sigma \cdot x_i)]\bfr_i^\top \bfe_j \bfe_j^\top \bfr_i\Big]\\
&=\sum_{i=1}^k r_{ii}^2\beta_2(i,\bfmu,\sigma)+\sum_{i=1}^k \sum_{j\neq i}r_{ij}^2\beta_0(i,\bfmu,\sigma)(1+\mu_{j}^2)\\
\end{aligned}
\end{equation}
In $B_0$, $\alpha_1(i,\bfmu,\sigma)=\mathbb{E}_{\bfx\sim\frac{1}{2}\mathcal{N}(\bfmu,\bfI_d)+\frac{1}{2}\mathcal{N}(-\bfmu,\bfI_d)}(x_i\phi'(x_i))=0$. By the equation in Page 30 of \cite{ZSJB17}, we have\\
\begin{equation}
\begin{aligned}
B_0=&\sum_{i\neq l}^k \mathbb{E}_{\bfx\sim\frac{1}{2}\mathcal{N}(\bfmu,\bfI_d)+\frac{1}{2}\mathcal{N}(-\bfmu,\bfI_d)}\Big[ \bfr_i^\top\Big(\phi'(\sigma\cdot x_i)\phi'(\sigma\cdot x_l)\Big(x_i^2\bfe_i\bfe_i^\top+x_l^2\bfe_l\bfe_l^\top+x_i x_l(\bfe_i \bfe_l^\top+\\
&\ \ \ \bfe_l \bfe_i^\top)+\sum_{j\neq i}x_jx_l\bfe_j\bfe_l^\top +\sum_{j\neq l}x_j x_i \bfe_j \bfe_i^\top+\sum_{j\neq i,l}\sum_{j'\neq i,l}x_j x_{j'} \bfe_j \bfe_{j'}^\top\Big)\Big)\bfr_l\Big]\\
=&\sum_{i\neq l}r_{ii}r_{li}\alpha_2(i,\bfmu,\sigma)\alpha_0(l,\bfmu,\sigma)+\sum_{i\neq l}r_{ij}r_{lj}\alpha_0(i,\bfmu,\sigma)\alpha_0(l,\bfmu,\sigma)(1+\mu_{j}^2)
\end{aligned}
\end{equation}
Therefore,
\begin{equation}
\begin{aligned}
A_0+B_0&=\sum_{i=1}^k \Big(r_{ii}\frac{\alpha_2(i,\bfmu,\sigma)}{\sqrt{1+\mu_{i}^2}}+\sum_{l\neq i}r_{li}\alpha_0(l,\bfmu,\sigma)\sqrt{1+\mu_{i}^2}\Big)^2-\sum_{i=1}^k r_{ii}^2\frac{\alpha_2^2(i,\bfmu,\sigma)}{1+\mu_{i}^2}\\
&-\sum_{i=1}^k\sum_{l\neq i} r_{li}^2\alpha_0(l,\bfmu,\sigma)^2(1+\mu_{i}^2)+\sum_{i=1}^k r_{ii}^2\beta_2(i,\bfmu,\sigma)+\sum_{i=1}^k \sum_{j\neq i}r_{ij}^2\beta_0(i,\bfmu,\sigma)(1+\mu_{j}^2)\\
&\geq \sum_{i=1}^k r_{ii}^2\Big(\beta_2(i,\bfmu,\sigma)-\frac{\alpha_2^2(i,\bfmu,\sigma)}{1+\mu_{i}^2}\Big)+\sum_{i=1}^k \sum_{j\neq i}r_{ij}^2\Big(\beta_0(i,\bfmu,\sigma)-\alpha_0^2(i,\bfmu,\sigma)\Big)(1+\mu_{j}^2)\\
&\geq \rho(\bfmu,\sigma)||\bfR||_F^2
\end{aligned}
\end{equation}

\subsubsection{Proof of Lemma \ref{lm: fraction bound}}\label{subsec: pf lm_fraction}
\noindent Following the equation (92) in Lemma 8 of \cite{FCL20} and by (\ref{xi})
\begin{equation}||\nabla^2\ell(\bfW)-\nabla^2\ell(\bfW')||\leq\sum_{j=1}^K\sum_{l=1}^K|\xi_{j,l}(\bfW)-\xi_{j,l}(\bfW')|\cdot||\bfx\bfx^\top||\end{equation}
By Lagrange's inequality, we have
\begin{equation}|\xi_{j,l}(\bfW)-\xi_{j,l}(\bfW')|\leq (\max_k |T_{j,k,l}|)\cdot||\bfx||\cdot\sqrt{K}||\bfW-\bfW'||_F\end{equation}
From Lemma \ref{lm: smoothness}, we know
\begin{equation}\max_k|T_{j,k,l}|\leq C_7\end{equation}
By Property \ref{prop: bound2}, we have \begin{equation}\mathbb{E}_{\bfx\sim\sum_{l=1}^L\lambda_l\mathcal{N}(\bfmu_l,\bfSg_l)}[||\bfx||^{2t}||]\leq d^t(2t-1)!!\sum_{l=1}^L\lambda_l(\|\bfmu_l\|_\infty+\|\bfSg_l\|)^{2t}\end{equation}
Therefore, for some constant $C_{12}>0$
\begin{equation}
\begin{aligned}
&\mathbb{E}_{\bfx\sim\sum_{l=1}^L\lambda_l\mathcal{N}(\bfmu_l,\bfSg_l)}[\sup_{\bfW\neq \bfW'}\frac{||\nabla^2\ell(\bfW)-\nabla^2\ell(\bfW')||}{||\bfW-\bfW'||_F}]
\leq K^{\frac{5}{2}}\mathbb{E}[||\bfx||_2^3]\\
\leq &K^{\frac{5}{2}}\sqrt{d \sum_{l=1}^L\lambda_l(\|\bfmu\|_\infty+\|\bfSg_l\|)^2}\sqrt{3d^2 \sum_{l=1}^L\lambda_l(\|\bfmu_l\|_\infty+\|\bfSg_l\|)^4}\\
=&C_{12}\cdot d^{\frac{3}
{2}}K^{\frac{5}{2}}\sqrt{\sum_{l=1}^L\lambda_l(\|\bfmu_l\|_\infty+\|\bfSg_l\|)^2\sum_{l=1}^L\lambda_l(\|\bfmu_l\|_\infty+\|\bfSg_l\|)^4}\label{C5_result}
\end{aligned}
\end{equation}
\\
\subsubsection{Proof of Lemma \ref{lm: smoothness} }\label{subsec: pf lm_smoothness}
\noindent Let $\bfa=(\bfa_1^\top,\cdots,\bfa_K^\top)^\top\in\mathbb{R}^{dK}$. Let $\Delta_{j,l}\in\mathbb{R}^{d\times d}$ be the $(j,l)$-th block of $\nabla^2\bar{f}(\bfW)-\nabla^2\bar{f}(\bfW^*\bfP)\in\mathbb{R}^{dK\times dK}$. By definition, 
\begin{equation}
||\nabla^2\bar{f}(\bfW)-\nabla^2\bar{f}(\bfW^*\bfP)||=\max_{||\bfa||=1}{\sum_{j=1}^{K}\sum_{l=1}^K \bfa_j^\top\Delta_{j,l}\bfa_l}\label{C1_1st}
\end{equation}
Denote $\bfP=(\bfp_1,\cdots,\bfp_K)\in\mathbb{R}^{K\times K}$. By the mean value theorem and (\ref{xi}), 
\begin{equation}
\begin{aligned}
\Delta_{j,l}&=\frac{\partial^2 \bar{f}(\bfW)}{\partial \bfw_j\partial\bfw_l}-\frac{\partial^2 \bar{f}(\bfW^*\bfP)}{\partial \bfw_j^*\partial\bfw_l^*}=\mathbb{E}_{\bfx\sim\sum_{l=1}^L\lambda_l\mathcal{N}(\bfmu_l,\sigma_l^2\bfI_d)}[(\xi_{j,l}(\bfW)-\xi_{j,l}(\bfW^*\bfP))\cdot \bfx\bfx^\top]\\
&=\mathbb{E}_{\bfx\sim\sum_{l=1}^L\lambda_l\mathcal{N}(\bfmu_l,\bfSg_l)}[\sum_{k=1}^K\left\langle \frac{\partial\xi_{j,l}(\bfW')}{\partial\bfw'_k},\bfw_k-\bfW^*\bfp_k\right\rangle\cdot\bfx\bfx^\top]\\
&=\mathbb{E}_{\bfx\sim\sum_{l=1}^L\lambda_l\mathcal{N}(\bfmu_l,\bfSg_l)}[\sum_{k=1}^K\left\langle T_{j,l,k}\cdot\bfx,\bfw_k-\bfW^*\bfp_k\right\rangle\cdot\bfx\bfx^\top]\label{delta}
\end{aligned}
\end{equation}
where $\bfW'=\gamma\bfW+(1-\gamma)\bfW^*\bfP$ for some $\gamma\in(0,1)$ and $T_{j,l,k}$ is defined such that $\frac{\partial\xi_{j,l}(\bfW')}{\partial\bfw'_k}=T_{j,l,k}\cdot x\in\mathbb{R}^d$. Then we provide an upper bound for $\xi_{j,l}$.
Since that $y=1 \text{ or }0$, we first compute the case in which $y=1$. From (\ref{xi}) we can obtain
\begin{equation}
\begin{aligned}
\xi_{j,l}(\bfW)=\left\{
\begin{array}{rcl}
\frac{1}{K^2}\phi'(\bfw_j^\top \bfx)\phi'(\bfw_l^\top \bfx)\cdot\frac{1}{H^2(\bfW)},\ \ \ \ \ \ \ \ \ \ \ \ \ \ & j\neq l\\
\frac{1}{K^2}\phi'(\bfw_j^\top \bfx)\phi'(\bfw_l^\top \bfx)\cdot\frac{1}{H^2(\bfW)}-\frac{1}{K}\phi''(\bfw_j^\top \bfx)\cdot\frac{1}{H(\bfW)}, &j=l
\end{array}\right.\label{xi_y1}
\end{aligned}
\end{equation}
We can bound $\xi_{j,l}(\bfW)$ by bounding each component of (\ref{xi_y1}). Note that we have\\
\begin{equation}
    \begin{aligned}
    \frac{1}{K^2}\phi'(\bfw_j^\top \bfx)\phi'(\bfw_l^\top \bfx)\cdot\frac{1}{H^2(\bfW)}\leq \frac{1}{K^2}\frac{\phi(\bfw_j^\top\bfx)\phi(\bfw_l^\top\bfx)(1-\phi(\bfw_j^\top\bfx))(1-\phi(\bfw_l^\top\bfx))}{\frac{1}{K^2}\phi(\bfw_j^\top\bfx)\phi(\bfw_l^\top\bfx)}\leq 1\label{xi_1st}
    \end{aligned}
\end{equation}
\begin{equation}
    \frac{1}{K}\phi''(\bfw_j^\top \bfx)\cdot\frac{1}{H(\bfW)}\leq \frac{1}{K}\frac{\phi(\bfw_j^\top\bfx)(1-\phi(\bfw_j^\top\bfx))(1-2\phi(\bfw_j^\top\bfx))}{\frac{1}{K}\phi(\bfw_j^\top\bfx)}\leq 1\label{xi_2nd}
\end{equation}
where (\ref{xi_1st}) holds for any $j,l\in[K]$. The case $y=0$ can be computed with the same upper bound by substituting $(1-H(\bfW))=\frac{1}{K}\sum_{j=1}^K(1-\phi(\bfw_j^\top\bfx))$ for $H(\bfW)$ in (\ref{xi_y1}), (\ref{xi_1st}) and (\ref{xi_2nd}). Therefore, there exists a constant $C_9>0$, such that
\begin{equation}
    |\xi_{j,l}(\bfW)|\leq C_9\label{xi_bound}
\end{equation}
We then need to calculate $T_{j,l,k}$. Following the analysis of $\xi_{j,l}(\bfW)$, we only consider the case of $y=1$ here for simplicity.
\begin{equation}
\begin{aligned}
T_{j,l,k}=\frac{-2}{K^3H^3(\bfW')}\phi'({\bfw'}_j^\top \bfx)\phi'({\bfw'}_l^\top  \bfx)\phi'({\bfw'}_k^\top \bfx),\ \ \ \text{where }j,l,k\text{ are not equal to each other}\label{T_jlk}
\end{aligned}
\end{equation}
\begin{equation}
T_{j,j,k}=\left\{
    \begin{array}{rcl}
    \frac{-2}{K^3H^3(\bfW')}\phi'({\bfw'}_j^\top \bfx)\phi'({\bfw'}_j^\top  \bfx)\phi'({\bfw'}_k^\top \bfx)+\frac{1}{K^2H^2(\bfW')}\phi''({\bfw'}_j^\top \bfx)\phi'({\bfw'}_k^\top \bfx),& j\neq k\\
    \frac{-2}{K^3H^3(\bfW')}(\phi'({\bfw'}_j^\top \bfx))^3+\frac{3}{K^2H^2(\bfW')}\phi''({\bfw'}_j^\top \bfx)\phi'({\bfw'}_j^\top \bfx)-\frac{\phi'''({\bfw'}_j^\top \bfx)}{KH(\bfW')}, &j=k\end{array}\right.\label{Tjjk}
\end{equation}

\begin{equation}
\begin{aligned}
     \bfa_j^\top \Delta_{j,l}\bfa_l&=\mathbb{E}_{\bfx\sim\sum_{l=1}^L\mathcal{N}(\bfmu_l,\bfSg_l)}[(\sum_{k=1}^K T_{j,l,k}\left\langle \bfx,\bfw_k-\bfW^*\bfp_k\right\rangle)\cdot(\bfa_j^\top \bfx)(\bfa_l^\top \bfx)]\\
     &\leq \sqrt{\mathbb{E}_{\bfx\sim\sum_{l=1}^L\mathcal{N}(\bfmu_l,\bfSg_l)}[\sum_{k=1}^K T_{j,k,l}^2]\cdot\mathbb{E}[\sum_{k=1}^K(\left\langle \bfx, \bfw_k-\bfW^*\bfp_k\right\rangle(\bfa_j^\top \bfx)(\bfa_l^\top \bfx))^2]}\\
     &\leq \sqrt{\mathbb{E}_{\bfx\sim\sum_{l=1}^L\mathcal{N}(\bfmu_l,\bfSg_l)}[\sum_{k=1}^K T_{j,k,l}^2]}\sqrt{\sum_{k=1}^K\sqrt{\mathbb{E}((\bfw_k-\bfW^*\bfp_k)^\top \bfx)^4}\cdot\sqrt{\mathbb{E}[(\bfa_j^\top \bfx)^4(\bfa_l^\top \bfx)^4]}}\\
     &\leq C_8\sqrt{\mathbb{E}_{\bfx\sim\sum_{l=1}^L\mathcal{N}(\bfmu_l,\bfSg_l)}[\sum_{k=1}^K T_{j,k,l}^2]}\sqrt{\sum_{k=1}^K||\bfw_k-\bfW^*\bfp_k||_2^2\cdot||\bfa_j||_2^2\cdot||\bfa_l||_2^2}\\
     &\ \ \cdot\Big(\sum_{l=1}^L\lambda_l(\|\bfmu_l\|+\|\bfSg_l^\frac{1}{2}\|)^4\sum_{l=1}^L\lambda_l(\|\bfmu_l\|+\|\bfSg_l^\frac{1}{2}\|)^8\Big)^{\frac{1}{4}}\label{d4d8}
\end{aligned}
\end{equation}
for some constant $C_8>0$. All the three inequalities of (\ref{d4d8}) are derived from Cauchy-Schwarz inequality. Note that we have
\begin{equation}
\begin{aligned}
\Big|\frac{-2}{K^3H^3(\bfW)}(\phi'(\bfw_j^\top \bfx))^2\phi'(\bfw_k^\top \bfx)\Big|&\leq\frac{2\phi^2(\bfw_j^\top \bfx)(1-\phi(\bfw_j^\top \bfx))^2\phi(\bfw_k^\top \bfx)(1-\phi(\bfw_k^\top \bfx))}{K^3\frac{1}{K^3}\phi^2(\bfw_j^\top \bfx)\phi(\bfw_k^\top \bfx)}\\
&=2(1-\phi(\bfw_j^\top \bfx))^2(1-\phi(\bfw_k^\top \bfx))\leq2\label{Tjjk_1st}
\end{aligned}
\end{equation}
\begin{equation}
\begin{aligned}
\Big|\frac{-2}{K^3H^3(\bfW)}\phi'(\bfw_j^\top \bfx)\phi'(\bfw_l^\top\bfx)\phi'(\bfw_k^\top \bfx)\Big|\leq2\label{Tjlk_term}
\end{aligned}
\end{equation}
\begin{equation}
\begin{aligned}
&\Big|\frac{3}{K^2H^2(\bfW)}\phi''(\bfw_j^\top \bfx)\phi'(\bfw_k^\top \bfx)\Big|\\
\leq&\Big|\frac{3\phi(\bfw_j^\top \bfx)(1-\phi(\bfw_j^\top \bfx))(1-2\phi(\bfw_j^\top \bfx))\phi(\bfw_k^\top \bfx)(1-\phi(\bfw_k^\top \bfx))}{K^2\frac{1}{K^2}\phi(\bfw_j^\top \bfx)\phi(\bfw_k^\top \bfx)}\Big|\\
=&\Big|3(1-\phi(\bfw_j^\top \bfx))(1-2\phi(\bfw_j^\top \bfx))(1-\phi(\bfw_k^\top \bfx))\Big|\leq3\label{Tjjk_2nd}
\end{aligned}
\end{equation}
\begin{equation}
\begin{aligned}
\Big|\frac{\phi'''(\bfw_j^\top \bfx)}{KH(\bfW)}\Big|\leq\Big|\frac{\phi(\bfw_j^\top \bfx)(1-\phi(\bfw_j^\top \bfx))(1-6\phi(\bfw_j^\top \bfx)+6\phi^2(\bfw_j^\top \bfx))}{K\frac{1}{K}\phi(\bfw_j^\top \bfx)}\Big|\leq1\label{Tjjk_3rd}
\end{aligned}
\end{equation}
Therefore, by combining (\ref{T_jlk}), (\ref{Tjjk}) and (\ref{Tjjk_1st}) to (\ref{Tjjk_3rd}), we have
\begin{equation}
    |T_{j,l,k}|\leq C_7\ \ \ \Rightarrow\ \ \ T^2_{j,l,k}\leq C_7^2, \forall j,l,k\in[K], \label{Tbound}
\end{equation}
for some constants $C_7>0$. By (\ref{C1_1st}), (\ref{delta}), (\ref{d4d8}), (\ref{Tbound}) and the Cauchy-Schwarz's Inequality, we have
\begin{equation}
\begin{aligned}
&\|\nabla^2\bar{f}(\bfW)-\nabla^2\bar{f}(\bfW^*\bfP)\|\\
\leq& C_8 \sqrt{C_7^2K}||\bfW-\bfW^*\bfP||_F \Big(\sum_{l=1}^L\lambda_l(\|\bfmu_l\|+\|\bfSg_l^\frac{1}{2}\|)^4\sum_{l=1}^L\lambda_l(\|\bfmu_l\|+\|\bfSg_l^\frac{1}{2}\|)^8\Big)^{\frac{1}{4}}\\
&\ \ \cdot\max_{||\bfa||=1}\sum_{j=1}^K\sum_{l=1}^K||\bfa_j||_2||\bfa_l||_2\\
\leq& C_8 \sqrt{C_7^2K}\cdot||\bfW-\bfW^*\bfP||_F\cdot \Big(\sum_{l=1}^L\lambda_l(\|\bfmu_l\|+\|\bfSg_l^\frac{1}{2}\|)^4\sum_{l=1}^L\lambda_l(\|\bfmu_l\|+\|\bfSg_l^\frac{1}{2}\|)^8\Big)^{\frac{1}{4}}\cdot\Big(\sum_{j=1}^K||\bfa_j||\Big)^2\\
\leq& C_8 \sqrt{C_7^2K^3}\cdot||\bfW-\bfW^*\bfP||_F\cdot \Big(\sum_{l=1}^L\lambda_l(\|\bfmu_l\|+\|\bfSg_l^\frac{1}{2}\|)^4\sum_{l=1}^L\lambda_l(\|\bfmu_l\|+\|\bfSg_l^\frac{1}{2}\|)^8\Big)^{\frac{1}{4}}
\end{aligned}
\end{equation}
Hence, we have
\begin{equation}
\begin{aligned}
&||\nabla^2\bar{f}(\bfW)-\nabla^2\bar{f}(\bfW^*\bfP)||\\
\leq&  C_5 K^{\frac{3}{2}}\Big(\sum_{l=1}^L\lambda_l(\|\bfmu_l\|+\|\bfSg_l^\frac{1}{2}\|)^4\sum_{l=1}^L\lambda_l(\|\bfmu_l\|+\|\bfSg_l^\frac{1}{2}\|)^8\Big)^{\frac{1}{4}}||\bfW-\bfW^*\bfP||_F
\end{aligned}
\end{equation}
for some constant $C_5>0$.\\\\

\subsubsection{Proof of Lemma \ref{lm: convex} }\label{sucsec: lm_convexity}
\noindent From \cite{FCL20}, we know\\

\begin{equation}
\begin{aligned}
    \nabla^2\bar{f}(\bfW^*\bfP)&\succeq\min_{||\bfa||=1}\frac{4}{K^2}\mathbb{E}_{\bfx\sim\sum_{l=1}^L\lambda_l\mathcal{N}(\bfmu_l,\bfSg_l)}\Big[\Big(\sum_{j=1}^K\phi'({\bfw_{\pi^*(j)}^*}^\top\bfx)(\bfa_{\pi^*(j)}^\top\bfx)\Big)^2\Big]\cdot\bfI_{dK}\\
    &= \min_{||\bfa||=1}\frac{4}{K^2}\mathbb{E}_{\bfx\sim\sum_{l=1}^L\lambda_l\mathcal{N}(\bfmu_l,\bfSg_l)}\Big[\Big(\sum_{j=1}^K\phi'({\bfw_{j}^*}^\top\bfx)(\bfa_{j}^\top\bfx)\Big)^2\Big]\cdot\bfI_{dK}\label{Hessian_lower}
\end{aligned}
\end{equation}

\noindent with $\bfa=(\bfa_1^\top,\cdots,\bfa_K^\top)^\top\in\mathbb{R}^{dK}$, where $\bfP$ is a specific permutation matrix and $\{\pi^*(j)\}_{j=1}^K$ is the indices permuted by $\bfP$. Similarly, 

\begin{equation}
\begin{aligned}
    \nabla^2 \bar{f}(\bfW^*\bfP)&\preceq \Big(\max_{||\bfa||=1}\bfa^\top\nabla^2 \bar{f}(\bfW^*)\bfa\Big)\cdot\bfI_{dK}\preceq C_4\cdot\max_{||\bfa||=1}\mathbb{E}_{\bfx\sim\sum_{l=1}^L\lambda_l\mathcal{N}(\bfmu_l,\bfSg_l)}\Big[\sum_{j=1}^K(\bfa_{\pi^*(j)}^\top\bfx)^2\Big]\cdot\bfI_{dK}\\
    &=C_4\cdot\max_{||\bfa||=1}\mathbb{E}_{\bfx\sim\sum_{l=1}^L\lambda_l\mathcal{N}(\bfmu_l,\bfSg_l)}\Big[\sum_{j=1}^K(\bfa_{j}^\top\bfx)^2\Big]\cdot\bfI_{dK}\label{Hessian_upper}
\end{aligned}
\end{equation}

\noindent for some constant $C_4>0$. By applying Property \ref{prop: bound}, we can derive the upper bound in (\ref{Hessian_upper}) as\\
\begin{equation}C_4\cdot\mathbb{E}_{\bfx\sim\sum_{l=1}^L\lambda_l\mathcal{N}(\bfmu_l,\bfSg_l)}\Big[\sum_{j=1}^K(\bfa_j^\top\bfx)^2\Big]\cdot\bfI_{dK}\preceq C_4\cdot \sum_{l=1}^L\lambda_l(\|\bfmu_l\|+\|\bfSg_l^\frac{1}{2}\|)^2\cdot\bfI_{dK}\end{equation}
To find a lower bound for (\ref{Hessian_lower}), we can first transfer the expectation  of the Gaussian Mixture Model to the weight sum of the expectations over general Gaussian distributions.
\begin{equation}
\begin{aligned}
&\min_{||\bfa||=1}\mathbb{E}_{\bfx\sim\sum_{l=1}^L\lambda_l\mathcal{N}(\bfmu_l,\bfSg_l)}\Big[\Big(\sum_{j=1}^K\phi'({\bfw_j^*}^\top\bfx)(\bfa_j^\top\bfx)\Big)^2\Big]\\
=&\min_{||\bfa||=1}\sum_{l=1}^L\lambda_l\mathbb{E}_{\bfx\sim\mathcal{N}(\bfmu_l,\bfSg_l)}\Big[\Big(\sum_{j=1}^K\phi'({\bfw_j^*}^\top\bfx)(\bfa_j^\top\bfx)\Big)^2\Big]\label{low_derivation}
\end{aligned}
\end{equation}
Denote $\bfU\in\mathbb{R}^{d\times k}$ as the orthogonal basis of $\bfW^*$. For any vector $\bfa_i\in\mathbb{R}^d$, there exists two vectors $\bfb_i\in\mathbb{R}^K$ and $\bfc_i\in\mathbb{R}^{d-K}$ such that\\
\begin{equation}
\bfa_i=\bfU \bfb_i+\bfU_\perp \bfc_i\label{decomposition}
\end{equation}
where $\bfU_\perp\in\mathbb{R}^{d\times(d-K)}$ denotes the complement of $\bfU$. We also have $\bfU_\perp^\top \bfmu_l=0$ by Property \ref{prop: column space}. Plugging (\ref{decomposition}) into RHS of (\ref{low_derivation}), and then we have
\begin{equation}
\begin{aligned}
&\mathbb{E}_{\bfx\sim\mathcal{N}(\bfmu_l,\bfSg_l)} \Big[\Big(\sum_{i=1}^K \bfa_i^\top \bfx\cdot\phi'({\bfw_i^*}^\top \bfx)\Big)^2\Big]\\
=&\mathbb{E}_{\bfx\sim\mathcal{N}(\bfmu_l,\bfSg_l)}\Big[\Big(\sum_{i=1}^K (\bfU \bfb_i+\bfU_\perp \bfc_i)^\top \bfx\cdot\phi'({\bfw_i^*}^\top \bfx)\Big)^2\Big]=A+B+C\label{C2_ABC}
\end{aligned}
\end{equation}
\begin{equation}
   \begin{aligned}
   A=\mathbb{E}_{\bfx\sim\mathcal{N}(\bfmu_l,\bfSg_l)} \Big[\Big(\sum_{i=1}^K  \bfb_i^\top \bfU^\top \bfx\cdot\phi'({\bfw_i^*}^\top \bfx)\Big)^2\Big]\label{C2_A}
   \end{aligned}
\end{equation}
\begin{equation}
\begin{aligned}
C&=\mathbb{E}_{\bfx\sim\mathcal{N}(\bfmu_l,\bfSg_l)} \Big[2\Big(\sum_{i=1}^K \bfc_i^\top \bfU_\perp^\top \bfx\cdot\phi'({\bfw_i^*}^\top \bfx)\Big)\cdot\Big(\sum_{i=1}^K \bfb_i^\top  \bfU^\top \bfx\cdot \phi'({\bfw_i^*}^\top \bfx)\Big)\Big]\\
&=\sum_{i=1}^K\sum_{j=1}^K\mathbb{E}_{\bfx\sim\mathcal{N}(\bfmu_l,\bfSg_l)}\Big[ 2\bfc_i^\top \bfU_\perp^\top \bfx\Big]\mathbb{E}_{\bfx\sim\mathcal{N}(\bfmu_l,\bfSg_l)}\Big[\bfb_i^\top \bfU^\top \bfx \cdot\phi'({\bfw_i^*}^\top \bfx)\phi'({\bfw_j^*}^\top \bfx)\Big]\\
&=\sum_{i=1}^K\sum_{j=1}^K\Big[ 2\bfc_i^\top \bfU_\perp^\top \bfmu_l\Big]\mathbb{E}_{\bfx\sim\mathcal{N}(\bfmu_l,\bfSg_l)}\Big[\bfb_i^\top \bfU^\top \bfx \cdot\phi'({\bfw_i^*}^\top \bfx)\phi'({\bfw_j^*}^\top \bfx)\Big]=0\label{C2_C}
\end{aligned}
\end{equation}
where the last step is by $\bfU_{\perp}^\top\bfmu_l=0$ by Property \ref{prop: column space}.
\begin{equation}
\begin{aligned}
  B=&\mathbb{E}_{\bfx\sim\mathcal{N}(\bfmu_l,\bfSg_l)}\Big[(\sum_{i=1}^K \bfc_i^\top \bfU_\perp^\top \bfx\cdot\phi'({\bfw_i^*}^\top \bfx))^2\Big]\\
  =&\mathbb{E}_{\bfx\sim\mathcal{N}(\bfmu_l,\bfSg_l)}[(\bft^\top \bfs)^2]\ \ \ \ \ \ \ \ \ \ \ \ \text{by defining }\bft=\sum_{i=1}^k \phi'({\bfw_i^*}^\top \bfx)\bfc_i\in\mathbb{R}^{d-K}\text{ and }\bfs=\bfU_\perp^\top \bfx\\
  =&\sum_{i=1}^K\mathbb{E}[t_i^2s_i^2]+\sum_{i\neq j}\mathbb{E}[t_it_js_is_j]\\
  =&\sum_{i=1}^K\mathbb{E}[t_i^2]\sum_{k=1}^d (\bfU_\perp)_{ik}^2\sigma_{lk}^2+\Big(\sum_{i=1}^K\mathbb{E}[t_i^2](\bfU_{\perp}^\top\bfmu_l)_i^2+\sum_{i\neq j}\mathbb{E}[t_it_j](\bfU_\perp^\top\bfmu_l)_i\cdot(\bfU_\perp^\top\bfmu_l)_j\Big)\\
  =&\mathbb{E}[\sum_{i=1}^{d-K}t_i^2\cdot \sum_{k=1}^d (\bfU_\perp)_{ik}^2\sigma_{lk}^2]+\mathbb{E}[(\bft^\top \bfU_\perp^\top \bfmu_l)^2]=\mathbb{E}[\sum_{i=1}^{d-K}t_i^2\cdot \sum_{k=1}^d (\bfU_\perp)_{ik}^2\sigma_{lk}^2]\label{C2_B}
\end{aligned}
\end{equation}
The last step is by $\bfU_{\perp}^\top\bfmu_l=0$. The 4th step is because that $s_i$ is independent of $t_i$, thus $\mathbb{E}[t_it_js_is_j]=\mathbb{E}[t_it_j]\mathbb{E}[s_is_j]$\\
\begin{equation}\mathbb{E}[s_is_j]=\left\{
    \begin{array}{rcl}
    (\bfU_\perp^\top \bfmu_l)_i \cdot(\bfU_\perp^\top \bfmu_l)_j, &\text{if }i\neq j\\
    (\bfU_\perp^\top\bfmu_l)_i^2+\sum_{k=1}^d (\bfU_\perp)_{ik}^2\sigma_{lk}^2, & \text{if }i=j
    \end{array}\right.\end{equation}
Since $\Big(\sum_{i=1}^k \bfr_i^\top \bfx\cdot \phi'(\sigma\cdot x_i)\Big)^2$ is an even function for any $\bfr_i\in\mathbb{R}^{d},\ i\in[k]$, so from Property \ref{prop: even} we have
\begin{equation}\mathbb{E}_{\bfx\sim\mathcal{N}(\bfmu_l,\bfSg_l)}\Big[(\sum_{i=1}^k \bfr_i^\top \bfx\cdot\phi'(\sigma\cdot x_i))^2\Big] = \mathbb{E}_{\bfx\sim\frac{1}{2}\mathcal{N}(\bfmu_l,\bfSg_l)+\frac{1}{2}\mathcal{N}(-\bfmu_l,\bfSg_l)}\Big[(\sum_{i=1}^k \bfr_i^\top \bfx\cdot\phi'(\sigma\cdot x_i))^2\Big]\end{equation}
Combining Lemma \ref{lm: rho} and Property \ref{prop: even}, we next follow the derivation for the standard Gaussian distribution in Page 36 of \cite{ZSJB17} and generalize the result to a Gaussian distribution with an arbitrary mean and variance as follows.\\
\begin{equation}
\begin{aligned}
A&=\mathbb{E}_{\bfx\sim\mathcal{N}(\bfmu_l,\bfSg_l)} \Big[\Big(\sum_{i=1}^K  \bfb_i^\top \bfU^\top \bfx\cdot\phi'({\bfw_i^*}^\top \bfx)\Big)^2\Big]\\
&\geq \int (2\pi)^{-\frac{K}{2}}|\bfU^\top\bfSg_l\bfU|^{-\frac{1}{2}}\Big[\Big(\sum_{i=1}^K  \bfb_i^\top \bfz\cdot\phi'({\bfv_i}^\top \bfz)\Big)^2\Big]\exp\big(-\frac{1}{2}\|\bfSg_l^{-1}\|\|\bfz-\bfU^\top\bfmu_l\|^2\big)d\bfz\\
&=\int (2\pi)^{-\frac{K}{2}}|\bfU^\top\bfSg_l\bfU|^{-\frac{1}{2}}\Big[\Big(\sum_{i=1}^K  \bfb_i^\top {\bfV^{\dagger}}^\top\bfs\cdot\phi'(s_i)\Big)^2\Big]\exp\big(-\frac{1}{2}\|\bfSg_l^{-1}\|\|{\bfV^{\dagger}}^\top\bfs-\bfU^\top\bfmu_l\|^2\big)\Big|\text{det}(\bfV^{\dagger})\Big|d\bfs\\
&\geq\int (2\pi)^{-\frac{K}{2}}|\bfU^\top\bfSg_l\bfU|^{-\frac{1}{2}}\Big[\Big(\sum_{i=1}^k  \bfb_i^\top {\bfV^{\dagger}}^\top\bfs\cdot\phi'(s_i)\Big)^2\Big]\exp\big(-\frac{\|\bfSg_l^{-1}\|\|\bfs-{\bfV}^\top\bfU^\top\bfmu_l\|^2}{2\delta_K^2(\bfW^*)}\big)\Big|\text{det}(\bfV^{\dagger})\Big|d\bfs\\
&\geq\int (2\pi)^{-\frac{K}{2}}|\bfU^\top\bfSg_l\bfU|^{-\frac{1}{2}}\Big[\Big(\sum_{i=1}^k  \bfb_i^\top {\bfV^{\dagger}}^\top(\delta_K(\bfW^*)\|\bfSg_l^{-1}\|^{-\frac{1}{2}})\bfg\cdot\phi'(\delta_K(\bfW^*)\|\bfSg_l^{-1}\|^{-\frac{1}{2}}\cdot g_i)\Big)^2\Big]\\
&\ \ \ \ \cdot\exp\big(-\frac{||\bfg-\frac{{\sqrt{\|\bfSg_l^{-1}\|}\bfW^*}^\top\bfmu_l}{\delta_K(\bfW^*)}||^2}{2}\big)\Big|\text{det}(\bfV^{\dagger})\Big|\|\bfSg_l^{-1}\|^{-\frac{K}{2}}\delta_K^{K}(\bfW^*)d\bfg\\
&=\frac{\|\bfSg_l^{-1}\|^{-1}}{\tau^K\eta}\mathbb{E}_\bfg\Big[(\sum_{i=1}^K (\bfb_i^\top {\bfV^{\dagger}}^\top \delta_K(\bfW^*)) \bfg\cdot \phi'(\|\bfSg_l^{-1}\|^{-\frac{1}{2}}\delta_K(\bfW^*)\cdot g_i))^2\Big]\\
&\geq\frac{\|\bfSg_l^{-1}\|^{-1}}{\tau^K\kappa^2\eta}\rho(\frac{{\bfW^*}^\top\bfmu_l}{\|\bfSg_l^{-1}\|^{-\frac{1}{2}}\delta_K(\bfW^*)}, \|\bfSg_l^{-1}\|^{-\frac{1}{2}}\delta_K(\bfW^*))||\bfb||^2.\label{A_derivation}
\end{aligned}
\end{equation}
The second step is by letting $\bfz=\bfU^\top\bfx\sim\mathcal{N}(\bfU^\top\bfmu_l,\bfU^\top\bfSg \bfU)$, $\bfy^\top\bfU^\top\bfSg_l^{-1}\bfU\bfy\leq \|\bfSg_l^{-1}\|\|\bfy\|^2$ for any $\bfy\in\mathbb{R}^K$. The third step is by letting $\bfs=\bfV^\top\bfz$. The last to second step follows from $\bfg=\frac{\bfs}{\|\bfSg_l^{-1}\|^{-\frac{1}{2}}\delta_K(\bfW^*)}$, where $\bfg\sim\mathcal{N}(\frac{{\bfW^*}^\top\bfmu_l}{\|\bfSg_l^{-1}\|^{-\frac{1}{2}}\delta_K(\bfW^*)},\boldsymbol{I}_K)$ and the last inequality is by Lemma \ref{lm: rho}. Similarly, we extend the derivation in Page 37 of \cite{ZSJB17} for the standard Gaussian distribution to a general Gaussian distribution as follows.\\
\begin{equation}
    \begin{aligned}
    B=\sum_{k=1}^d (\bfU_\perp)_{ik}^2\sigma_{lk}^2\mathbb{E}_{\bfx\sim\mathcal{N}(\bfmu_l,\bfSg_l)}[||\bft||^2]\geq\frac{\|\bfSg_l^{-1}\|^{-1}}{\eta\tau^K\kappa^2}\rho(\frac{{\bfW^*}^\top\bfmu_l}{\|\bfSg_l^{-1}\|^{-\frac{1}{2}}|\delta_K(\bfW^*)}, \|\bfSg_l^{-1}\|^{-\frac{1}{2}}\delta_K(\bfW^*))||\bfc||^2\label{B_derivation}
    \end{aligned}
\end{equation}
Combining (\ref{C2_ABC}) - (\ref{C2_B}), (\ref{A_derivation}) and (\ref{B_derivation}), we have
\begin{equation}\min_{||\boldsymbol{a}||=1}\mathbb{E}_{\bfx\sim\mathcal{N}(\bfmu_l,\bfSg_l)}\Big[(\sum_{i=1}^k \bfa_i^\top \bfx\cdot\phi'({\bfw_i^*}^\top \bfx))^2\Big]\geq\frac{\|\bfSg_l^{-1}\|^{-1}}{\eta\tau^K\kappa^2}\rho(\frac{{\bfW^*}^\top\bfmu_l}{\delta_K(\bfW^*)\|\bfSg_l^{-1}\|^{-\frac{1}{2}}},\delta_K(\bfW^*)\|\bfSg_l^{-1}\|^{-\frac{1}{2}}).\end{equation}
\\
For the Gaussian Mixture Model $\bfx\sim\sum_{l=1}^L\mathcal{N}(\bfmu_l,\bfSg)$, we have
\begin{equation}
\begin{aligned}
&\min_{\|\bfa\|=1}\mathbb{E}_{\bfx\sim\sum_{l=1}^L\lambda_l\mathcal{N}(\bfmu_l,\bfSg_l)}\Big[(\sum_{i=1}^k \bfa_i^\top \bfx\cdot\phi'({\bfw_i^*}^\top \bfx))^2\Big]\\
\geq &\sum_{l=1}^L\lambda_l\frac{\|\bfSg_l^{-1}\|^{-1}}{\eta\tau^K\kappa^2}\rho(\frac{{\bfW^*}^\top\bfmu_l}{\delta_K(\bfW^*)\|\bfSg_l^{-1}\|^{-\frac{1}{2}}},\delta_K(\bfW^*)\|\bfSg_l^{-1}\|^{-\frac{1}{2}})
\end{aligned}
\end{equation}
Therefore, 
\begin{equation}
\begin{aligned}&\frac{4}{K^2}\sum_{l=1}^L\lambda_l\frac{\|\bfSg_l^{-1}\|^{-1}}{\eta\tau^K\kappa^2}\rho(\frac{{\bfW^*}^\top\bfmu_l}{\delta_K(\bfW^*)\|\bfSg_l^{-1}\|^{-\frac{1}{2}}},\delta_K(\bfW^*)\|\bfSg_l^{-1}\|^{-\frac{1}{2}})\cdot\bfI_{dK}\\
\preceq&\nabla^2\bar{f}(\bfW^*\bfP)\preceq C_4\cdot\sum_{l=1}^L\lambda_l(\|\bfmu_l\|+\|\bfSg_l^\frac{1}{2}\|)^2\cdot\bfI_{dK}
\end{aligned}\end{equation}
From (\ref{smooth}) in Lemma \ref{lm: smoothness}, since that we have the condition $\|\bfW-\bfW^*\bfP\|_F\leq r$ and (\ref{radius}), we can obtain
\begin{equation}
    \begin{aligned}
        &||\nabla^2 \bar{f}(\bfW)-\nabla^2 \bar{f}(\bfW^*\bfP)||\\
        \leq& C_5 K^{\frac{3}{2}}\Big(\sum_{l=1}^L\lambda_l(\|\bfmu_l\|+\|\bfSg_l^{\frac{1}{2}}\|)^4\sum_{l=1}^L\lambda_l(\|\bfmu_l\|+\|\bfSg_l^{\frac{1}{2}}\|)^8\Big)^{\frac{1}{4}}||\bfW-\bfW^*\bfP||_F\\
        \leq& \frac{4\epsilon_0}{K^2}\sum_{l=1}^L\lambda_l\frac{\|\bfSg_l^{-1}\|^{-1}}{\eta\tau^K\kappa^2}\rho(\frac{{\bfW^*}^\top\bfmu_l}{\delta_K(\bfW^*)\|\bfSg_l^{-1}\|^{-\frac{1}{2}}},\delta_K(\bfW^*)\|\bfSg_l^{-1}\|^{-\frac{1}{2}}),
    \end{aligned}
\end{equation}
where $\epsilon_0\in(0,\frac{1}{4})$. 
Then we have
\begin{equation}
\begin{aligned}
    ||\nabla^2\bar{f}(\bfW)||&\geq ||\nabla^2\bar{f}(\bfW^*\bfP)||-||\nabla^2 \bar{f}(\bfW)-\nabla^2\bar{f}(\bfW^*\bfP)||\\
    &\geq \frac{4(1-\epsilon_0)}{K^2}\sum_{l=1}^L\lambda_l\frac{\|\bfSg_l^{-1}\|^{-1}}{\eta\tau^K\kappa^2}\rho(\frac{{\bfW^*}^\top\bfmu_l}{\delta_K(\bfW^*)\|\bfSg_l^{-1}\|^{-\frac{1}{2}}},\delta_K(\bfW^*)\|\bfSg_l^{-1}\|^{-\frac{1}{2}})\label{f_lower}
\end{aligned}
\end{equation}
\begin{equation}
\begin{aligned}
    ||\nabla^2\bar{f}(\bfW)||&\leq ||\nabla^2\bar{f}(\bfW^*)||+||\nabla^2 \bar{f}(\bfW)-\nabla^2\bar{f}(\bfW^*\bfP)||\\
    &\leq C_4\cdot \sum_{l=1}^L\lambda_l(\|\bfmu_l\|+\|\bfSg^{\frac{1}{2}}\|)^2+ \frac{4}{K^2}\sum_{l=1}^L\lambda_l\frac{\|\bfSg_l^{-1}\|^{-1}}{\eta\tau^K\kappa^2}\rho(\frac{{\bfW^*}^\top\bfmu_l}{\delta_K(\bfW^*)\|\bfSg_l^{-\frac{1}{2}}\|},\delta_K(\bfW^*)\|\bfSg_l^{-\frac{1}{2}}\|)\\
    &\lesssim C_4\cdot \sum_{l=1}^L\lambda_l(\|\bfmu_l\|+\|\bfSg_l^{\frac{1}{2}}\|)^2\label{f_upper}
\end{aligned}
\end{equation}
The last inequality of (\ref{f_upper}) holds since $C_4\cdot \sum_{l=1}\lambda_l(\|\bfmu_l\|+\|\bfSg_l^{\frac{1}{2}}\|)^2=\Omega(\max_l\{\|\bfSg_l\|\})$, $\frac{4}{K^2}\sum_{l=1}^L\lambda_l\frac{\|\bfSg_l^{-1}\|^{-1}}{\eta\tau^K\kappa^2}\rho(\frac{{\bfW^*}^\top\bfmu_l}{\delta_K(\bfW^*)\|\bfSg_l^{-1}\|^{-\frac{1}{2}}},\delta_K(\bfW^*)\|\bfSg_l^{-1}\|^{-\frac{1}{2}})=O(\frac{\max_l\{\|\bfSg_l\|\}}{K^2})$  and $\Omega(\max_l\{\|\bfSg_l\|\})\geq O(\frac{\max_l\{\|\bfSg_l\|\}}{K^2})$. Combining (\ref{f_lower}) and (\ref{f_upper}), we have
\begin{equation}
\begin{aligned}
&\frac{4(1-\epsilon_0)}{K^2}\sum_{l=1}^L\lambda_l\frac{\|\bfSg_l^{-1}\|^{-1}}{\eta\tau^K\kappa^2}\rho(\frac{{\bfW^*}^\top\bfmu_l}{\delta_K(\bfW^*)\|\bfSg_l^{-1}\|^{-\frac{1}{2}}},\delta_K(\bfW^*)\|\bfSg_l^{-1}\|^{-\frac{1}{2}})\cdot\bfI\\
\preceq&\nabla^2\bar{f}(\bfW)\preceq C_4\cdot \sum_{l=1}^L\lambda_l(\|\bfmu_l\|+\sigma_l)^2\cdot\bfI
\end{aligned}
\end{equation}\\

\subsubsection{Proof of Lemma \ref{lm: bernstein}}\label{subsec: pf lm_bernstein}
\noindent Let $N_\epsilon$ be the $\epsilon$-covering number of the Euclidean ball $\mathbb{B}(\bfW^*\bfP,r)$. It is known that $\log{N_\epsilon}\leq dK\log(\frac{3r}{\epsilon})$ from \cite{V10}. Let $\mathcal{W}_\epsilon=\{\bfW_1,...,\bfW_{N_\epsilon}\}$ be the $\epsilon$-cover set with $N_\epsilon$ elements. For any $\bfW\in\mathbb{B}(\bfW^*\bfP,r)$, let $j(\bfW)=\mathop{\arg\min}\limits_{j\in[N_\epsilon]}||\bfW-\bfW_{j(\bfW)}||_F\leq\epsilon$ for all $\bfW\in \mathbb{B}(\bfW^*\bfP,r)$.\\
Then for any $\bfW\in\mathbb{B}(\bfW^*\bfP,r)$, we have 
\begin{equation}
\begin{aligned}
  &\|\nabla^2 f_n(\bfW)-\nabla^2 \bar{f}(\bfW)\|\\
  \leq&\frac{1}{n}||\sum_{i=1}^n[\nabla^2\ell(\bfW;\bfx_i)-\nabla^2\ell(\bfW_{j(\bfW)};\bfx_i)]||\\
  &+||\frac{1}{n}\sum_{i=1}^n \nabla^2\ell(\bfW_{j(\bfW)};\bfx_i)-\mathbb{E}_{\bfx\sim\sum_{l=1}^L\lambda_l\mathcal{N}(\bfmu_l,\bfSg_l)}[\nabla^2\ell(\bfW_{j(\bfW)};\bfx_i)]||\\
    &+||\mathbb{E}_{\bfx\sim\sum_{l=1}^L\lambda_l\mathcal{N}(\bfmu_l,\bfSg_l)}[\nabla^2\ell(\bfW_{j(\bfW)};\bfx_i)]-\mathbb{E}_{\bfx\sim\sum_{l=1}^L\lambda_l\mathcal{N}(\bfmu_l,\bfSg_l)}[\nabla^2\ell(\bfW;\bfx_i)]||\label{distance3parts}
\end{aligned}
\end{equation}
Hence, we have 
\begin{equation}\mathbb{P}\Big(\sup_{\bfW\in\mathbb{B}(\bfW^*\bfP,r)}||\nabla^2 f_n(\bfW)-\nabla^2 \bar{f}(\bfW)||\geq t\Big)\leq\mathbb{P}(A_t)+\mathbb{P}(B_t)+\mathbb{P}(C_t)\label{Prob_bound}
\end{equation}
where $A_t$, $B_t$ and $C_t$ are defined as
\begin{equation}
A_t=\{\sup_{\bfW\in\mathbb{B}(\bfW^*\bfP,r)}\frac{1}{n}||\sum_{i=1}^n[\nabla^2\ell(\bfW;\bfx_i)-\nabla^2\ell(\bfW_{j(\bfW)};\bfx_i)]||\geq\frac{t}{3}\}\label{A_t}
\end{equation}
\begin{equation}B_t=\{\sup_{\bfW\in\mathbb{B}(\bfW^*\bfP,r)}||\frac{1}{n}\sum_{i=1}^n \nabla^2\ell(\bfW_{j(\bfW)};\bfx_i)-\mathbb{E}_{\bfx\sim\sum_{l=1}^L\lambda_l\mathcal{N}(\bfmu_l,\bfSg_l)}[\nabla^2\ell(\bfW_{j(\bfW)};\bfx_i)]||\geq\frac{t}{3}\}\label{B_t}
\end{equation}
\begin{equation}
\begin{aligned}
C_t=&\{\sup_{\bfW\in\mathbb{B}(\bfW^*\bfP,r)}||\mathbb{E}_{\bfx\sim\sum_{l=1}^L\lambda_l\mathcal{N}(\bfmu_l,\bfSg_l)}[\nabla^2\ell(\bfW_{j(\bfW)};\bfx_i)]\\
&-\mathbb{E}_{\bfx\sim\sum_{l=1}^L\lambda_l\mathcal{N}(\bfmu_l,\bfSg_l)}[\nabla^2\ell(\bfW;\bfx_i)]||\geq\frac{t}{3}\}\label{C_t}
\end{aligned}
\end{equation}
Then we bound $\mathbb{P}(A_t)$, $\mathbb{P}(B_t)$, and $\mathbb{P}(C_t)$ separately. \\\\
1) \textbf{Upper bound on }$\mathbb{P}(B_t)$. By Lemma 6 in \cite{FCL20}, we obtain\\
\begin{equation}
\begin{aligned}
&\Big|\Big|\frac{1}{n}\sum_{i=1}^n\nabla^2\ell(\bfW;\bfx_i)-\mathbb{E}_{\bfx\sim\sum_{l=1}^L\lambda_l\mathcal{N}(\bfmu_l,\bfSg_l)}[\nabla^2 \ell(\bfW;\bfx_i)]\Big|\Big|\\
\leq& 2\sup_{\bfv\in \bfV_{\frac{1}{4}}}\Big|\left\langle  \bfv,(\frac{1}{n}\sum_{i=1}^n\nabla^2\ell(\bfW;\bfx_i)-\mathbb{E}_{\bfx\sim\sum_{l=1}^L\lambda_l\mathcal{N}(\bfmu_l,\bfSg_l)}[\nabla^2\ell(\bfW;\bfx_i)])\bfv\right\rangle\Big|\label{bound_double}
\end{aligned}
\end{equation}
where $\bfV_{\frac{1}{4}}$ is a $\frac{1}{4}$-cover of the unit-Euclidean-norm ball $\mathbb{B}(\boldsymbol{0},1)$ with $\log|\bfV_{\frac{1}{4}}|\leq dK\log{12}$. Taking the union bound over $\mathcal{W}_\epsilon$ and $\bfV_{\frac{1}{4}}$, we have
\begin{equation}
\begin{aligned}
\mathbb{P}(B_t)\leq &\mathbb{P}\Big(\sup_{\bfW\in\mathcal{W_\epsilon}, \bfv\in \bfV_{\frac{1}{4}}} \Big|\frac{1}{n}\sum_{i=1}^n G_i\Big|\geq\frac{t}{6}\Big)\\ \leq& \exp(dK(\log{\frac{3r}{\epsilon}+\log{12}}))\sup_{\bfW\in\mathcal{W}_\epsilon, \bfv\in \bfV_{\frac{1}{4}}}\mathbb{P}(|\frac{1}{n}\sum_{i=1}^n G_i|\geq\frac{t}{6})\label{P_Bt_initial}
\end{aligned}
\end{equation}
where $G_i=\left\langle \bfv,(\nabla^2\ell(\bfW,\bfx_i)-\mathbb{E}_{\bfx\sim\sum_{l=1}^L\lambda_l\mathcal{N}(\bfmu_l,\bfSg_l)}[\nabla^2\ell(\bfW,\bfx_i)]\bfv)\right\rangle$ and $\mathbb{E}[G_i]=0$. Here $\bfv=(\bfu_1^\top,\cdots,\bfu_K^\top)^\top\in\mathbb{R}^{dK}$.\\
\begin{equation}
\begin{aligned}
   |G_i|&=\Big|\sum_{j=1}^K\sum_{l=1}^K\Big[\xi_{j,l}\bfu_j^\top \bfx \bfx^\top \bfu_l-\mathbb{E}_{\bfx\sim\sum_{l=1}^L\lambda_l\mathcal{N}(\bfmu_l,\bfSg_l)}(\xi_{j,l}\bfu_j^\top \bfx \bfx^\top \bfu_l)\Big]\Big|\\
   &\leq C_9\cdot \Big[ \sum_{j=1}^K(\bfu_j^\top \bfx)^2+\sum_{j=1}^K\mathbb{E}_{\bfx\sim\sum_{l=1}^L\lambda_l\mathcal{N}(\bfmu_l,\bfSg_l)}(\bfu_j^\top \bfx)^2\Big]\label{G_i}
\end{aligned}
\end{equation}
for some $C_9>0$. The first step of (\ref{G_i}) is by (\ref{Hessian}). The last step is by (\ref{xi_bound}) and the Cauchy-Schwarz's Inequality.\\ %{\color{blue}$\sum_{j=1}^K\sum_{l=1}^K(u_j^\top x)(u_l^\top x)\leq \sqrt{\sum_{j=1}^K(u_j^\top x)^2\sum_{l=1}^K(u_l^\top x)^2}=\sum_{j=1}^K(u_j^\top x)^2$}
\begin{equation}
\begin{aligned}
    \mathbb{E}[|G_i|^p]&\leq\sum_{l=1}^p\binom{p}{l}C_9\cdot\mathbb{E}_{\bfx\sim\sum_{l=1}^L\lambda_l\mathcal{N}(\bfmu_l,\bfSg_l)}\Big[(\sum_{j=1}^K(\bfu_j^\top \bfx)^{2})^{l}\Big]\\
    &\ \cdot\Big(\sum_{j=1}^K\mathbb{E}_{\bfx\sim\sum_{l=1}^L\lambda_l\mathcal{N}(\bfmu_l,\bfSg_l)}(\bfu_j^\top \bfx)^{2}\Big)^{p-l}\\
    &= \sum_{l=1}^p \binom{p}{l}C_9\cdot \mathbb{E}_{\bfx\sim\sum_{l=1}^L\lambda_l\mathcal{N}(\bfmu_l,\bfSg_l)}\Big[\sum_{l_1+\cdots+l_K=l}\frac{l!}{\prod_{j=1}^K l_j!}\prod_{j=1}^K (\bfu_j^\top \bfx)^{2l_j}\Big]\\
    &\ \ \ \cdot\Big(\sum_{j=1}^K\mathbb{E}_{\bfx\sim\sum_{l=1}^L\lambda_l\mathcal{N}(\bfmu_l,\bfSg_l)}(\bfu_j^\top \bfx)^{2}\Big)^{p-l}\\
    &=\sum_{l=1}^p \binom{p}{l}C_9\cdot \Big[\sum_{l_1+\cdots+l_K=l}\frac{l!}{\prod_{j=1}^K l_j!}\prod_{j=1}^K \mathbb{E}_{\bfx\sim\sum_{l=1}^L\lambda_l\mathcal{N}(\bfmu_l,\bfSg_l)}(\bfu_j^\top \bfx)^{2l_j}\Big]\\
    &\ \ \ \cdot\Big(\sum_{j=1}^K\mathbb{E}_{\bfx\sim\sum_{l=1}^L\lambda_l\mathcal{N}(\bfmu_l,\bfSg_l)}(\bfu_j^\top \bfx)^{2}\Big)^{p-l}\\
    &=C_9\cdot\sum_{l=1}^p\binom{p}{l}\Big(\sum_{j=1}^K\mathbb{E}_{\bfx\sim\sum_{l=1}^L\lambda_l\mathcal{N}(\bfmu_l,\bfSg_l)}(\bfu_j^\top \bfx)^{2}\Big)^{l}\\
    &\ \cdot\Big(
    \sum_{j=1}^K\mathbb{E}_{\bfx\sim\sum_{l=1}^L\lambda_l\mathcal{N}(\bfmu_l,\bfSg_l)}(\bfu_j^\top \bfx)^{2}\Big)^{p-l}\\
    &=C_9 \cdot\Big(\sum_{j=1}^K\mathbb{E}_{\bfx\sim\sum_{l=1}^L\lambda_l\mathcal{N}(\bfmu_l,\bfSg_l)}(\bfu_j^\top \bfx)^2\Big)^{p}\\
    &\leq C_9\cdot\Big(\sum_{j=1}^K 1!!||\bfu_j||^2 \sum_{l=1}^L\lambda_l(\|\bfmu_l\|+\|\bfSg_l^\frac{1}{2}\|)^2\Big)^p \ \\
    &\leq C_9\cdot \Big(\sum_{l=1}^L\lambda_l(\|\bfmu_l\|+\|\bfSg_l^\frac{1}{2}\|)^2\Big)^p \label{E_Gi^p}
\end{aligned}
\end{equation}
where the first step is by the triangle inequality and the Binomial theorem, and the second step comes from the Multinomial theorem. The second to last inequality in (\ref{E_Gi^p}) results from Property \ref{prop: bound}. The last inequality is because $\bfv\in\bfV_{\frac{1}{4}}$, $\sum_{j=1}^K||u_j||^2=||\bfv||^2\leq1$.
\begin{equation}
\begin{aligned}
    \mathbb{E}[\exp(\theta G_i)]&=1+\theta\mathbb{E}[G_i]+\sum_{p=2}^\infty\frac{\theta^p\mathbb{E}[|G_i|^p]}{p!}\\
    &\leq1+\sum_{p=2}^\infty \frac{|e\theta|^p}{p^p}C_9\cdot \Big(\sum_{l=1}\lambda_l(\|\bfmu_l\|+\|\bfSg_l^\frac{1}{2}\|)^2\Big)^p  \\
    &\leq1+C_9\cdot|e\theta|^2\Big(\sum_{l=1}^L\lambda_l(\|\bfmu_l\|+\|\bfSg_l^\frac{1}{2}\|)^2\Big)^2  \label{G_i_mgf}
\end{aligned}
\end{equation}
where the first inequality holds from  $p!\geq(\frac{p}{e})^p$ and (\ref{E_Gi^p}), and the third line holds provided that \begin{equation}\label{eqn:maxp}
\max_{p\geq2}\{\frac{\frac{|e\theta|^{(p+1)}}{{(p+1)}^{(p+1)}}\cdot \Big(\sum_{l=1}^L\lambda_l(\|\bfmu_l\|+\|\bfSg_l^\frac{1}{2}\|)^2\Big)^{p+1}}{\frac{|e\theta|^p}{p^p}\cdot \Big(\sum_{l=1}^L\lambda_l(\|\bfmu_l\|+\|\bfSg_l^\frac{1}{2}\|)^2\Big)^p}\}\leq\frac{1}{2}
\end{equation}
Note that the quantity inside the maximization in (\ref{eqn:maxp})  achieves its maximum when $p=2$, because it is monotonously decreasing. Therefore, (\ref{eqn:maxp})
holds if  $\theta\leq\frac{27}{4e}\sum_{l=1}^L\lambda_l(\|\bfmu_l\|+\|\bfSg_l^\frac{1}{2}\|)^2$.  Then
\begin{equation}
\begin{aligned}\label{B_t_quadratic}
\mathbb{P}\Big(\frac{1}{n}\sum_{i=1}^n G_i\geq\frac{t}{6}\Big)&=\mathbb{P}\Big(\exp(\theta\sum_{i=1}^n G_i)\geq\exp(\frac{n\theta t}{6})\Big)\leq  e^{-\frac{n\theta t}{6}}\prod_{i=1}^n\mathbb{E}[\exp(\theta G_i)]\\
&\leq \exp(C_{10}\theta^2n \Big(\sum_{l=1}^L\lambda_l(\|\bfmu_l\|+\|\bfSg_l^\frac{1}{2}\|)^2\Big)^2-\frac{n\theta t}{6})
\end{aligned}
\end{equation}
for some constant $C_{10}>0$. The first inequality follows from Markov's Inequality. When $\theta=\min\{\frac{t}{12 C_{10} \Big(\sum_{l=1}^L\lambda_l(\|\bfmu_l\|+\|\bfSg_l^\frac{1}{2}\|)^2\Big)^2},\frac{27}{4e}\sum_{l=1}^L\lambda_l(\|\bfmu_l\|+\|\bfSg_l^\frac{1}{2}\|)^2\}$, we have a modified Bernstein's Inequality for the Gaussian Mixture Model as follows
\begin{equation}
\begin{aligned}
\mathbb{P}(\frac{1}{n} \sum_{i=1}^n G_i\geq\frac{t}{6})\leq  \exp\Big(\max\{&-\frac{C_{10} n t^2}{144\Big(\sum_{l=1}^L\lambda_l(\|\bfmu_l\|+\|\bfSg_l^\frac{1}{2}\|)^2\Big)^2},\\
&-C_{11} n \sum_{l=1}^L\lambda_l(\|\bfmu_l\|+\|\bfSg_l^\frac{1}{2}\|)^2\cdot t\}\Big)\label{half_bernstein}
\end{aligned}
\end{equation}
for some constant $C_{11}>0$. We can obtain the same bound for $\mathbb{P}(-\frac{1}{n}\sum_{i=1}^n G_i\geq \frac{t}{6})$ by replacing $G_i$ as $-G_i$. Therefore, we have
\begin{equation}
\begin{aligned}
\mathbb{P}(|\frac{1}{n} \sum_{i=1}^n G_i|\geq\frac{t}{6})\leq2\exp\Big(\max\{&-\frac{C_{10} n t^2}{144\Big(\sum_{l=1}^L\lambda_l(\|\bfmu_l\|+\|\bfSg_l^\frac{1}{2}\|)^2\Big)^2},\\
&-C_{11}n\sum_{l=1}^L\lambda_l(\|\bfmu_l\|+\|\bfSg_l^\frac{1}{2}\|)^2\cdot t\}\Big)\label{entire_bernstein}
\end{aligned}
\end{equation}
Thus, as long as 
\begin{equation}\label{t_B_t} t\geq C_6\cdot \max\{\sum_{l=1}^L\lambda_l(\|\bfmu_l\|+\|\bfSg_l^\frac{1}{2}\|)^2\sqrt{\frac{dK\log{\frac{36r}{\epsilon}}+\log{\frac{4}{\delta}}}{n}},\frac{dK\log{\frac{36r}{\epsilon}}+\log{\frac{4}{\delta}}}{\sum_{l=1}^L\lambda_l(\|\bfmu_l\|+\|\bfSg_l^\frac{1}{2}\|)^2 n}\}
\end{equation}
for some large constant $C_6>0$, we have $\mathbb{P}(B_t)\leq\frac{\delta}{2}.$\\\\
2) \textbf{Upper bound on }$\mathbb{P}(A_t)$ and $\mathbb{P}(C_t)$. 
From Lemma \ref{lm: fraction bound}, we can obtain
\begin{equation}
\begin{aligned}
    &\sup_{\bfW\in\mathbb{B}(\bfW^*\bfP,r)}||\mathbb{E}_{\bfx\sim\sum_{l=1}^L\lambda_l\mathcal{N}(\bfmu_l,\bfSg_l)}[\nabla^2\ell(\bfW_{j(\bfW)};\bfx)]-\mathbb{E}_{\bfx\sim\sum_{l=1}^L\lambda_l\mathcal{N}(\bfmu_l,\bfSg_l)}[\nabla^2\ell(\bfW;\bfx)]||\\
    &\leq \sup_{\bfW\in\mathbb{B}(\bfW^*\bfP,r)}\frac{||\mathbb{E}_{\bfx\sim\sum_{l=1}^L\lambda_l\mathcal{N}(\bfmu_l,\bfSg_l)}[\nabla^2\ell(\bfW_{j(\bfW)};\bfx)]-\mathbb{E}_{\bfx\sim\sum_{l=1}^L\lambda_l\mathcal{N}(\bfmu_l,\bfSg_l)}[\nabla^2\ell(\bfW;\bfx)]||}{||\bfW-\bfW_{j(\bfW)}||_F}\\
    &\ \ \ \cdot\sup_{\bfW\in\mathbb{B}(\bfW^*\bfP,r)}||\bfW-\bfW_{j(\bfW)}||_F\\
    &\leq C_{12}\cdot d^{\frac{3}
{2}}K^{\frac{5}{2}}\sqrt{\sum_{l=1}^L\lambda_l(\|\bfmu_l\|_\infty+\|\bfSg_l^\frac{1}{2}\|)^2\sum_{l=1}^L\lambda_l(\|\bfmu_l\|_\infty+\|\bfSg_l^\frac{1}{2}\|)^4}\cdot\epsilon
\end{aligned}
\end{equation}
Therefore, $C_t$ holds if \begin{equation}t\geq C_{12}\cdot d^{\frac{3}
{2}}K^{\frac{5}{2}}\sqrt{\sum_{l=1}^L\lambda_l(\|\bfmu_l\|_\infty+\|\bfSg_l^\frac{1}{2}\|)^2\sum_{l=1}^L\lambda_l(\|\bfmu_l\|_\infty+\|\bfSg_l^\frac{1}{2}\|)^4}\cdot\epsilon\end{equation}
We can bound the $A_t$ as below.
\begin{equation}
\begin{aligned}
    &\mathbb{P}\Big(\sup_{\bfW\in\mathbb{B}(\bfW^*\bfP,r)}\frac{1}{n}||\sum_{i=1}^n[\nabla^2\ell(\bfW_{j(\bfW)};\bfx_i)-\nabla^2\ell(\bfW;\bfx_i)]||\geq\frac{t}{3}\Big)\\
    &\leq\frac{3}{t}\mathbb{E}_{\bfx\sim\sum_{l=1}^L\lambda_l\mathcal{N}(\bfmu_l,\bfSg_l)}\Big[\sup_{\bfW\in\mathbb{B}(\bfW^*\bfP,r)}\frac{1}{n}||\sum_{i=1}^n[\nabla^2\ell(\bfW_{j(\bfW)};\bfx_i)-\nabla^2\ell(\bfW;\bfx_i)]||\Big]\\
    &=\frac{3}{t}\mathbb{E}_{\bfx\sim\sum_{l=1}^L\lambda_l\mathcal{N}(\bfmu_l,\bfSg_l)}\Big[\sup_{\bfW\in\mathbb{B}(\bfW^*\bfP,r)}||\nabla^2\ell(\bfW_{j(\bfW)};\bfx_i)-\nabla^2\ell(\bfW;\bfx_i)||\Big]\\
    &\leq\frac{3}{t}\mathbb{E}\Big[\sup_{\bfW\in\mathbb{B}(\bfW^*\bfP,r)}\frac{||\nabla^2\ell(\bfW_{j(\bfW)};\bfx_i)-\nabla^2\ell(\bfW;\bfx_i)||}{||\bfW-\bfW_{j(\bfW)}||_F}\Big]\cdot\sup_{\bfW\in\mathbb{B}(\bfW^*\bfP,r)}||\bfW-\bfW_{j(\bfW)}||_F\\
    &\leq \frac{C_{12}\cdot d^{\frac{3}
{2}}K^{\frac{5}{2}}\sqrt{\sum_{l=1}^L\lambda_l(\|\bfmu_l\|_\infty+\|\bfSg_l^\frac{1}{2}\|)^2\sum_{l=1}^L\lambda_l(\|\bfmu_l\|_\infty+\|\bfSg_l^\frac{1}{2}\|)^4}\cdot\epsilon}{t},
\end{aligned}
\end{equation}
where the first inequality is by Markov's inequality, and the last inequality comes from Lemma \ref{lm: fraction bound}. Thus, taking 
\begin{equation}t\geq \frac{C_{12}\cdot d^{\frac{3}
{2}}K^{\frac{5}{2}}\sqrt{\sum_{l=1}^L\lambda_l(\|\bfmu_l\|_\infty+\|\bfSg_l^\frac{1}{2}\|)^2\sum_{l=1}^L\lambda_l(\|\bfmu_l\|_\infty+\|\bfSg_l^\frac{1}{2}\|)^4}\cdot\epsilon}{\delta}\label{bound_A_t}
\end{equation}
ensures that $\mathbb{P}(A_t)\leq\frac{\delta}{2}$.\\\\
3) \textbf{Final step}\\
Let $\epsilon=\frac{\delta}{C_{12}\cdot d^{\frac{3}
{2}}K^{\frac{5}{2}}\sqrt{\sum_{l=1}^L\lambda_l(\|\bfmu_l\|_\infty+\|\bfSg_l^\frac{1}{2}\|)^2\sum_{l=1}^L\lambda_l(\|\bfmu_l\|_\infty+\|\bfSg_l^\frac{1}{2}\|)^4}\cdot ndK}$ and $\delta=d^{-10}$, then from (\ref{t_B_t}) and (\ref{bound_A_t}) we need
\begin{equation}
\begin{aligned}
t>&\max\{\frac{1}{ndK},\ \ C_6\cdot \sum_{l=1}^L\lambda_l(\|\bfmu_l\|+\|\bfSg_l^\frac{1}{2}\|)^2\\
&\ \ \cdot\sqrt{\frac{dK\log(36rnd^{\frac{25}{2}}K^{\frac{7}{2}}\sqrt{\sum_{l=1}^L\lambda_l(\|\bfmu_l\|_\infty+\|\bfSg_l^\frac{1}{2}\|)^2\sum_{l=1}^L\lambda_l(\|\bfmu_l\|_\infty+\|\bfSg_l^\frac{1}{2}\|)^4})+\log{\frac{4}{\delta}}}{n}},\\
&\frac{dK\log(36rnd^{\frac{25}{2}}K^{\frac{7}{2}}\cdot\sqrt{\sum_{l=1}^L\lambda_l(\|\bfmu_l\|_\infty+\|\bfSg_l^\frac{1}{2}\|)^2\sum_{l=1}^L\lambda_l(\|\bfmu_l\|_\infty+\|\bfSg_l^\frac{1}{2}\|)^4})+\log{\frac{4}{\delta}}}{\sum_{l=1}^L\lambda_l(\|\bfmu_l\|+\|\bfSg_l^\frac{1}{2}\|)^2n}\}
\end{aligned}
\end{equation}
So by setting $t=\sum_{l=1}^L\lambda_l(\|\bfmu_l\|+\|\bfSg_l^\frac{1}{2}\|)^2\sqrt{\frac{dK\log{n}}{n}}$, as long as $n\geq C'\cdot dK\log{dK}$, we have \begin{equation}\mathbb{P}(\sup_{\bfW\in\mathbb{B}(\bfW^*\bfP,r)}||\nabla^2 f_n(\bfW)-\nabla^2\bar{f}(\bfW)||\geq C_6\cdot \sum_{l=1}^L\lambda_l(\|\bfmu_l\|+\|\bfSg_l^\frac{1}{2}\|)^2\sqrt{\frac{dK\log{n}}{n}})\leq d^{-10}\end{equation}\\

\subsection{Proof of Lemma \ref{lemma: convergence} and its supportive lemmas}\label{sec: convergence}
%\tcr{This proof is too simplified. Please include more details. One shall be able to understand the proof  without reading the reference. Even if we follow the idea of that paper, we need to describe explicitely what the proof idea and logic sequence are. Again, nobody will first read the proof of the reference and then read our paper.}

%We first present a lemma used in proving Lemma \ref{lemma: convergence} in Section \ref{subsec: lm_g_Bernstein} and then prove   Lemma \ref{lemma: convergence} in Section \ref{subsec: pf lm2}. 

%First, similar to the proof of \textbf{Lemma }\ref{lm: bernstein}, we can derive the following lemma in \textbf{Subsection }\ref{subsec: lm_g_Bernstein}, of which the proof can be found in \textbf{Section }\ref{subsec: pf lm_g_Bernstein}. The proof of \textbf{Lemma }\ref{lemma: convergence} is provided in \textbf{Subsection }\ref{subsec: pf lm2} based on \textbf{Lemma }\ref{lm: gradient_Bernstein}.
\subsubsection{Proof of Lemma \ref{lm: gradient_Bernstein}}
\iffalse
\begin{lemma}\label{lm: gradient_Bernstein}
If $r$ is defined in (\ref{radius}) for $\epsilon_0\in(0,\frac{1}{4})$, then with probability at least $1-d^{-10}$, we have\footnote{$\nabla \tilde{f}_n(\bfW)$ is defined as $\frac{1}{n} \sum_{i=1}^n ( \nabla  l(\bfW, \bfx_i, y_i) +\nu_i )$ in algorithm \ref{gd}}

\begin{equation}\sup_{\bfW\in\mathbb{B}(\bfW^*\bfP,r)}||\nabla \tilde{f}_n(\bfW)-\nabla \tilde{f}(\bfW)||\leq C_{13}\cdot\sqrt{K \sum_{l=1}^L\lambda_l(\|\bfmu_l\|+\|\bfSg_l\|)^2}\sqrt{\frac{d\log{n}}{n}}(1+\xi)\end{equation}
for some constant $C_{13}>0$, where $\bfP$ is a permutation matrix. 
\end{lemma}
\fi

\noindent Note that $\nabla \tilde{f}_n(\bfW)=\nabla f_n(\bfW)+\frac{1}{n}\sum_{i=1}^n \nu_i$, $\nabla \tilde{f}(\bfW)=\nabla \bar{f}(\bfW)+\mathbb{E}[\nu_i]=\nabla \bar{f}(\bfW)$. Therefore, we have
\begin{equation}
    \sup_{\bfW\in\mathbb{B}(\bfW^*\bfP,r)}||\nabla \tilde{f}_n(\bfW)-\nabla \tilde{f}(\bfW)||\leq \sup_{\bfW\in\mathbb{B}(\bfW^*\bfP,r)}||\nabla f_n(\bfW)-\nabla \bar{f}(\bfW)||+\|\frac{1}{n}\sum_{i=1}^n \nu_i\|
\end{equation}
\noindent Then, similar to the idea of the proof of Lemma \ref{lm: bernstein}, we adopt an $\epsilon$-covering net of the ball $\mathbb{B}(\bfW^*,r)$ to build a relationship between any arbitrary point in the ball and the points in the covering set. We can then divide the distance between $\nabla f_n(\bfW)$ and $\nabla \bar{f}(\bfW)$ into three parts, similar to (\ref{distance3parts}). (\ref{A_t'}) to (\ref{C_t'}) can be derived in a similar way as   (\ref{A_t}) to (\ref{C_t}), with ``$\nabla^2$'' replaced by ``$\nabla$''. Then we need to bound $\mathbb{P}(A_t')$, $\mathbb{P}(B_t')$ and $\mathbb{P}(C_t')$ respectively, where $A_t'$, $B_t'$ and $C_t'$ are defined below.
\begin{equation}
A_t'=\{\sup_{\bfW\in\mathbb{B}(\bfW^*\bfP,r)}\frac{1}{n}||\sum_{i=1}^n[\nabla\ell(\bfW;\bfx_i)-\nabla\ell(\bfW_{j(\bfW)};\bfx_i)]||\geq\frac{t}{3}\}\label{A_t'}
\end{equation}
\begin{equation}B_t'=\{\sup_{\bfW\in\mathbb{B}(\bfW^*\bfP,r)}||\frac{1}{n}\sum_{i=1}^n \nabla\ell(\bfW_{j(\bfW)};\bfx_i)-\mathbb{E}_{\bfx\sim\sum_{l=1}^L\lambda_l\mathcal{N}(\bfmu_l,\bfSg_l)}[\nabla\ell(\bfW_{j(\bfW)};\bfx_i)]||\geq\frac{t}{3}\}\label{B_t'}
\end{equation}
\begin{equation}
\begin{aligned}
C_t'=&\{\sup_{\bfW\in\mathbb{B}(\bfW^*\bfP,r)}||\mathbb{E}_{\bfx\sim\sum_{l=1}^L\lambda_l\mathcal{N}(\bfmu_l,\bfSg_l)}[\nabla\ell(\bfW_{j(\bfW)};\bfx_i)]\\
&-\mathbb{E}_{\bfx\sim\sum_{l=1}^L\lambda_l\mathcal{N}(\bfmu_l,\bfSg_l)}[\nabla\ell(\bfW;\bfx_i)]||\geq\frac{t}{3}\}\label{C_t'}
\end{aligned}
\end{equation}
(a) Upper bound of $\mathbb{P}(B_t')$. Applying Lemma 3 in \cite{MBM16}, we have
\begin{equation}
    \begin{aligned}
    &||\frac{1}{n}\sum_{i=1}^n \nabla\ell(\bfW_{j(\bfW)};\bfx_i)-\mathbb{E}_{\bfx\sim\sum_{l=1}^L\lambda_l\mathcal{N}(\bfmu_l,\bfSg_l)}[\nabla\ell(\bfW_{j(\bfW)};\bfx_i)]||\\
    \leq & 2\sup_{\bfv\in V_{\frac{1}{2}}}\Big|\left\langle \frac{1}{n}\sum_{i=1}^n \nabla\ell(\bfW_{j(\bfW)};\bfx_i)-\mathbb{E}_{\bfx\sim\sum_{l=1}^L\lambda_l\mathcal{N}(\bfmu_l,\bfSg_l)}[\nabla\ell(\bfW_{j(\bfW)};\bfx_i)], \bfv\right\rangle\Big|
    \end{aligned}
\end{equation}
Define $G_i'=\left\langle \bfv,(\nabla\ell(\bfW,\bfx_i)-\mathbb{E}_{\bfx\sim\sum_{l=1}^L\lambda_l\mathcal{N}(\bfmu_l,\bfSg_l)}[\nabla\ell(\bfW,\bfx_i)])\right\rangle$. Here $\bfv \in\mathbb{R}^{d}$. To compute $\nabla\ell(\bfW,\bfx_i)$, we require the derivation in Property \ref{prop: fx}. Then we can have an upper bound of $\zeta(\bfW)$ in (\ref{gradient}).
\begin{equation}
    \zeta(\bfW)=\left\{
    \begin{array}{rcl}\Big|-\frac{1}{K}\frac{1}{H(\bfW)}\phi'(\bfw_j^\top\bfx)\Big|\leq \frac{\phi(\bfw_j^\top\bfx)(1-\phi(\bfw_j^\top\bfx))}{K\cdot\frac{1}{K}\phi(\bfw_j^\top\bfx)}\leq 1, &y=1\\
    \Big|\frac{1}{K}\frac{1}{1-H(\bfW)}\phi'(\bfw_j^\top\bfx)\Big|\leq \frac{\phi(\bfw_j^\top\bfx)(1-\phi(\bfw_j^\top\bfx))}{K\cdot\frac{1}{K}(1-\phi(\bfw_j^\top\bfx))}\leq 1, &y=0
    \end{array}\right.
\end{equation}
Then we have an upper bound of $G_i'$.
\begin{equation}
\begin{aligned}
   |G_i'|&=\Big|\zeta_{j,l}\bfv^\top \bfx -\mathbb{E}_{\bfx\sim\sum_{l=1}^L\lambda_l\mathcal{N}(\bfmu_l,\bfSg_l)}[\zeta\bfv^\top \bfx ]\Big|\\
   &\leq |\bfv^\top \bfx|+\mathbb{E}_{\bfx\sim\sum_{l=1}^L\lambda_l\mathcal{N}(\bfmu_l,\bfSg_l)}[|\bfv^\top \bfx|]\label{G_i'}
\end{aligned}
\end{equation}
Following the idea of (\ref{E_Gi^p}) and (\ref{G_i_mgf}), and by $\bfv\in V_{\frac{1}{2}}$, we have
\begin{equation}
    \mathbb{E}[|G_i'|^p]\leq O\Big(\Big(\sum_{l=1}^L\lambda_l(\|\bfmu_l\|+\|\bfSg_l^\frac{1}{2}\|)^2\Big)^{\frac{p}{2}}\Big)
\end{equation}
\begin{equation}
    \mathbb{E}[\exp(\theta G_i')]\leq 1+O\Big(|e\theta^2|\sum_{l=1}^L\lambda_l(\|\bfmu_l\|+\|\bfSg_l^\frac{1}{2}\|)^2\Big)\label{Gi'_mgf}
\end{equation}
where (\ref{Gi'_mgf}) holds if $\theta\leq \frac{27}{4e}\sqrt{\sum_{l=1}^L\lambda_l(\|\bfmu_l\|+\|\bfSg_l\|)^2}$. Following the derivation of (\ref{P_Bt_initial}) and (\ref{B_t_quadratic}) to (\ref{t_B_t}), we have 
\begin{equation}\label{entire_bernstein_Gi'}
\begin{aligned}
&\mathbb{P}(|\frac{1}{n} \sum_{i=1}^n G_i'|\geq\frac{t}{6})\\
\leq&2\exp\Big(\max\big\{-\frac{C_{14} n t^2}{144\sum_{l=1}^L\lambda_l(\|\bfmu_l\|+\|\bfSg_l^\frac{1}{2}\|)^2},-C_{15}n\sqrt{\sum_{l=1}^L\lambda_l(\|\bfmu_l\|+\|\bfSg_l^\frac{1}{2}\|)^2}\cdot t\big\}\Big)
\end{aligned}
\end{equation}
for some constant $C_{14}>0$ and $C_{15}>0$. Moreover, we can obtain $\mathbb{P}(B_t')\leq\frac{\delta}{2}$ as long as 
\begin{equation}\label{t_B_t'} 
t\geq C_{13}\cdot \max\{\sqrt{\sum_{l=1}^L\lambda_l(\|\bfmu_l\|+\|\bfSg_l^\frac{1}{2}\|)^2}\sqrt{\frac{dK\log{\frac{18r}{\epsilon}}+\log{\frac{4}{\delta}}}{n}},\frac{dK\log{\frac{18r}{\epsilon}}+\log{\frac{4}{\delta}}}{\sqrt{\sum_{l=1}^L\lambda_l(\|\bfmu_l\|+\|\bfSg_l^\frac{1}{2}\|)^2}\cdot n}\}
\end{equation}
(b) For the upper bound of $\mathbb{P}(A_t')$ and $\mathbb{P}(C_t')$, we can first derive
\begin{equation}
    \begin{aligned}
    &\mathbb{E}_{\bfx\sim\sum_{l=1}^L\lambda_l\mathcal{N}(\bfmu_l,\bfSg_l)}\Big[ \sup_{\bfW\neq \bfW'\in\mathbb{B}(\bfW^*\bfP,r)}\frac{||\nabla\ell(\bfW,\bfx)-\nabla\ell(\bfW',\bfx)||}{||\bfW-\bfW'||_F}\Big]\\
    \leq &\mathbb{E}_{\bfx\sim\sum_{l=1}^L\lambda_l\mathcal{N}(\bfmu_l,\bfSg_l)}\Big[ \sup_{\bfW\neq \bfW'\in\mathbb{B}(\bfW^*\bfP,r)}\frac{|\zeta(\bfW)-\zeta(\bfW')|\cdot||\bfx||}{||\bfW-\bfW'||_F}\Big]\\
    \leq &\mathbb{E}_{\bfx\sim\sum_{l=1}^L\lambda_l\mathcal{N}(\bfmu_l,\bfSg_l)}\Big[ \sup_{\bfW\neq \bfW'\in\mathbb{B}(\bfW^*\bfP,r)}\frac{\max_{1\leq j,l\leq K}\{|\xi_{j,l}({\bfW''})|\}\cdot||\bfx||^2\sqrt{K}||\bfW-\bfW'||_F}{||\bfW-\bfW'||_F}\Big]\\
    \leq &\mathbb{E}_{\bfx\sim\sum_{l=1}^L\lambda_l\mathcal{N}(\bfmu_l,\bfSg_l)}\Big[ \sup_{\bfW\neq \bfW'\in\mathbb{B}(\bfW^*\bfP,r)}\frac{C_9\cdot||\bfx||^2\sqrt{K}||\bfW-\bfW'||_F}{||\bfW-\bfW'||_F}\Big]\\
    \leq & C_9\cdot3\sqrt{K}d \cdot \sum_{l=1}^L\lambda_l(\|\bfmu_l\|_\infty+\|\bfSg_l^\frac{1}{2}\|)^2
    \end{aligned}
\end{equation}
The first inequality is by (\ref{gradient}). The second inequality is by the Mean Value Theorem. The third step is by (\ref{xi_bound}). The last inequality is by Property \ref{prop: bound2}. Therefore, following the steps in part (2) of Lemma \ref{lm: bernstein}, we can conclude that $C_t'$ holds if 
\begin{equation}\label{t_C_t'}
    t\geq 3C_9\cdot \sqrt{K}d \cdot \sum_{l=1}^L\lambda_l(\|\bfmu_l\|_\infty+\|\bfSg_l^\frac{1}{2}\|)^2\cdot\epsilon
\end{equation}
Moreover, from (\ref{bound_A_t}) in Lemma \ref{lm: bernstein} we have that
\begin{equation}\label{t_A_t'}
    t\geq\frac{18C_9\cdot \sqrt{K}d \cdot \sum_{l=1}^L\lambda_l(\|\bfmu_l\|_\infty+\|\bfSg_l\|)^2\cdot\epsilon}{\delta}
\end{equation}
ensures $\mathbb{P}(A_t')\leq\frac{\delta}{2}$. Therefore, let $\epsilon=\frac{\delta}{18C_9\cdot \sqrt{K}d \cdot \sum_{l=1}^L\lambda_l(\|\bfmu_l\|_\infty+\|\bfSg_l\|)^2\cdot\epsilon\cdot ndK}$, $\delta=d^{-10}$ and $t=C_{13}\sqrt{K\sum_{l=1}^L\lambda_l(\|\bfmu_l\|+\|\bfSg_l\|)^2}\sqrt{\frac{d\log{n}}{n}}$, if $n\geq C''\cdot dK\log{dK}$ for some constant $C''>0$, we have
\begin{equation}
    \mathbb{P}(\sup_{\bfW\in\mathbb{B}(\bfW^*\bfP,r)}||\nabla f_n(\bfW)-\nabla \bar{f}(\bfW)||)\geq C_{13}\cdot \sqrt{K\sum_{l=1}^L\lambda_l(\|\bfmu_l\|+\|\bfSg_l\|)^2}\sqrt{\frac{d\log{n}}{n}}\leq d^{-10}
\end{equation}

\noindent By Hoeffding's inequality in \cite{V10} and Property \ref{prop: sub-Gaussian}, we have
\begin{equation}
\begin{aligned}
    &\mathbb{P}\Big(\frac{1}{n}\sum_{i=1}^n \|\nu_i\|_F\geq C_{13}\cdot\sqrt{\sum_{l=1}^L\lambda_l(\|\bfmu_l\|+\|\bfSg_l^\frac{1}{2}\|)^2}\sqrt{\frac{dK\log n}{n}}\xi\Big)\\
    \lesssim& \exp(-C_{13}^2\cdot\sum_{l=1}^L\lambda_l(\|\bfmu_l\|+\|\bfSg_l^\frac{1}{2}\|)^2\frac{\xi^2 dK\log n}{dK\xi^2})\\
    \lesssim& d^{-10}
\end{aligned}
\end{equation}
Therefore, 
\begin{equation}
    \begin{aligned}
        &\sup_{\bfW\in\mathbb{B}(\bfW^*\bfP,r)}||\nabla \tilde{f}_n(\bfW)-\nabla \tilde{f}(\bfW)||\\
        \leq & C_{13}\cdot \sqrt{K\sum_{l=1}^L\lambda_l(\|\bfmu_l\|+\|\bfSg_l^\frac{1}{2}\|)^2}\sqrt{\frac{d\log{n}}{n}}+\frac{1}{n}\sum_{i=1}^n \|\nu_i\|\\
        \leq& C_{13}\cdot \sqrt{K\sum_{l=1}^L\lambda_l(\|\bfmu_l\|+\|\bfSg_l^\frac{1}{2}\|)^2}\sqrt{\frac{d\log{n}}{n}}+\frac{1}{n}\sum_{i=1}^n \|\nu_i\|_F\\
        \leq& C_{13}\cdot \sqrt{K\sum_{l=1}^L\lambda_l(\|\bfmu_l\|+\|\bfSg_l^\frac{1}{2}\|)^2}\sqrt{\frac{d\log{n}}{n}}(1+\xi)\\
    \end{aligned}
\end{equation}

\subsection{Proof of Lemma \ref{lemma: tensor bound} and its supportive lemmas}\label{sec: tensor bound}
We need Lemma \ref{lm: p2} to Lemma \ref{lm: solution to 1st order moment}, which are stated in Section \ref{subsec: useful lemms lm3}, for the proof of Lemma \ref{lemma: tensor bound}. Section \ref{subsec: pf lm3} summarizes the proof of Lemma \ref{lemma: tensor bound}. The proofs of Lemma \ref{lm: p2} to Lemma \ref{lm: M1} are provided in Section \ref{subsec: pf lm_P2} to Section \ref{subsec: pf lm_M1}. Lemma \ref{lm: bound on subspace estimation} and Lemma \ref{lm: solution to 1st order moment} are cited from \cite{ZSJB17}.   Although \cite{ZSJB17} considers the standard Gaussian distribution, the proofs of Lemma \ref{lm: bound on subspace estimation} and \ref{lm: solution to 1st order moment} hold for any data distribution. Therefore, these two lemmas can be applied here directly.\\

The tensor initialization in \cite{ZSJB17} only holds for the standard Gaussian distribution.    We exploit a more general definition of tensors from \cite{JSA14} for the tensor initialization in our algorithm. We also develop new error bounds for the initialization.

\subsubsection{Proof of Lemma \ref{lm: p2}}\label{subsec: pf lm_P2}
\noindent From Assumption \ref{assumption1}, if the Gaussian Mixture Model is a symmetric probability distribution defined in (\ref{symmetric_GMM}), then by Definition \ref{def: M}, we have
\\
\begin{equation}
\begin{aligned}
    &||\widehat{\bfQ}_3(\bfI,\bfI,\boldsymbol{\alpha})-\bfQ_3(\bfI,\bfI,\boldsymbol{\alpha})||\\
    =&\Big|\Big|\frac{1}{n}\sum_{i=1}^n\Big[y_i\cdot  p(\bfx)^{-1}\sum_{l=1}^L\lambda_l(2\pi|\bfSg_l|)^{-\frac{d}{2}}\exp(-\frac{1}{2}(\bfx-\bfmu_l)\bfSg_l^{-1}(\bfx-\bfmu_l))\\
    &\cdot\Big(\big((\bfx-\bfmu_l)\bfSg_l^{-1}\big)^{\otimes 3}-\big((\bfx-\bfmu_l)\bfSg_l^{-1}\big)\widetilde{\otimes}\bfSg_l^{-1}\Big)\Big](\bfI,\bfI,\boldsymbol{\alpha})\\
    -&\mathbb{E}\Big[y\cdot  p(\bfx)^{-1}\sum_{l=1}^L\lambda_l(2\pi|\bfSg_l|)^{-\frac{d}{2}}\exp(-\frac{1}{2}(\bfx-\bfmu_l)\bfSg_l^{-1}(\bfx-\bfmu_l))\\
    &\cdot\Big(\big((\bfx-\bfmu_l)\bfSg_l^{-1}\big)^{\otimes 3}-\big((\bfx-\bfmu_l)\bfSg_l^{-1}\big)\widetilde{\otimes}\bfSg_l^{-1}\Big)\Big](\bfI,\bfI,\boldsymbol{\alpha})\Big|\Big|\label{distance_M3}
\end{aligned}
\end{equation}
Following  \cite{ZSJB17}, 
  %a special outer product 
  $\widetilde{\otimes}$   is defined such that for any $\bfv\in\mathbb{R}^{d_1}$ and $\bfZ\in\mathbb{R}^{d_1\times d_2}$, 
\begin{equation}\label{outer_prod_sp}
    \bfv\widetilde{\otimes}\bfZ=\sum_{i=1}^{d_2}(\bfv\otimes\bfz_i\otimes\bfz_i+\bfz_i\otimes\bfv\otimes\bfz_i+\bfz_i\otimes\bfz_i\otimes\bfv),
\end{equation}
where $\bfz_i$ is the $i$-th column of $\bfZ$.
By Definition \ref{def: M}, we have
\begin{equation}
\begin{aligned}
    &\Big|\Big|\Big[y\cdot p(\bfx)^{-1}\sum_{l=1}^L\lambda_l(2\pi|\bfSg_l|)^{-\frac{d}{2}}\exp(-\frac{1}{2}(\bfx-\bfmu_l)\bfSg_l^{-1}(\bfx-\bfmu_l))\\
    &\cdot\Big(\big((\bfx-\bfmu_l)\bfSg_l^{-1}\big)^{\otimes 3}-\big((\bfx-\bfmu_l)\bfSg_l^{-1}\big)\widetilde{\otimes}\bfSg_l^{-1}\Big)\Big](\bfI,\bfI,\boldsymbol{\alpha})\Big|\Big|\\
    &\lesssim \Big|\Big| \frac{\sum_{l=1}^L\lambda_l(2\pi|\bfSg_l|)^{-\frac{d}{2}}\exp(-\frac{1}{2}(\bfx-\bfmu_l)\bfSg_l^{-1}(\bfx-\bfmu_l))\cdot\big((\bfx-\bfmu_l)\bfSg_l^{-1}\big)^{\otimes 2}\big(\boldsymbol{\alpha}^\top\bfSg_l^{-1}(\bfx-\bfmu_l)\big)}{\sum_{l=1}^L\lambda_l(2\pi|\bfSg_l|)^{-\frac{d}{2}}\exp(-\frac{1}{2}(\bfx-\bfmu_l)\bfSg_l^{-1}(\bfx-\bfmu_l))}\Big|\Big|\\
    &\lesssim  ||\sigma_{\min}^{-6}(\bfx^\top\boldsymbol{\alpha})\bfx\bfx^\top||\label{P_2_upper}
\end{aligned}
\end{equation}
The first step of (\ref{P_2_upper}) is because $(\bfx-\bfmu_l)\bfSg_l)^{\otimes 2}(\boldsymbol{\alpha}^\top \bfSg_l^{-1}(\bfx-\bfmu_l))$ is the dominant term of the entire expression, and $y\leq 1$.  The second step is because the expression can be considered as a normalized weighted summation of $((\bfx-\bfmu_l)\bfSg_l)^{\otimes 2}(\boldsymbol{\alpha}^\top \bfSg_l^{-1}(\bfx-\bfmu_l))$ and $(\bfx^\top\boldsymbol{\alpha})\bfx\bfx^\top$ is its dominant term.
%\tcr{why does this hold?}\\
Define $S_m(\bfx)=(-1)^m\frac{\nabla^m_\bfx p(\bfx)}{p(\bfx)}$, where $p(\bfx)$ is the probability density function of the random variable $\bfx$. From Definition \ref{def: M}, we can verify that 
\begin{equation}\label{MjSm}
    \bfQ_j=\mathbb{E}[y\cdot S_m(\bfx)]\ \ \ j\in\{1,2,3\}
\end{equation}
Then define $Gp_i=\left\langle \bfv,([y_i\cdot S_3(\bfx_i)](\bfI_d,\bfI_d,\boldsymbol{\alpha})-\mathbb{E}\big[[y_i\cdot S_3(\bfx_i)](\bfI_d,\bfI_d,\boldsymbol{\alpha})\big]\bfv)\right\rangle$, where $||\bfv||=1$, then $\mathbb{E}[Gp_i]=0$. Similar to the proof of (\ref{G_i}), (\ref{E_Gi^p}), and (\ref{G_i_mgf}) in Lemma \ref{lm: bernstein}, we have
\begin{equation}|Gp_i|^p\lesssim \big|\sigma_{\min}^{-6}(\bfx_i^\top\boldsymbol{\alpha})(\bfx_i^\top\bfv)^2+\mathbb{E}_{\bfx\sim\sum_{l=1}^L\mathcal{N}(\bfmu_l,\bfSg_l)}[\sigma_{\min}^{-6}(\bfx_i^\top\boldsymbol{\alpha})(\bfx_i^\top\bfv)^2]\big|^p
\end{equation}
\begin{equation}
\begin{aligned}
\mathbb{E}[|Gp_i|^p]&\lesssim \big(\mathbb{E}_{\bfx\sim\sum_{l=1}^L\mathcal{N}(\bfmu_l,\bfSg_l)}[\sigma_{\min}^{-6}(\bfx_i^\top\boldsymbol{\alpha})(\bfx_i^\top\bfv)^2]\big)^p\\
&\leq \sigma_{\min}^{-6p}\mathbb{E}_{\bfx\sim\sum_{l=1}^L\mathcal{N}(\bfmu_l,\bfSg_l)}[(\bfx^\top\boldsymbol{\alpha})^2]^{\frac{p}{2}}\mathbb{E}_{\bfx\sim\sum_{l=1}^L\mathcal{N}(\bfmu_l,\bfSg_l)}[(\bfx^\top\bfv)^4]^{\frac{p}{2}}\\
&\leq \tau^{6p}\sqrt{D_2(\Psi)D_4(\Psi)}^p
\end{aligned}
\end{equation}
\begin{equation}
\begin{aligned}
\mathbb{E}[\exp(\theta Gp_i)]&\lesssim 1+\sum_{p=2}^\infty\frac{\theta^p\mathbb{E}[|Gp_i|^p]}{p!}\lesssim 1+\sum_{p=2}^\infty \frac{|e\theta|^p\tau^{6p}(D_2(\Psi)D_4(\Psi))^{\frac{p}{2}}}{p^p}\\
&\lesssim 1+\theta^2\tau^{12}D_2(\Psi)D_4(\Psi)
\end{aligned}
\end{equation}
%\tcr{why does the third inequality of the above equation hold?}
Hence, similar to the derivation of (\ref{B_t_quadratic}), we have \begin{equation}\mathbb{P}\Big(\frac{1}{n}\sum_{i=1}^n Gp_i\geq t\Big)\leq \exp\Big(-n\theta t+C_{16}n\theta^2\big(\tau^{6}\sqrt{D_2(\Psi)D_4(\Psi)}\big)^2\Big)\end{equation}
for some constant $C_{16}>0$. Let $\theta=\frac{t}{2C_{16}\big(\tau^{6}\sqrt{D_2(\Psi)D_4(\Psi)}\big)^2}$ and $t=\delta_1^2(\bfW^*)\cdot\big(\tau^{6}\sqrt{D_2(\Psi)D_4(\Psi)}\big)\cdot\sqrt{\frac{d\log{n}}{n}}$, then we have
\begin{equation}||\widehat{\bfQ}_3(\bfI_d,\bfI_d,\boldsymbol{\alpha})-\bfQ_3(\bfI_d,\bfI_d,\boldsymbol{\alpha})||\leq \delta_1(\bfW^*)^2\cdot\big(\tau^6\sqrt{D_2(\Psi)D_4(\Psi)}\big)\cdot\sqrt{\frac{d\log{n}}{n}}\label{sym_case}
\end{equation}
with probability at least $1-2n^{-\Omega(\delta_1^4(\bfW^*) d)}$.\\
If the Gaussian Mixture Model is not a symmetric distribution which is defined in (\ref{symmetric_GMM}), we would have a similar result as follows.\\
\begin{equation}||\widehat{\bfQ}_2-\bfQ_2||=\Big|\Big|\frac{1}{n}\sum_{i=1}^n[y_i\cdot S_2(\bfx)]-\mathbb{E}[y\cdot S_2(\bfx)]\Big|\Big|\end{equation}
\begin{equation}||y_i\cdot S_2(\bfx_i)||\lesssim ||\sigma_{\min}^{-4}\frac{1}{K}\sum_{j=1}^K\phi({\bfw_j^*}^\top \bfx_i)\bfx_i\bfx_i^\top||\end{equation}
Then define $Gp_i'=\left\langle \bfv,([y_i\cdot S_2(\bfx_i)]-\mathbb{E}\big[y_i\cdot S_2(\bfx_i)\big]\bfv)\right\rangle$, where $||\bfv||=1$, then $\mathbb{E}[Gp_i']=0$. Similar to the proof of (\ref{G_i}), (\ref{E_Gi^p}) and (\ref{G_i_mgf}) in Lemma \ref{lm: bernstein}, we have
\begin{equation}|Gp_i'|^p\lesssim \big|\sigma_{\min}^{-4}(\bfx_i^\top\bfv)^2+\mathbb{E}_{\bfx\sim\sum_{l=1}^L\mathcal{N}(\bfmu_l,\bfSg_l)}[\sigma_{\min}^{-4}(\bfx_i^\top\bfv)^2]\big|^p\end{equation}
\begin{equation}\mathbb{E}[|Gp_i'|^p]\lesssim \big(\mathbb{E}_{\bfx\sim\sum_{l=1}^L\mathcal{N}(\bfmu_l,\bfSg_l)}[\sigma_{\min}^{-4}(\bfx_i^\top\bfv)^2]\big)^p\leq \tau^{4p}D_2(\Psi)^p\end{equation}
\begin{equation}
\begin{aligned}\mathbb{E}[\exp(\theta Gp_i')]&\lesssim 1+\sum_{p=2}^\infty\frac{\theta^p\mathbb{E}[|Gp_i|^p]}{p!}\lesssim 1+\sum_{p=2}^\infty \frac{|e\theta|^p\tau^{4p}D_2(\Psi)^{p}}{p^p}\\
&\lesssim 1+\theta^2\tau^{8}D_2(\Psi)^2
\end{aligned}
\end{equation}
%\tcr{why does the third inequality of the above equation hold?}
Hence, similar to the derivation of (\ref{B_t_quadratic}), we have \begin{equation}\mathbb{P}\Big(\frac{1}{n}\sum_{i=1}^n Gp_i\geq t\Big)\leq \exp\Big(-n\theta t+C_{17}n\theta^2\big(\tau^{4}D_2(\Psi)\big)^2\Big)\end{equation}
for some constant $C_{17}>0$. Let $\theta=\frac{t}{2C_{17}\big(\tau^{4}D_2(\Psi)\big)^2}$ and $t=\delta_1^2(\bfW^*)\cdot\big(\tau^{4}D_2(\Psi)\big)\cdot\sqrt{\frac{d\log{n}}{n}}$, then we have
\begin{equation}
\begin{aligned}
&||\widehat{\bfQ}_2-\bfQ_2||\lesssim \delta_1^2(\bfW^*)\cdot\tau^{4}D_2(\Psi)\cdot\sqrt{\frac{d\log{n}}{n}}\\
\lesssim &\sqrt{\frac{d\log{n}}{n}}\cdot \delta_1^2(\bfW^*)\cdot\tau^{6}\sqrt{D_2(\Psi)D_4(\Psi)} \label{nonsym_case}
\end{aligned}
\end{equation}
with probability at least $1-2n^{-\Omega(\delta_1^4(\bfW^*)d)}$.\\\\

\subsubsection{Proof of Lemma \ref{lm: R3}}\label{subsec: pf lm_R3}
\noindent We consider each component of $y=\frac{1}{K}\sum_{i=1}^K\phi({\bfw_i^*}^\top \bfx)$. \\
Define $\bfT_i(\bfx): \mathbb{R}^d\rightarrow\mathbb{R}^{K\times K\times K}$ such that \begin{equation}\bfT_i(\bfx)=[\phi({\bfw_i^*}^\top \bfx)\cdot S_3(\bfx)](\widehat{\bfU},\widehat{\bfU},\widehat{\bfU})\end{equation}
We flatten $\bfT_i(\bfx): \mathbb{R}^{d}\rightarrow\mathbb{R}^{K\times K\times K}$ along the first dimension to obtain the function $\bfB_i(\bfx): \mathbb{R}^d\rightarrow \mathbb{R}^{K\times K^2}$. Similar to the derivation of the last step of Lemma E.8 in \cite{ZSJB17}, we can obtain $\|\bfT_i(\bfx)\|\leq \|\bfB_i(\bfx)\|$. By (\ref{distance_M3}), we have \\
\begin{equation}||\bfB_i(\bfx)||\lesssim \sigma_{\min}^{-6}\frac{1}{K}\sum_{j=1}^K\phi({\bfw_j^*}^\top \bfx_i)(\widehat{\bfU}^\top\bfx)^3\end{equation}
Define $Gr_i=\left\langle \bfv,\bfB_i(\bfx_i))-\mathbb{E}[ \bfB_i(\bfx_i)]\bfv)\right\rangle$, where $||\bfv||=1$, so $\mathbb{E}[Gr_i]=0$.  Similar to the proof of (\ref{G_i}), (\ref{E_Gi^p}) and (\ref{G_i_mgf}) in Lemma \ref{lm: bernstein}, we have
\begin{equation}|Gr_i|^p\lesssim \big|\sigma_{\min}^{-6}(\bfv^\top\widehat{\bfU}^\top\bfx)^3+\mathbb{E}_{\bfx\sim\sum_{l=1}^L\mathcal{N}(\bfmu_l,\bfSg_l)}[\sigma_{\min}^{-6}(\bfv^\top\widehat{\bfU}^\top\bfx)^3]\big|^p\end{equation}
\begin{equation}\mathbb{E}[|Gr_i|^p]\lesssim \big(\mathbb{E}_{\bfx\sim\sum_{l=1}^L\mathcal{N}(\bfmu_l,\bfSg_l)}[\sigma_{\min}^{-6}(\bfv^\top\widehat{\bfU}^\top\bfx)^3]\big)^p\lesssim \tau^{6p}\sqrt{D_6(\Psi)}^p\end{equation}
\begin{equation}
\begin{aligned}\mathbb{E}[\exp(\theta Gr_i)]&\lesssim 1+\sum_{p=2}^\infty\frac{\theta^p\mathbb{E}[|Gr_i|^p]}{p!}\lesssim 1+\sum_{p=2}^\infty \frac{|e\theta|^p\tau^{6p}D_6(\Psi)^{\frac{p}{2}}}{p^p}\\
&\leq 1+\theta^2(\tau^{12}\sqrt{D_6(\Psi)})^2
\end{aligned}
\end{equation}
Hence, similar to the derivation of (\ref{B_t_quadratic}), we have \begin{equation}\mathbb{P}\Big(\frac{1}{n}\sum_{i=1}^n Gr_i\geq t\Big)\leq \exp\Big(-n\theta t+C_{18}\theta^2\big(\tau^{6}\sqrt{D_6(\Psi)}\big)^2\Big)\end{equation}
for some constant $C_{18}>0$. Let $\theta=\frac{t}{C_{18}\big(\tau^{6}\sqrt{D_6(\Psi)}\big)^2}$ and $t=\delta_1^2(\bfW^*)\cdot\big(\tau^{6} \sqrt{D_6(\Psi)}\big)\cdot\sqrt{\frac{\log{n}}{n}}$, then we have
\begin{equation}||\widehat{\bfR}_3-\bfR_3||\lesssim \delta_1(\bfW^*)^2\cdot\big(\tau^{6} \sqrt{D_6(\Psi)}\big)\cdot\sqrt{\frac{\log{n}}{n}}\end{equation}
with probability at least $1-2n^{-\Omega(\delta_1^4(\bfW^*)) }$.\\
\\\\

\subsubsection{Proof of Lemma \ref{lm: M1}}\label{subsec: pf lm_M1}
\noindent From Definition \ref{def: M}, we have
\begin{equation}||\widehat{\bfQ}_1-\bfQ_1||=\Big|\Big|\frac{1}{n}\sum_{i=1}^n[y_i\cdot S_1(\bfx)]-\mathbb{E}[y\cdot S_1(\bfx)]\Big|\Big|.\end{equation}
Based on Definition \ref{def: M}, 
\begin{equation}
\begin{aligned}\Big|\Big|[y_i\cdot S_1(\bfx_i)]\Big|\Big| &   \lesssim \Big|\Big| \frac{\sum_{l=1}^L\lambda_l\lambda_l(2\pi\prod_{k=1}^d\sigma_{lk}^2)^{-\frac{d}{2}}\exp(-\frac{1}{2}(\bfx-\bfmu_l)\bfSg_l^{-1}(\bfx-\bfmu_l))\cdot(\bfx-\bfmu_l)\bfSg_l^{-1}}{\sum_{l=1}^L\lambda_l\lambda_l(2\pi\prod_{k=1}^d\sigma_{lk}^2)^{-\frac{d}{2}}\exp(-\frac{1}{2}(\bfx-\bfmu_l)\bfSg_l^{-1}(\bfx-\bfmu_l))}\Big|\Big|\\
&\lesssim \Big|\Big|\sigma_{\min}^{-2}\frac{1}{K}\sum_{j=1}^K\phi({\bfw_j^*}^\top \bfx_i)\bfx_i\Big|\Big|
\end{aligned}\end{equation}
Define $Gq_i=\left\langle \bfv,([y_i\cdot S_1(\bfx_i)]-\mathbb{E}\big[[y_i\cdot S_1(\bfx_i)]\big]\bfv)\right\rangle$, where $||\bfv||=1$, so $\mathbb{E}[Gq_i]=0$.  Similar to the proof of (\ref{G_i}), (\ref{E_Gi^p}), and (\ref{G_i_mgf}) in Lemma \ref{lm: bernstein}, we have
\begin{equation}|Gq_i|^p\lesssim \big|\sigma_{\min}^{-2}(\bfx_i^\top\bfv)+\mathbb{E}_{\bfx\sim\sum_{l=1}^L\mathcal{N}(\bfmu_l,\bfSg_l)}[\sigma_{\min}^{-2}(\bfx_i^\top\bfv)]\big|^p\end{equation}
\begin{equation}\mathbb{E}[|Gq_i|^p]\lesssim \big(\mathbb{E}_{\bfx\sim\sum_{l=1}^L\mathcal{N}(\bfmu_l,\bfSg_l)}[\sigma_{\min}^{-2}(\bfx_i^\top\bfv)]\big)^p\leq \tau^{2p}\sqrt{D_2(\Psi)}^p\end{equation}
\begin{equation}
    \begin{aligned}
        \mathbb{E}[\exp(\theta Gq_i)]&\lesssim 1+\sum_{p=2}^\infty\frac{\theta^p\mathbb{E}[|Gq_i|^p]}{p!}\lesssim 1+\sum_{p=2}^\infty \frac{|e\theta|^p\tau^{2p}D_2(\Psi)^{\frac{p}{2}}}{p^p}\\
        &\leq 1+\theta^2(\tau^{2}\sqrt{D_2(\Psi)})^2
    \end{aligned}
\end{equation}
Hence, similar to the derivation of (\ref{B_t_quadratic}), we have \begin{equation}|\mathbb{P}\Big(\frac{1}{n}\sum_{i=1}^n Gq_i\geq t\Big)\leq \exp\Big(-n\theta t+C_{19}\theta^2\big(\tau^{2}\sqrt{D_2(\Psi)}\big)^2\Big)\end{equation}
for some constant $C_{19}>0$. Let $\theta=\frac{t}{C_{19}\big(\tau^{2}\sqrt{D_2(\Psi)}\big)^2}$ and $t=\big(\tau^2 \sqrt{D_2(\Psi)}\big)\cdot\sqrt{\frac{d\log{n}}{n}}$, then we have
\begin{equation}||\widehat{\bfQ}_1-\bfQ_1||\lesssim \big(\tau^2\sqrt{D_2(\Psi))}\big)\cdot\sqrt{\frac{d\log{n}}{n}}\end{equation}
with probability at least $1-2n^{-\Omega(d)}$.

\end{document}